\documentclass{article}

\usepackage{graphicx}

\graphicspath{{figures/}}

\usepackage{amssymb}
\usepackage{svg}
\usepackage{arxiv}
\usepackage[utf8]{inputenc} 
\usepackage[T1]{fontenc}    
\usepackage{url}            
\usepackage{booktabs}       
\usepackage{amsfonts}       
\usepackage{nicefrac}       
\usepackage{microtype}      
\usepackage{lipsum}
\usepackage{graphicx}
\usepackage[colorlinks]{hyperref}
\usepackage{tabularx}
\usepackage{multirow}
\usepackage{amsmath} 
\usepackage{comment}
\usepackage{algorithm}
\usepackage{amsmath}
\usepackage{algpseudocode}
\usepackage{booktabs}
\usepackage{caption}
\usepackage{xcolor}
\usepackage{tcolorbox}
\tcbuselibrary{breakable} 
\usepackage{appendix}
\usepackage{subcaption}
\usepackage[
  style=numeric, 
  sorting=none,
  maxbibnames=10,    
  minbibnames=3,     
  maxcitenames=1     
]{biblatex}

\bibliography{template}

\graphicspath{ {./images/} }

\title{\textbf{SkyReels-V4}: Multi-modal Video-Audio Generation, Inpainting and Editing model}

\author{
 SkyReels Team \\
  Skywork AI \\
}

\begin{document}
\maketitle

\begin{abstract}
SkyReels-V4 is a unified multi-modal video foundation model for joint video–audio generation, inpainting, and editing. The model adopts a dual-stream Multimodal Diffusion Transformer (MMDiT) architecture, where one branch synthesizes video and the other generates temporally aligned audio, while sharing a powerful text encoder based on the Multimodal Large Language Models (MLLM). SkyReels-V4 accepts rich multi-modal instructions, including text, images, video clips, masks, and audio references. By combining the MLLM’s multi-modal instruction-following capability with in-context learning in the video-branch MMDiT, the model can inject fine-grained visual guidance under complex conditioning, while the audio-branch MMDiT simultaneously leverages audio references to guide sound generation. On the video side, we adopt a channel-concatenation formulation that unifies a wide range of inpainting-style tasks—such as image-to-video, video extension, and video editing—under a single interface, and naturally extends to vision-referenced inpainting and editing via multi-modal prompts. SkyReels-V4 supports up to 1080p resolution, 32 FPS, and 15-second duration, enabling high-fidelity, multi-shot, cinema-level video generation with synchronized audio. To make such high-resolution, long-duration generation computationally feasible, we introduce an efficiency strategy: Joint generation of low-resolution full sequences and high-resolution keyframes, followed by dedicated super-resolution and frame interpolation models. To our knowledge, SkyReels-V4 is the first 
video foundation model that simultaneously supports multi-modal input, joint video–audio generation, and a unified treatment of generation, inpainting, and editing, while maintaining strong efficiency and quality at cinematic resolutions and durations.
\end{abstract}

\section{Introduction}

From the earliest days of cinema, filmmakers have understood that compelling storytelling demands the seamless orchestration of sight and sound. When the Lumière Brothers first projected \textit{L'Arrivée d'un train en gare de La Ciotat} in 1895, audiences recoiled at the silent image of an approaching locomotive; yet it was not until \textit{The Jazz Singer} synchronized Al Jolson's voice with his moving lips in 1927 that cinema truly came alive. This historical evolution from silent film to ``talkies'' reflects a fundamental truth about human perception: vision provides spatial structure and compositional context, while audition conveys temporal rhythm, emotional texture, and narrative continuity. Neither modality alone suffices—their synergy creates the immersive experiences that define modern media.

Over the past year, the field of video generation has witnessed a decisive paradigm shift from unimodal synthesis toward \textbf{joint audio-video generation}. Proprietary commercial systems such as Veo-3.1~\cite{veo31}, Sora-2~\cite{sora2}, Kling-2.6~\cite{Kling26}, Gen-4.5~\cite{Gen45}, Seedance-1.5~\cite{seedance2025seedance15pronative}, Wan-2.6~\cite{Wan26} have transformed video generation capabilities into practical, utility-driven tools that natively produce synchronized audio alongside visual content. These systems mark a significant departure from earlier text-to-video (T2V) or video-to-audio (V2A) pipelines, which handled one modality at a time and often suffered from audio-visual asynchrony, lip-speech mismatches, and degraded unimodal quality. 

In parallel, substantial progress has been made in \textbf{multimodal-referenced video generation}, where models accept diverse conditioning inputs beyond text. For instance, Vidu~\cite{ViduQ2} pioneered Reference-to-Video generation, enabling coherent synthesis from multiple referenced images. Runway Aleph~\cite{runwayaleph} introduced state-of-the-art in-context video editing, performing a wide range of operations—adding, removing, and transforming objects, generating arbitrary scene angles, and modifying style and lighting—directly on input videos. In the audio-to-video domain, systems such as Omnihuman-1/1.5~\cite{omnihuman1,omnihuman15}, SkyReels-A3~\cite{SkyReels_A3,fei2025skyreels}, KlingAvatar~\cite{ding2025klingavatargroundingmultimodalinstructions,klingteam2025klingavatar20technicalreport}, and Multitalk~\cite{kong2025let} have demonstrated compelling talking-head synthesis and audio-driven animation. Recently, Kling-Omni~\cite{klingteam2025klingomnitechnicalreport} emerged as the first model to support both image and video references for video generation, yet it remains limited to visual synthesis without audio output. Alongside these developments, concurrent works including Kling-3.0~\cite{Kling30}, Seedance-2.0~\cite{seedance20}, and Vidu-Q3~\cite{viduQ3} have taken meaningful steps toward bridging this gap, each integrating multimodal inputs with joint video–audio generation within a unified model. Nevertheless, these systems still fall short of a fully comprehensive solution.

Despite these advances, \textbf{no existing system simultaneously unifies multimodal inputs (text, images, videos, masks, and audio references), joint video–audio generation, comprehensive inpainting, and editing capabilities within a single framework}. Current state-of-the-art models remain fundamentally fragmented: audio-driven systems such as Omnihuman-1/1.5 and Multitalk adopt shallow fusion mechanisms (e.g., cross-attention or lightweight adapters) that fail to fully align audio-visual representations, while multimodal-referenced models such as Kling-Omni focus exclusively on visual conditioning without native audio synthesis. Although recent efforts—Kling-3.0, Seedance-2.0, and Vidu-Q3—have taken meaningful steps toward joint video–audio generation under multimodal inputs, none of these systems natively integrates comprehensive inpainting and fine-grained editing within a unified architecture. The field therefore still lacks a unified foundation model capable of seamlessly handling generation, inpainting, and editing conditioned on arbitrary combinations of text, images, videos, masks and audio references.

To address these limitations, we present \textbf{SkyReels-V4}, a multi-modal video foundation model that jointly generates video and audio while unifying generation, inpainting, and editing within a single architecture. SkyReels-V4 is built on a dual-stream MMDiT (Multi-Modal Diffusion Transformer) design: one branch is dedicated to video synthesis, and the other to audio generation. The two branches share a common text encoder instantiated by a strong MLLM that provides multi-modal understanding and instruction-following across text, images, videos and audios. This shared MLLM backbone allows SkyReels-V4 to condition on diverse inputs—including text, images, videos, and audio references—in a unified, semantically coherent manner.

To support a broad set of video manipulation tasks, we design the video branch around a channel concatenation formulation that treats diverse operations as special cases of inpainting. Specifically, tasks such as image-to-video generation, video extension, and video editing are expressed via masked inputs and additional conditioning channels that are concatenated with the latent representation. This unified inpainting perspective allows SkyReels-V4 to handle heterogeneous workflows within a single model, while the underlying MLLM enables visually referenced inpainting: for example, altering a character's clothing based on a reference image, extending a shot while preserving composition from a provided frame, or editing specific regions indicated by a mask.

SkyReels-V4 is designed not only for flexibility but also for cinematic quality. The model supports video generation at up to 1080p resolution, 32 FPS, and 15-second duration, and it can handle multi-shot sequences suitable for film-like storytelling. Achieving such resolutions and lengths with diffusion-based architectures is computationally demanding; naive scaling leads to prohibitive memory and time costs. To overcome this, we introduce an efficiency strategy. Instead of direct 1080p generation, we present a joint low-resolution / high-resolution keyframe generation, where the model first produces a low-resolution full sequence and high-resolution keyframes, followed by specialized super-resolution and frame interpolation modules that reconstruct a temporally consistent, high-resolution video. This design enables SkyReels-V4 to achieve surprisingly high generation speed even for long, high-resolution videos with synchronized audio, making it practical for real-world creative and production environments.

To the best of our knowledge, SkyReels-V4 is the first system worldwide that simultaneously supports (i) rich multi-modal inputs (text, images, video, masks, and audio), (ii) joint video–audio generation, and (iii) a unified framework for generation, inpainting, and editing, while scaling effectively to high-resolution, long-duration outputs. This combination of capabilities positions SkyReels-V4 as a versatile foundation model for next-generation video creation and editing. 

Extensive experiments demonstrate the superior performance of SkyReels-V4 compared to current state-of-the-art methods. Our model achieves state-of-the-art results on the Artificial Analysis Arena~\cite{ArtificialAnalysis}. Comprehensive human evaluation through SkyReels-VABench reveals that SkyReels-V4 significantly outperforms proprietary commercial systems, with particular strengths in 
instruction following, motion quality, and complex multi-shot narratives. Additionally, the model demonstrates robust performance across diverse multimodal conditioning tasks including reference-to-video, motion-to-video, and video editing, validating its effectiveness as a unified foundation model for professional video-audio content creation.

In summary, our contributions are:
\begin{itemize}
    \item We introduce SkyReels-V4, a dual-stream MMDiT-based foundation model that jointly generates video and audio under multi-modal instruction and reference inputs.
    \item We propose a unified channel-concatenation inpainting framework for video, enabling image-to-video, video extension, video editing, and vision-referenced inpainting within a single architecture.
    \item We design an efficiency scheme — Joint low-res / high-res keyframe generation with super-resolution and interpolation—that makes 1080p, 32 FPS, 15-second, multi-shot video generation with synchronized audio computationally feasible.
    \item We demonstrate that SkyReels-V4 is, to our knowledge, the first model to unify multi-modal input, joint video–audio generation, and generation/inpainting/editing tasks at cinematic quality and speed, setting a new baseline for multi-modal video foundation models.
\end{itemize}

\section{Related Work}
\label{sec:related_work}
\subsection{Video Generative Models}
Diffusion models have transformed video generation, evolving from early 2D+1D architectures like Video Diffusion Models~\cite{ho2022videodiffusionmodels} and AnimateDiff~\cite{guo2024animatediffanimatepersonalizedtexttoimage} to DiT-based frameworks~\cite{peebles2023scalablediffusionmodelstransformers}. Sora~\cite{videoworldsimulators2024} demonstrated the effectiveness of large-scale training with spatiotemporal attention. While closed-source systems (Veo-3.1~\cite{veo31}, Kling-O1~\cite{klingteam2025klingomnitechnicalreport}, Sora-2~\cite{sora2}, Hailuo-2.3~\cite{Hailuo23}, Gen-4.5~\cite{Gen45}) lead commercially, open-source models---CogVideoX~\cite{cogvideox}, HunyuanVideo~\cite{hunyuanvideo1,hunyuanvideo2025}, WAN-2.1/2.2~\cite{wan2025wanopenadvancedlargescale}, SkyReels series~\cite{SkyReels-A1,SkyReelsV1,skyreelsv2,fei2025skyreelsA2,li2026skyreelsv3techniquereport}, LTX~\cite{LTX-video,ltx2}, MAGI-1~\cite{magi-1}---are rapidly narrowing the gap through data scaling and quality improvements.

\subsection{Video-Audio Generative Models}
Joint text-to-audio+video (T2AV) generation aims to synthesize synchronized audiovisual content from text. Commercial systems (Veo-3.1~\cite{veo31}, Sora-2~\cite{sora2}, Kling-3.0~\cite{Kling30}) show strong capabilities but lack transparency. Open-source approaches evolved from coupled U-Nets~\cite{ruan2023mmdiffusionlearningmultimodaldiffusion} to DiT-based methods: adapter-based (AV-DiT~\cite{wang2024avditefficientaudiovisualdiffusion}), expert-orchestration (MMDisCo~\cite{hayakawa2025mmdiscomultimodaldiscriminatorguidedcooperative}, Universe-1~\cite{wang2025universe1unifiedaudiovideogeneration}), and dual-stream architectures (Ovi~\cite{low2025ovitwinbackbonecrossmodal}, BridgeDiT~\cite{guan2025tamingtexttosoundingvideogeneration}, JavisDiT~\cite{liu2025javisditjointaudiovideodiffusion}) using cross-attention or flow matching---though these incur high computational costs. LTX-2~\cite{ltx2} proposes asymmetric streams for efficiency. Unified single-tower models like Apollo~\cite{wang2026apollounifiedmultitaskaudiovideo} process audio-video tokens jointly via Omni-Full Attention, enabling multitask training (T2AV/TI2AV/TI2V) with tighter coupling. Despite progress, synchronized speech-video synthesis and complete soundscapes remain underexplored, with precise spatio-temporal alignment an open challenge.

\begin{figure}[t]
\centering
\includegraphics[width=\linewidth]{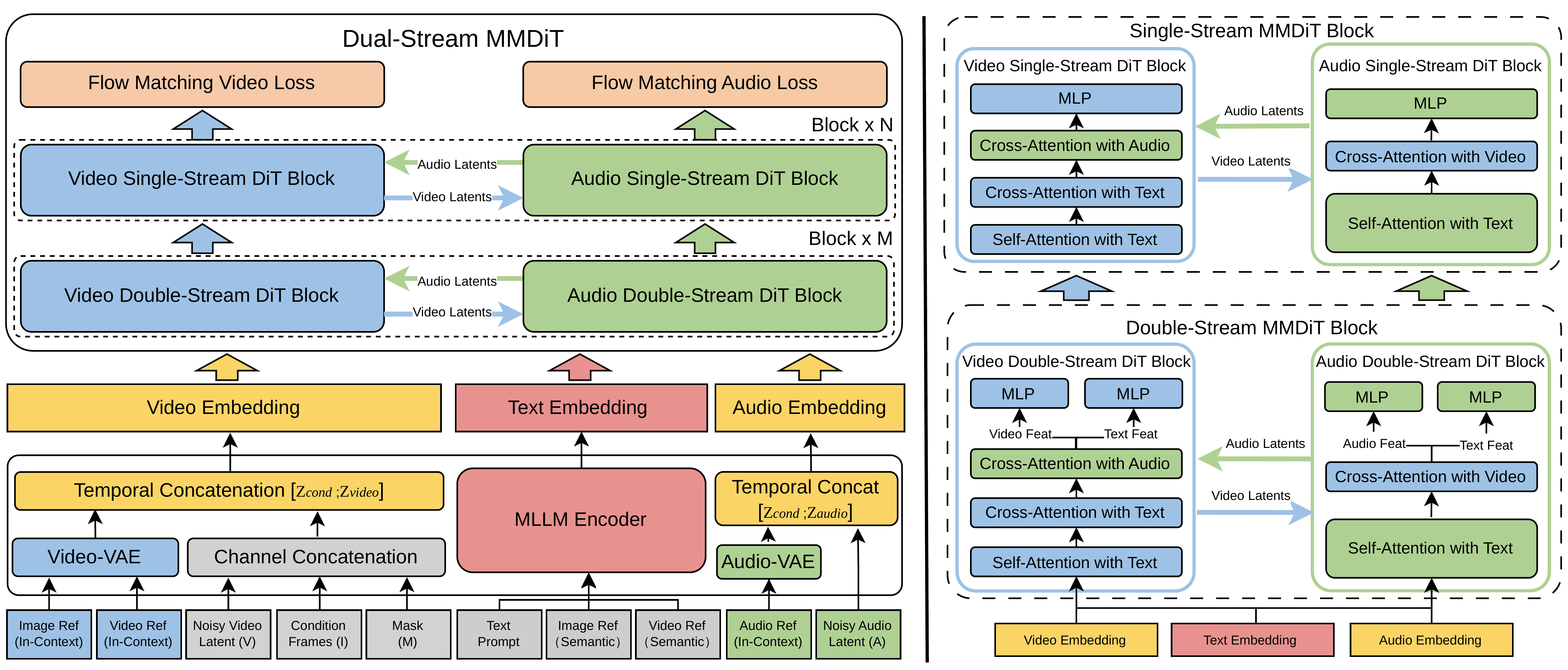}
\caption{Overview of the proposed method.}
\label{fig:overview}  
\end{figure}
  
\section{Model Design}

We present SkyReels-V4, a unified multi-modal video foundation model for joint video-audio generation, inpainting, and editing. The model adopts a dual-stream MMDiT architecture with rich multi-modal instruction following, enabling seamless integration of text, image, video, mask, and audio conditioning signals while maintaining computational efficiency across cinematic resolutions and durations. The overview of the model architecture is shown in Figure~\ref{fig:overview}.

\subsection{Dual-Stream MMDiT Architecture for Joint Video-Audio Generation}

Our architecture employs a symmetric twin backbone design with parallel video and audio branches, both constructed on an identical Multimodal Diffusion Transformer (MMDiT) framework. The video branch is initialized from a pretrained text-to-video model, while the audio branch is trained from scratch with matching architectural specifications. 

\textbf{Hybrid Dual-Stream and Single-Stream MMDiT Blocks.} Following the MMDiT design, each transformer block processes video (or audio) and text modalities through a hybrid architecture that balances modality alignment with parameter efficiency. The initial $M$ layers employ a \textit{Dual-Stream} design where video/audio and text tokens maintain separate parameters for adaptive layer normalization, QKV projections, and MLPs, but interact during joint self-attention:

\begin{align}
\mathbf{Q}_v, \mathbf{K}_v, \mathbf{V}_v &= \text{QKV}_v(\text{LayerNorm}_v(\mathbf{x}_v)), \\
\mathbf{Q}_t, \mathbf{K}_t, \mathbf{V}_t &= \text{QKV}_t(\text{LayerNorm}_t(\mathbf{x}_t)), \\
\mathbf{x}'_v, \mathbf{x}'_t &= \text{Attention}([\mathbf{Q}_v; \mathbf{Q}_t], [\mathbf{K}_v; \mathbf{K}_t], [\mathbf{V}_v; \mathbf{V}_t]),
\end{align}

\noindent where $\mathbf{x}_v$ and $\mathbf{x}_t$ denote video/audio and text token embeddings, respectively, and $[\cdot;\cdot]$ represents concatenation. This design facilitates strong cross-modal alignment during early layers. The subsequent $N$ layers transition to a \textit{Single-Stream} architecture that processes concatenated video (or audio) and text tokens with shared parameters, maximizing computational efficiency. This hybrid strategy achieves faster convergence than either pure approach.

\textbf{Reinforced Text Conditioning via Cross-Attention.} To address potential semantic dilution of text features in the single-stream stages, we augment the video block with an additional text cross-attention layer immediately following self-attention:

\begin{equation}
\mathbf{x}''_v = \mathbf{x}'_v + \text{Attention}(\mathbf{Q}=\mathbf{x}'_v, \mathbf{K}=\mathbf{x}_t, \mathbf{V}=\mathbf{x}_t),
\end{equation}

\noindent where the video stream queries the text embeddings, reinforcing textual guidance throughout the generation process. This cross-attention mechanism is crucial for maintaining fine-grained semantic control in later model stages.

\textbf{Bidirectional Audio-Video Cross-Attention.} To enable temporal synchronization between modalities, each transformer block incorporates paired cross-attention layers: the audio stream attends to video features, and the video stream reciprocally attends to audio features. This bidirectional mechanism exchanges synchronization cues throughout the entire network depth:

\begin{equation}
\begin{aligned}
\mathbf{a}'_i &= \mathbf{a}_i + \text{CrossAttn}(\mathbf{Q}=\mathbf{a}_i, \mathbf{K}=\mathbf{v}_i, \mathbf{V}=\mathbf{v}_i), \\
\mathbf{v}''_i &= \mathbf{v}'_i + \text{CrossAttn}(\mathbf{Q}=\mathbf{v}'_i, \mathbf{K}=\mathbf{a}'_i, \mathbf{V}=\mathbf{a}'_i),
\end{aligned}
\end{equation}

\noindent where $\mathbf{a}_i$ and $\mathbf{v}_i$ are audio and video features at layer $i$. The architectural symmetry ensures both modalities share the same latent dimension, eliminating the need for intermediate projection layers and preserving the attention structure from unimodal pretraining.

\textbf{Temporal Alignment via RoPE Scaling.} Despite architectural symmetry, the temporal resolutions differ: video latents span 21 frames while audio latents contain 218 tokens (44.1 kHz $\times$ 5s). To align these temporal scales, we apply Rotary Positional Embeddings (RoPE) to both modalities and scale the audio RoPE frequencies by $21/218 \approx 0.09633$ to match the video's coarser temporal resolution. This ensures that audio and video tokens attend to each other with temporally consistent correspondence. 

\textbf{Shared Multi-Modal Text Encoder.} We simplify prompt conditioning by employing a single frozen MLLM text encoder applied to a combined prompt that concatenates visual and acoustic descriptions. The resulting multi-modal embeddings are independently consumed by both audio and video branches via self-attention and cross-attention. This unified semantic context improves cross-modal alignment while simplifying training and inference, and crucially enables the model to process rich multi-modal instructions including text, images, video clips, and audio references.

\textbf{Training Objective.} We adopt a flow matching framework for training. Given a video latent $\mathbf{z}_v^0$ and audio latent $\mathbf{z}_a^0$, we sample timestep $t \sim \mathcal{U}(0,1)$ and construct noisy latents $\mathbf{z}_v^t = t\mathbf{z}_v^0 + (1-t)\boldsymbol{\epsilon}_v$ and $\mathbf{z}_a^t = t\mathbf{z}_a^0 + (1-t)\boldsymbol{\epsilon}_a$, where $\boldsymbol{\epsilon}_v, \boldsymbol{\epsilon}_a \sim \mathcal{N}(0, \mathbf{I})$. The model predicts the velocity field $\mathbf{v}_\theta$ that pushes noise toward data:

\begin{equation}
\mathcal{L}_{\text{flow}} = \mathbb{E}_{t,\mathbf{z}_v^0,\mathbf{z}_a^0,\boldsymbol{\epsilon}_v,\boldsymbol{\epsilon}_a} \left[ \left\| \mathbf{v}_\theta^v(t, \mathbf{z}_v^t, \mathbf{z}_a^t, \mathbf{c}) - (\mathbf{z}_v^0 - \boldsymbol{\epsilon}_v) \right\|^2 + \left\| \mathbf{v}_\theta^a(t, \mathbf{z}_a^t, \mathbf{z}_v^t, \mathbf{c}) - (\mathbf{z}_a^0 - \boldsymbol{\epsilon}_a) \right\|^2 \right],
\end{equation}

\noindent where $\mathbf{c}$ denotes the conditioning information (multi-modal embeddings and optional spatial-temporal masks). The joint training objective encourages both branches to learn synchronized generation while respecting their respective modality-specific characteristics. 

\subsection{Unified Video Inpainting via Channel Concatenation}

To enable diverse video generation and editing tasks within a single framework, we employ a flexible input conditioning mechanism applied to the video branch. The input to the video MMDiT is formed by concatenating three tensors along the channel dimension:

\begin{equation}
\mathbf{Z}_{\text{input}} = \text{Concat}(\mathbf{V}, \mathbf{I}, \mathbf{M}),
\end{equation}

\noindent where $\mathbf{V} \in \mathbb{R}^{T \times H \times W \times C}$ is the noisy video latent, $\mathbf{I} \in \mathbb{R}^{T \times H \times W \times C}$ contains VAE-encoded conditional frames (with non-conditional frames filled with black image latents), and $\mathbf{M} \in \mathbb{R}^{T \times H \times W \times 1}$ is a binary mask specifying which spatiotemporal regions are conditions (value 1) versus regions to be generated (value 0).

This formulation unifies multiple generation tasks through different mask configurations:

\begin{itemize}
    \item \textbf{Text-to-Video (T2V):} $\mathbf{M} = \mathbf{0}$ (all frames generated)
    \item \textbf{Image-to-Video (I2V):} $M_{t=0} = 1, M_{t>0} = 0$ (first frame conditioned)
    \item \textbf{Video Extension:} $M_{t<k} = 1, M_{t \geq k} = 0$ (first $k$ frames conditioned)
    \item \textbf{Start-End Frame Interpolation:} $M_{t=0} = M_{t=T-1} = 1$, others 0
    \item \textbf{Video Editing:} $M_{t,h,w} = 1$ for preserved regions, 0 for edited regions (arbitrary spatiotemporal masks)
\end{itemize}

This unified formulation naturally accommodates both fixed foreground/background masks and dynamic per-frame editing masks, enabling precise control over spatial and temporal modifications. 

Crucially, the inpainting mechanism is applied exclusively to the video stream. During inpainting and editing tasks, the audio branch generates audio from scratch conditioned on the (partially conditioned or edited) video content, ensuring acoustic consistency with the generated or modified visual content. This design allows the audio to adapt seamlessly to video modifications while maintaining temporal synchronization through the bidirectional cross-attention mechanism.

\subsection{Multi-Modal In-Context Learning for Vision-Referenced Generation and Editing}

Beyond text and inpainting masks, our framework supports rich multi-modal conditioning through reference images and video clips, enabling complex vision-referenced generation tasks such as multi-identity video generation and identity-preserving video editing under multi-modal prompts.

\textbf{Multi-Modal Instruction Following via MLLM.} Reference visual inputs (images or video frames) are jointly processed with the text prompt through the MLLM text encoder to extract semantically enriched multi-modal embeddings. The MLLM's instruction-following capability enables the model to understand complex compositional requests that combine visual references with textual descriptions (e.g., ``generate a video of person A from the reference @image\_1 speaking <dialogue>hello, how are you?</dialogue> in the style of person B's @video\_1''). These multi-modal embeddings are consumed by both the video and audio branches.

\textbf{In-Context Visual Conditioning via Self-Attention.} To provide explicit visual reference signals beyond semantic guidance, we inject reference visuals directly into the video self-attention mechanism. Each reference image or video frame is encoded via the VAE, padded to a uniform spatial resolution, and concatenated along the temporal dimension. These condition latents $\mathbf{Z}_{\text{cond}}$ are prepended to the noisy video latents $\mathbf{Z}_{\text{video}}$ before self-attention:

\begin{equation}
\mathbf{Z}_{\text{attn}} = [\mathbf{Z}_{\text{cond}}; \mathbf{Z}_{\text{video}}],
\end{equation}

\noindent where the concatenated sequence undergoes joint self-attention. This in-context learning enables the model to directly reference fine-grained visual patterns (e.g., identity characteristics, texture details, pose variations) from the conditions when generating or editing video content.

\textbf{Temporal Positional Disambiguation via Offset 3D RoPE.} To distinguish condition latents from noisy video latents and organize multiple reference visuals, we employ 3D Rotary Positional Embeddings with temporal index offsets. Condition latents receive negative temporal indices, sequentially encoding each reference visual before the generated video frames:

\begin{equation}
\text{RoPE}_{\text{temporal}}(\mathbf{Z}_{\text{cond},i}) = \text{RoPE}(t = -N_{\text{cond}} + i), \quad \text{RoPE}_{\text{temporal}}(\mathbf{Z}_{\text{video},j}) = \text{RoPE}(t = j),
\end{equation}

\noindent where $N_{\text{cond}}$ is the total number of condition tokens and $i, j$ index the condition and video tokens respectively. Spatial indices are preserved across all tokens, ensuring that attention relationships respect both spatial and temporal structure. This offset-based positional encoding provides an effective inductive bias for distinguishing conditioning context from generation targets without introducing task-specific architectural modifications, and naturally extends to multiple reference visuals of varying types (images, short clips, etc.).

\textbf{Audio Reference Conditioning.} Similarly, audio references (e.g., speech samples, musical themes, ambient soundscapes) are encoded and processed as in-context conditions for the audio branch. By combining multi-modal semantic guidance from the MLLM with in-context visual patterns from the video branch and audio patterns from audio references, the model achieves fine-grained control over both visual and acoustic generation.

\subsection{Data Pipeline}

Our data pipeline consists of three main components: data collection, data processing, and captioning. This pipeline handles three modalities—images, videos, and audio—to support multimodal model training.

\subsubsection{Data Collection}

Our training data comprises both real-world and synthetic data across three modalities: images, videos, and audio.

\textbf{Real-world Data.} 
We collect real-world data from two primary sources: publicly available datasets and licensed in-house data. Public datasets include images (LAION~\cite{laion}, Flickr~\cite{flickr}, etc.), videos (WebVid-10M\cite{Bain21}, Koala-36M~\cite{wang2025koala36mlargescalevideodataset}, OpenHumanVid~\cite{li2025openhumanvidlargescalehighqualitydataset}, etc.), and audio (Emilia~\cite{he2024emiliaextensivemultilingualdiverse}, AudioSet~\cite{jort_audioset_2017}, VGGSound~\cite{chen2020vggsoundlargescaleaudiovisualdataset}, SoundNet~\cite{aytar2016soundnet}, etc.). Our in-house licensed data encompasses authorized movies, TV series, short videos, and web series.

\textbf{Synthetic Data.}
We generate synthetic data to address sparse scenarios and generation tasks inadequately covered by real-world data. We focus on three key areas: multilingual text generation, multilingual speech synthesis and multimodal inpainting/editing tasks.

For text generation, we construct synthetic data covering multiple languages, including Chinese, English, Japanese, Korean, German, French, etc. Our synthetic image-text data includes simple text rendering and context-aware text generation that preserves font characteristics. For video-text data, we generate basic text effect videos and context-aware text with attributes matching reference styles (font, size, color) and motion characteristics (trajectories, special effects).

To enhance speech generation and multilingual coverage, we employ multiple TTS models spanning various languages. We curate diverse text corpora to ensure the model learns pronunciations beyond common characters, including rare and uncommon scripts.

For multimodal inpainting and editing tasks, paired training data is inherently unavailable in real-world datasets. We therefore construct these data through a sophisticated pipeline involving visual segmentation models, image/video editing models, and controllable generation techniques. 

\subsubsection{Data Processing}

Our data processing pipeline is tailored to three data types: images, pure audio, and videos (with or without audio).

\textbf{Image Data Processing.}
The image processing pipeline consists of three stages: deduplication, filtering, and balancing. Deduplication employs strict image-level matching. Filtering includes basic quality metrics (resolution, IQA scores, aspect ratio, etc.) and content quality criteria (watermarks, stamps, logos, overly small text, etc.). For data balancing, we adopt stage-specific strategies: during pretraining, we compute image embedding similarities, perform clustering, and balance cluster proportions; during supervised fine-tuning (SFT), we define entity and scene categories, matching them against captions for fine-grained balancing.

\textbf{Audio Data Processing.}
The audio processing pipeline includes: category classification, quality filtering, duration control, content recognition, and audio captioning. First, we classify audio into four categories—sound effects, music, speech, and singing—using Qwen3-Omni~\cite{Qwen3-Omni}. Next, we perform quality filtering based on SNR, MOS score, clipping ratio, and audio bandwidth. We use voice activity detection (VAD) to select audio with silence ratios below 0.2. For duration control, we segment long audio clips into 15-second chunks and concatenate short clips by category to reach 15 seconds. For speech and singing categories, we employ Whisper to transcribe spoken and sung content. Finally, we uniformly caption all audio using Qwen3-Omni.

\textbf{Video Data Processing.}
Video processing consists of four stages: preprocessing (segmentation and deduplication), filtering, balancing, and audio-video synchronization for videos with audio tracks.

\textit{Preprocessing.} Traditional methods using PyDetect and TransNet-V2~\cite{soucek2020transnetv2} produce scene-cut clips that often lack narrative coherence. Instead, we adopt intelligent segmentation that combines TransNet's shot boundary predictions via VLM to extract semantically complete segments, including both long takes and multi-shot clips. In later training stages, we further apply video highlight detection to identify segments with richer content. We deduplicate segmented clips using VideoCLIP embeddings~\cite{wang2024videoclipxladvancinglongdescription}.

\textit{Filtering.} We filter videos based on three quality dimensions: basic quality (duration, resolution, aesthetic score, blur, contrast, exposure), content quality (watermarks, logos, text overlays, synthetic artifacts, content type/source issues), and motion quality (camera stability, motion magnitude/speed, frame drops).

\textit{Balancing.} To improve training efficiency, we balance data along two dimensions: conceptual diversity and motion diversity. We define a taxonomy of concept labels covering different subjects and scene types, matching them against video content for concept balancing. We further balance motion types by defining key motion patterns for each subject or scene category.

\textit{Audio-Video Synchronization.} Following the audio pipeline, we obtain preliminary audio captions. For videos containing speech or singing, we determine whether the video is person-free (background audio), single-person, or multi-person based on the first frame, and apply audio-visual synchronization filtering accordingly. We adopt the widely-used SyncNet~\cite{raina2025syncnetcorrelatingobjectivetime} model, which uses a ConvNet architecture to learn joint embeddings between sound and mouth images, to filter speech videos lacking sufficient audio-video synchronization. We adapt the model to handle millions of video samples and produce scalar confidence and offset values. We retain only clips satisfying $|\text{offset}| \leq 3 \land \text{confidence} > 1.5$ with a minimum mean volume of -60 decibels. Finally, we integrate audio and video captions into unified descriptions.

\subsubsection{Captioning}

We generate three types of captions: short captions, long captions, and structured captions. Short captions provide concise descriptions of video content and audio information. Long captions offer comprehensive descriptions of environment, subjects, lighting, atmosphere, and other nuanced details. Structured captions follow a standardized descriptive order with special tokens to denote in-video text(<text></text>), sound effects(<sfx></sfx>), speech content(<dialogue></dialogue>), singing content(<singing></singing>), and background music(<bgm></bgm>). In final training stages, we exclusively use structured captions. To align user prompts with this format, we employ a prompt enhancer that reformats free-form input into the structured representation.

\section{Training Strategy}

We adopt a progressive multi-stage training paradigm that systematically develops the model's capabilities across multiple modalities and tasks. Our training pipeline consists of three major phases: Video Pretrain, Audio Pretrain, and Video-Audio Joint Training, followed by supervised fine-tuning. This structured approach enables the model to learn spatial concepts, temporal dynamics, audio generation, and multi-modal alignment in a stable and efficient manner. Table~\ref{tab:training_schedule} summarizes our complete multi-stage training schedule, including tasks, resolutions, data volumes, and training epochs for each stage.

\subsection{Video Pretrain}

The video pretraining phase follows a progressive strategy that gradually increases spatial resolution, temporal length, and task complexity. We begin with text-to-image (T2I) training to establish strong semantic understanding and visual concept learning, which we find significantly accelerates subsequent video training convergence.

\textbf{Stage 1: Text-to-Image Foundation.} We first train the T2I task at 256px resolution using 3 billion images for 3 epochs. This stage enables the model to learn the correspondence between text and visual content, establishing a solid foundation for spatial composition and concept formation.

\textbf{Stage 2: Initial Video Learning.} We introduce text-to-video (T2V) generation while maintaining T2I training. At 256px resolution and 16 fps, we train on 1 billion images and 400 million videos for 3 epochs, with video durations ranging from 2 to 10 seconds. Training at lower resolution allows the model to more rapidly converge on motion dynamics and temporal coherence.

\textbf{Stage 3: Inpainting Capabilities.} We expand the task set to include image inpainting, image-to-video (I2V), video-to-video (V2V), and video editing tasks, each comprising 5\% of the training mix. This stage trains for 2 epochs with video durations extended to 2--15 seconds, enabling the model to learn spatial and temporal inpainting capabilities.

\textbf{Stage 4: Mixed Resolution Scaling.} We employ mixed-resolution training at 256px and 480px, maintaining 16 fps and 2--15 second durations. Training on 100 million images and 100 million videos, we keep the inpainting task ratio unchanged, allowing the model to gradually adapt to higher resolution generation.

\textbf{Stage 5: High Resolution Training.} We further scale to mixed resolutions of 480px, 720px, and 1080px at 16 fps, with video durations of 3--15 seconds. This stage uses 50 million images and 50 million videos, substantially improving the model's high-resolution generation quality.

\textbf{Stage 6: Multi-modal Condition Pretrain.} We introduce image reference and video reference conditioning for both generation and inpainting tasks, comprising 20\% of the training data each, with the remaining 60\% dedicated to T2V. This stage trains on 20 million images and 50 million videos, equipping the model with flexible multi-modal conditioning capabilities.

\subsection{Audio Pretrain}

The audio backbone is pretrained from scratch on hundreds of thousands of hours of primarily speech data, with durations up to 15 seconds. During pretraining, we use variable-length audio to maximize coverage of diverse acoustics, providing the audio backbone with broad exposure to natural variability in duration and content. The long-form raw audio enables the model to generate consistent audio that respects speaker traits such as pitch and emotion.

\subsection{Video-Audio Joint Training}

Following the completion of video and audio pretraining, we enter the joint training phase, simultaneously training three tasks: text-to-video (T2V), text-to-audio-video (T2AV), and text-to-audio (T2A). In this phase, we allocate half of the video pretrain data to T2AV joint training while incorporating T2A data to enable synchronized audio-visual generation.

\subsection{Video-Audio Supervised Fine-tuning}

In the final SFT stage, we focus exclusively on joint generation data, training on 5 million videos with multi-modal condition support (image, video, and audio), which comprises 20\% of the data. We conclude with a final fine-tuning step on 1 million manually curated high-quality videos, further refining generation quality, motion coherence, and audio-visual alignment.

\begin{table}[t]
\centering
\caption{Complete training schedule across all stages. The progressive strategy gradually increases resolution, temporal length, and task complexity.}
\label{tab:training_schedule}
\resizebox{\textwidth}{!}{
\begin{tabular}{l|l|l|l|l}
\hline
\textbf{Task} & \textbf{Stage} & \textbf{Resolution} & \textbf{Data Volume} & \textbf{Epochs} \\
\hline
\multicolumn{5}{c}{\textit{Video Pretrain}} \\
\hline
T2I & Stage 1 & 256px & 3B images & 3 \\
\hline
T2I + T2V & Stage 2 & 256px, 16fps, 2-10s & 1B images / 400M videos & 3 \\
\hline
T2I + T2V + Inpaint & Stage 3 & 256px, 16fps, 2-15s & 1B images / 400M videos & 2 \\
(Image Inpaint, I2V, V2V, Edit) & & (Inpaint: 5\% each) & & \\
\hline
Mixed Tasks & Stage 4 & 256/480px, 16fps, 2-15s & 100M images / 100M videos & 2 \\
(T2I, T2V, Inpaint) & & (Inpaint ratio unchanged) & & \\
\hline
Mixed Tasks & Stage 5 & 480/720/1080px, & 50M images / 50M videos & 2 \\
(T2I, T2V, Inpaint) & & 16fps, 3-15s & & \\
\hline
Multi-modal Condition & Stage 6 & 480/720/1080px, & 20M images / 50M videos & 2 \\
(Image/Video Ref: 20\% each) & & 16fps, 3-15s & & \\
(T2V: 60\%) & & & & \\
\hline
\multicolumn{5}{c}{\textit{Audio Pretrain}} \\
\hline
Audio Backbone & Pretrain & Variable length, up to 15s & Hundreds of thousands of hours & 3 \\
\hline
\multicolumn{5}{c}{\textit{Video-Audio Joint Training}} \\
\hline
T2V + T2AV + T2A & Joint Pretrain & 720/1080px, 16fps, 5-15s & 50\% video data + T2A data & 2 \\
\hline
\multicolumn{5}{c}{\textit{Video-Audio Supervised Fine-tuning}} \\
\hline
T2AV + Multi-modal & SFT Stage 1 & 720/1080px, 16fps, 5-15s & 5M videos (Multi-modal: 20\%) & 3 \\
\hline
T2AV + Multi-modal & SFT Stage 2 & 720/1080px, 16fps, 5-15s & 1M curated videos & 3 \\
\hline
\end{tabular}
}
\end{table}
\subsection{Video Super-Resolution and Frame Interpolation (Refiner)}

\begin{figure}[t]
\centering
\includegraphics[width=\linewidth]{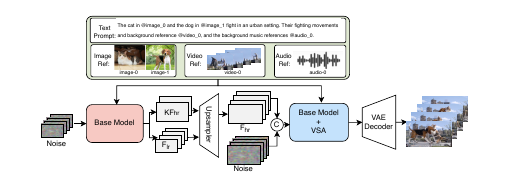}
\caption{The pipeline of the video super-resolution and frame interpolation method. F denotes the output latent of our base model. KF demotes the key frames latent of our base model.}
\label{fig:vsr-architecture}
\end{figure}

To further enhance visual quality and temporal smoothness of generated videos, we introduce a dedicated Refiner module that jointly performs video super-resolution (VSR) and frame interpolation, as shown in Fig.~\ref{fig:vsr-architecture}. This cascaded architecture operates on the outputs of the base multi-modal video generation model, leveraging both low-resolution results and high-resolution key-frame  results to synthesize high-fidelity, fine-grained visual details while simultaneously increasing temporal resolution.

\textbf{Architecture and Design.} We initialize the Refiner weights from the pre-trained video generation model to ensure effective knowledge transfer and training stability. The Refiner accepts three categories of inputs: (1) multi-modal visual conditions (image, video, and audio references at high resolution), (2) multi-modal text instructions consistent with the base model, and (3) the base model outputs, which include low-resolution predictions for all frames and high-resolution predictions for keyframes. To support the efficient inference strategy described earlier, we incorporate a joint prediction task during post-training, where the base model learns to simultaneously predict all frames at low resolution and keyframes at high resolution. 
With these inputs, we first linearly interpolate the low-resolution latents to the target high resolution. Then, for keyframe positions, we replace the interpolated results with the directly predicted high-resolution keyframe latents from the base model. Finally, these combined latents are concatenated with high-resolution noisy latents along the channel dimension as input to the DiT model.

To support multi-task generation, inpainting, and editing capabilities in the Refiner, we design a unified framework. For inpainting tasks, we incorporate the high-resolution version of the source video, using it to replace the interpolated regions where inpainting is not required. A spatial mask guides the model to distinguish between regions requiring refinement and those that should remain unchanged. This design enables the Refiner to handle both unconditional super-resolution and conditional inpainting across multiple modalities.

\textbf{Computational Efficiency.} To address the computational overhead imposed by long temporal contexts and high-resolution inputs, we adopt Video Sparse Attention (VSA)~\cite{zhang2025vsafastervideodiffusion}, a trainable sparse attention mechanism designed for video diffusion transformers. VSA employs a hierarchical two-stage approach: a coarse stage that aggregates spatial-temporal cubes to identify critical token regions through lightweight pooled attention, and a fine stage that applies dense attention only within the selected top-K cubes. This design eliminates the need to compute full quadratic attention while maintaining hardware efficiency through block-sparse layouts compatible with modern GPU kernels. By exploiting spatio-temporal redundancy in a learnable manner, VSA enables us to reduce attention computational cost by approximately $3\times$ while preserving generation quality, making it practical to process high-resolution video sequences during both training and inference.

\textbf{Training Data and Configuration.} For dataset construction, we curate 1 million high-quality video clips spanning diverse scenarios and resolutions from 1K to 4K. We incorporate high-resolution images alongside video data to enhance the model's capability for generating fine visual details. The task composition mirrors the base model's multi-task structure, maintaining consistent ratios for generation, inpainting, and editing tasks. All weights of the Refiner are fully trainable throughout the training process, following the flow matching paradigm.

\section{Model Performance}
We evaluate model performance on a public arena leaderboard to assess overall user preference in an open-ended setting. Beyond this, we conduct comprehensive human assessments spanning five key dimensions: Instruction Following, Audio-Visual Synchronization, Visual Quality, Motion Quality, and Audio Quality, providing a fine-grained analysis of model capabilities. Furthermore, our multimodal inpainting and editing framework unlocks a wide range of practical applications, including but not limited to subject insertion, object removal, background replacement, and reference-guided video synthesis. We showcase representative examples of these applications in Appendix~\ref{appendix:applications}.

\subsection{Artificial Analysis Arena}
Artificial Analysis~\cite{ArtificialAnalysis} is a widely recognized benchmarking platform for evaluating generative models across image and video generation domains. The platform operates an open arena where models are scored by the public, with Elo scores calculated from pairwise comparisons to reflect user preferences.
We evaluate our model on the Artificial Analysis Video Arena across four generation tracks: text-to-video with audio, text-to-video without audio, image-to-video with audio, and image-to-video without audio. These tracks are designed to comprehensively assess the quality of both joint video–audio synthesis and video-only generation under different conditioning modalities. Our model is benchmarked against notable external baselines including Veo 3.1, Kling 3.0, grok-imagine-video, Sora-2, Vidu-Q3, Wan 2.6, etc.

\begin{figure}[htbp]
    \centering
    \begin{subfigure}[b]{0.48\textwidth}
        \centering
        \includegraphics[width=\textwidth]{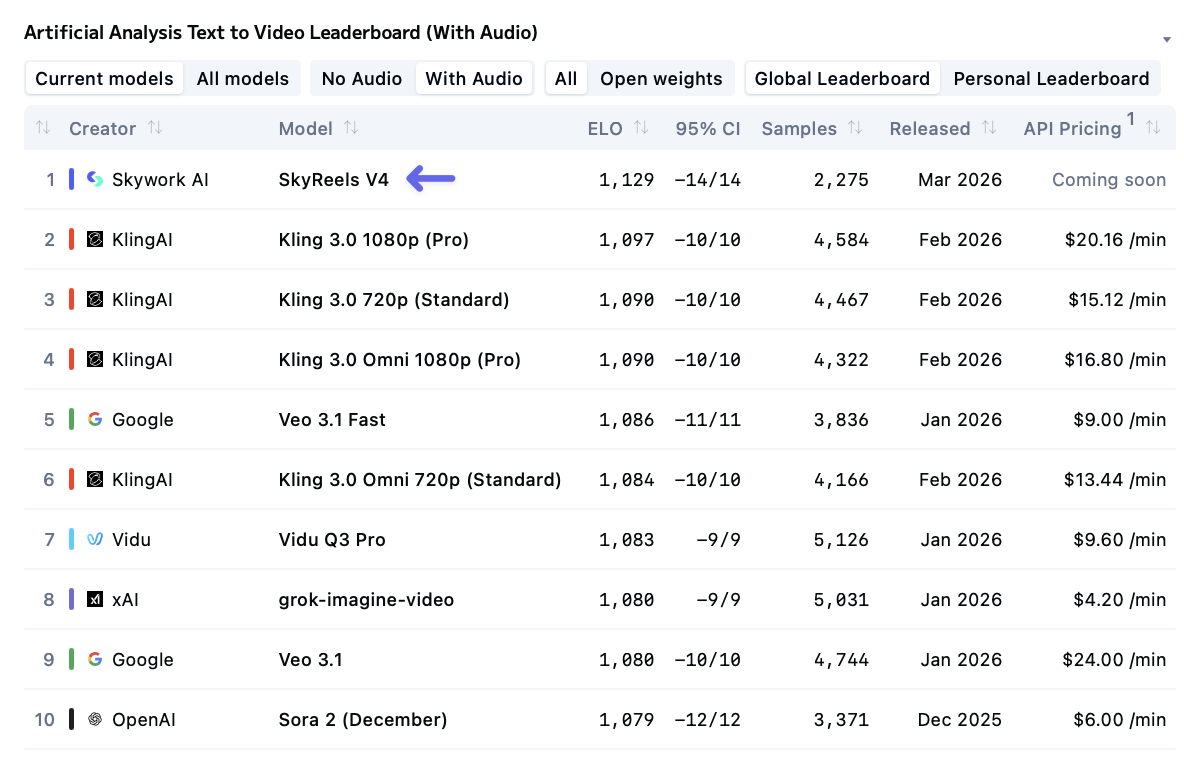}
        \caption{Text-to-Video with Audio (Rank \textbf{\#1})}
        \label{fig:aa_t2va}
    \end{subfigure}
    \hfill
    \begin{subfigure}[b]{0.48\textwidth}
        \centering
        \includegraphics[width=\textwidth]{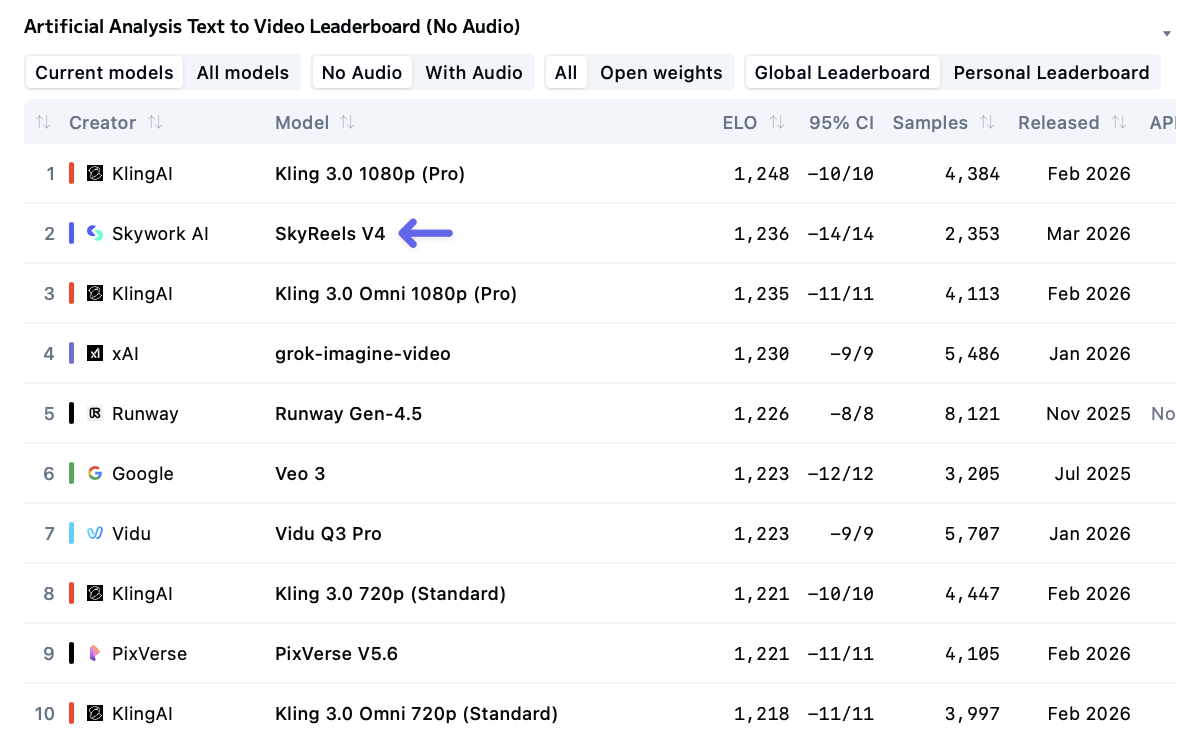}
        \caption{Text-to-Video without Audio (Rank \textbf{\#2})}
        \label{fig:aa_t2v}
    \end{subfigure}

    \vspace{1em}

    \begin{subfigure}[b]{0.48\textwidth}
        \centering
        \includegraphics[width=\textwidth]{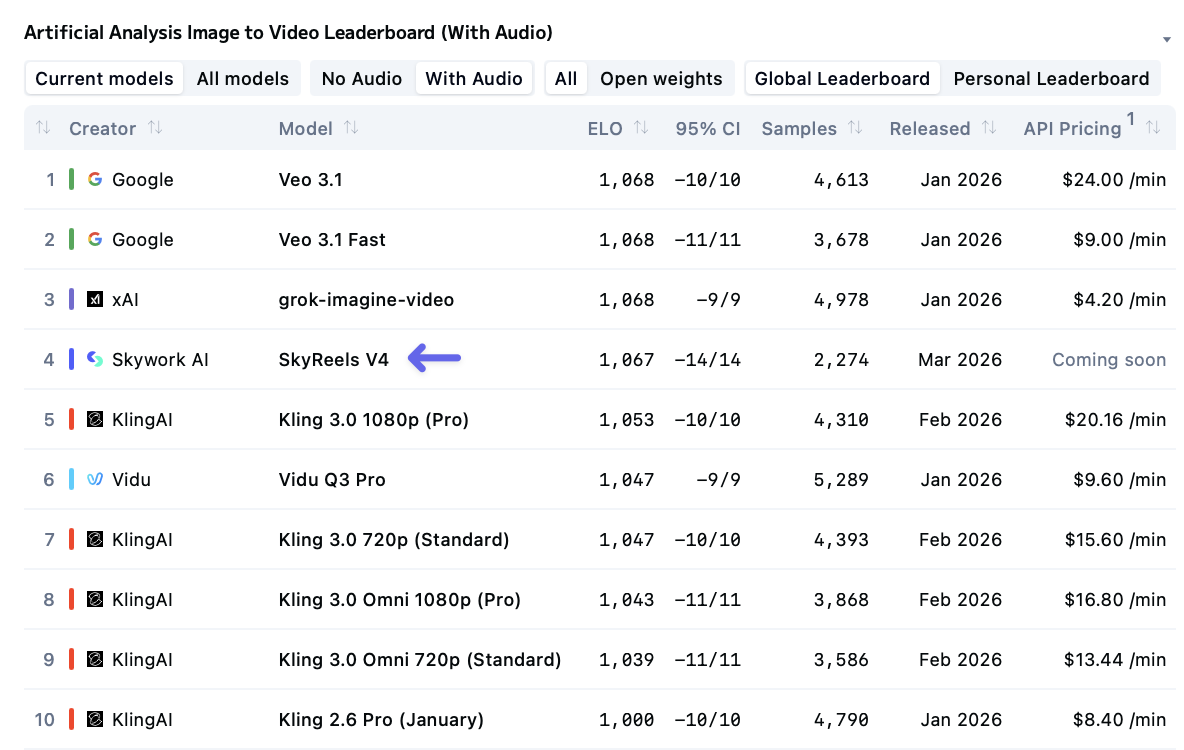}
        \caption{Image-to-Video with Audio (Rank \textbf{\#4})}
        \label{fig:aa_i2va}
    \end{subfigure}
    \hfill
    \begin{subfigure}[b]{0.48\textwidth}
        \centering
        \includegraphics[width=\textwidth]{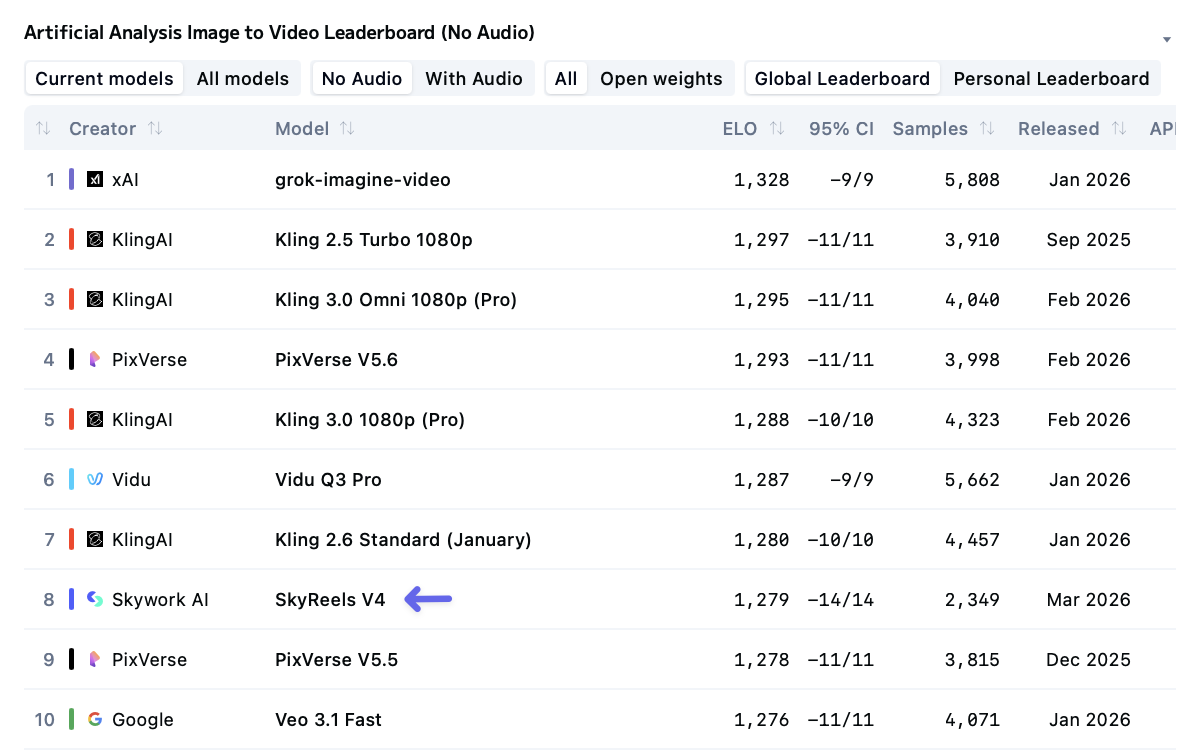}
        \caption{Image-to-Video without Audio (Rank \textbf{\#7})}
        \label{fig:aa_i2v}
    \end{subfigure}

    \caption{Artificial Analysis Video Arena Leaderboards across four generation tracks (as of 2026-03-18). SkyReels V4 ranks \textbf{\#1} in Text-to-Video with Audio, \textbf{\#2} in Text-to-Video without Audio, \textbf{\#4} in Image-to-Video with Audio, and \textbf{\#7} in Image-to-Video without Audio, demonstrating strong and competitive audiovisual generation quality as evaluated by public user preferences.}
    \label{fig:aa_leaderboard}
\end{figure}

\textbf{Results:} As shown in Figure~\ref{fig:aa_leaderboard}, our model achieves strong and competitive performance across all four tracks (as of 2026-03-18), as evaluated by public user preferences. Specifically, SkyReels V4 ranks \textbf{1st} in text-to-video with audio generation, surpassing Kling 3.0 and Veo 3.1, and ranks \textbf{2nd} in text-to-video without audio generation. For image-conditioned generation, our model ranks \textbf{4th} in image-to-video with audio and \textbf{7th} in image-to-video without audio, demonstrating broad multimodal video generation capability.

\subsection{Human Assessments}

To comprehensively assess the joint video-audio generation capabilities, we introduce \textbf{SkyReels-VABench}, a novel human evaluation benchmark designed to evaluate state-of-the-art text-to-video+audio models in the market.

\subsubsection{Benchmark Design}

SkyReels-VABench extends our previous SkyReels-Bench~\cite{skyreelsv2} by incorporating comprehensive audio dimensions and multi-shot video scenarios. The benchmark comprises 2000+ carefully curated prompts spanning diverse content categories including advertising, social media content, narrative storytelling, educational content, and entertainment. The prompts are designed to test models across varying complexity levels, from single-shot scenarios to complex multi-shot sequences with sophisticated audio requirements.

\textbf{Language Coverage:} The benchmark includes prompts in multiple languages, with particular emphasis on Chinese and English to assess cross-lingual generation capabilities.

\textbf{Content Diversity:} Prompts cover a wide range of subjects (humans, animals, objects, abstract concepts), environments (indoor, outdoor, natural, urban), and temporal dynamics (static, slow-motion, fast-action sequences).

\textbf{Audio Complexity:} The benchmark tests various audio modalities including speech (monologue, dialogue, narration), singing (various genres and vocal styles), sound effects (environmental, mechanical, natural), and background music (various genres and emotional tones).

\subsubsection{Evaluation Metrics}

Our evaluation framework encompasses five primary dimensions:
\begin{table*}[htbp]
\centering
\caption{Comprehensive Evaluation Dimensions for Audio-Visual Generation}
\label{tab:evaluation_dimensions}
\resizebox{\textwidth}{!}{
\begin{tabular}{@{}llp{10cm}@{}}
\toprule
\textbf{Dimension} & \textbf{Sub-dimension} & \textbf{Evaluation Criteria} \\ 
\midrule
\multirow{10}{*}{\parbox{3cm}{\textbf{Instruction Following}}} 
& \multicolumn{2}{l}{\emph{Video Instruction Following}} \\
& Subject description & Accurate representation of subjects, attributes, and appearances \\
& Subject interaction & Correct execution of actions, interactions, and motion dynamics \\
& Camera movement & Proper execution of camera operations (pan, tilt, zoom, dolly) \\
& Style and aesthetics & Adherence to visual styles, color palettes, and artistic directions \\
& Multi-shot consistency & Correct shot transitions, cross-shot coherence, and reference accuracy \\
\cmidrule(lr){2-3}
& \multicolumn{2}{l}{\emph{Audio Instruction Following}} \\
& Semantic adherence & Fidelity to audio content and characteristics \\
& Temporal accuracy & Correct timing and duration of audio events \\
& Speaker attributes & Speaker-visual matching, vocal characteristics, emotional tone, content accuracy \\
\midrule
\multirow{4}{*}{\parbox{3cm}{\textbf{Audio-Visual Synchronization}}} 
& Lip-sync accuracy & Precise speech-mouth synchronization and correct speaker identification \\
& Sound effect alignment & Temporal correspondence between visual events and sound effects \\
& Atmospheric matching & Coherence between BGM, scene atmosphere, and emotional tone \\
& Spatial audio & Sound spatialization matching visual source locations \\
\midrule
\multirow{4}{*}{\textbf{Visual Quality}} 
& Visual clarity & Sharpness, definition, and resolution \\
& Color accuracy & Natural color balance and saturation without distortion \\
& Compositional quality & Aesthetic composition, framing, and visual balance \\
& Structural integrity & Absence of visual artifacts and corruptions \\
\midrule
\multirow{5}{*}{\textbf{Motion Quality}} 
& Physical plausibility & Adherence to physical laws (gravity, inertia, momentum) \\
& Motion fluidity & Smooth transitions without abrupt discontinuities \\
& Motion stability & Absence of jittering, deformation, and flickering \\
& Temporal consistency & Consistency of dynamic elements across frames \\
& Motion vividness & Action, camera, atmospheric, and emotional expressiveness \\
\midrule
\multirow{5}{*}{\textbf{Audio Quality}} 
& Absence of artifacts & No clipping, truncation, distortion, or glitches \\
& Spatial soundstage & Appropriate stereo imaging and spatial rendering \\
& Timbre realism & Natural and realistic tonal qualities \\
& Signal clarity & Clean audio with appropriate signal-to-noise ratio \\
& Dynamic range & Appropriate audio level variation without compression artifacts \\
\bottomrule
\end{tabular}
}
\end{table*}

\begin{figure}[htbp]
    \centering
    \includegraphics[width=\linewidth]{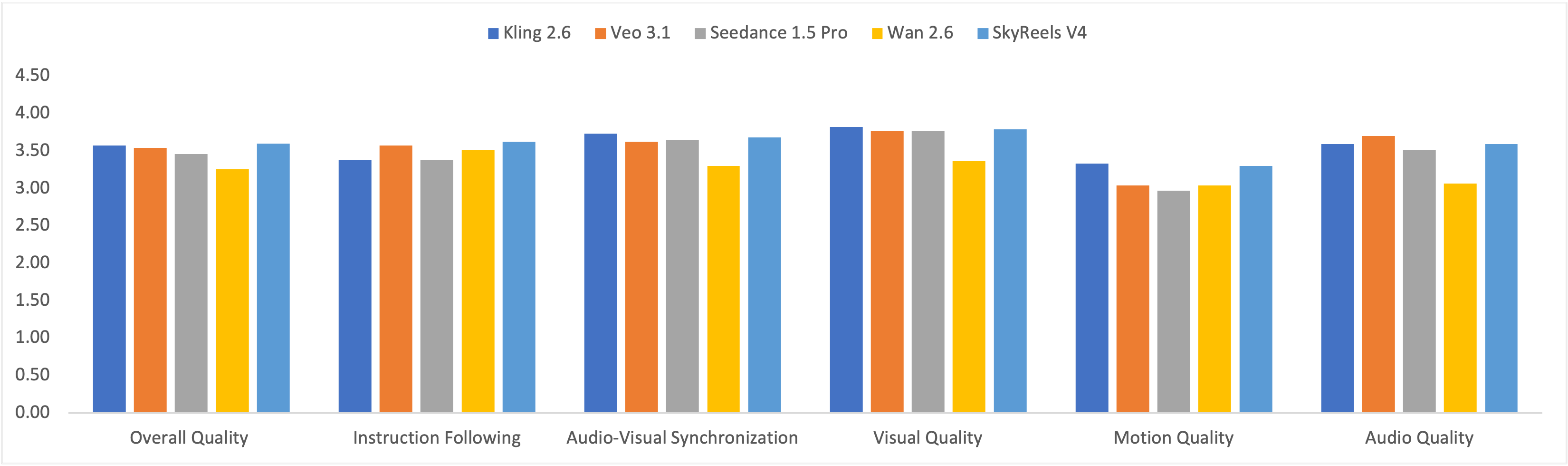}
    \caption{Absolute scoring results (5-point Likert scale) comparing SkyReels V4 against baselines. Higher scores indicate better performance.}
    \label{fig:absolute_all}
\end{figure}

\begin{figure}[htbp]
    \centering
    
    \includegraphics[width=0.8\linewidth]{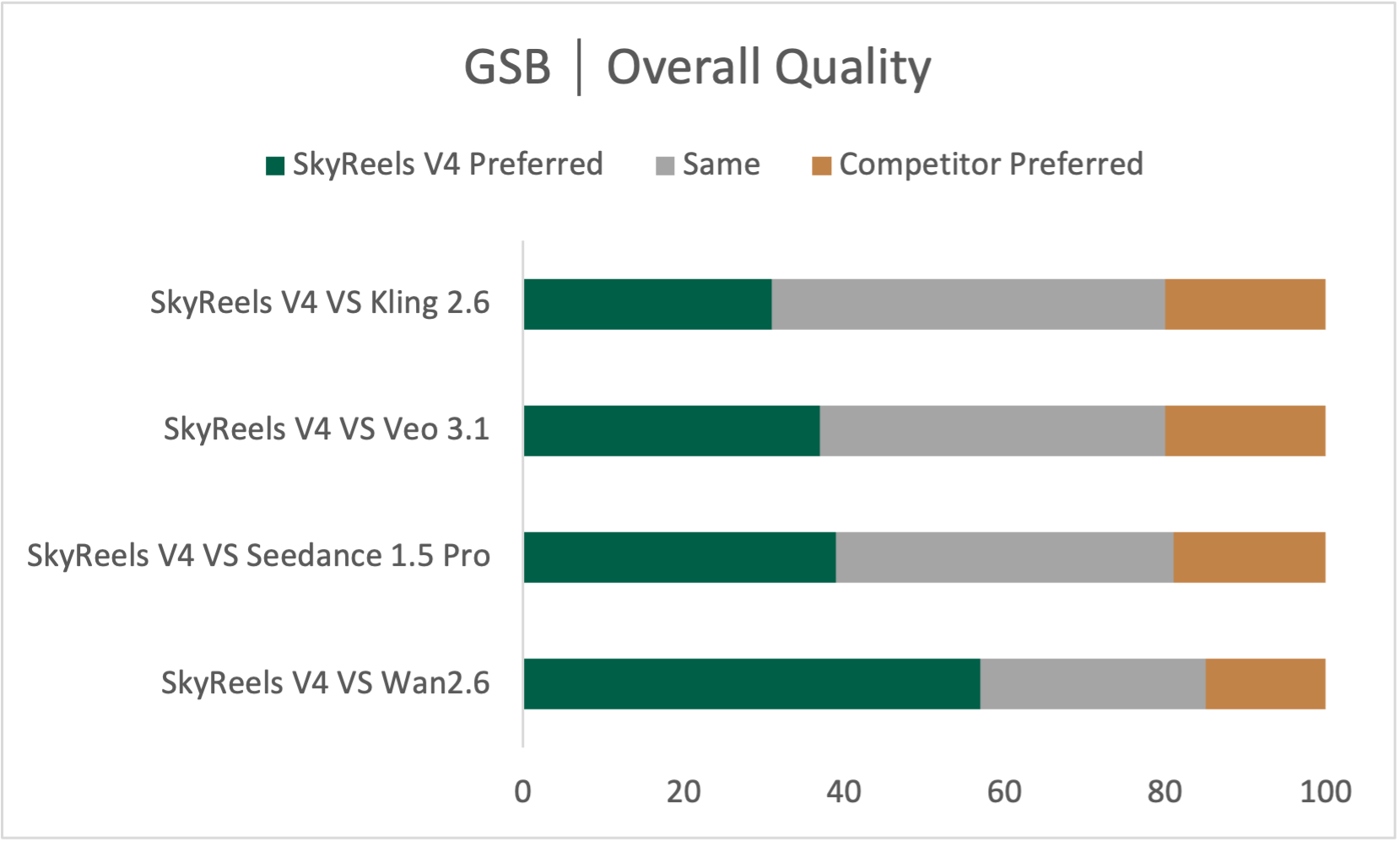}
    \caption*{GSB overall quality comparison: SkyReels V4 vs.\ all baselines. Each bar shows the proportion of ``Good'', ``Same'', and ``Bad'' ratings.}
    
    \vspace{0.8em}
    
    \begin{subfigure}[b]{0.49\linewidth}
        \centering
        \includegraphics[width=\linewidth]{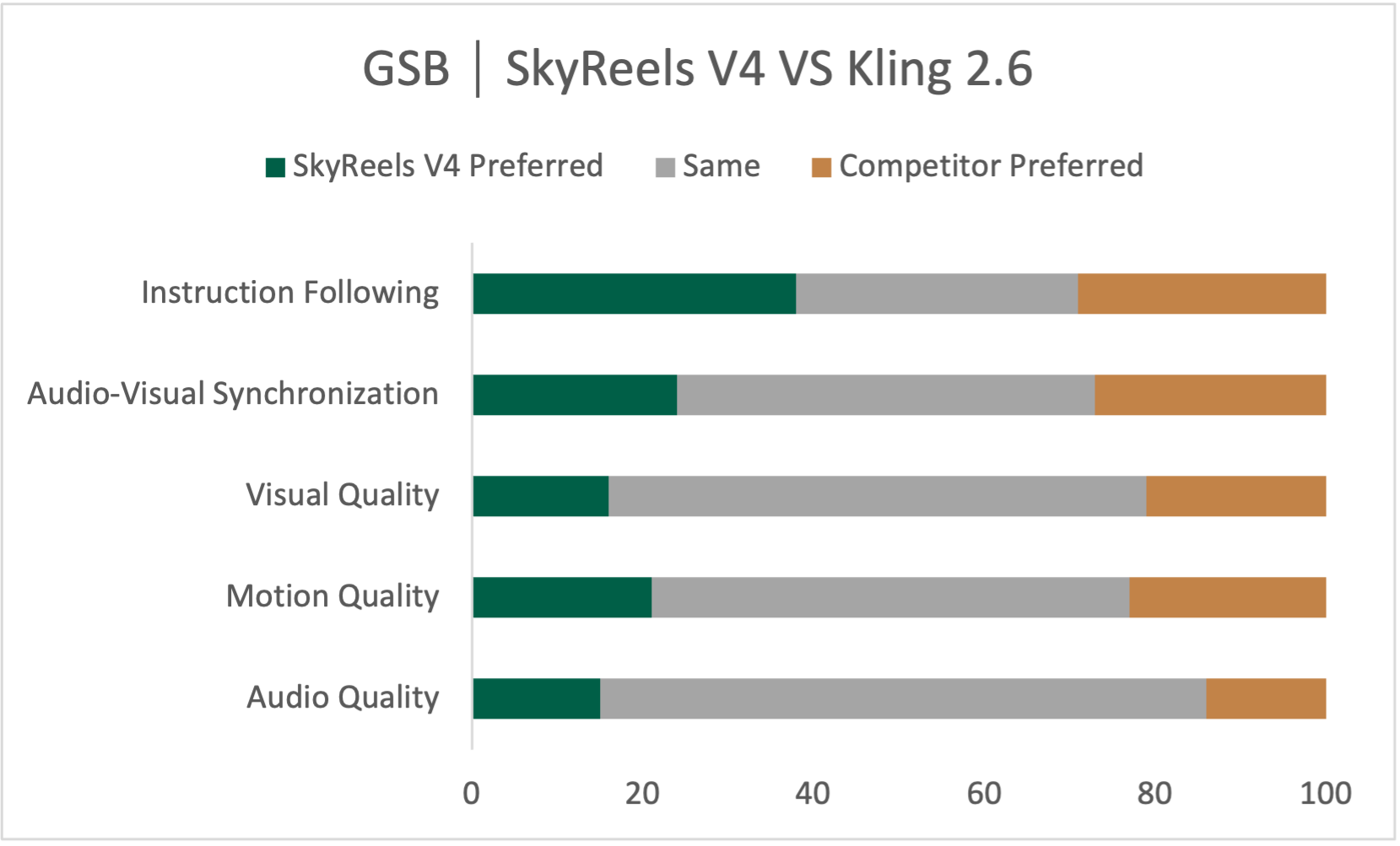}
        \caption{SkyReels V4 vs.\ Kling 2.6}
        \label{fig:gsb_kling}
    \end{subfigure}
    \hfill
    \begin{subfigure}[b]{0.49\linewidth}
        \centering
        \includegraphics[width=\linewidth]{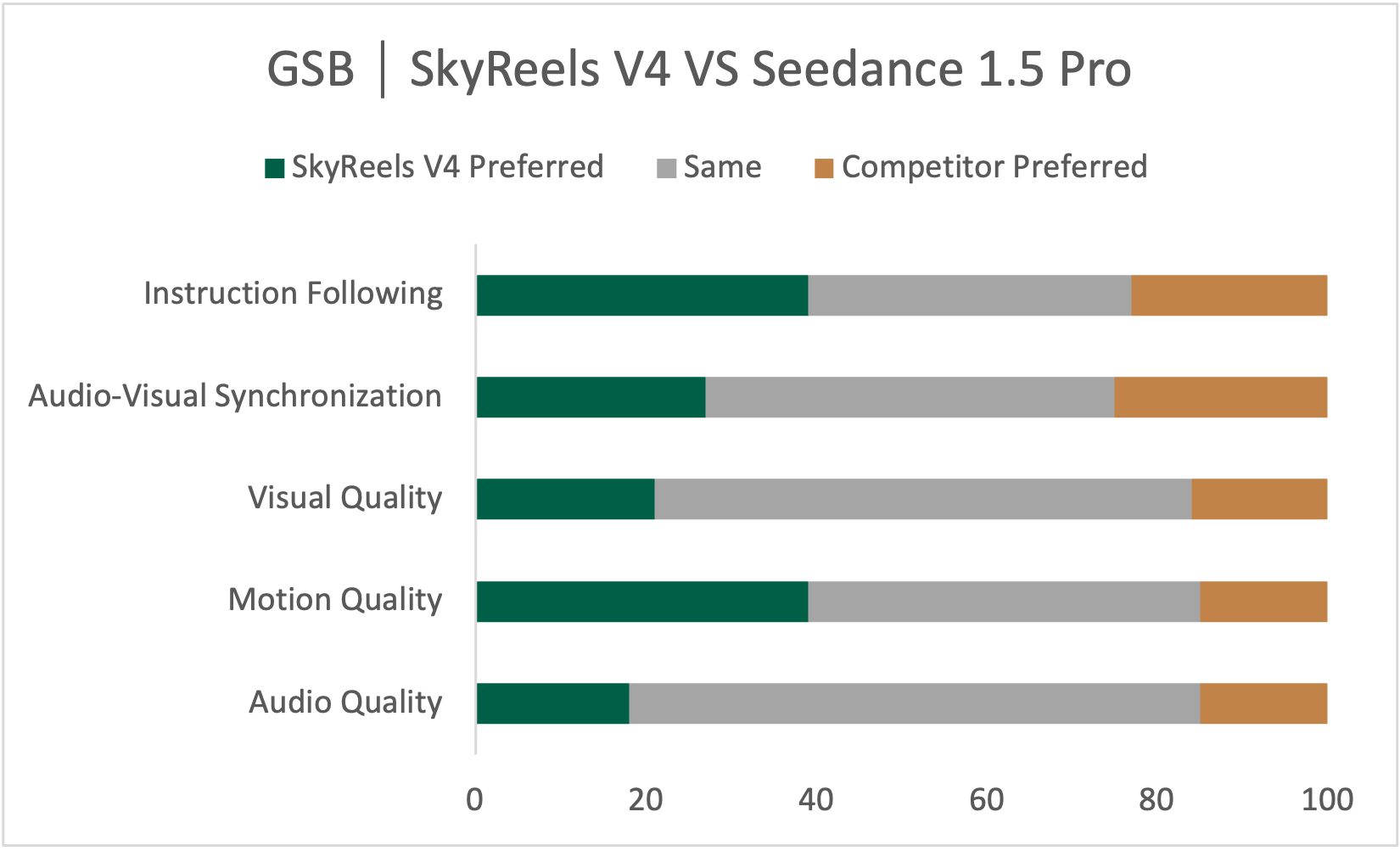}
        \caption{SkyReels V4 vs.\ Seedance 1.5 Pro}
        \label{fig:gsb_seedance}
    \end{subfigure}
    
    \vspace{0.5em}
    
    \begin{subfigure}[b]{0.49\linewidth}
        \centering
        \includegraphics[width=\linewidth]{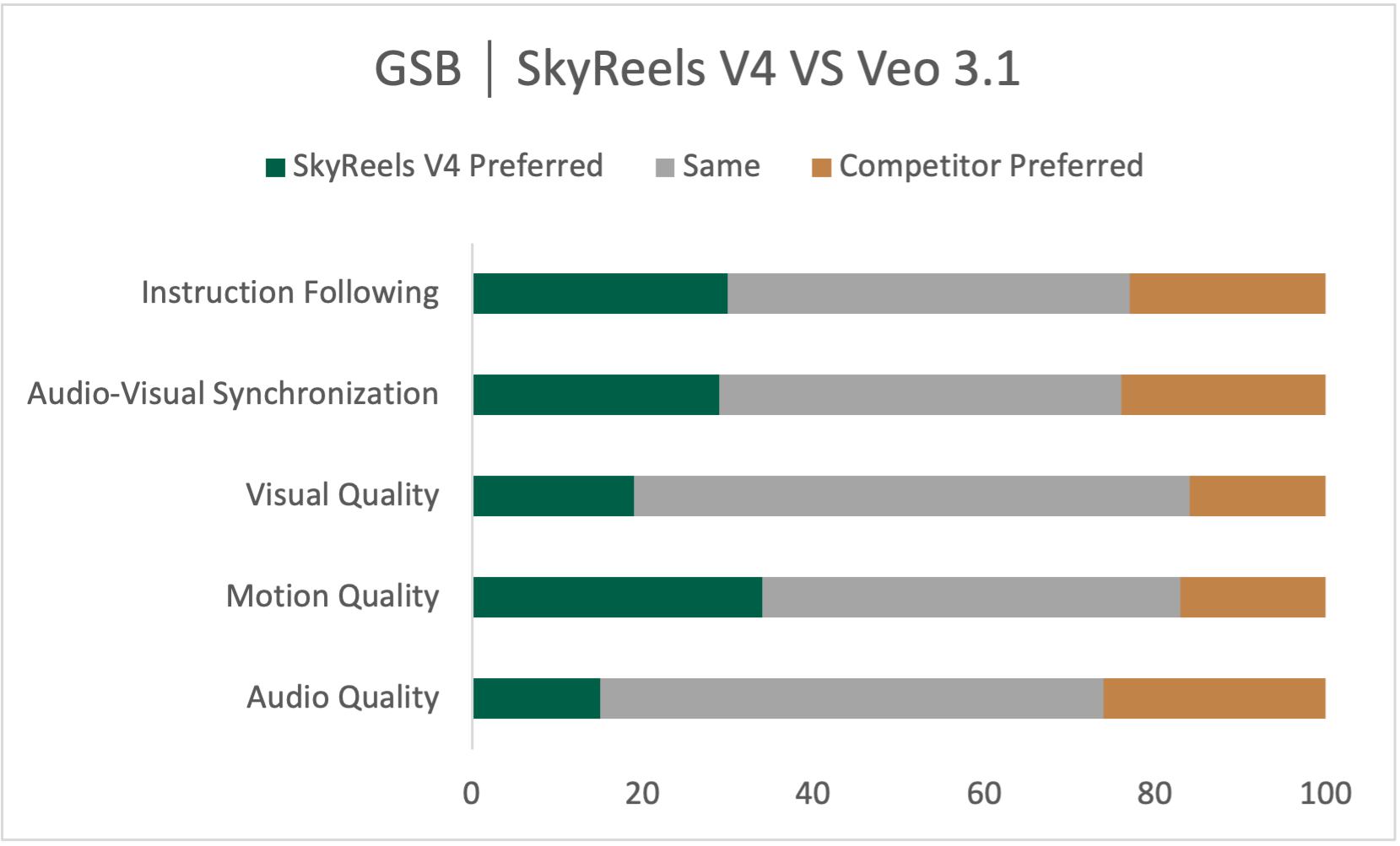}
        \caption{SkyReels V4 vs.\ Veo 3.1}
        \label{fig:gsb_veo}
    \end{subfigure}
    \hfill
    \begin{subfigure}[b]{0.49\linewidth}
        \centering
        \includegraphics[width=\linewidth]{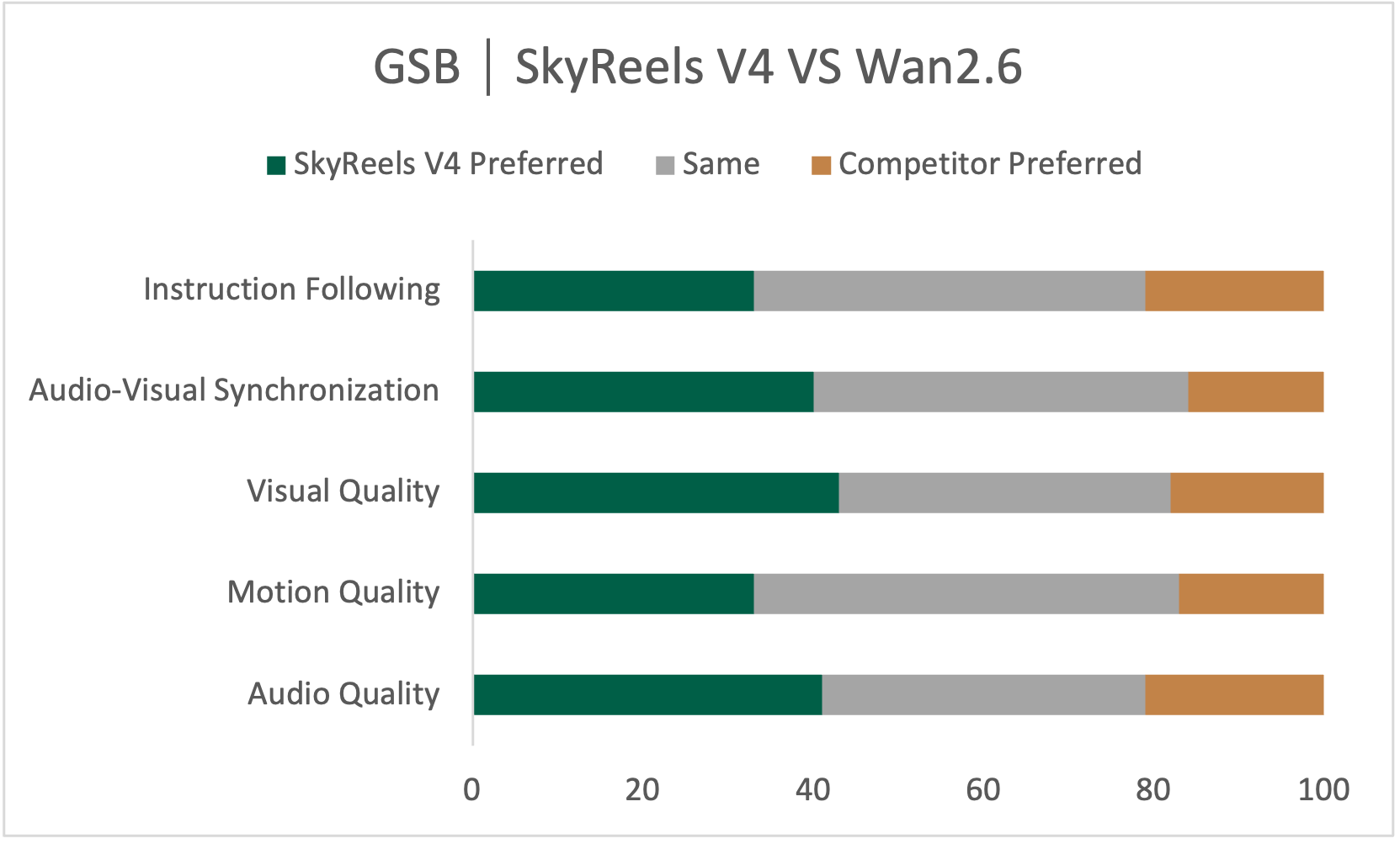}
        \caption{SkyReels V4 vs.\ Wan 2.6}
        \label{fig:gsb_wan}
    \end{subfigure}
    
    \caption{GSB comparison results. \textbf{Top:} Overall quality comparison between SkyReels V4 and all baselines. \textbf{Bottom:} Per-dimension GSB comparison across five evaluation dimensions: Prompt Following, Audio-Visual Synchronization, Visual Quality, Motion Quality, and Audio Quality.}
    \label{fig:gsb_all}
\end{figure}

\subsubsection{Evaluation Methodology}

We employ a dual-metric evaluation protocol conducted by a panel of 50 professional evaluators with backgrounds in video production, audio engineering, and content creation:

\textbf{Absolute Scoring:} Evaluators rate each dimension using a 5-point Likert scale (1 = Extremely Dissatisfied, 2 = Dissatisfied, 3 = Neutral, 4 = Satisfied, 5 = Extremely Satisfied), enabling standardized performance comparison across models.

\textbf{Good-Same-Bad (GSB) Comparison:} Pairwise comparisons between model outputs enable more granular quality differentiation. For each prompt, evaluators compare outputs from different models and assign one of three labels: "Good" (clearly better), "Same" (comparable quality), or "Bad" (clearly worse).

\subsubsection{Baselines}

We compare our model against state-of-the-art video-audio generation systems, including 
\begin{itemize}
    \item Veo 3.1 (Google)
    \item Kling 2.6 (Kuaishou)
    \item Seedance 1.5 Pro (ByteDance)
    \item Wan 2.6 (Alibaba)
\end{itemize}

\subsubsection{Results}

\paragraph{Absolute Scoring.}
We first evaluate all models using the absolute scoring protocol, where human evaluators rate each dimension on a 5-point Likert scale. As shown in Figure~\ref{fig:absolute_all}, SkyReels V4 achieves the highest overall average score among all competing models. The per-dimension breakdown reveals a nuanced picture of SkyReels V4's strengths: it demonstrates particularly strong performance in Prompt Following and Motion Quality. For Visual Quality, SkyReels V4 performs comparably to the strongest competing models. While SkyReels V4 shows relatively modest advantages in Audio-Visual Synchronization and Audio Quality, it nonetheless maintains state-of-the-art performance in these dimensions as well, underscoring its overall competitiveness across the full evaluation spectrum.

\paragraph{Good-Same-Bad (GSB) Comparison.}
To further validate our model's superiority, we conduct pairwise GSB comparisons between SkyReels V4 and each baseline. As illustrated in Figure~\ref{fig:gsb_all}, SkyReels V4 consistently achieves a higher proportion of ``Good'' ratings against all competing models in terms of overall quality. The per-dimension GSB results for each pairwise comparison are presented, demonstrating that SkyReels V4 outperforms Kling 2.6, Seedance 1.5 Pro, Veo 3.1, and Wan 2.6 across the majority of evaluation dimensions.

\section{Conclusion}
In this work, we present \textbf{SkyReels-V4}, a unified multi-modal video foundation model that jointly generates video and audio while supporting generation, inpainting, and editing within a single architecture. Built upon a dual-stream MMDiT design with a shared MLLM-based text encoder, SkyReels-V4 accepts rich multi-modal conditioning inputs---including text, images, video clips, masks, and audio references---and produces high-fidelity, synchronized video--audio outputs at cinematic quality (up to 1080p, 32 FPS, 15 seconds). To support diverse video creation tasks, we employ channel-concatenation to unify generation, inpainting, and editing by reformulating them as inpainting problems under specific mask configurations, while leveraging temporal-concatenation to flexibly incorporate multi-modal references such as images, video clips, and audio. Additionally, our joint low-resolution/high-resolution keyframe generation strategy enables efficient generation at scale.

Extensive evaluations validate SkyReels-V4's effectiveness. On the Artificial Analysis Arena, our model ranks among the top systems in the text-to-video-with-audio track. On our proposed SkyReels-VABench, SkyReels-V4 achieves the highest overall average score, with particularly strong performance in Prompt Following and Motion Quality, while maintaining state-of-the-art performance across all evaluation dimensions. Pairwise comparisons further confirm that SkyReels-V4 consistently outperforms competing baseline systems.

To the best of our knowledge, SkyReels-V4 is the first model to simultaneously unify multi-modal inputs, joint video--audio generation, and generation/inpainting/editing capabilities at cinematic quality and scale. We hope this work serves as a foundation for future research in multi-modal video generation systems.

\section{Contributors}
We gratefully acknowledge all contributors for their dedicated efforts. The following lists recognize participants by their primary contribution roles:
\begin{itemize}
    \item \textbf{Project Sponsor:} Yahui Zhou
    \item \textbf{Project Leader:} Guibin Chen (guibin.chen@kunlun-inc.com)
    \item \textbf{Contributors:}
    \begin{itemize}
        \item \textit{Infrastructure:} Hao Zhang, Zhiheng Xu, Weiming Xiong, Yuzhe Jin, Zhuangzhuang Liu, Wenyan Liu
        \item \textit{Data \& Video Understanding:} Mingyuan Fan, Yiming Wang, Mingshan Chang, Jiahua Wang, Yuqiang Xie, Peng Zhao, Xuanyue Zhong, Fuxiang Zhang, Peiyu Wang
        \item \textit{Video Model Training:} Dixuan Lin, Jiangping Yang, Sheng Chen, Chaofeng Ao, Yunjie Yu, Jujie He, Yuhao Feng, Shiwen Tu, Chaojie Wang, Rui Yan, Wei Shen, Jingchen Wu, Weikai Xu
        \item \textit{Audio Model Training:} Zhengcong Fei, Zheng Chen, Tuanhui Li, Baoxuan Gu, Kaifei Wang, Xuchen Song, Max W. Y. Lam, Chien-Hung Liu
        \item \textit{Multi-modal Training:} Youqiang Zhang, Debang Li, Nuo Pang, Yikun Dou, Xiaopeng Sun, Jingtao Xu, Binjie Mao, Liang Zeng, Haoxiang Guo
        \item \textit{Model Evaluation:} Binglu Zhang, Yu Shen, Tianhui Xiong, Bin Peng
    \end{itemize}
\end{itemize}




\printbibliography

\newpage

\appendix
\section{Application Examples}
\label{appendix:applications}

\begin{table}[h!]
\centering
\caption{Summary of video generation, inpainting, and editing tasks}
\label{tab:task_overview}
\begin{tabular}{|l|l|p{6cm}|}
\hline
\textbf{Main Task} & \textbf{Subtask} & \textbf{Description} \\
\hline
\multirow{2}{*}{\textbf{Generation}} & Image + Audio Ref & Generate videos from multiple reference images and audio inputs \\
& Image + Motion Ref & Generate videos from image and video/motion reference (poses, trajectories) \\
\hline
\multirow{2}{*}{\textbf{Inpainting}} & Region Inpainting & Inpaint subjects, attributes, or backgrounds in video regions \\
& Reference-Guided & Inpaint using reference image guidance for style consistency \\
\hline
\multirow{10}{*}{\textbf{Editing}} & Element Removal & Remove watermarks, subtitles, and logos intelligently \\
& Subject Manipulation & Add, delete, or modify subjects in videos \\
& Attribute Editing & Edit local attributes (color, texture, shape, etc.) \\
& Background Editing & Modify backgrounds while preserving foreground \\
& Style Transfer & Transform videos into different artistic styles \\
& Camera Control & Modify shot angle, shot type, and camera position \\
& Scene Attributes & Edit weather, lighting, tone, and time of day \\
& Subject + Motion Ref & Combine subject and motion from different references \\
& Subject + Expression Ref & Transfer facial expressions from reference video \\
& Background + Video Ref & Combine background and video references \\
& First-Frame + Effect Ref & Apply effects from reference to first-frame \\
\hline
\end{tabular}
\end{table}

This appendix demonstrates typical application cases of our model in video-audio generation, inpainting, and editing. Our model supports flexible multimodal reference inputs, capable of processing various modalities including images, audio, and motion information. The following sections are organized into three main categories: Generation, Inpainting, and Editing.

\newpage
\subsection{Reference-based Generation}
\label{appendix:generation}

\subsubsection{Multiple Image and Audio Reference Generation}
\label{appendix:multi-reference-generation}

\begin{figure}[h]
\renewcommand{\arraystretch}{1.5}
\sloppy
\textbf{Instruction:} \textit{In an elegantly decorated indoor environment with warm, intimate lighting, two people sit facing each other across a dark wooden table. The camera first focuses on @Actor-0, looking weary, says softly, <dialogue>I'm a little tired, I'm going back to my room to rest.</dialogue>@Audio-0. @Actor-1 sits across from @Actor-0, hands clasped on the table, and says with determination, <dialogue>I will pay a visit to your parents tomorrow.</dialogue>@Audio-1. Then @Actor-0 in another room, by the window, holds her phone to her ear and speaks, <dialogue>Mom, Li Zeting said he's coming to our house tomorrow.</dialogue>@Audio-0. The scene shifts to another location—a warm-toned home interior, with a red fabric sofa visible in the background. @Actor-2, sitting tensely with phone to her ear, worries, <dialogue>But with our family's situation, do you think he might look down on us?</dialogue>@Audio-2. <bgm>Soft, sorrowful music plays during the first two shots, transitioning to slow, melancholic tense music for the last two shots.</bgm>}

\vspace{1.0em}

\begin{tabular}{@{}m{2em} p{\textwidth}@{}}


\raggedleft\rotatebox{90}{\bfseries Ref. Image} &
\begin{minipage}{\linewidth}
\raggedright
\begin{tabular}{@{}lll@{}}
\includegraphics[width=0.22\linewidth]{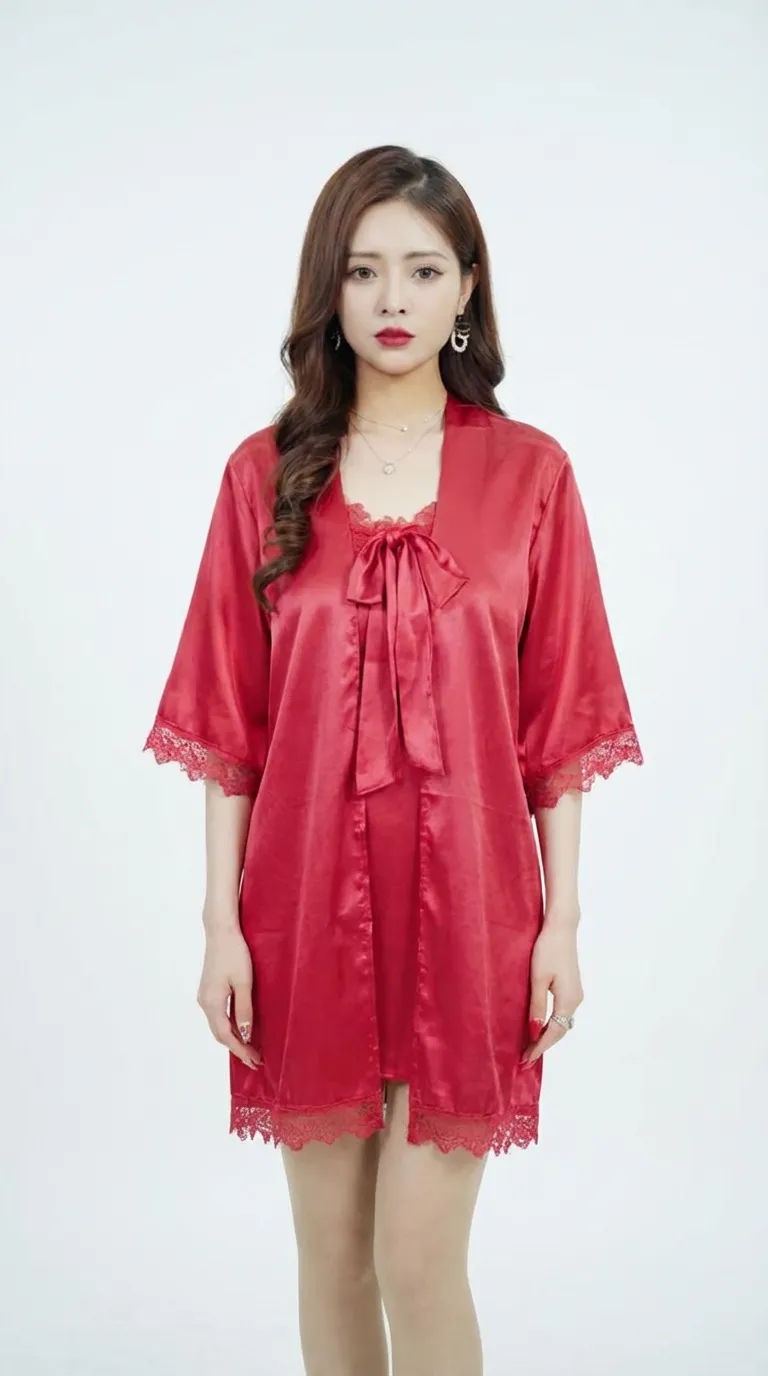} &
\includegraphics[width=0.22\linewidth]{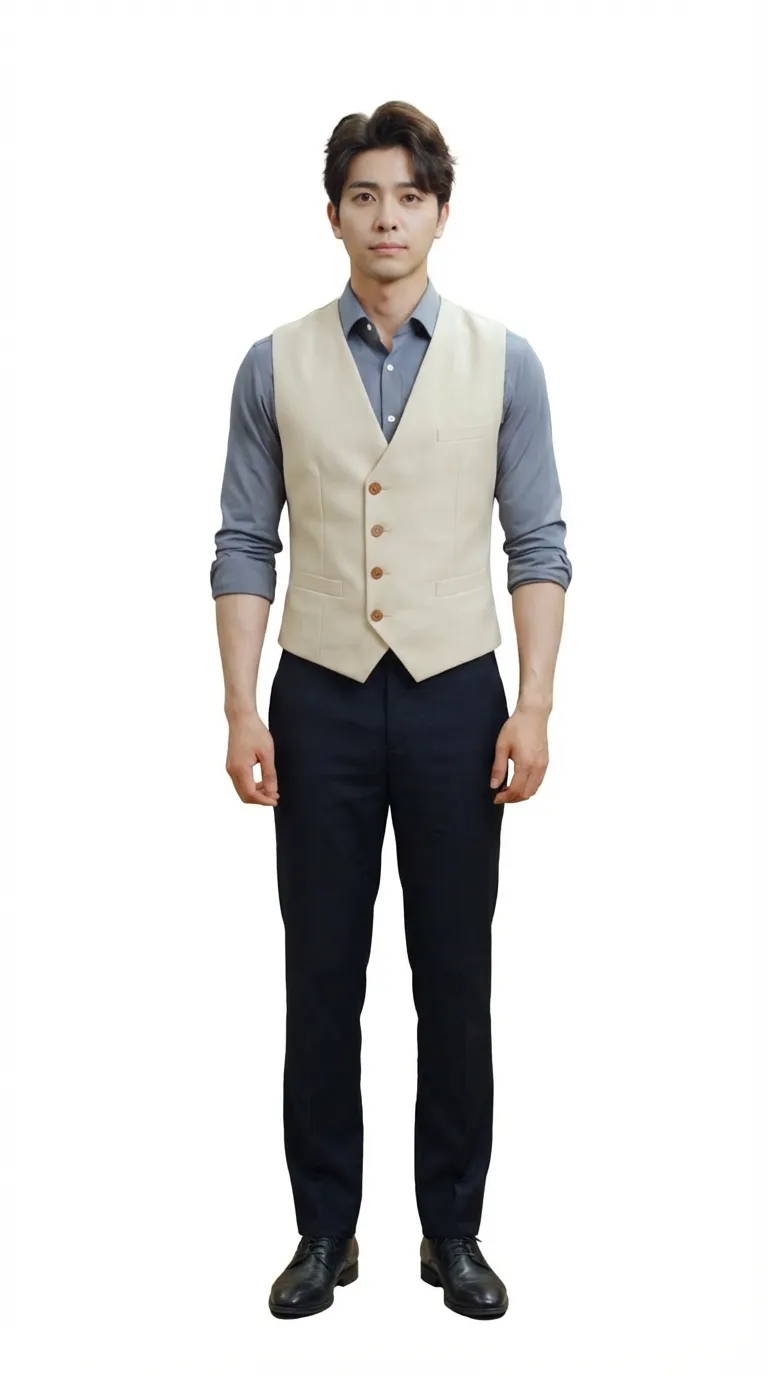} &
\includegraphics[width=0.22\linewidth]{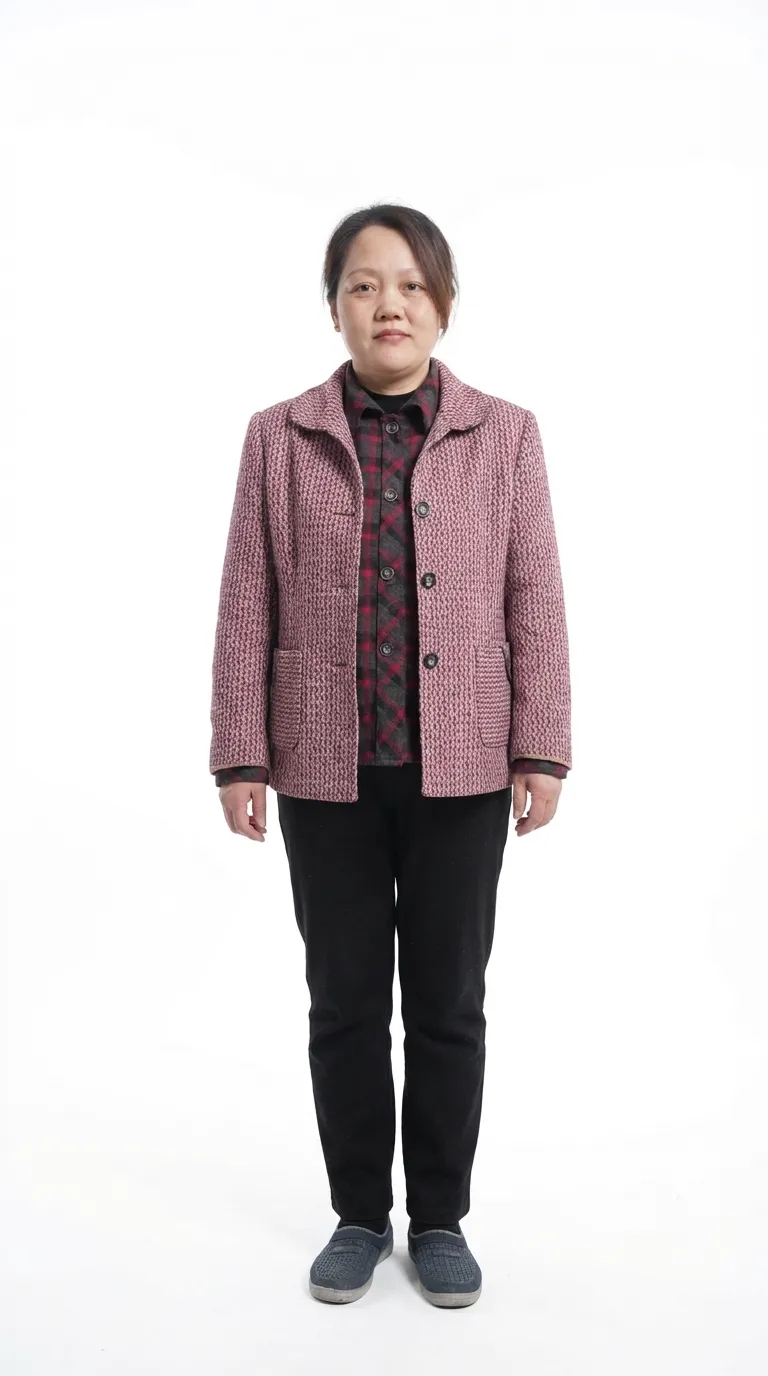} \\
@Actor-0 & @Actor-1 & @Actor-2
\end{tabular}
\end{minipage} \\

\raggedleft\rotatebox{90}{\bfseries Ref. Audio} &
\begin{minipage}{\linewidth}
\raggedright
\begin{tabular}{@{}lll@{}}
\includegraphics[width=0.22\linewidth]{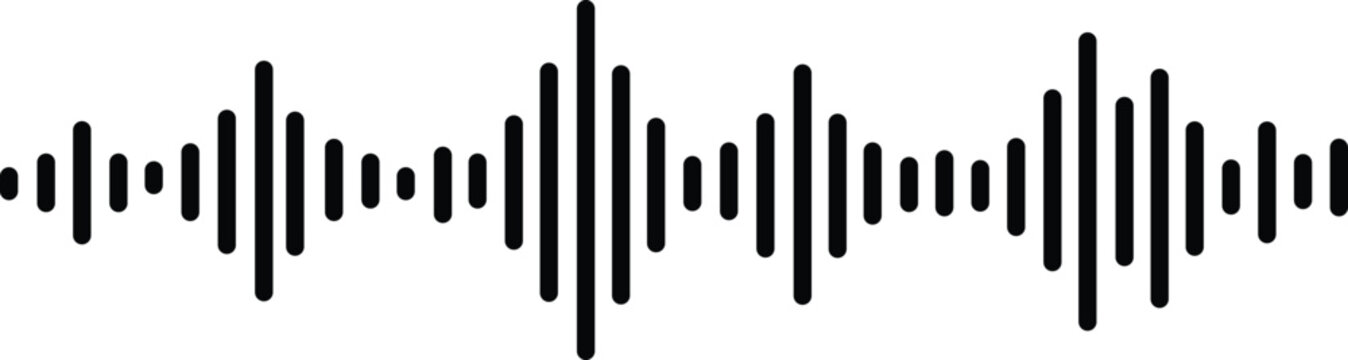} &
\includegraphics[width=0.22\linewidth]{figures/mo2v/audio.jpg} &
\includegraphics[width=0.22\linewidth]{figures/mo2v/audio.jpg} \\
@Audio-0 & @Audio-1 & @Audio-2
\end{tabular}
\end{minipage} \\


\raggedleft\rotatebox{90}{\bfseries Output Video} &
\begin{minipage}{\linewidth}
\raggedright
\includegraphics[width=0.87\linewidth]{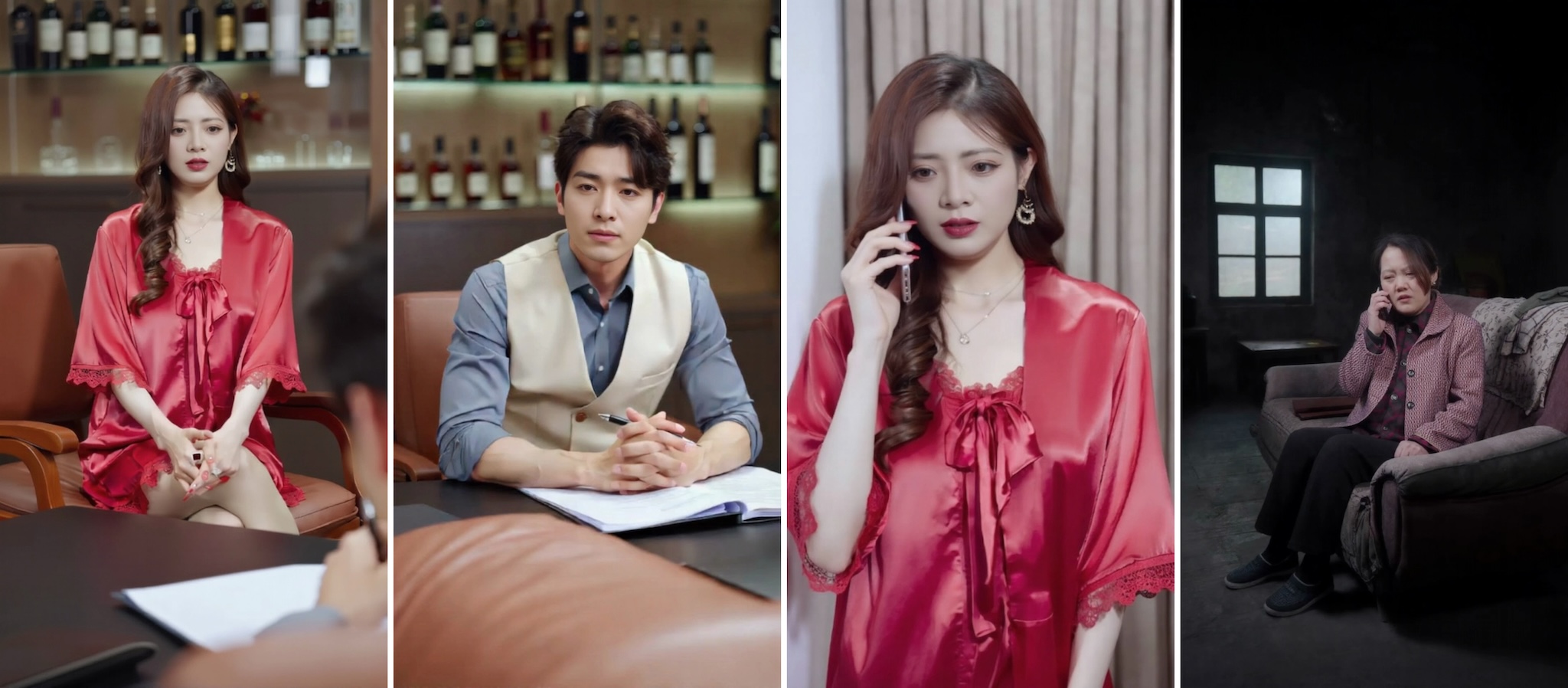}
\end{minipage} \\

\end{tabular}
\caption{Example of multiple images and audios reference.}
\label{fig:mo2v_res1}
\end{figure}

Our model can simultaneously accept multiple reference images and audio inputs to generate videos that are stylistically consistent with the references and audio-matched. The result is shown in Fig.~\ref{fig:mo2v_res1}.

\newpage
\subsubsection{Image Reference and Motion Reference Generation}
\label{appendix:motion-reference-generation}

The model supports using image references to determine content and style, while simultaneously using motion references (e.g., pose sequences, trajectories) to control the dynamic characteristics of the generated video.


\begin{flushleft}
    \textbf{Instruction:} \textit{Animate the person in @image\_1 using the movements from @video\_1.}%
    \vspace{0.5em}
    \begin{tabular}{m{1.2em} m{0.22\linewidth} m{0.22\linewidth} m{0.22\linewidth}}

        \centering\rotatebox{90}{\textbf{Ref. Image}} &
        \includegraphics[width=\linewidth]{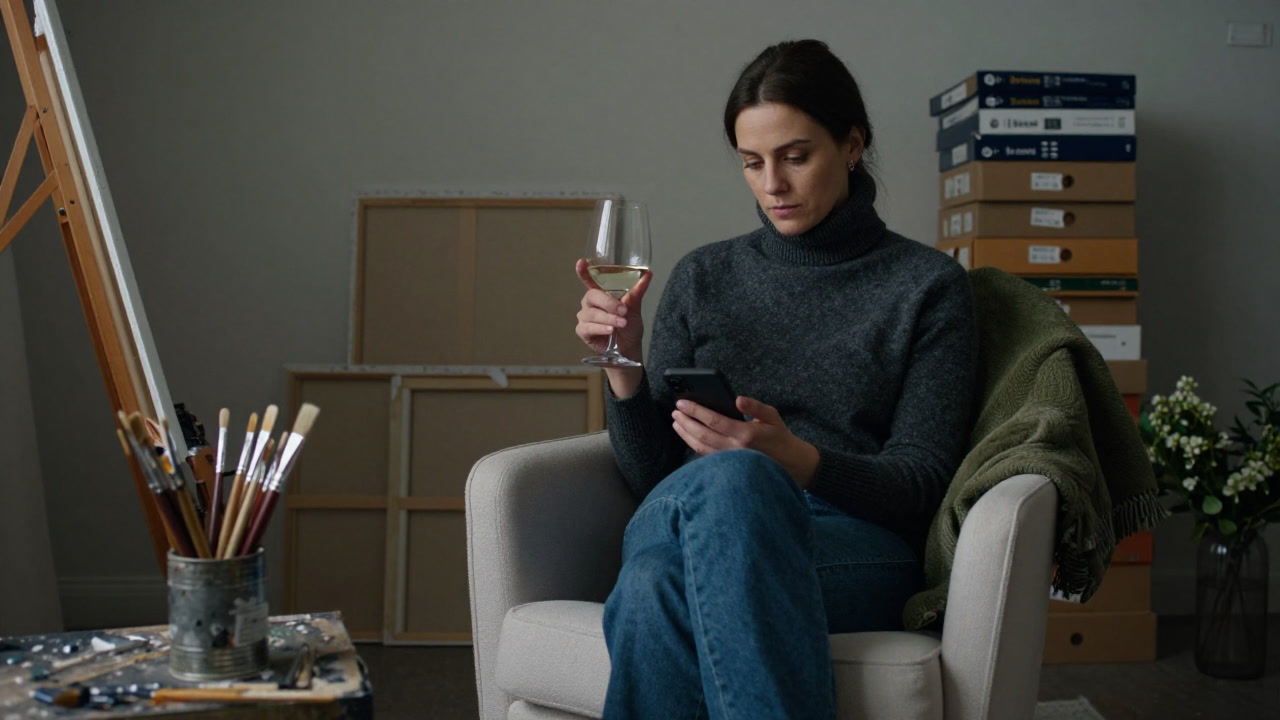} & & \\[2pt]

        \centering\rotatebox{90}{\textbf{Input Video}} &
        \includegraphics[width=\linewidth]{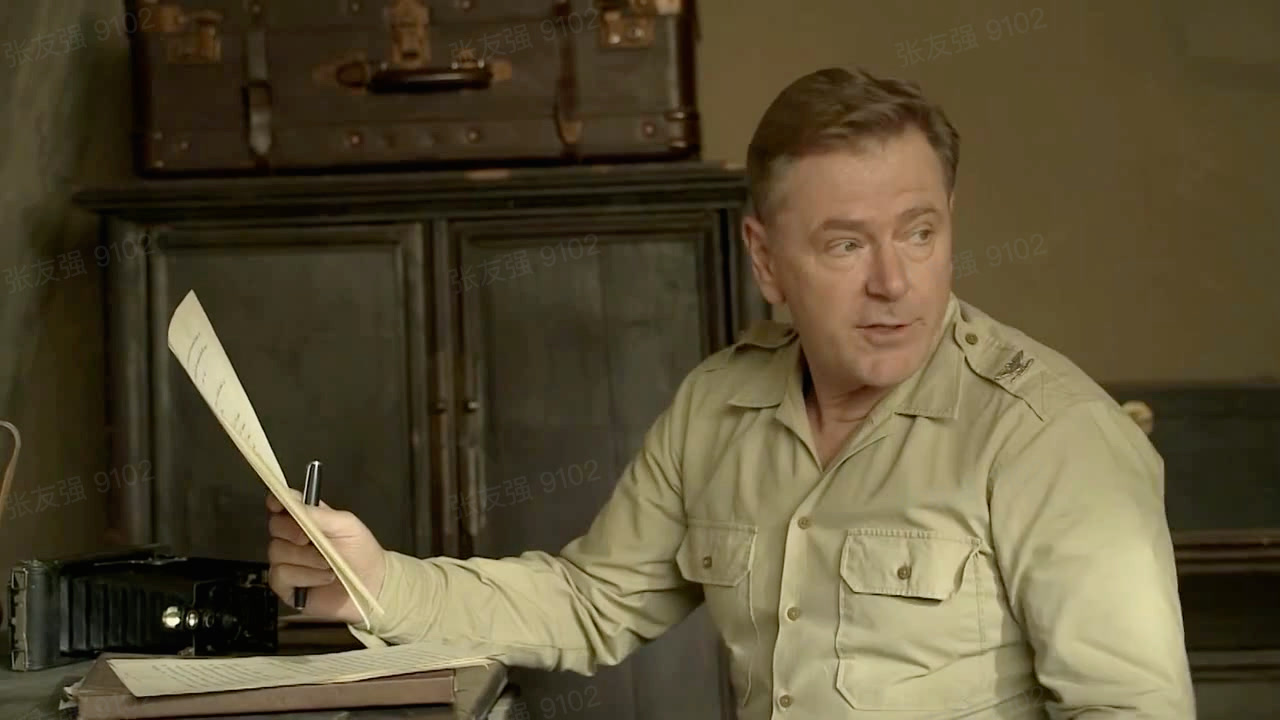} &
        \includegraphics[width=\linewidth]{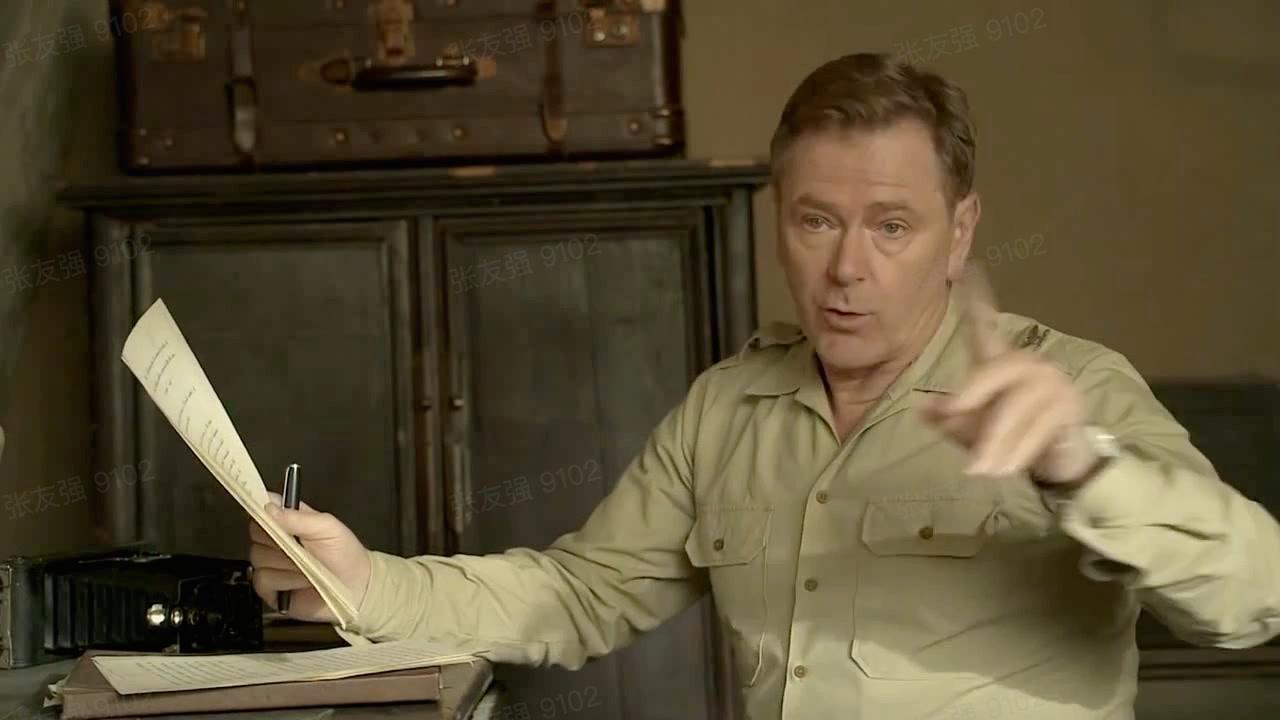} &
        \includegraphics[width=\linewidth]{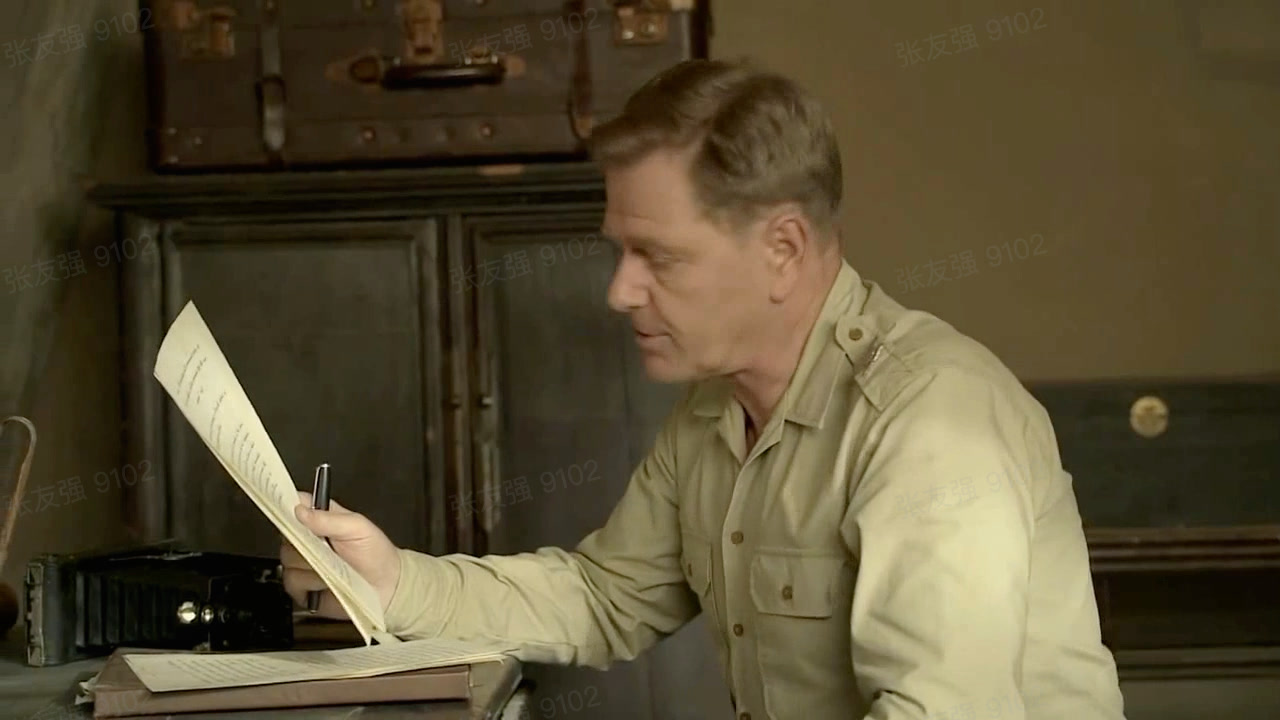} \\[2pt]

        \centering\rotatebox{90}{\textbf{Output Video}} &
        \includegraphics[width=\linewidth]{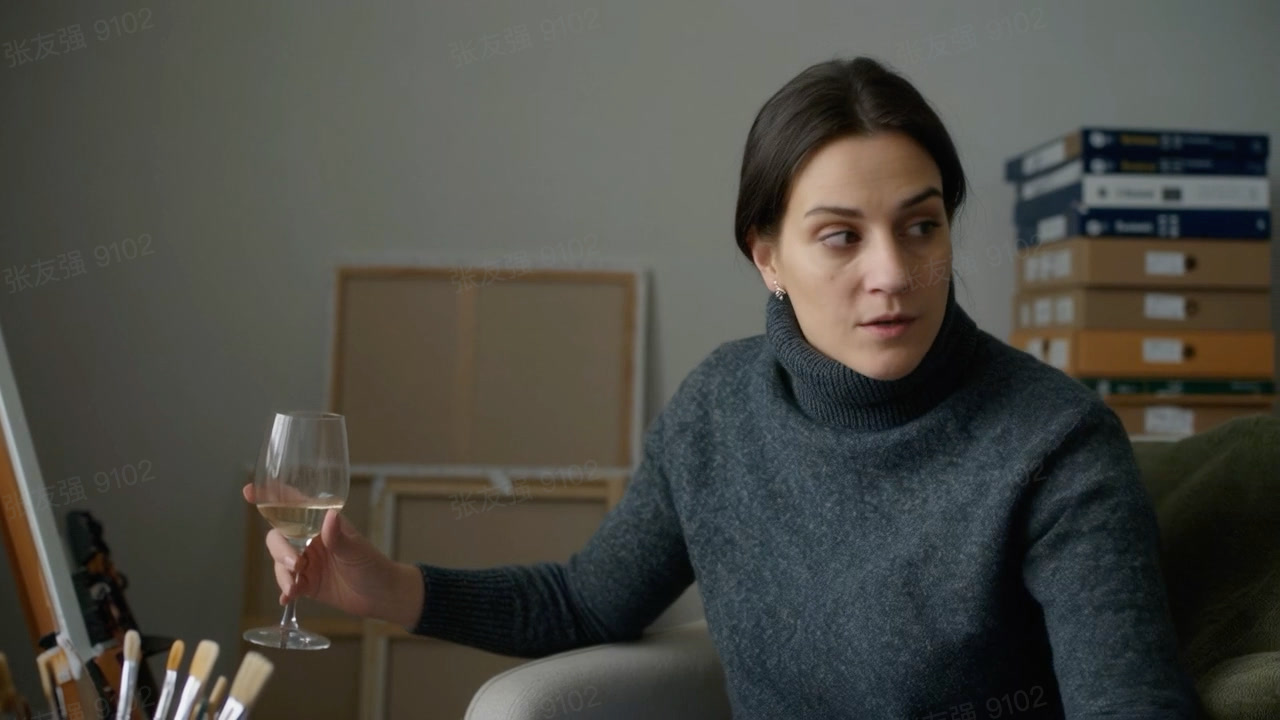} &
        \includegraphics[width=\linewidth]{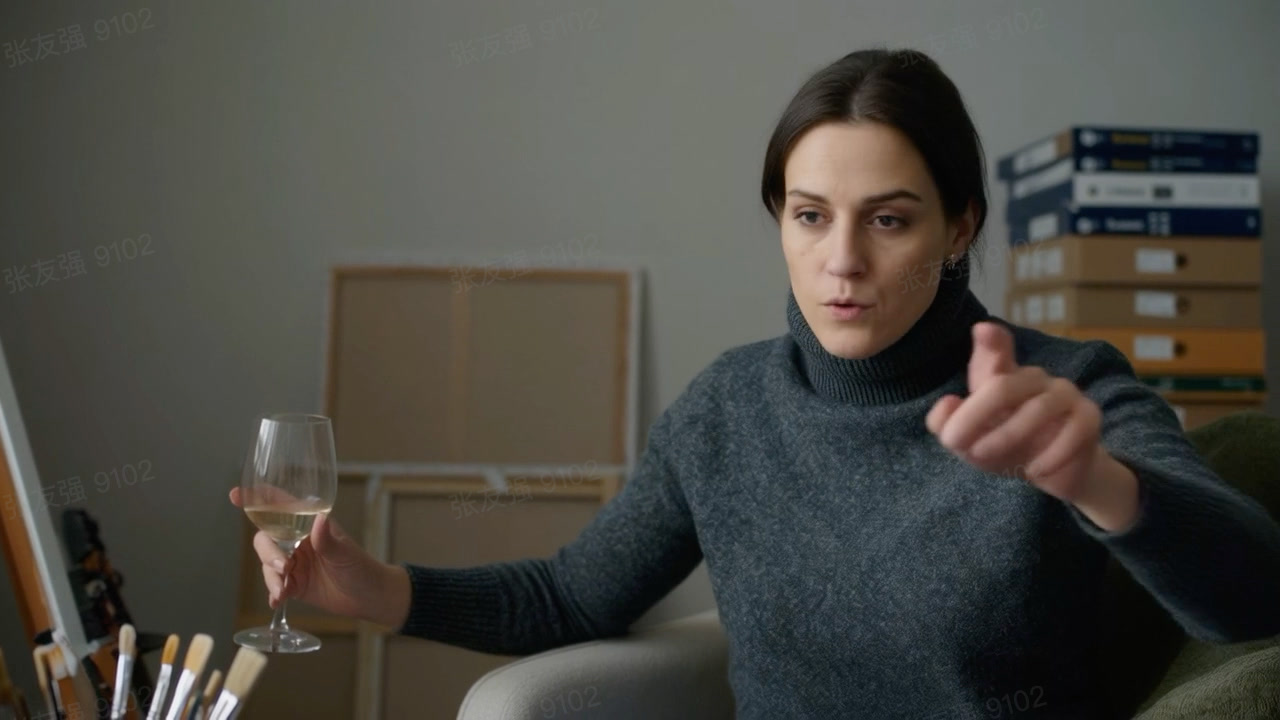} &
        \includegraphics[width=\linewidth]{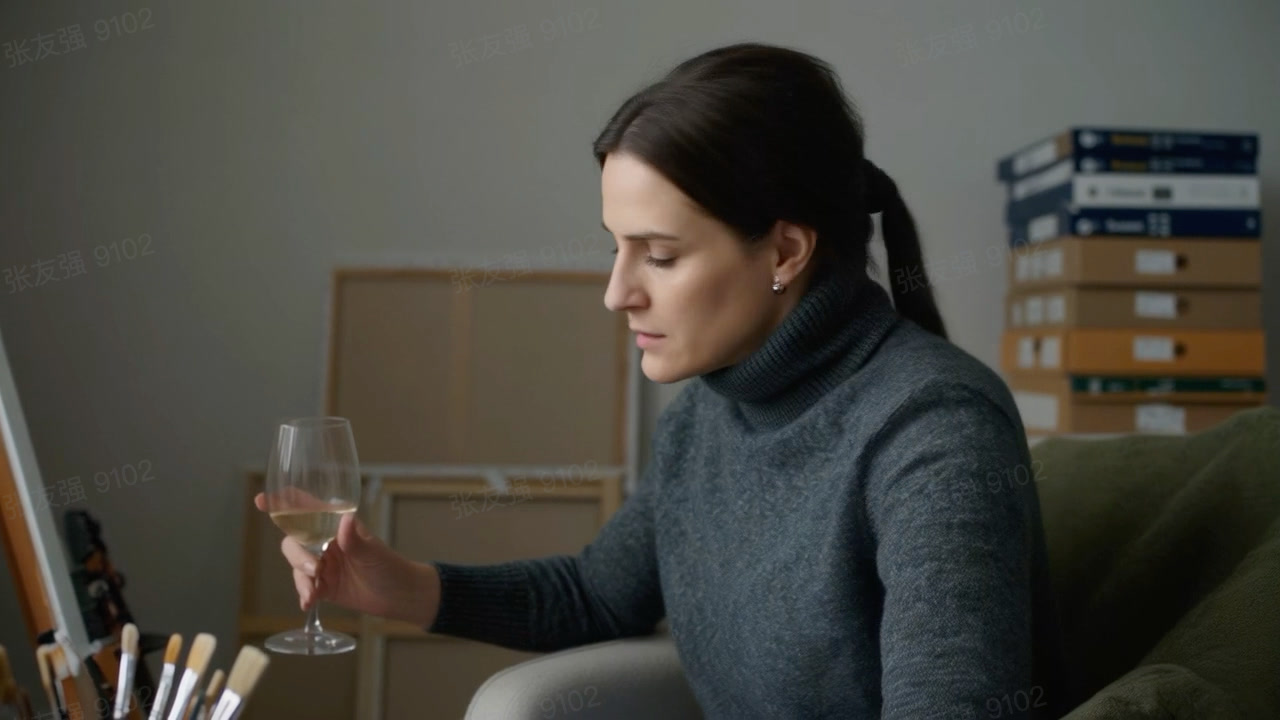} \\
    \end{tabular}
\end{flushleft}

\begin{flushleft}
    \textbf{Instruction:} \textit{The medical professional in @image\_1 and the curly-haired woman from @image\_2 execute the  dance movements demonstrated in @video\_1, all set within the same stage environment as @video\_1.}%
    \vspace{0.5em}
    \begin{tabular}{m{1.2em} m{0.22\linewidth} m{0.22\linewidth} m{0.22\linewidth}}
        

        \centering\rotatebox{90}{\textbf{Ref. Image}} &
        \includegraphics[width=\linewidth]{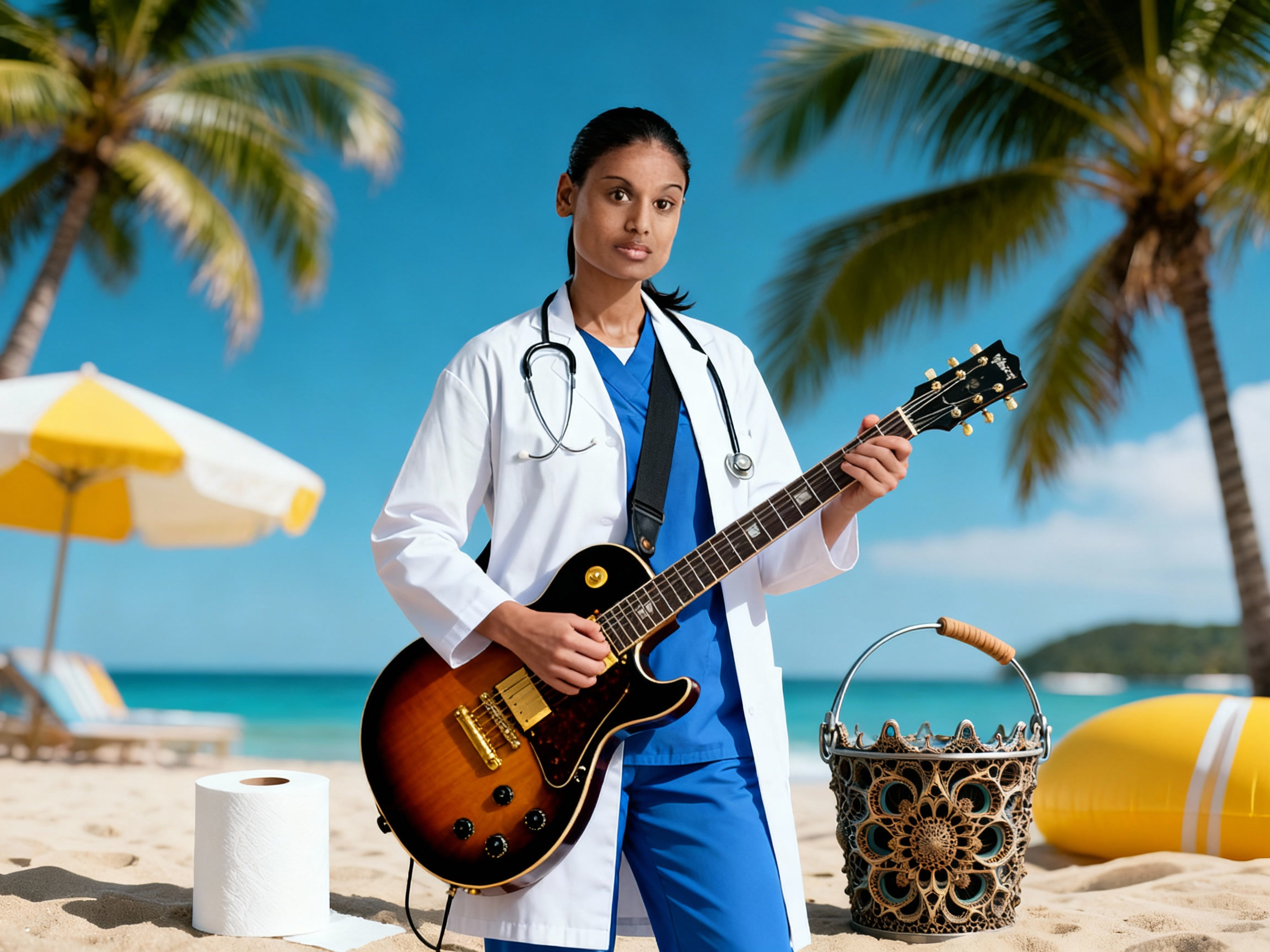} &
        \includegraphics[width=\linewidth]{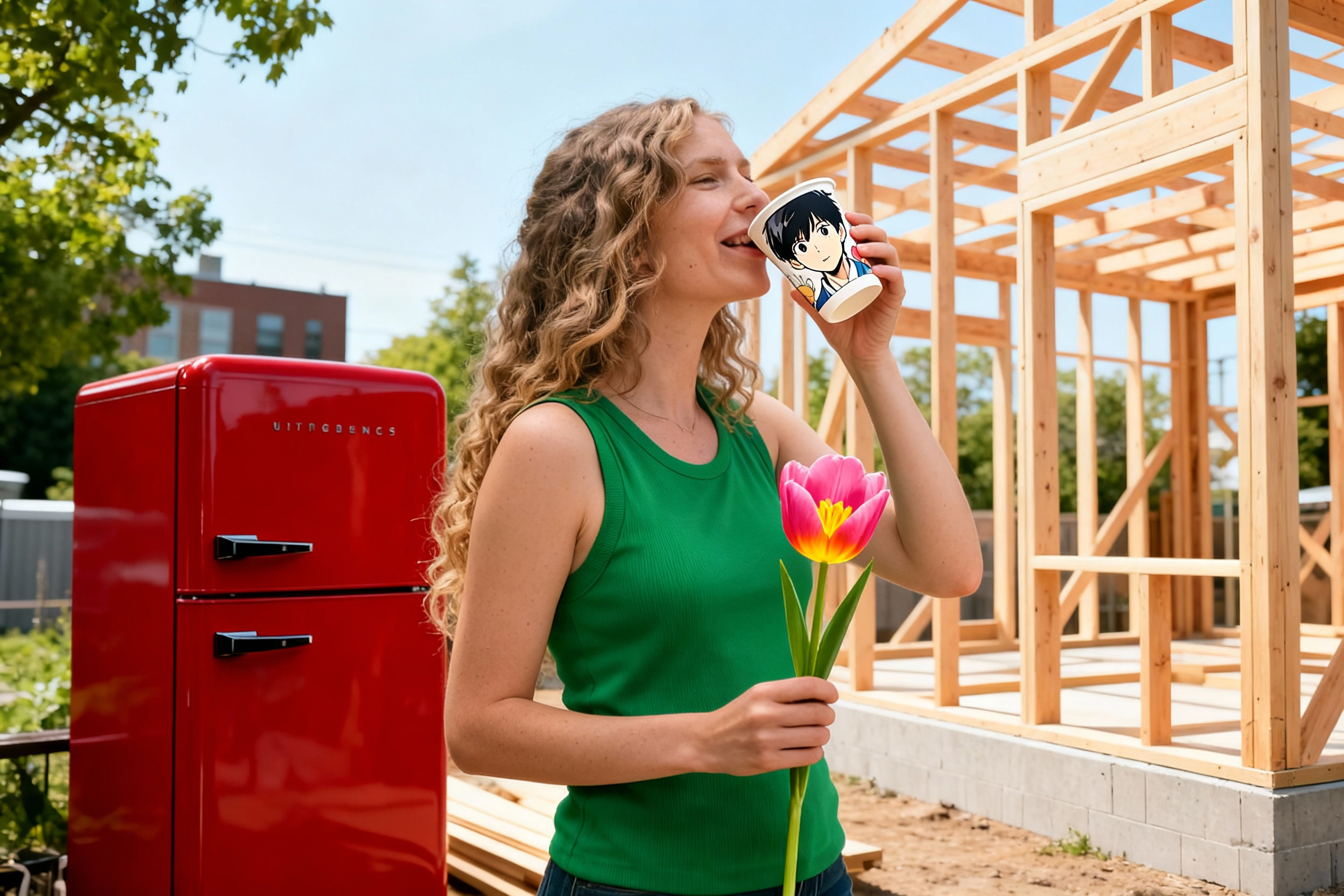} & \\[2pt]

        \centering\rotatebox{90}{\textbf{Input Video}} &
        \includegraphics[width=\linewidth]{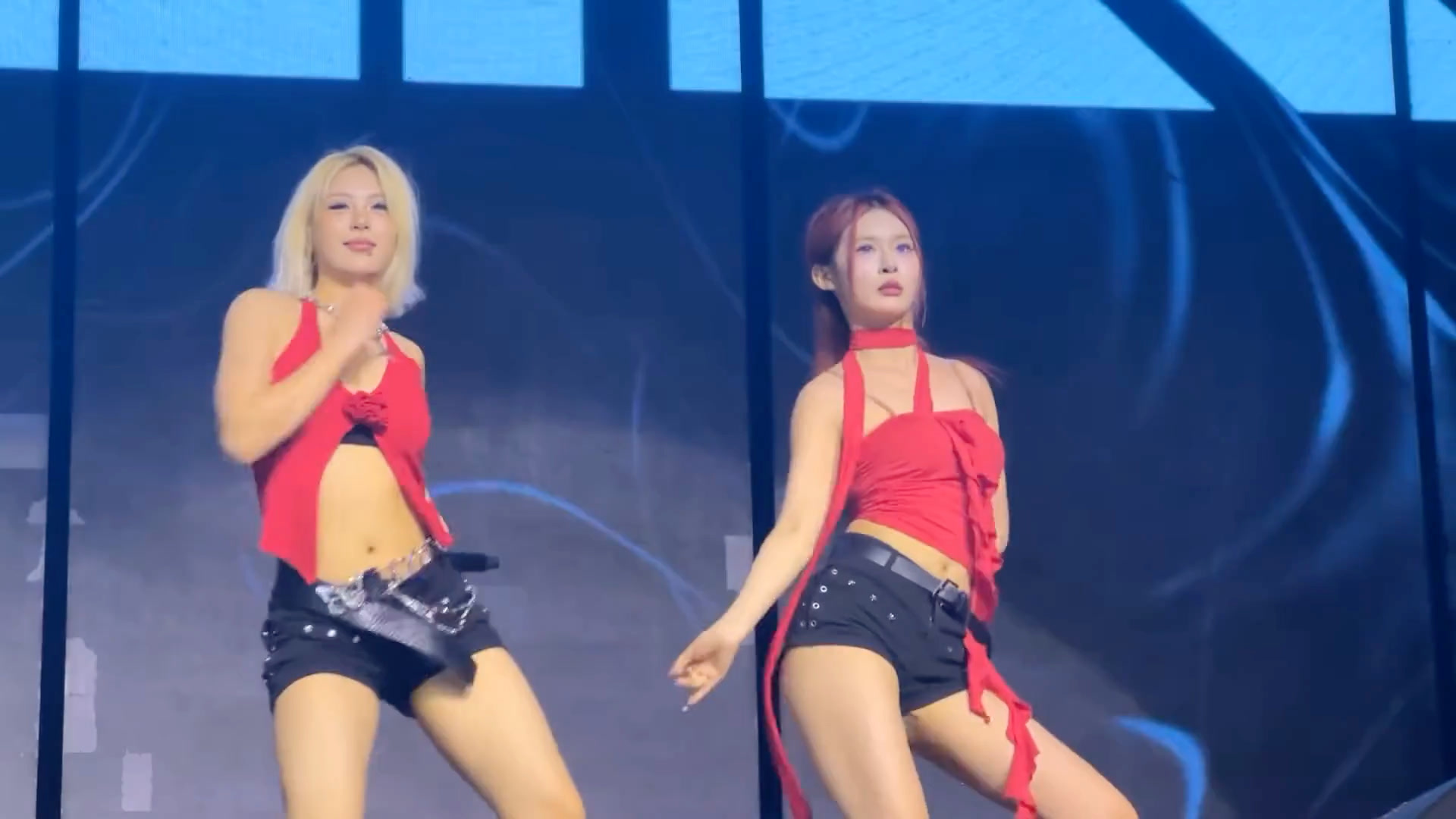} &
        \includegraphics[width=\linewidth]{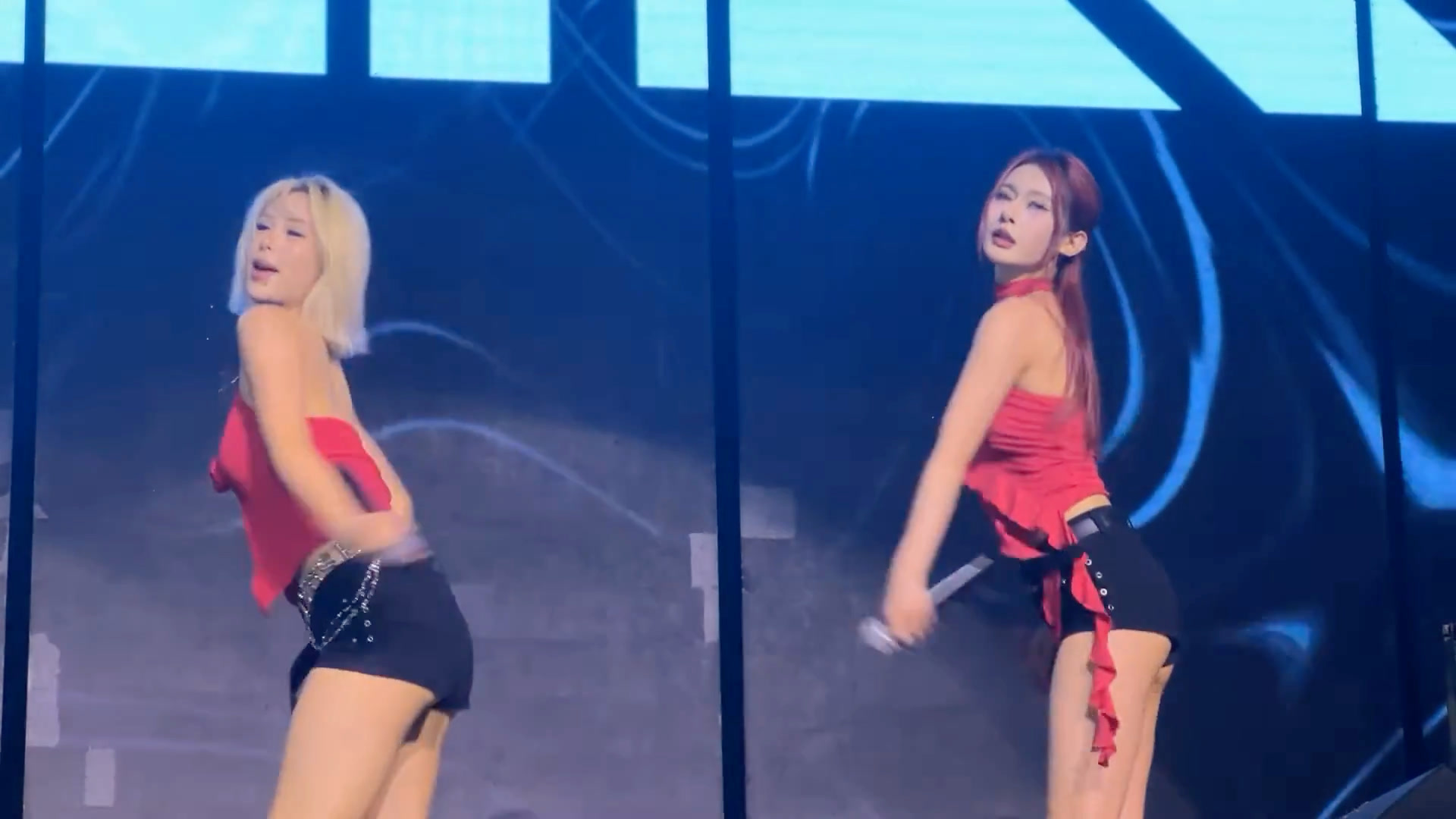} &
        \includegraphics[width=\linewidth]{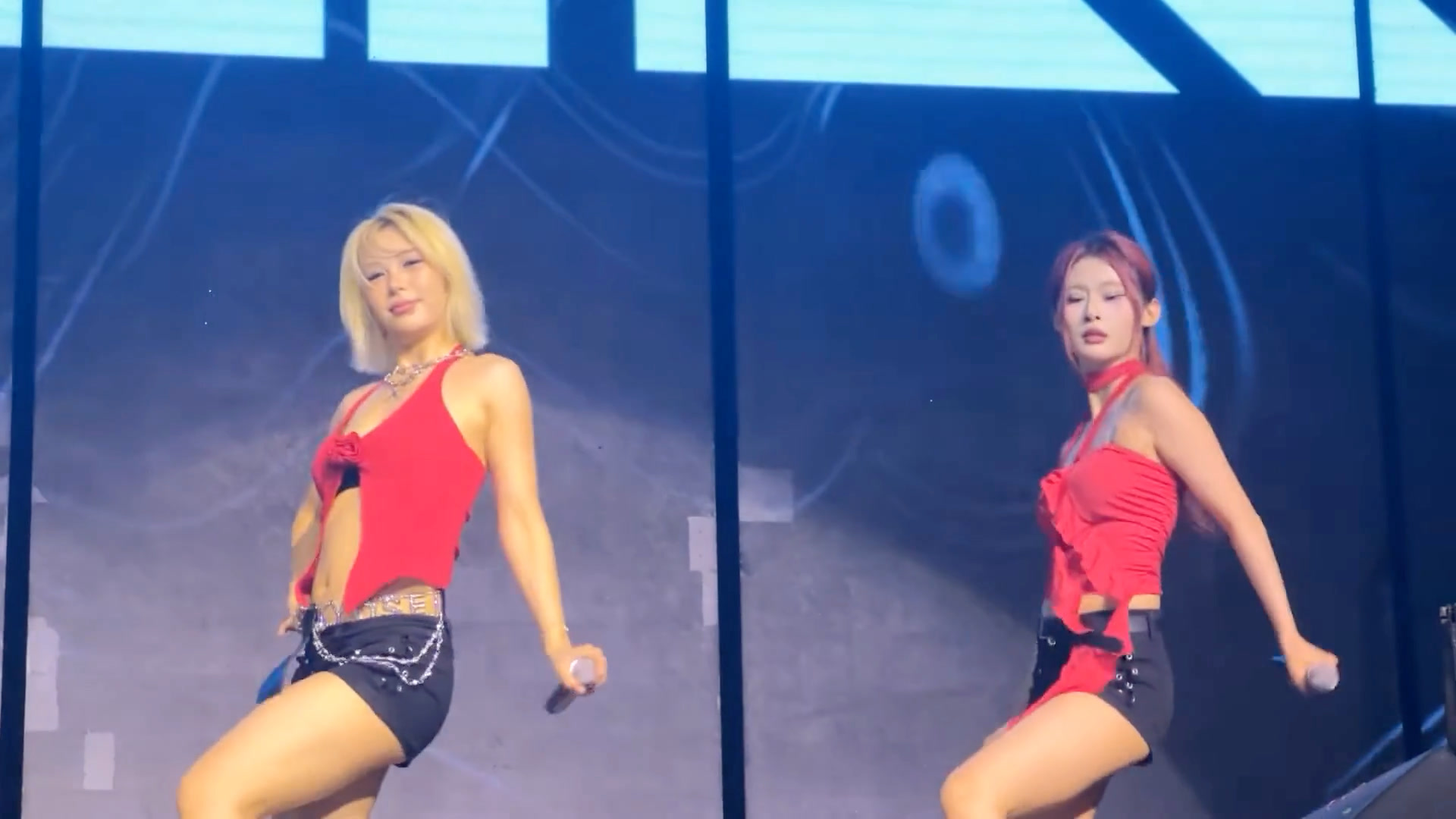} \\[2pt]

        \centering\rotatebox{90}{\textbf{Output Video}} &
        \includegraphics[width=\linewidth]{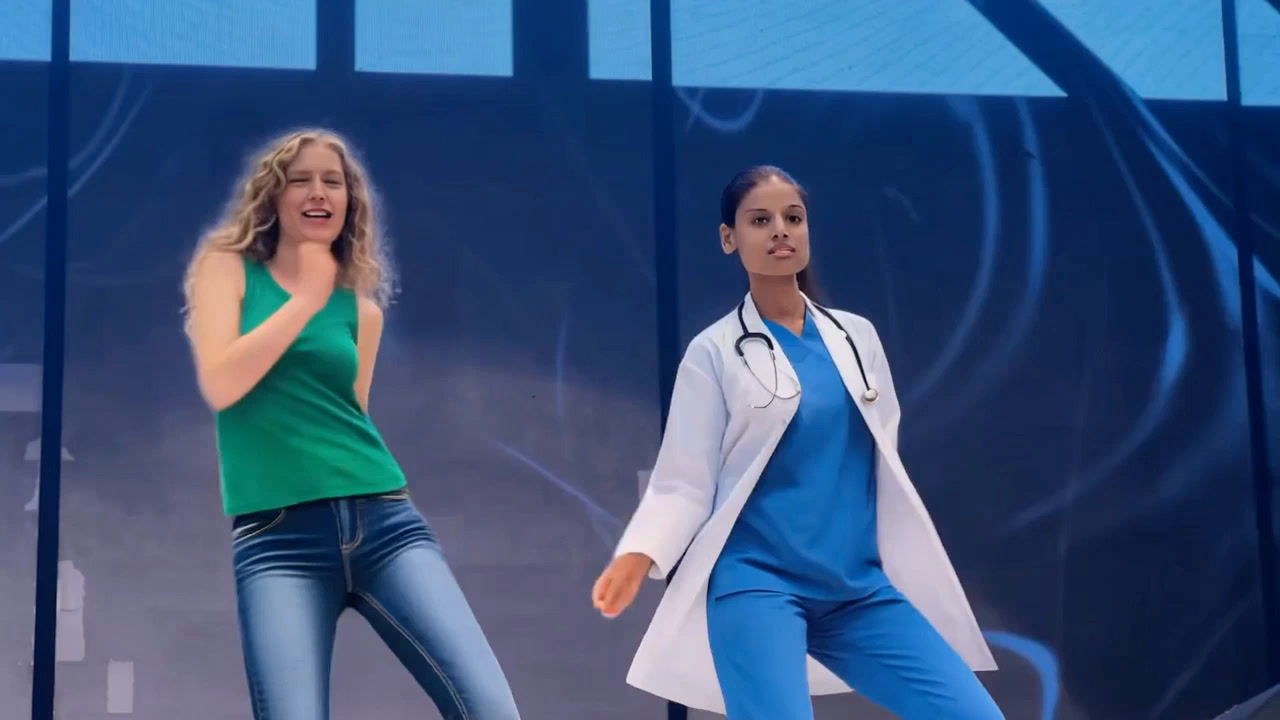} &
        \includegraphics[width=\linewidth]{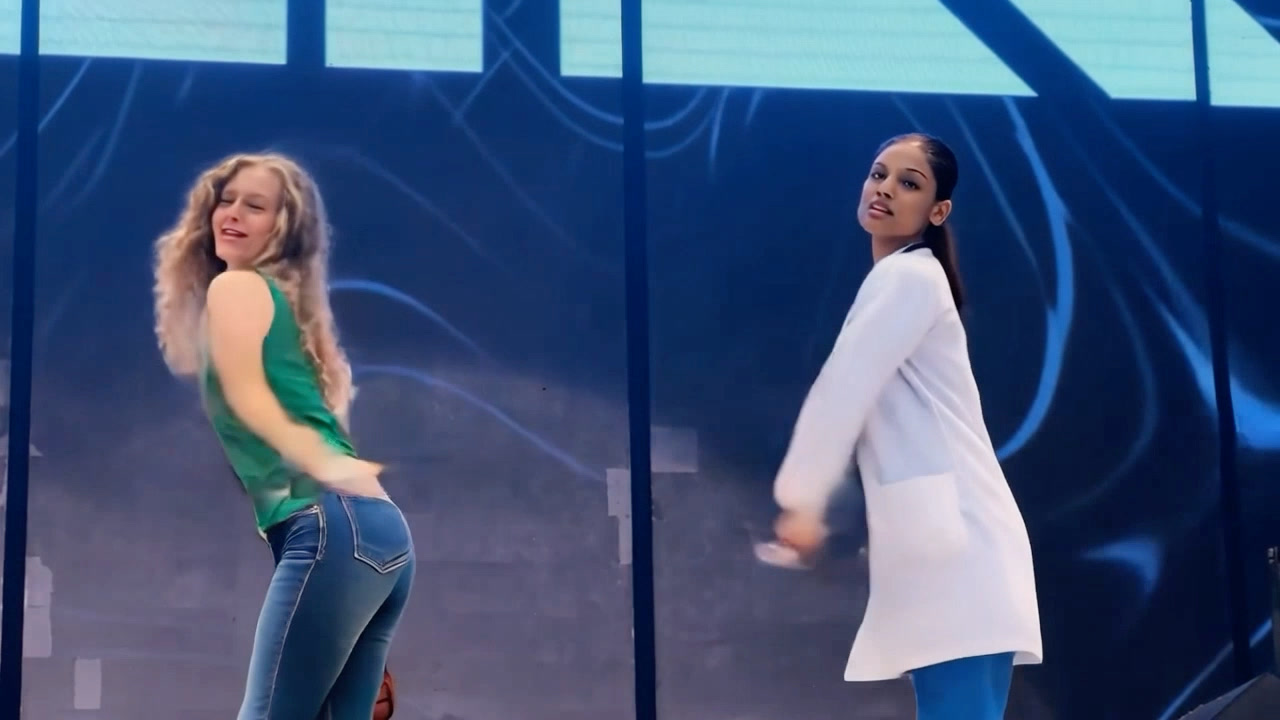} &
        \includegraphics[width=\linewidth]{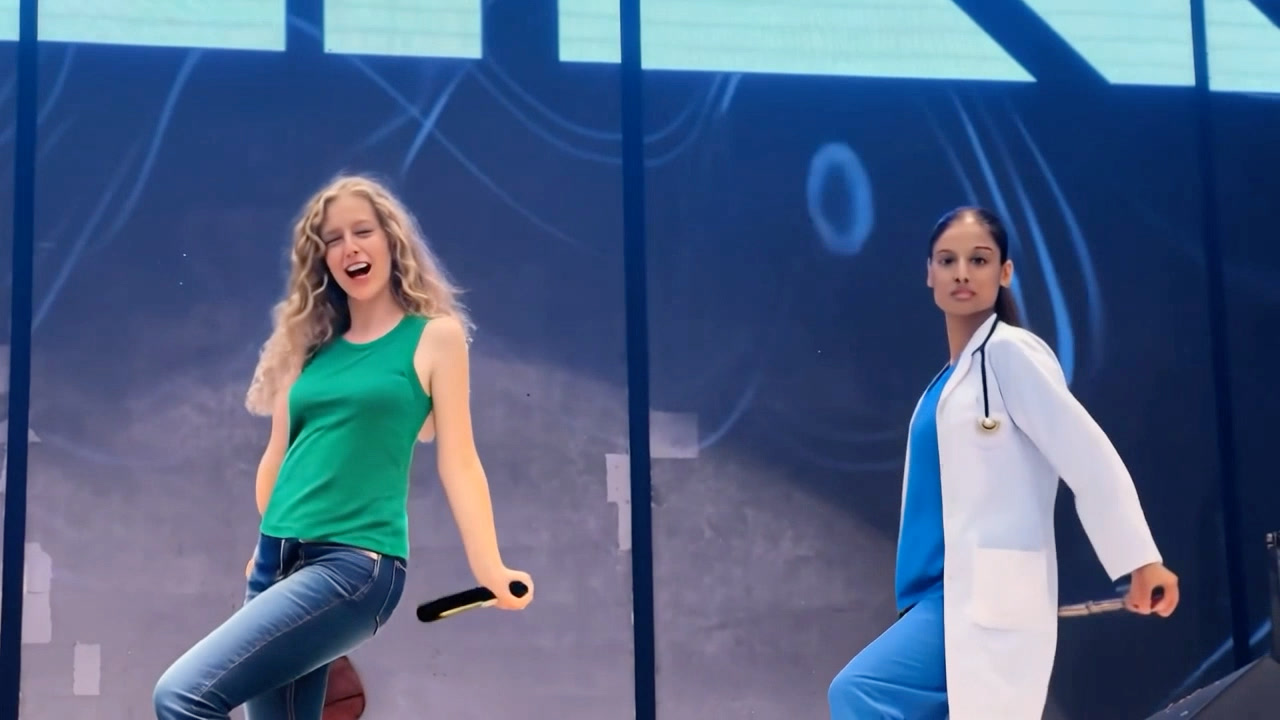} \\
    \end{tabular}
\end{flushleft}


\begin{figure}[h!]
    \centering
    \caption{Examples of motion transfer in video reference.}
    \label{fig:gen_motion_ref_2}
\end{figure}

\newpage
\subsection{Video Inpainting}
\label{appendix:inpainting}

\subsubsection{Subject/Attribute/Background Inpainting}
\label{appendix:object-inpainting}

The model can precisely inpaint specific regions in videos, including subject replacement, attribute modification, and background replacement.


\begin{flushleft}
    \textbf{Instruction:} \textit{Replace the subject in the mask area in @video\_1 with a majestic elk standing in the same field.}%
    \vspace{0.5em}
    \begin{tabular}{m{1.2em} m{0.22\linewidth} m{0.22\linewidth} m{0.22\linewidth}}

        \centering\rotatebox{90}{\textbf{Input Video}} &
        \includegraphics[width=\linewidth]{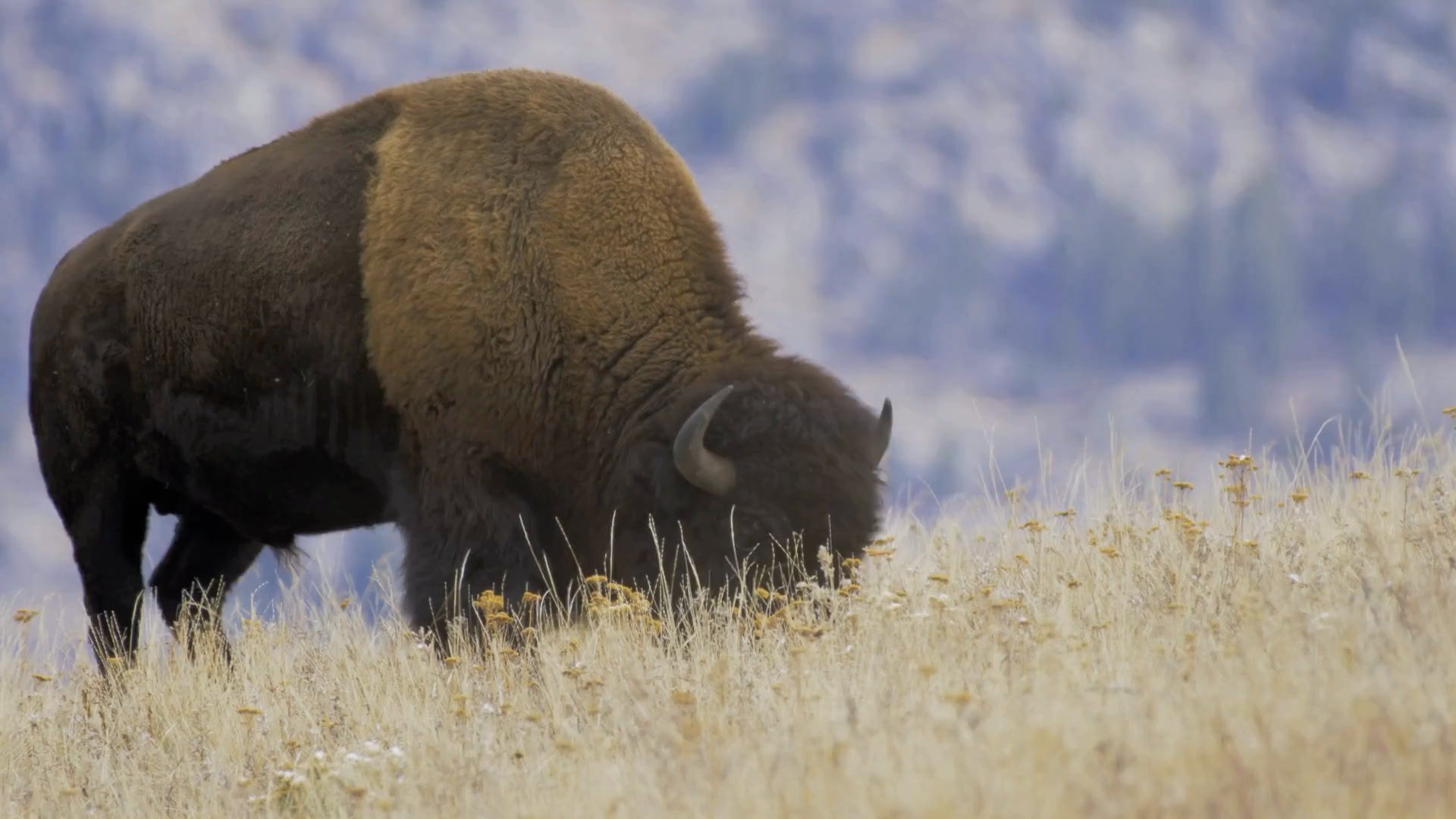} &
        \includegraphics[width=\linewidth]{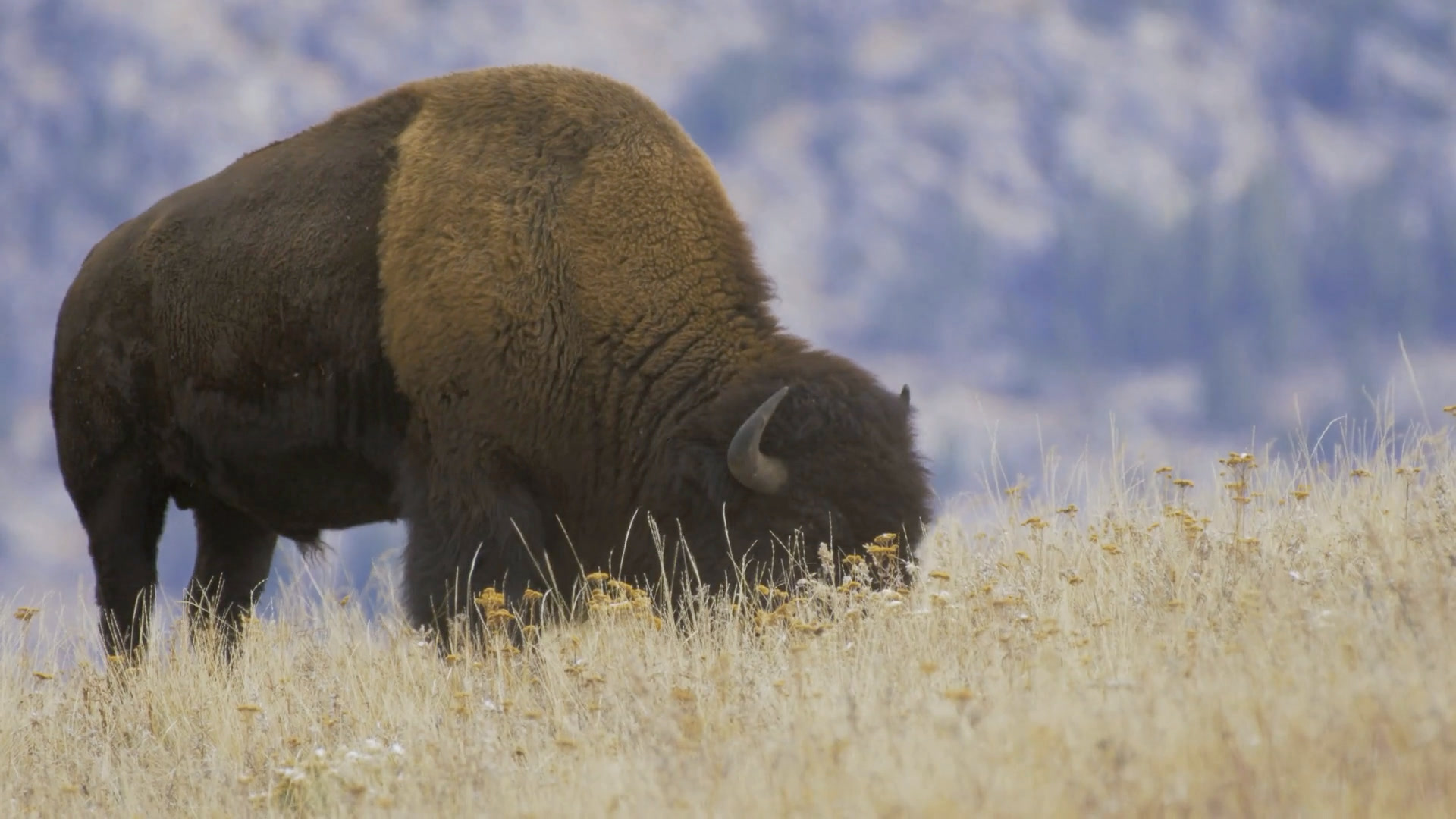} &
        \includegraphics[width=\linewidth]{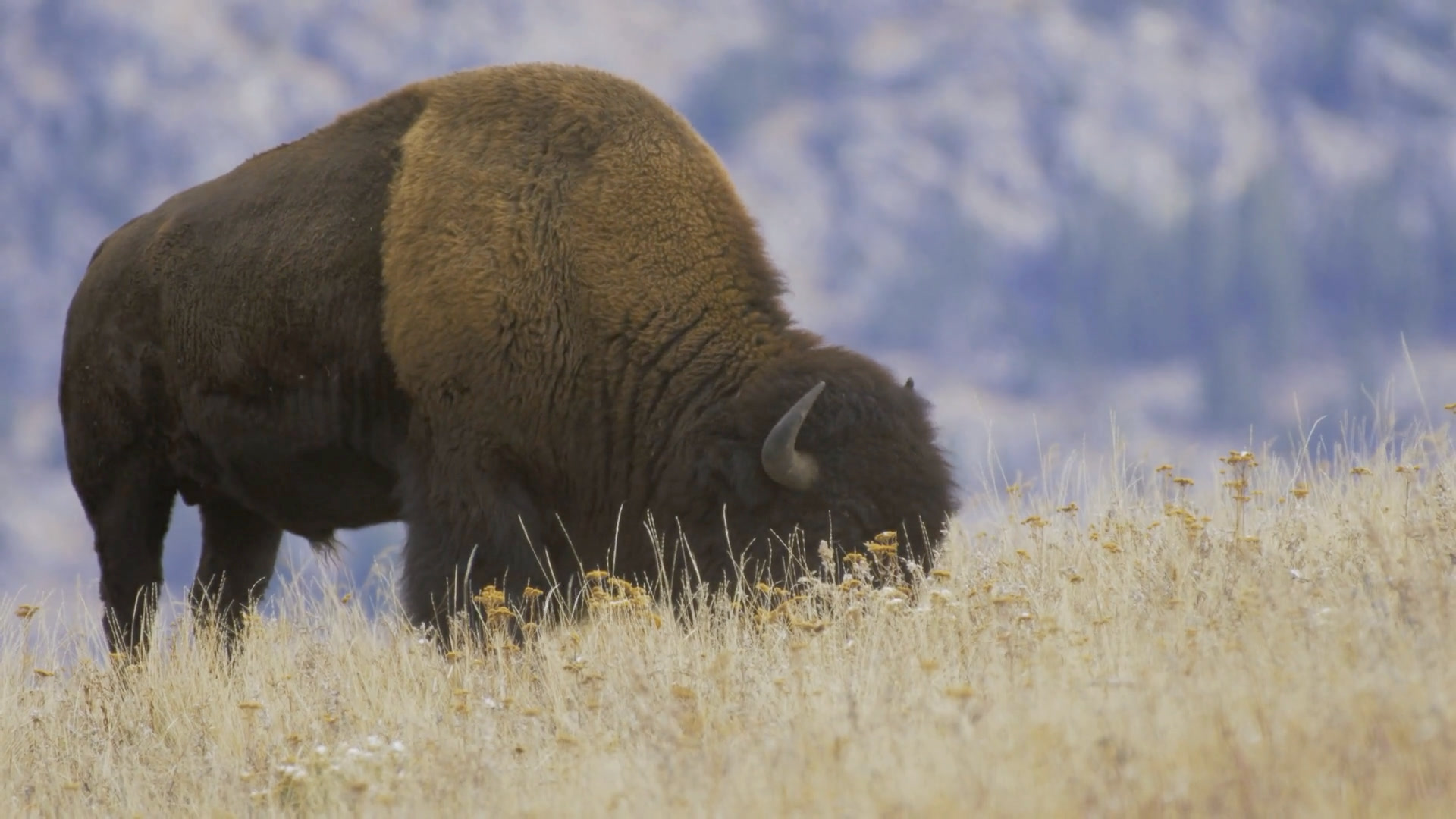} \\[2pt]

        \centering\rotatebox{90}{\textbf{Masked Video}} &
        \includegraphics[width=\linewidth]{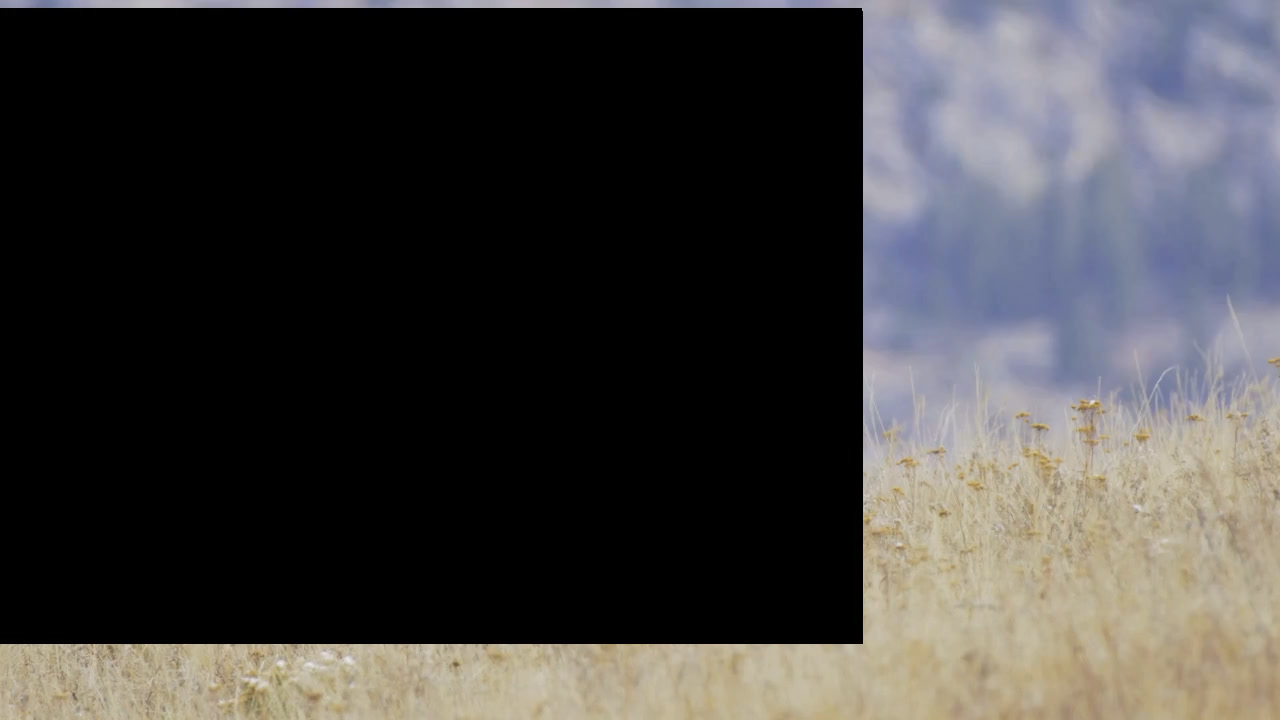} &
        \includegraphics[width=\linewidth]{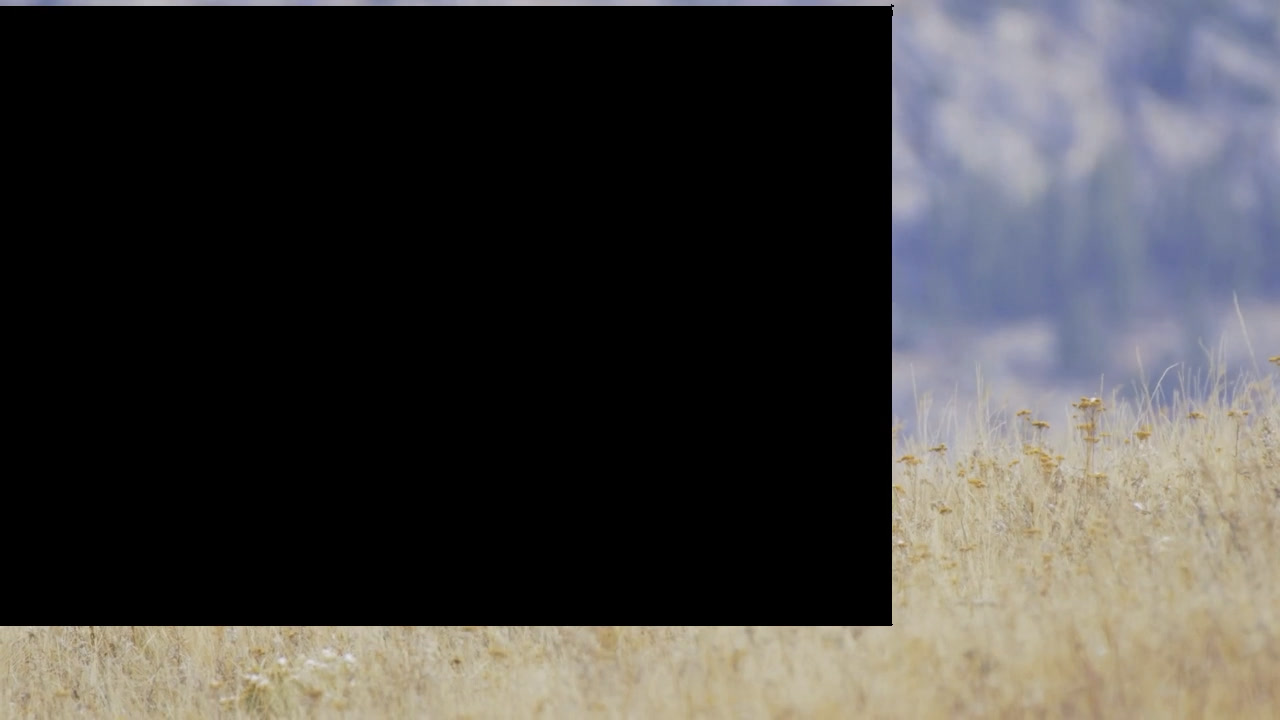} &
        \includegraphics[width=\linewidth]{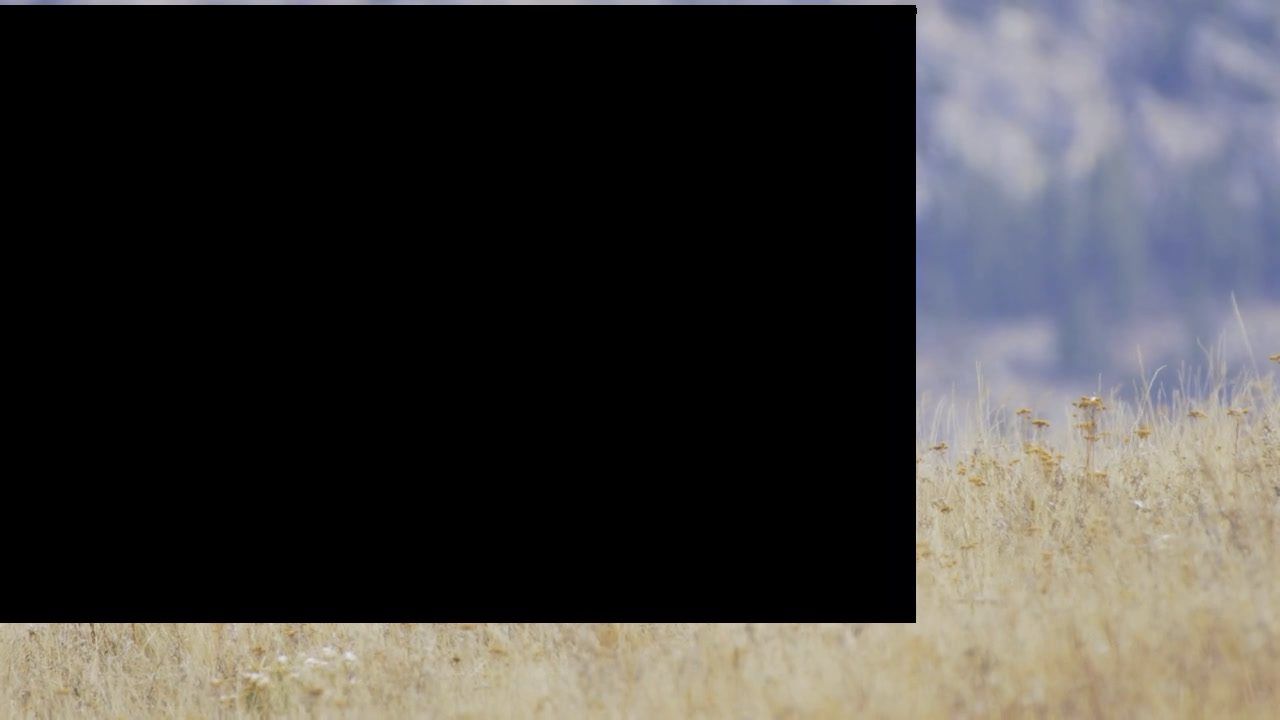} \\[2pt]

        \centering\rotatebox{90}{\textbf{Output Video}} &
        \includegraphics[width=\linewidth]{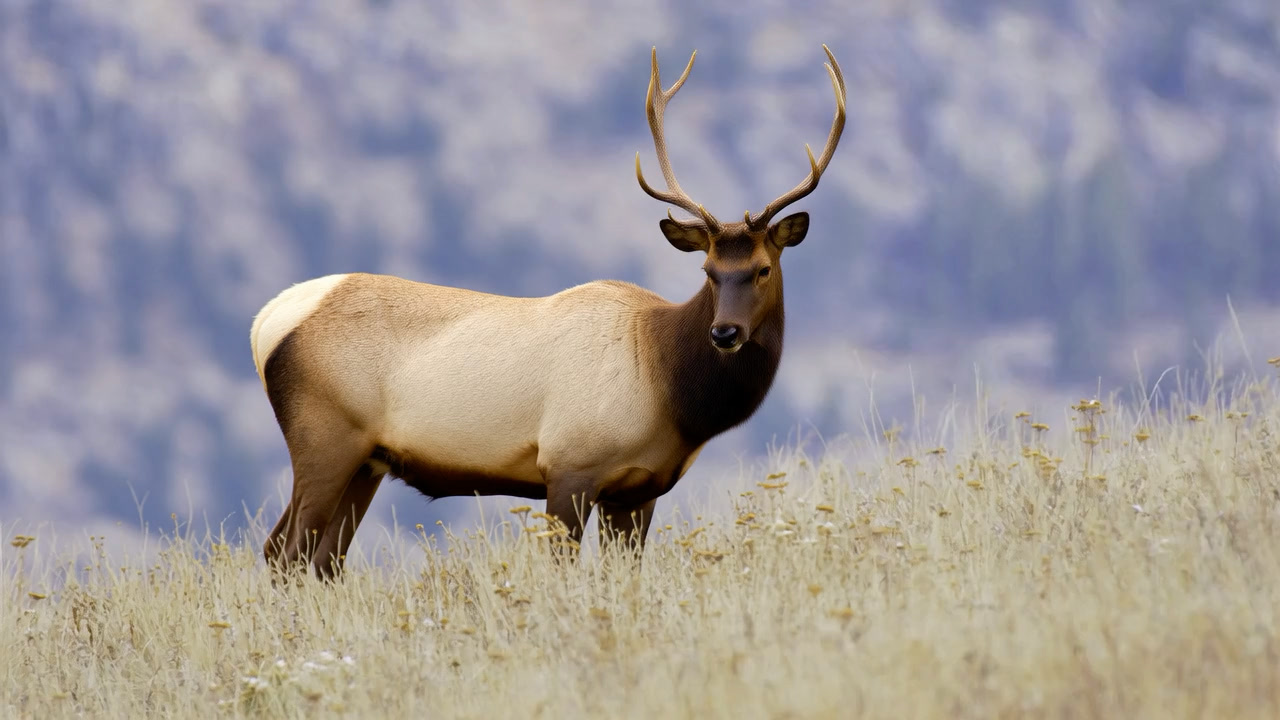} &
        \includegraphics[width=\linewidth]{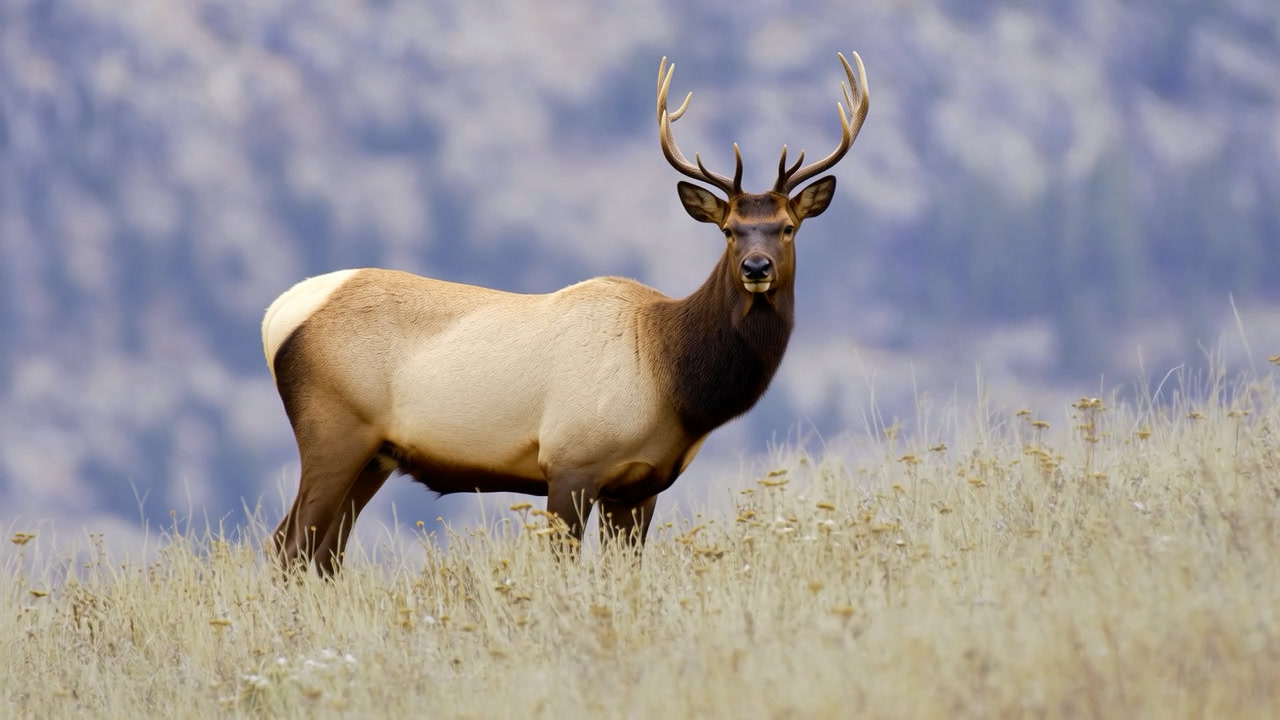} &
        \includegraphics[width=\linewidth]{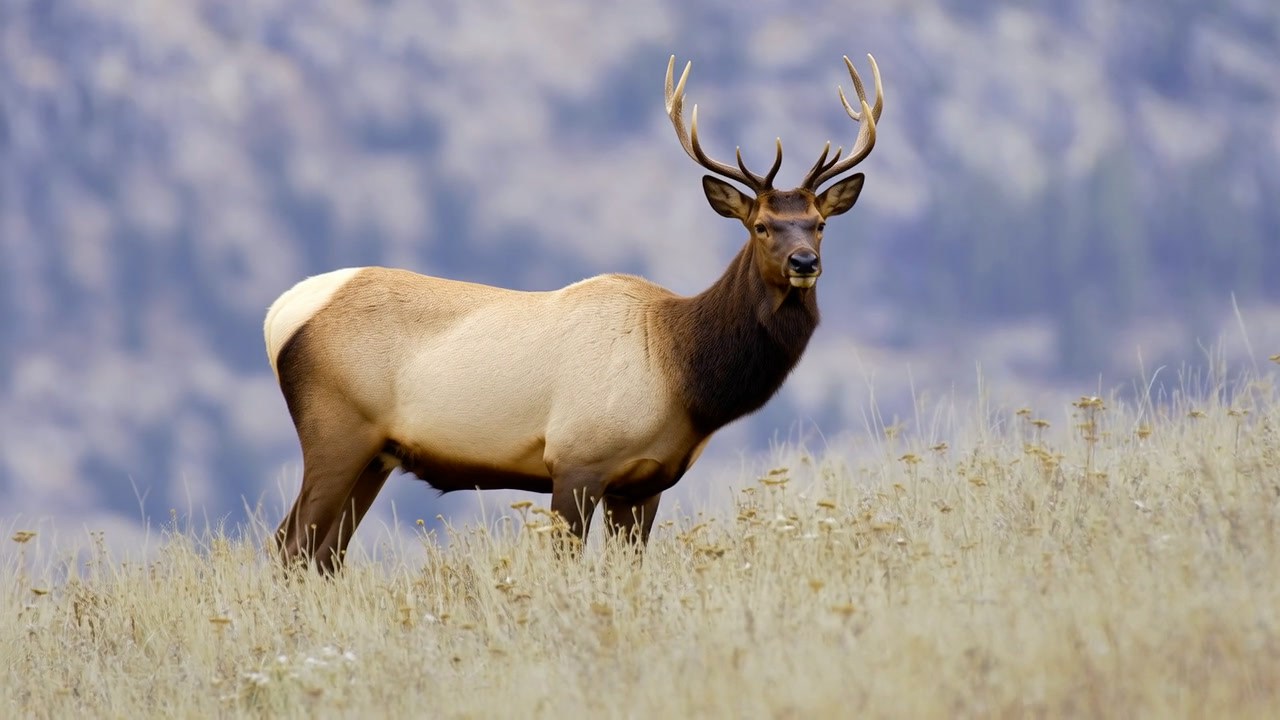} \\
    \end{tabular}
\end{flushleft} 


\begin{flushleft}
    \textbf{Instruction:} \textit{Change the color of the tie in the masked area of @video\_1 to blue.}%
    \vspace{0.5em}
    \begin{tabular}{m{1.2em} m{0.22\linewidth} m{0.22\linewidth} m{0.22\linewidth}}

        \centering\rotatebox{90}{\textbf{Input Video}} &
        \includegraphics[width=\linewidth]{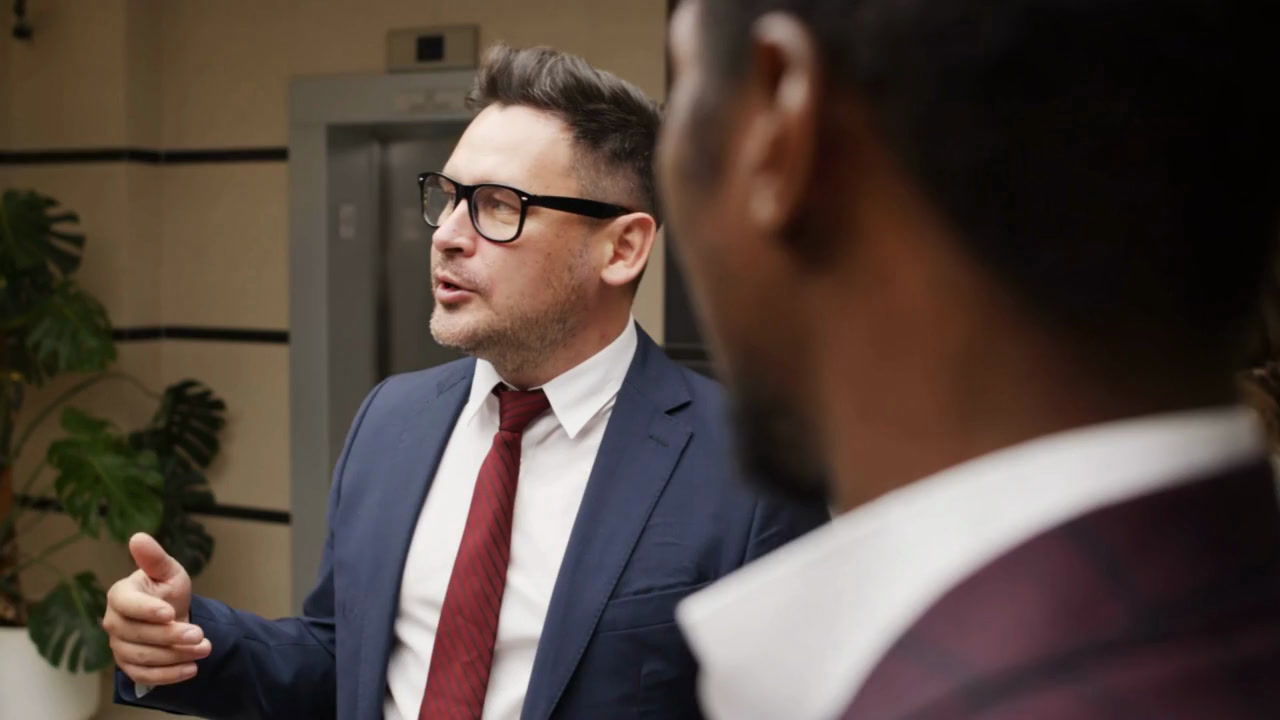} &
        \includegraphics[width=\linewidth]{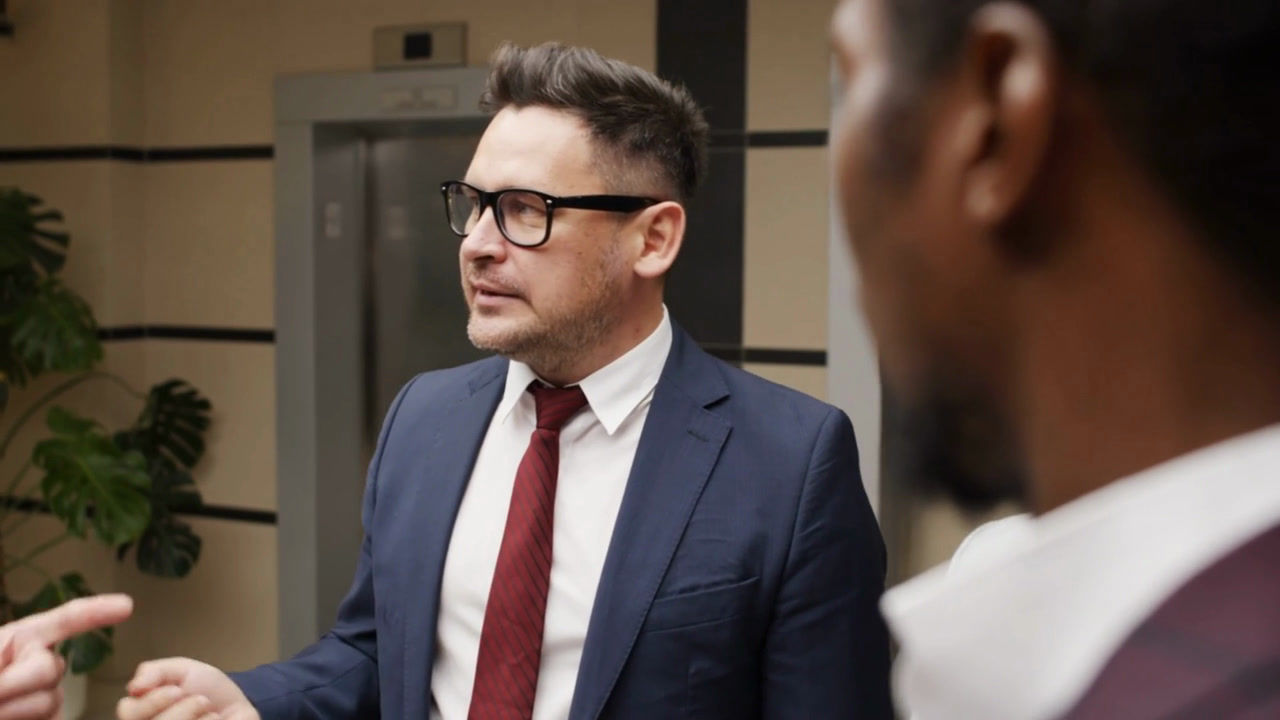} &
        \includegraphics[width=\linewidth]{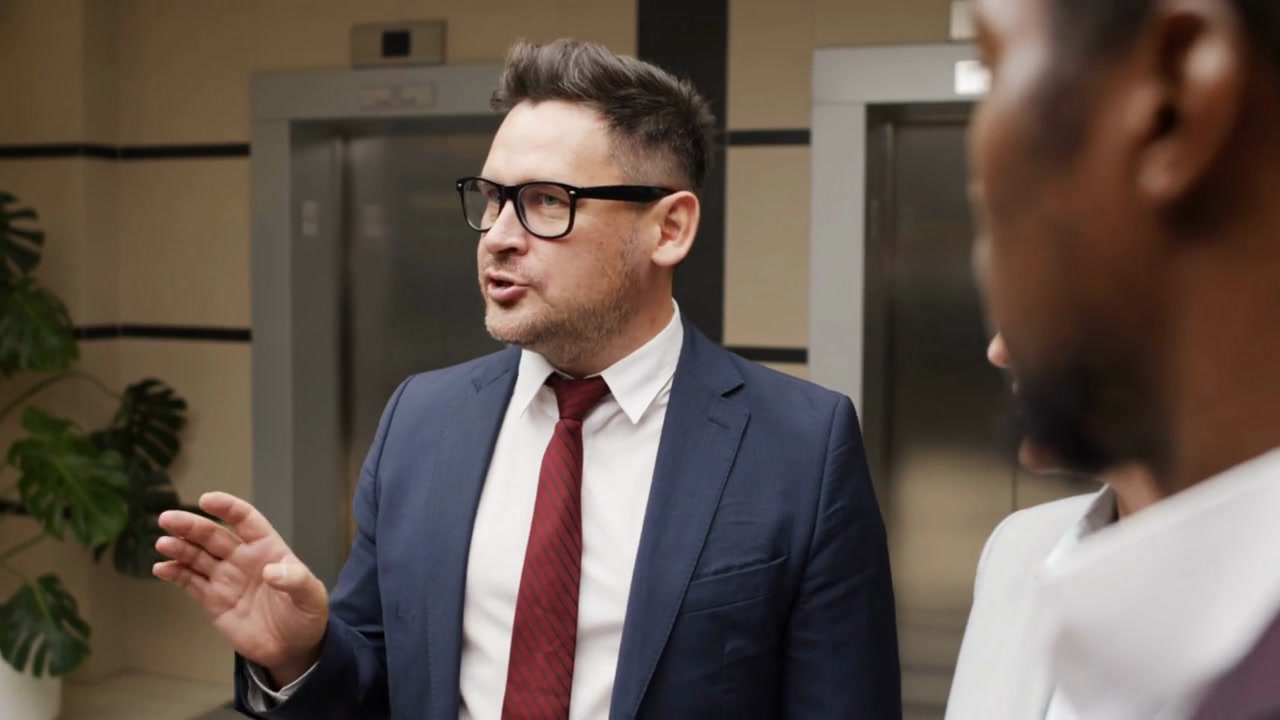} \\[2pt]

        \centering\rotatebox{90}{\textbf{Masked Video}} &
        \includegraphics[width=\linewidth]{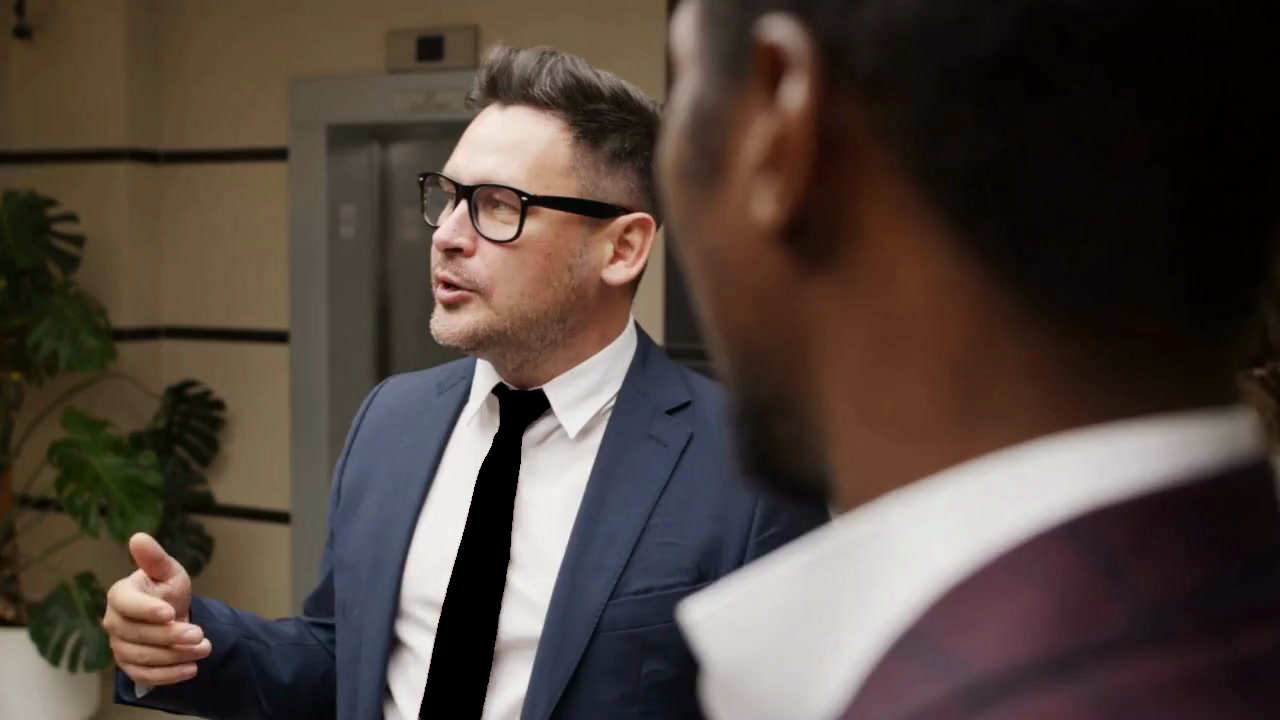} &
        \includegraphics[width=\linewidth]{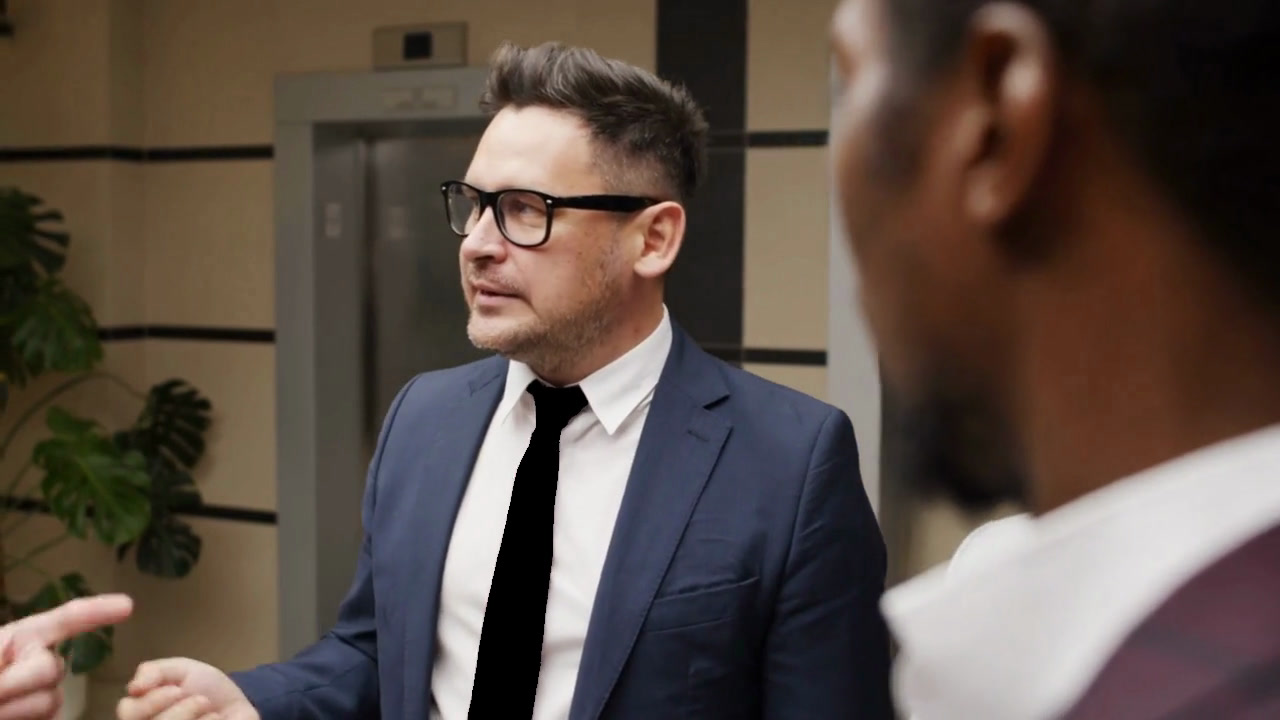} &
        \includegraphics[width=\linewidth]{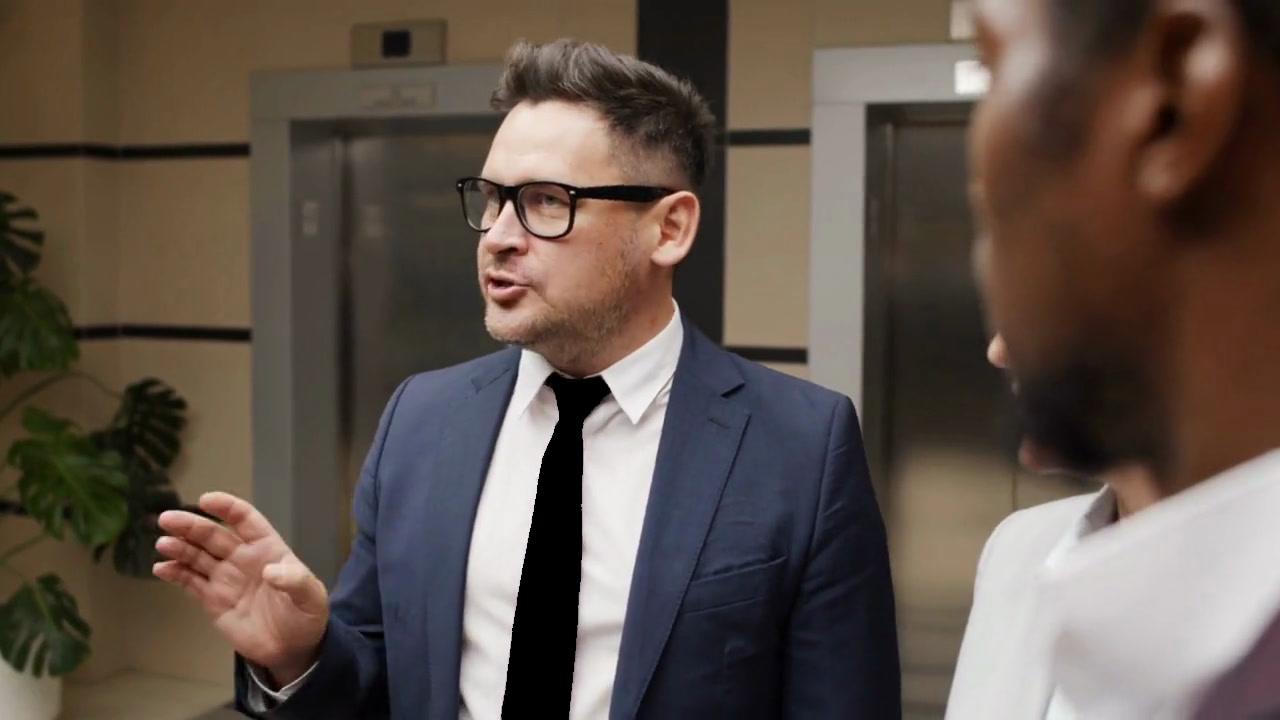} \\[2pt]

        \centering\rotatebox{90}{\textbf{Output Video}} &
        \includegraphics[width=\linewidth]{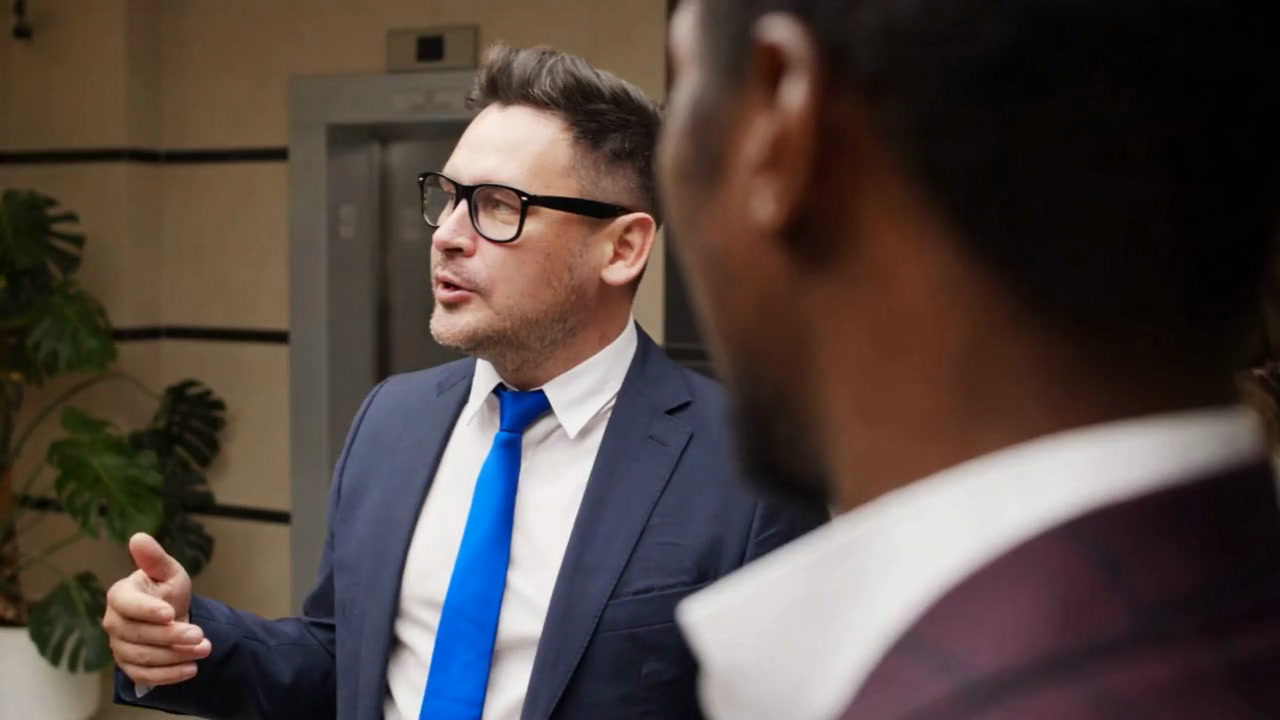} &
        \includegraphics[width=\linewidth]{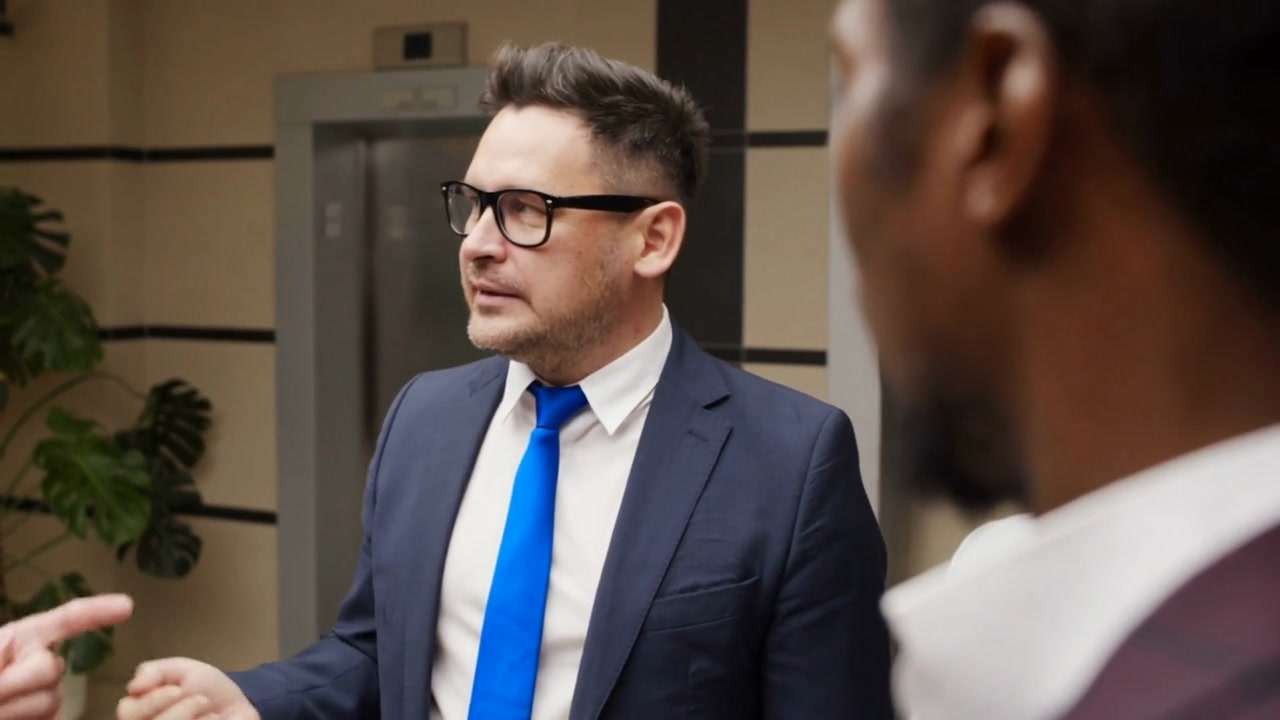} &
        \includegraphics[width=\linewidth]{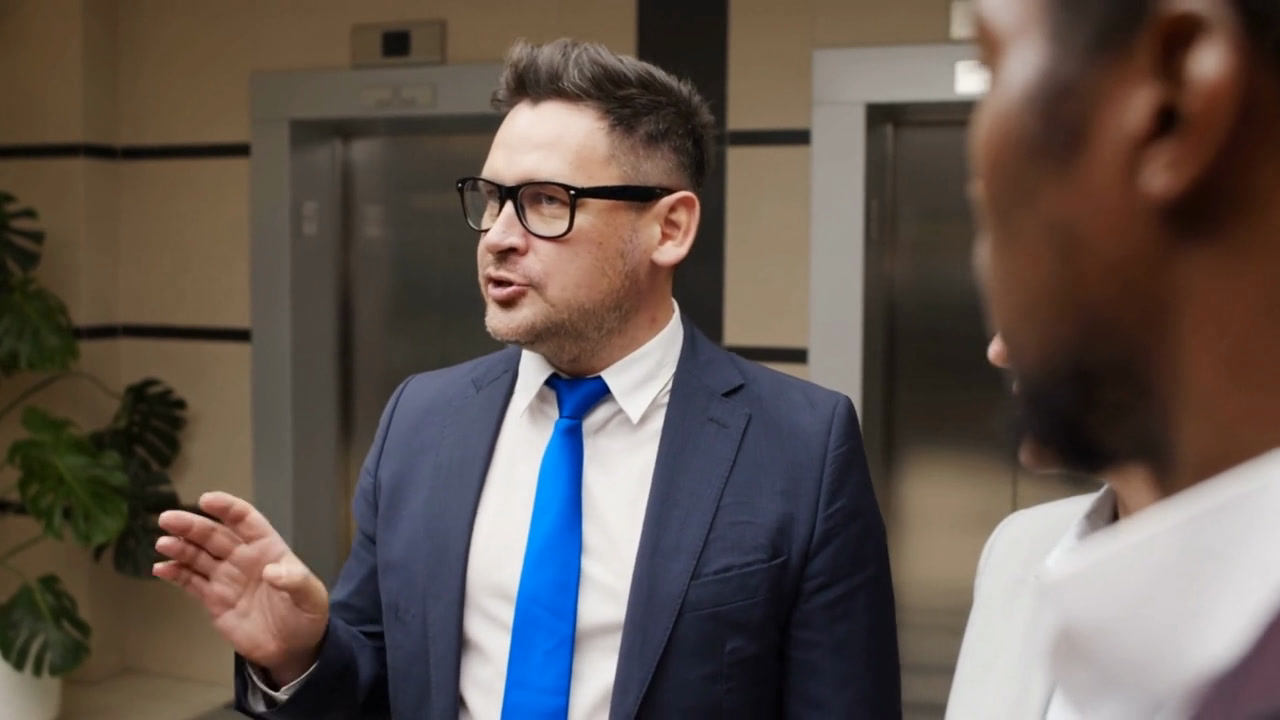} \\
    \end{tabular}
\end{flushleft}



\newpage
\begin{flushleft}
    \textbf{Instruction:} \textit{Replace the background in the masked area of @video\_1 with a stunning cinematic view of the Amalfi Coast in Italy during a warm golden hour sunset.}%
    \vspace{0.5em}
    \begin{tabular}{m{1.2em} m{0.22\linewidth} m{0.22\linewidth} m{0.22\linewidth}}

        \centering\rotatebox{90}{\textbf{Input Video}} &
        \includegraphics[width=\linewidth]{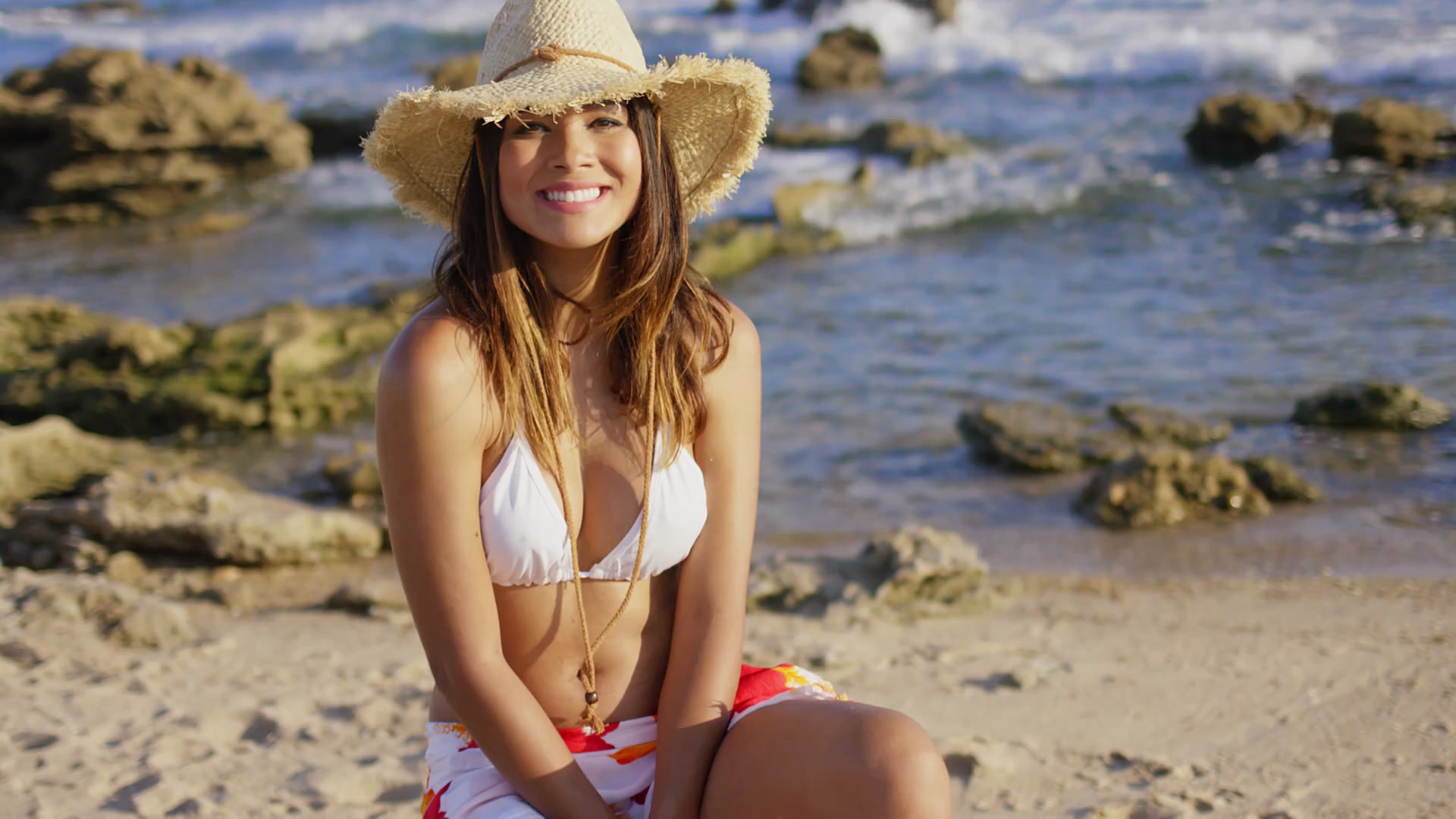} &
        \includegraphics[width=\linewidth]{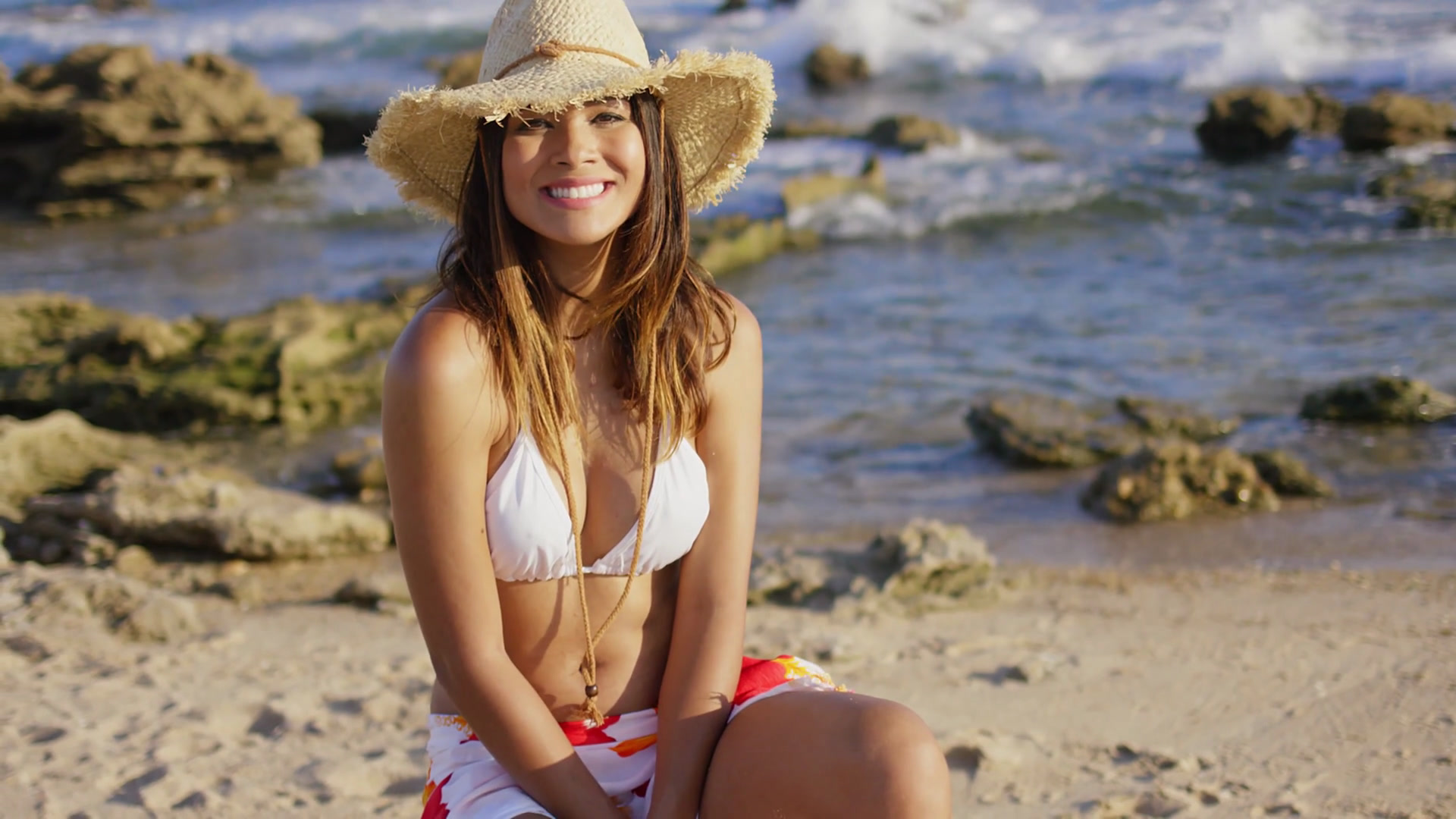} &
        \includegraphics[width=\linewidth]{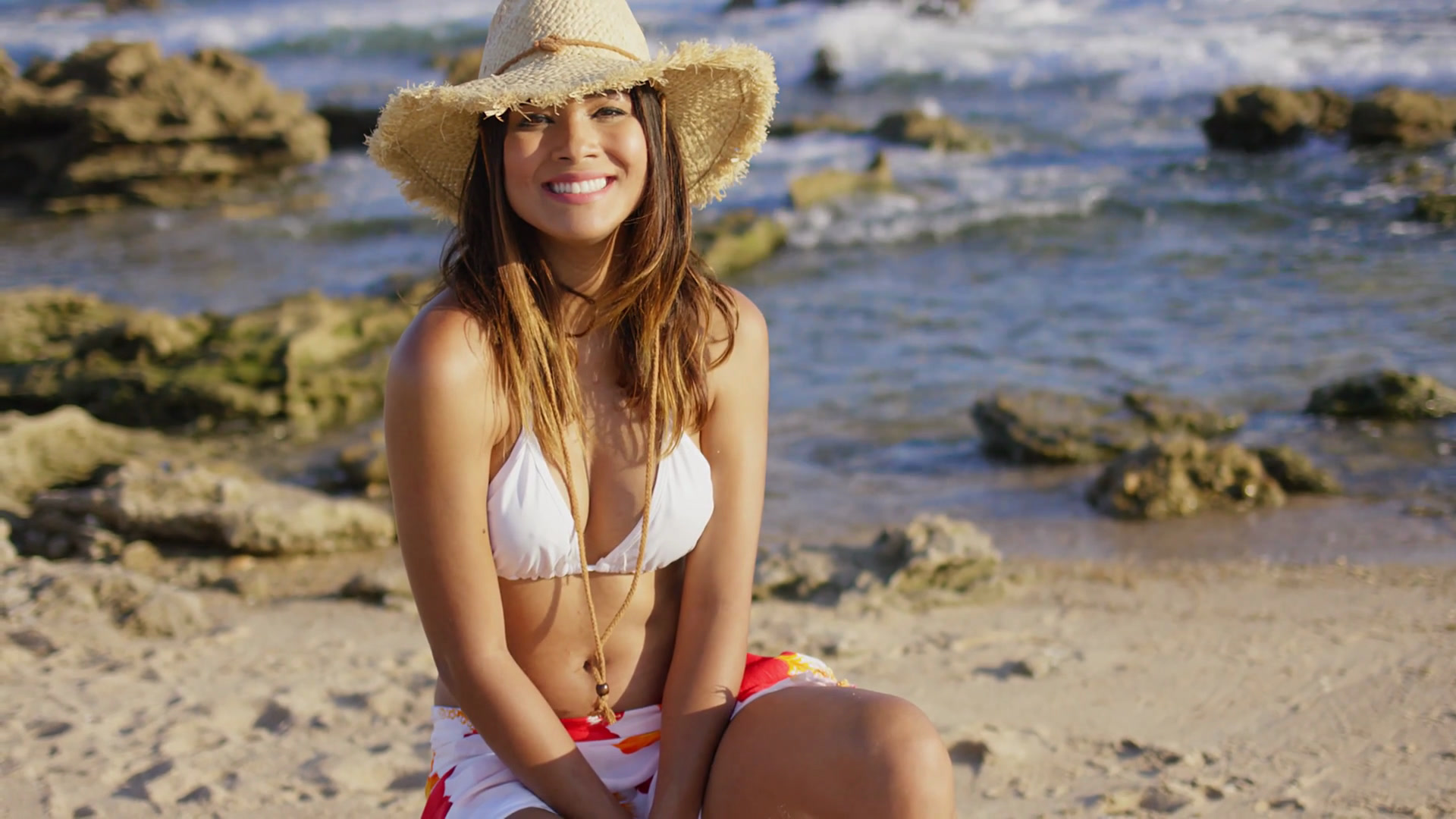} \\[2pt]

        \centering\rotatebox{90}{\textbf{Masked Video}} &
        \includegraphics[width=\linewidth]{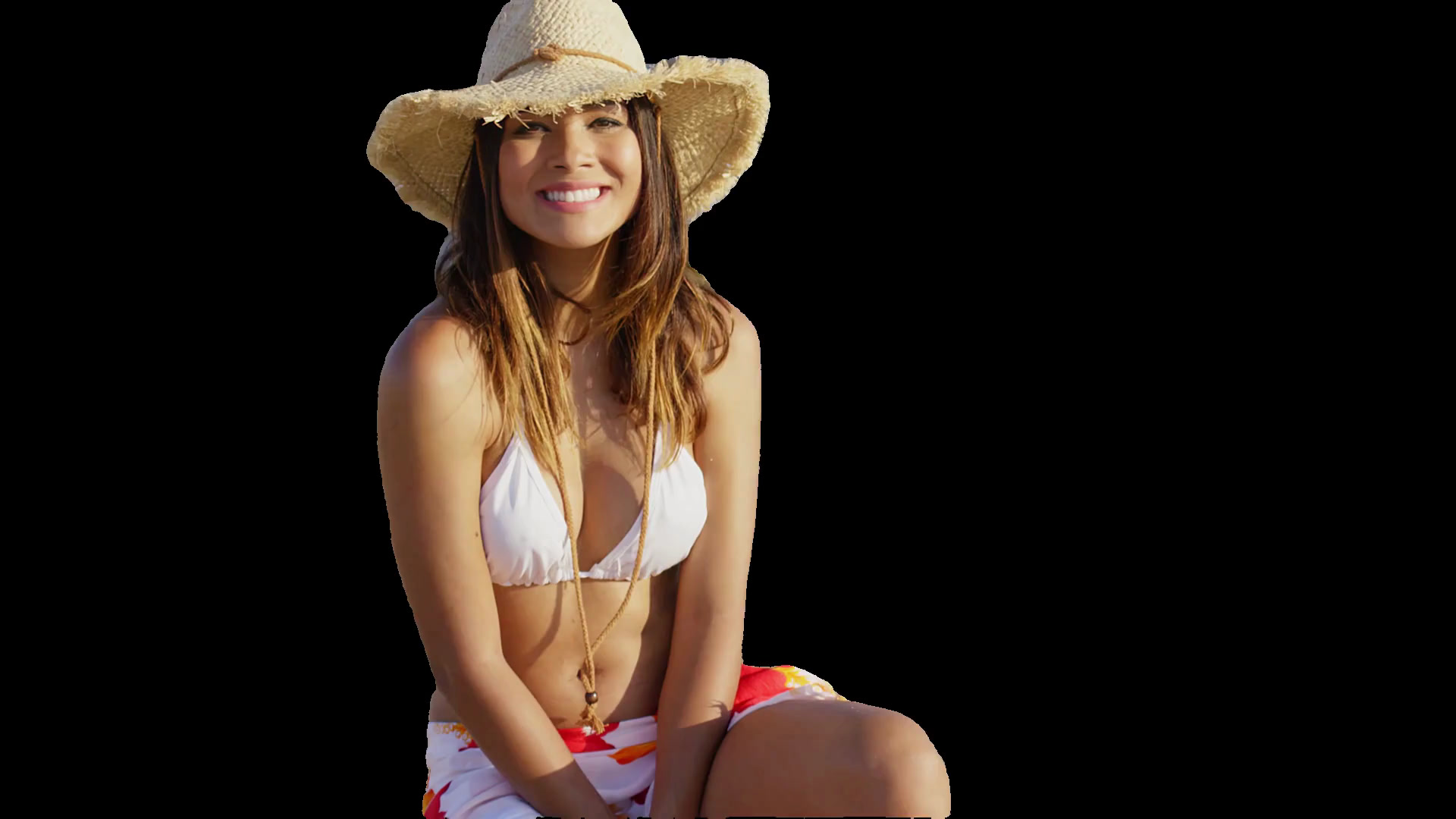} &
        \includegraphics[width=\linewidth]{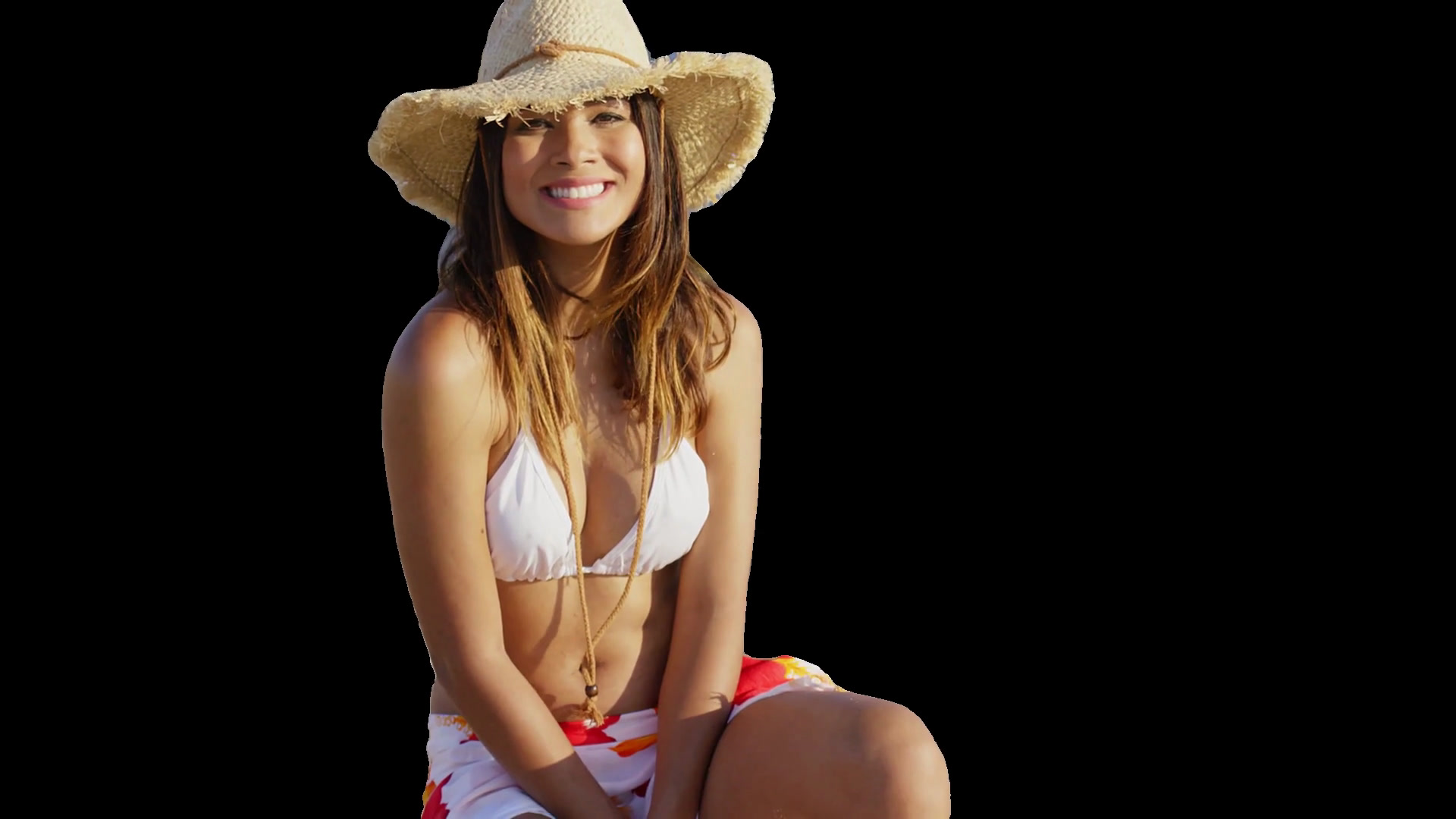} &
        \includegraphics[width=\linewidth]{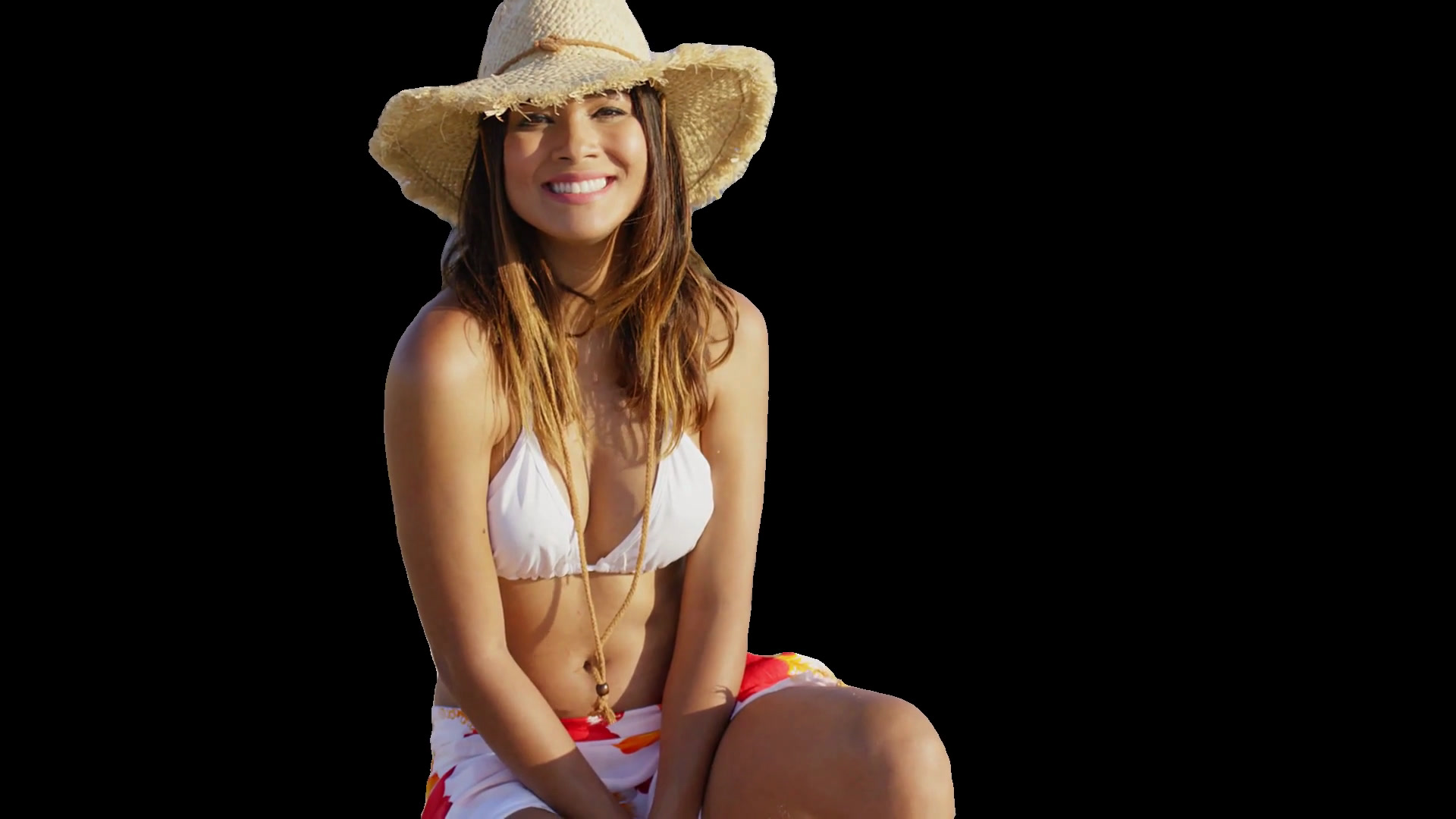} \\[2pt]

        \centering\rotatebox{90}{\textbf{Output Video}} &
        \includegraphics[width=\linewidth]{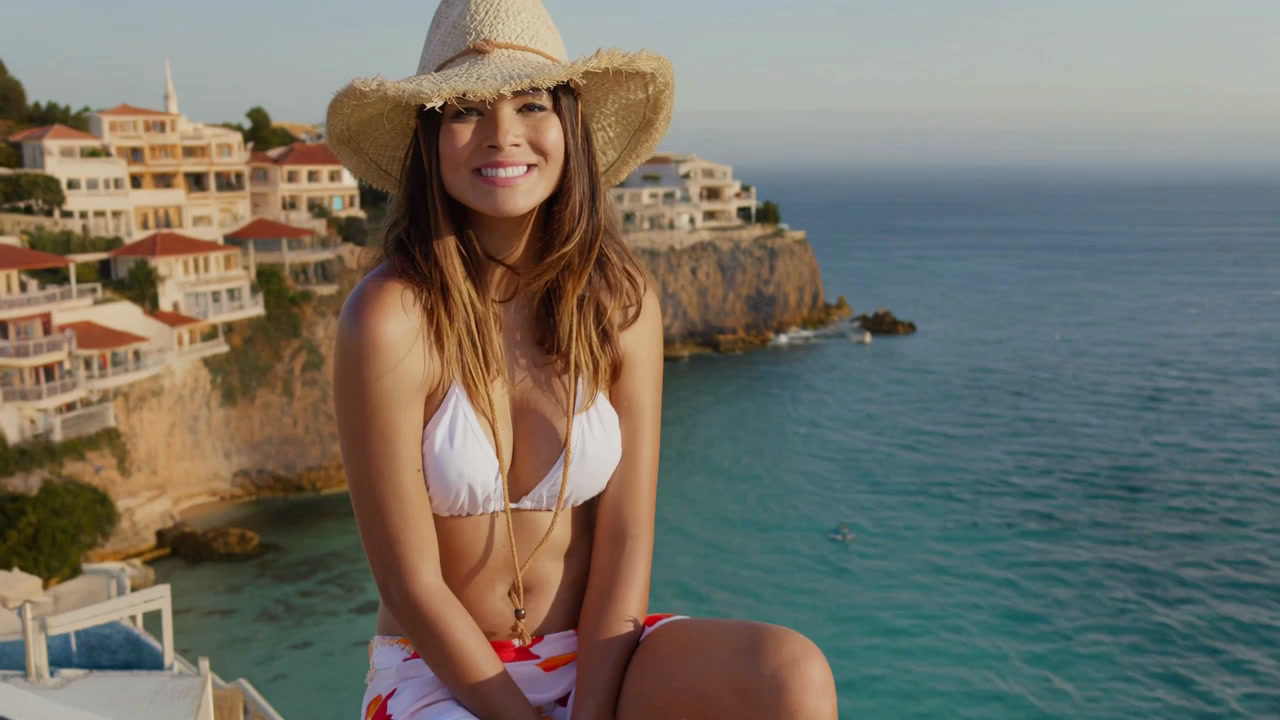} &
        \includegraphics[width=\linewidth]{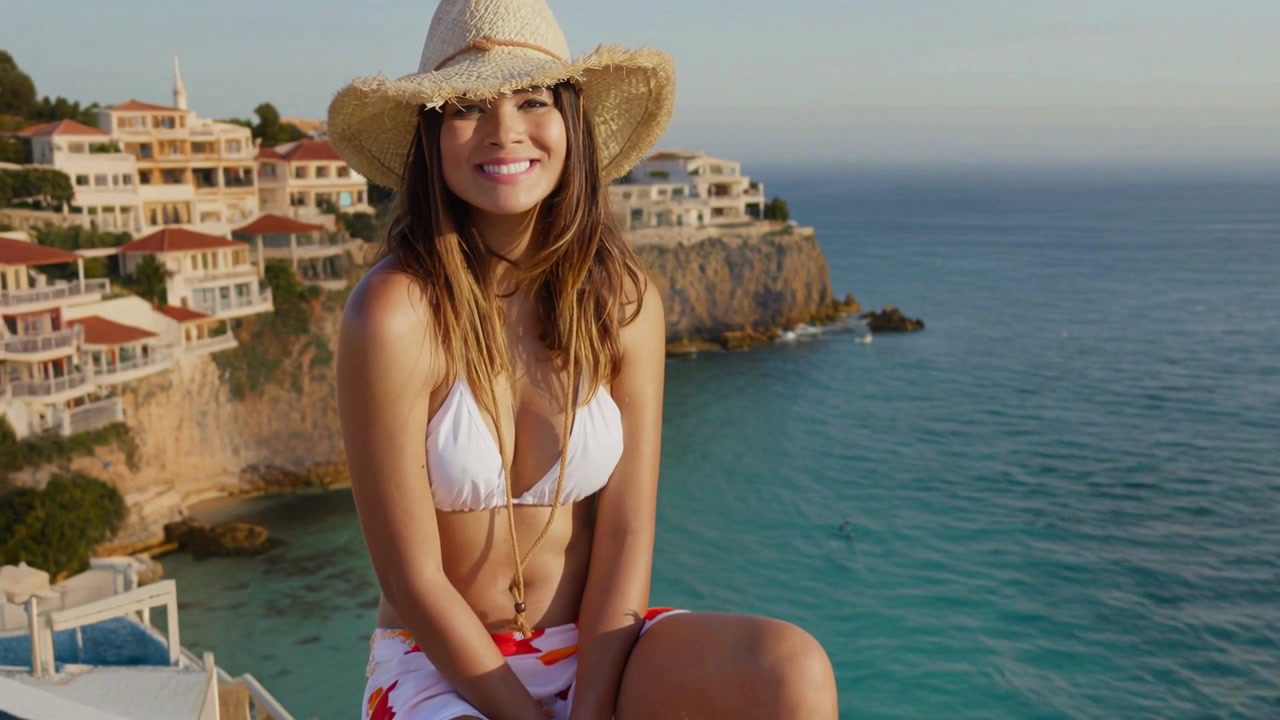} &
        \includegraphics[width=\linewidth]{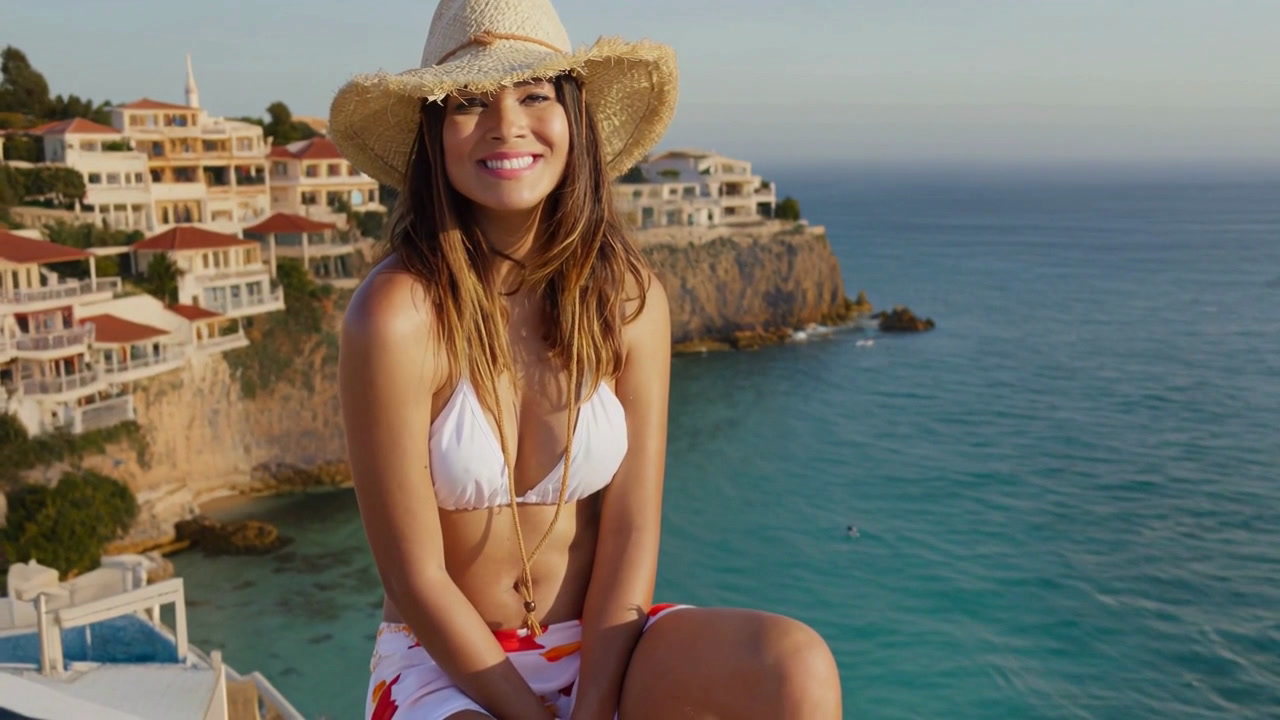} \\
    \end{tabular}
\end{flushleft}


\begin{figure}[h!]
    \centering
    \caption{Examples of subject/attribute/background inpainting.}
    \label{fig:inpaint_object_1}
\end{figure}

\newpage
\subsubsection{Image Reference Inpainting}
\label{appendix:image-reference-inpainting}

The model supports using reference images to guide the inpainting process, ensuring that the inpainted content is consistent with the reference style.

\begin{flushleft}
    \textbf{Instruction:} \textit{Add the man from @image\_1 to the left mask area of @video\_1.}%
    \vspace{0.5em}
    \begin{tabular}{m{1.2em} m{0.22\linewidth} m{0.22\linewidth} m{0.22\linewidth}}

        \centering\rotatebox{90}{\textbf{Ref. Image}} &
        \includegraphics[width=\linewidth]{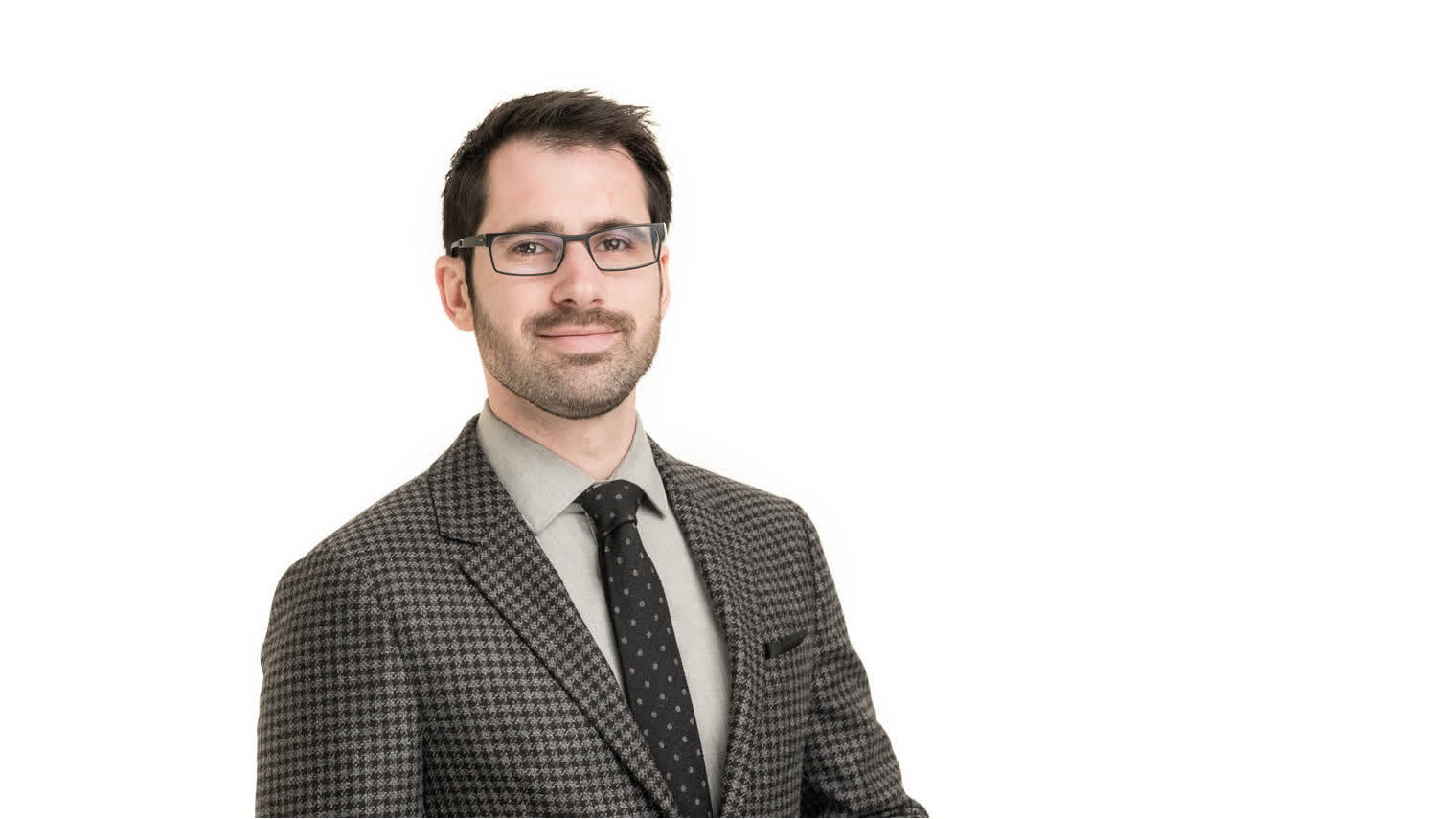} & & \\[2pt]

        \centering\rotatebox{90}{\textbf{Input Video}} &
        \includegraphics[width=\linewidth]{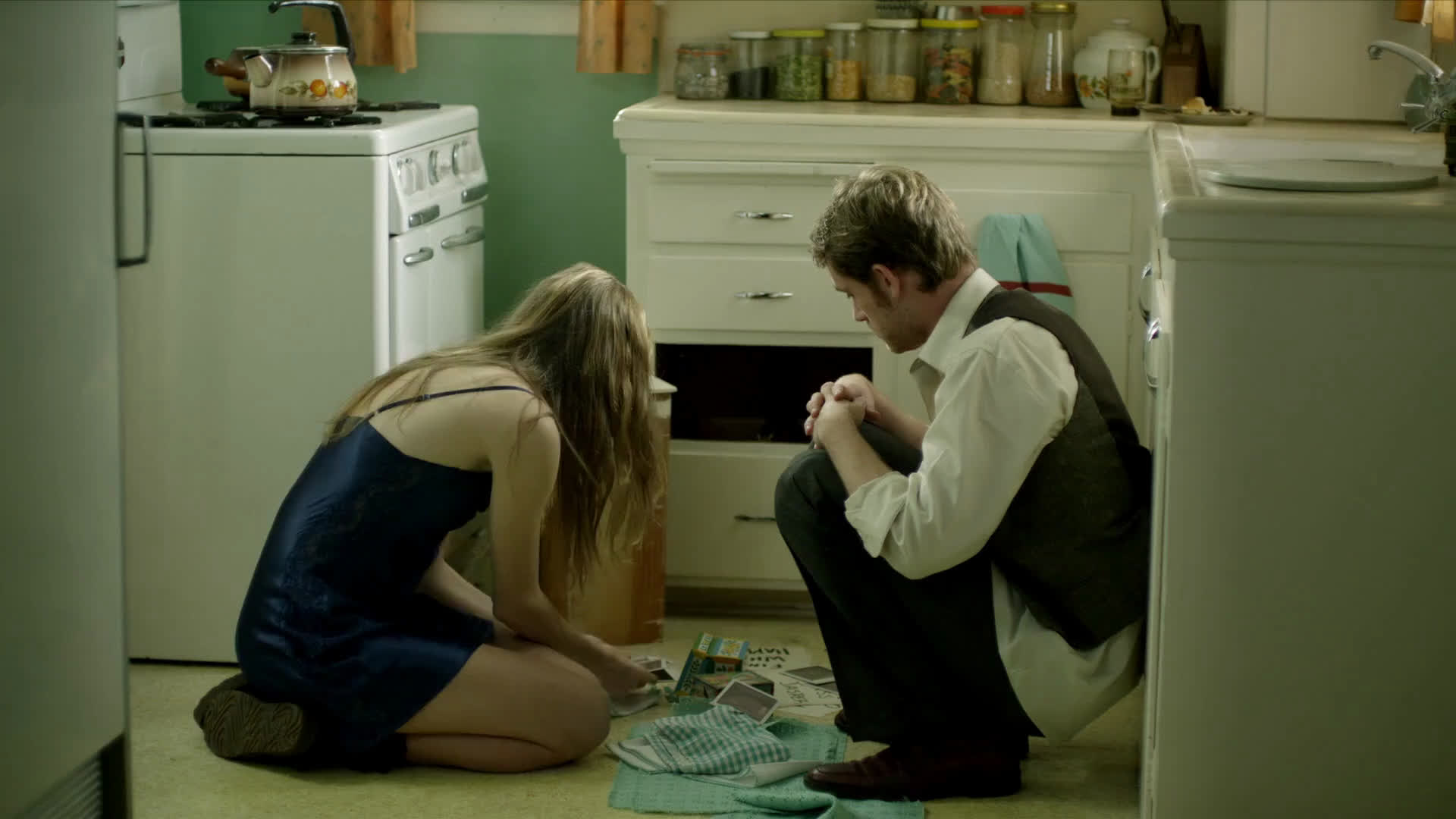} &
        \includegraphics[width=\linewidth]{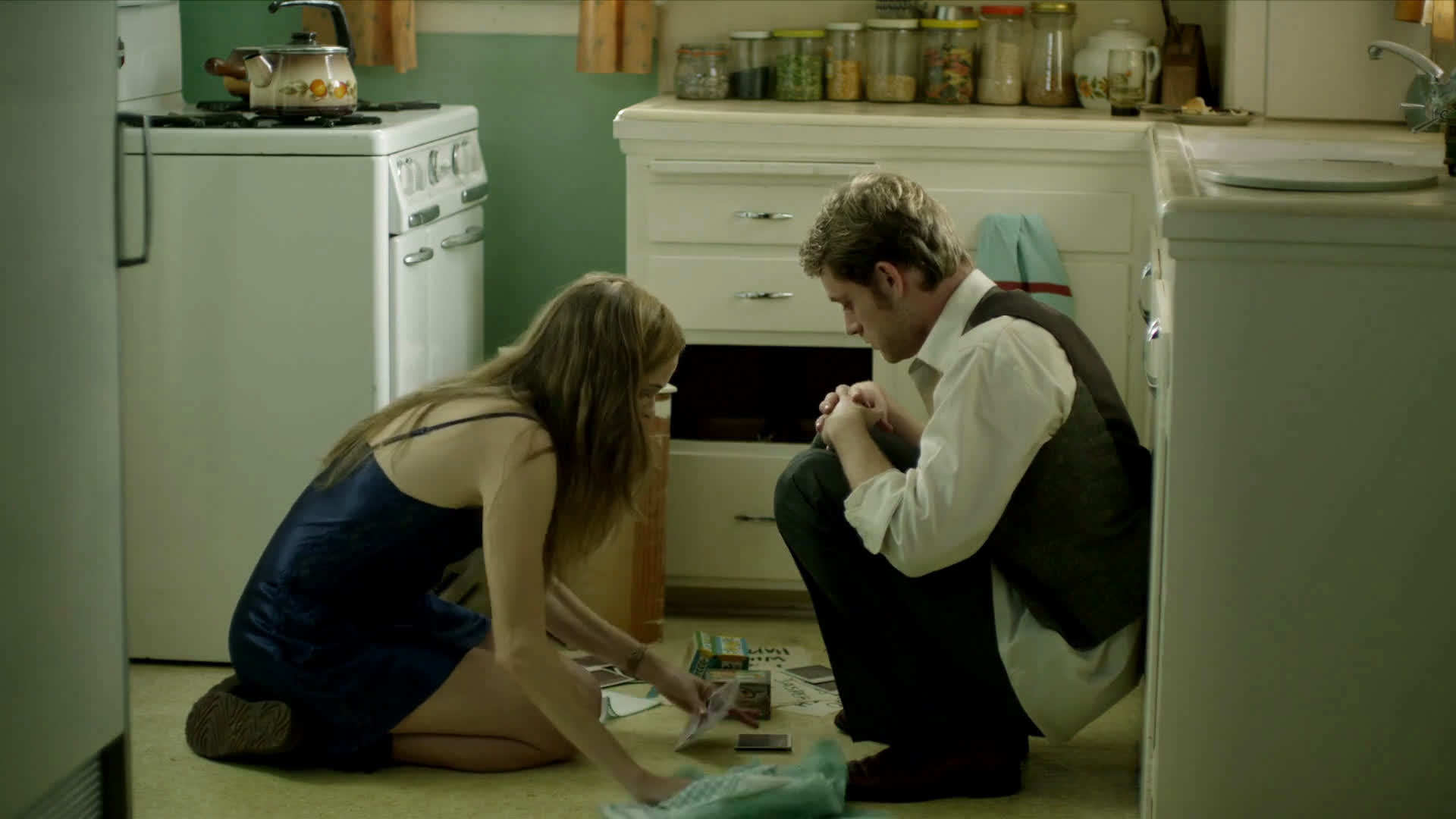} &
        \includegraphics[width=\linewidth]{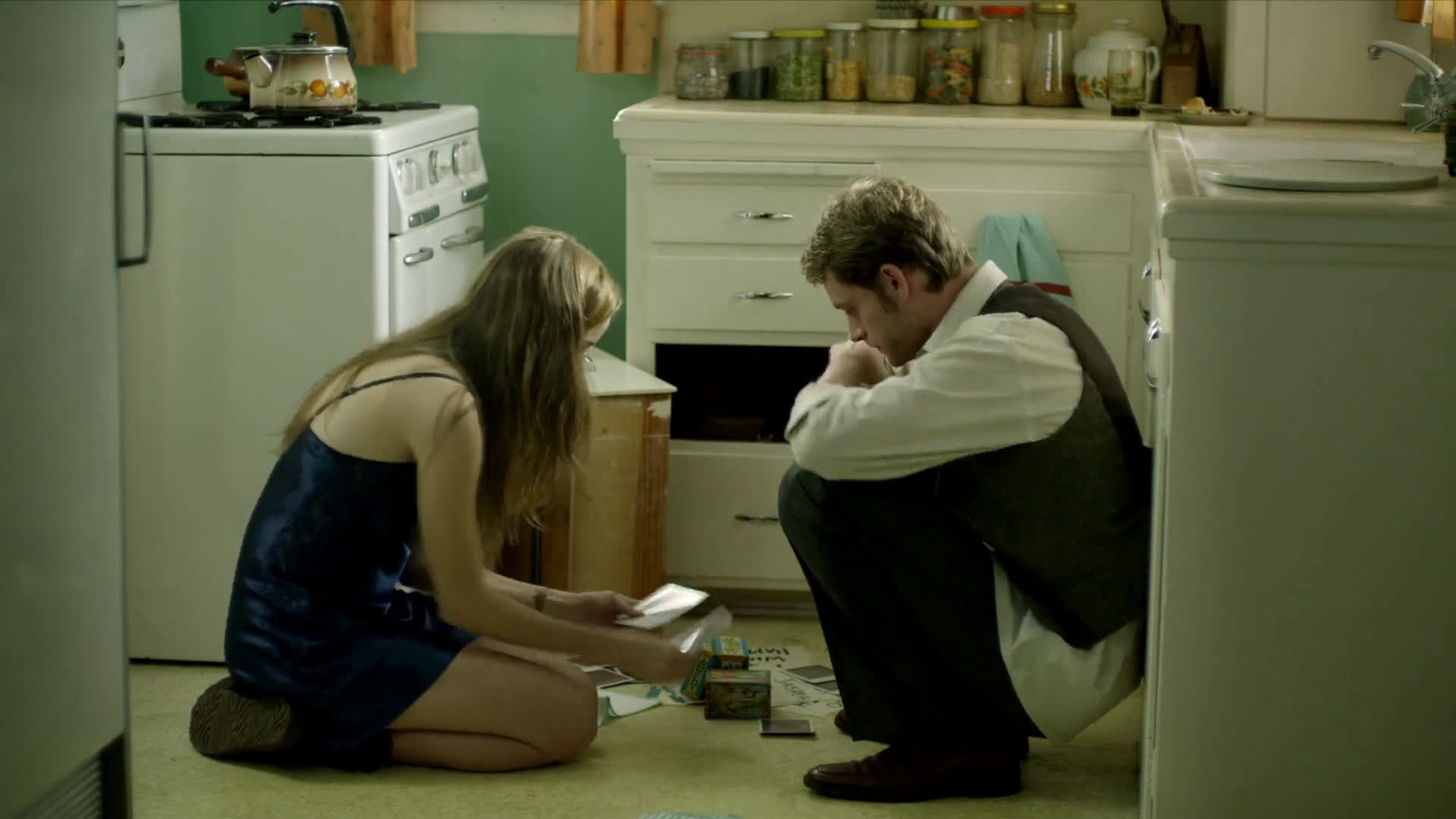} \\[2pt]

        \centering\rotatebox{90}{\textbf{Masked Video}} &
        \includegraphics[width=\linewidth]{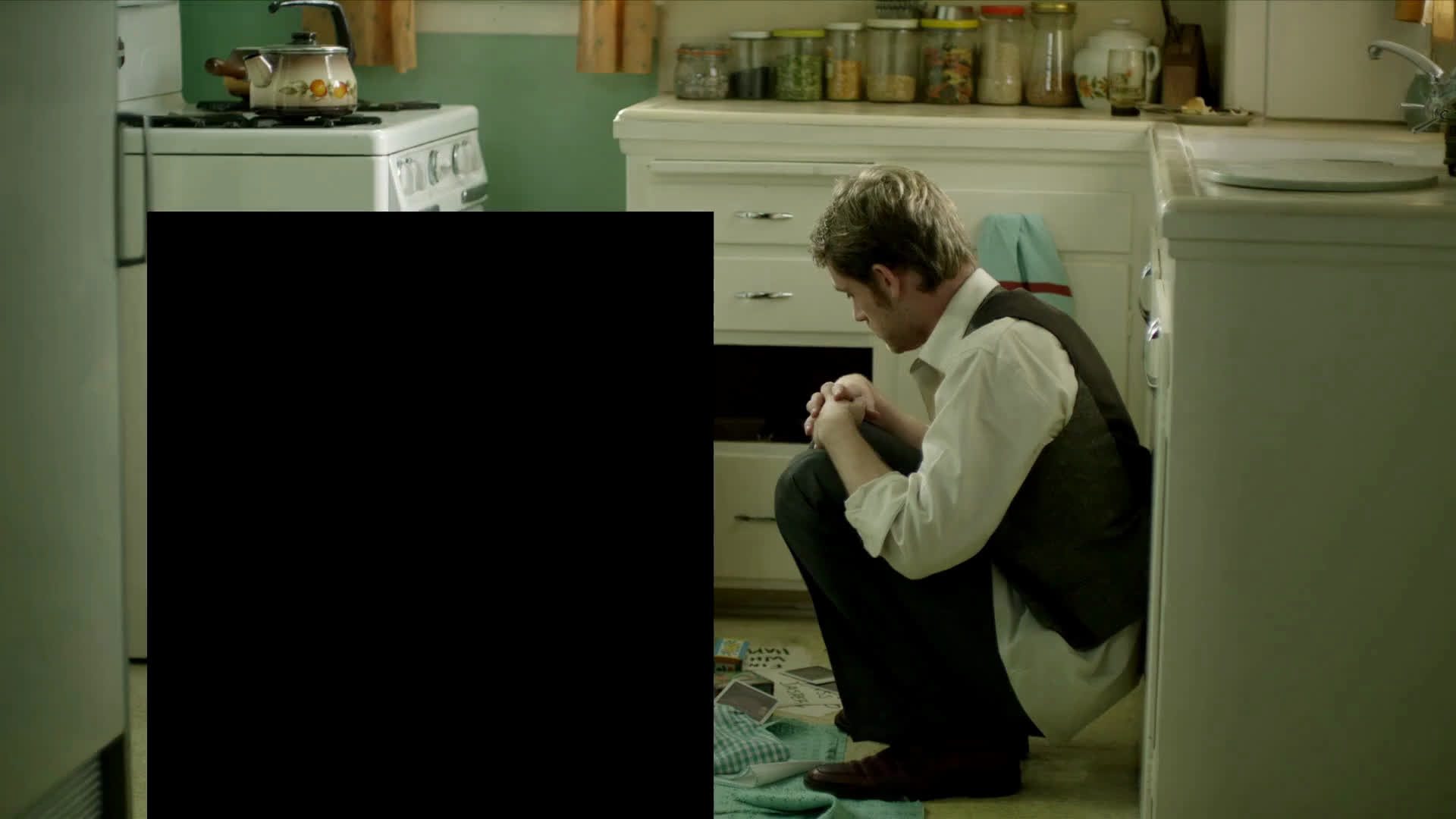} &
        \includegraphics[width=\linewidth]{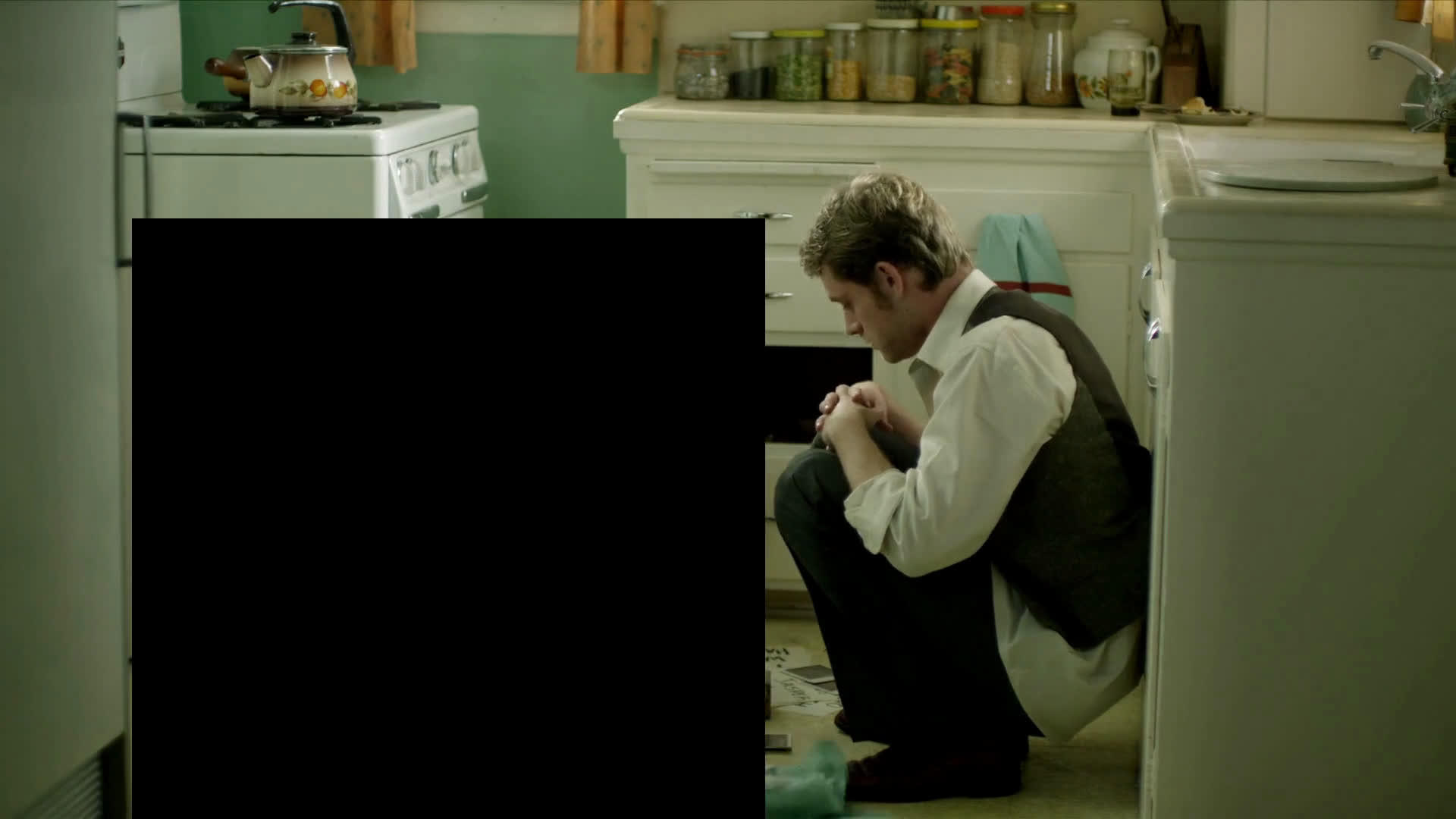} &
        \includegraphics[width=\linewidth]{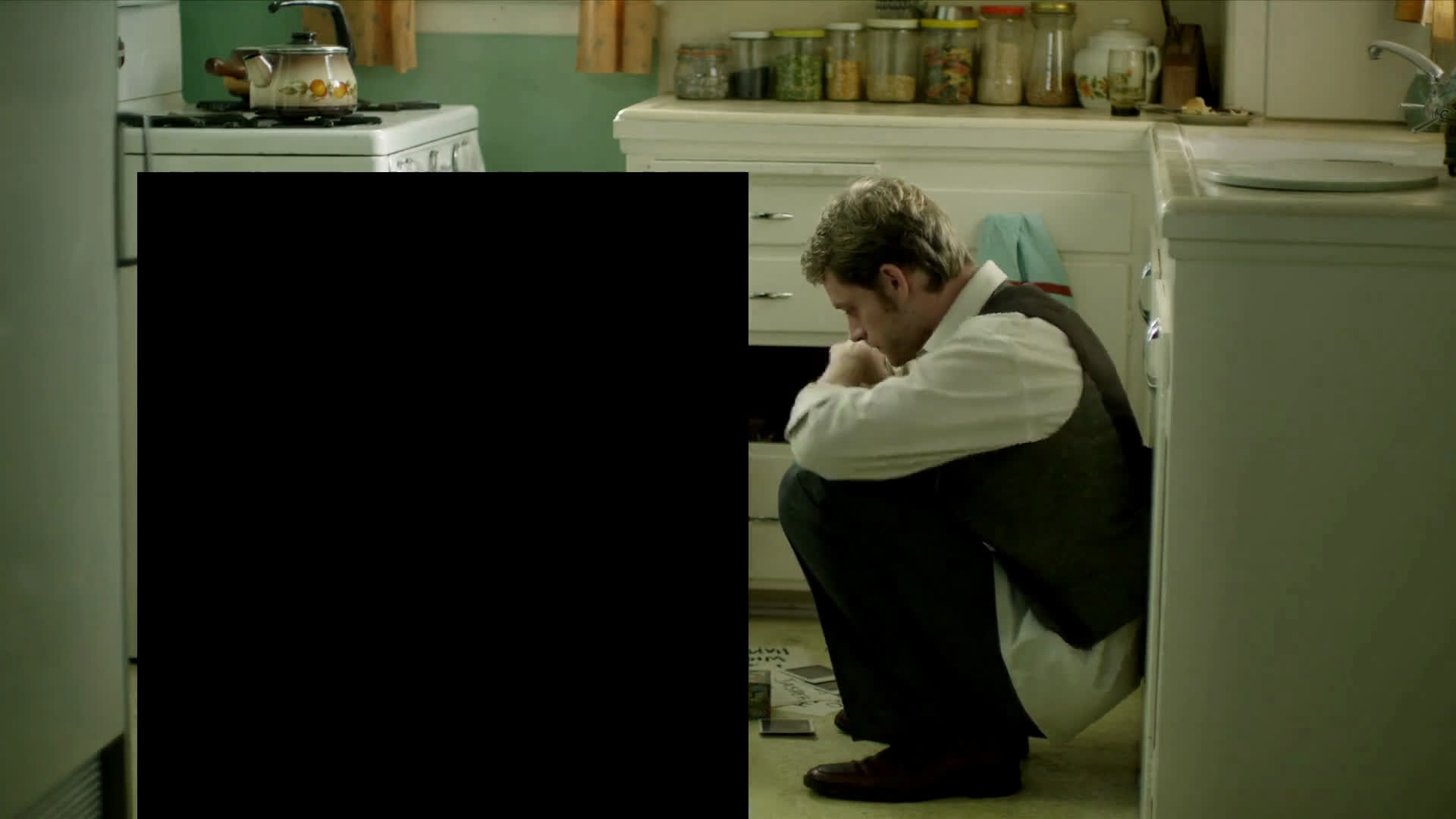} \\[2pt]

        \centering\rotatebox{90}{\textbf{Output Video}} &
        \includegraphics[width=\linewidth]{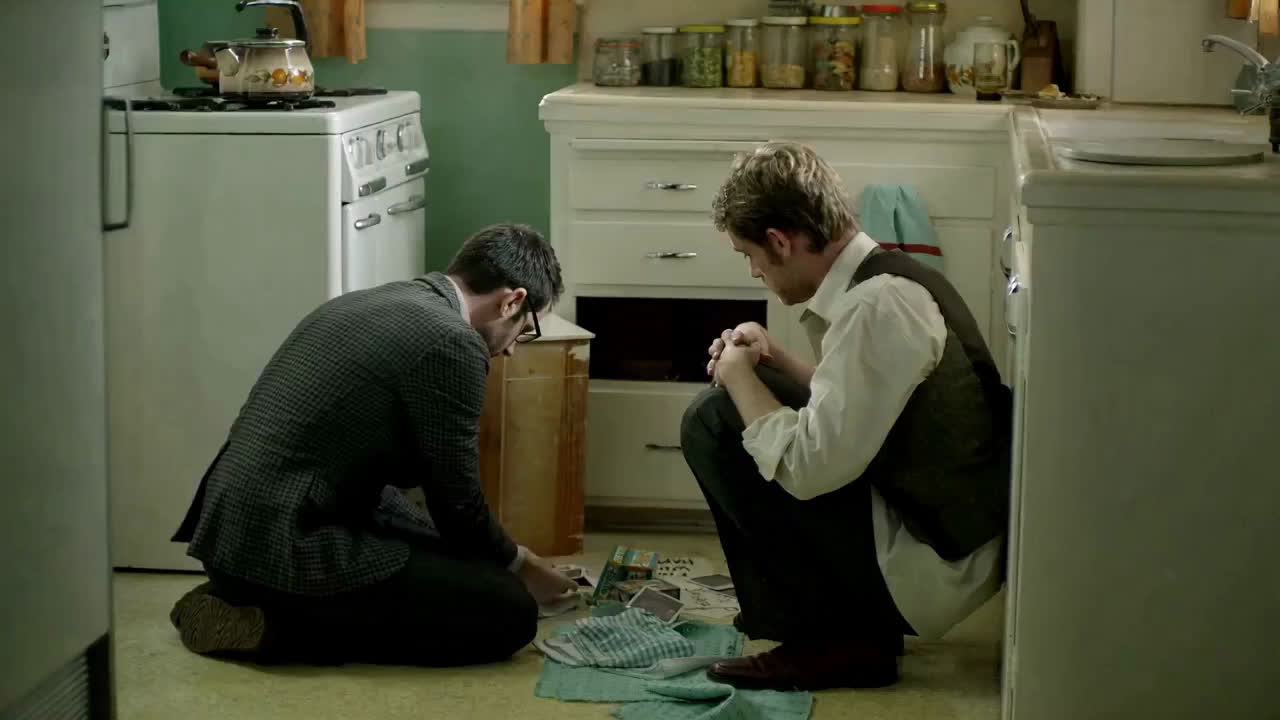} &
        \includegraphics[width=\linewidth]{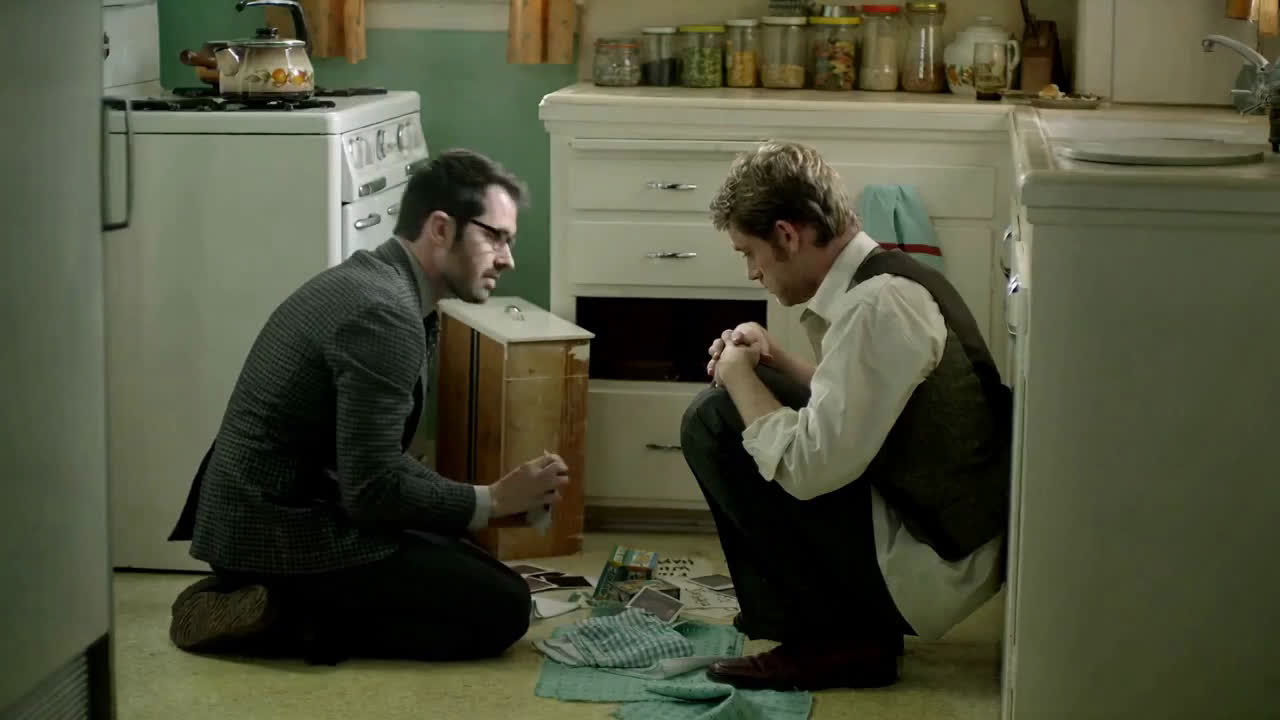} &
        \includegraphics[width=\linewidth]{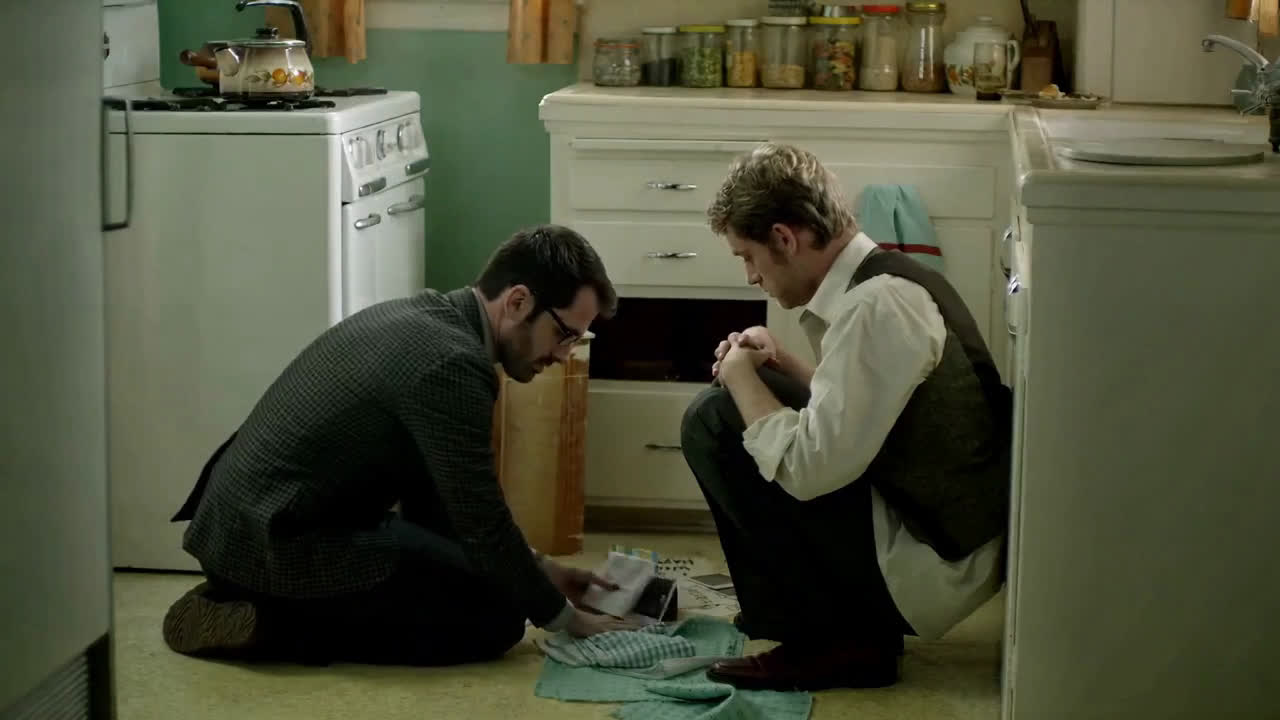} \\
    \end{tabular}
\end{flushleft}



\begin{flushleft}
    \textbf{Instruction:} \textit{Replace the right mask area in @video\_1 with the cat from @image\_1 and the left mask area in @video\_1 with the woman from @image\_2, ensuring a harmonious and natural scene.}%
    \vspace{0.5em}
    \begin{tabular}{m{1.2em} m{0.22\linewidth} m{0.22\linewidth} m{0.22\linewidth}}
        
        \centering\rotatebox{90}{\textbf{Ref. Image}} &
        \includegraphics[width=\linewidth]{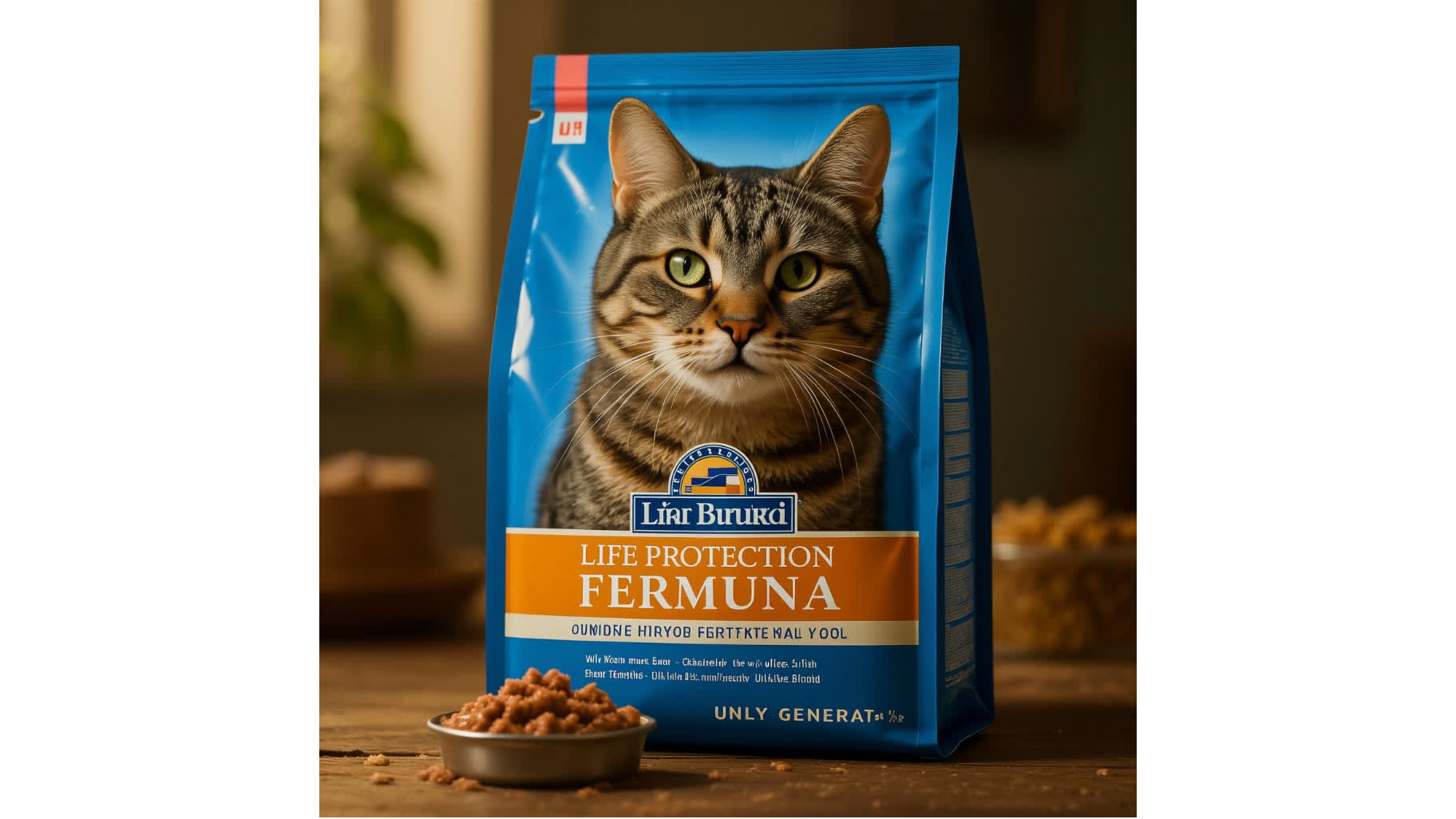} &
        \includegraphics[width=\linewidth]{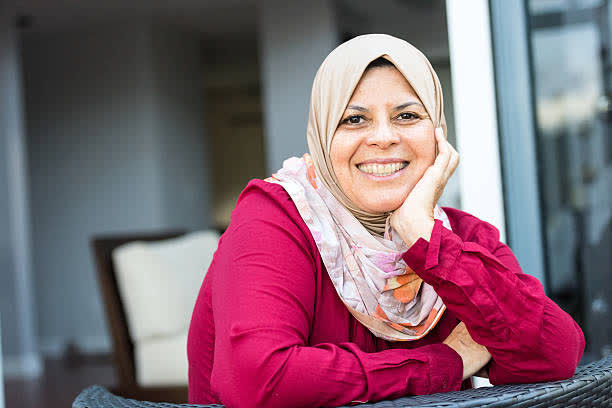} & \\[2pt]

        \centering\rotatebox{90}{\textbf{Input Video}} &
        \includegraphics[width=\linewidth]{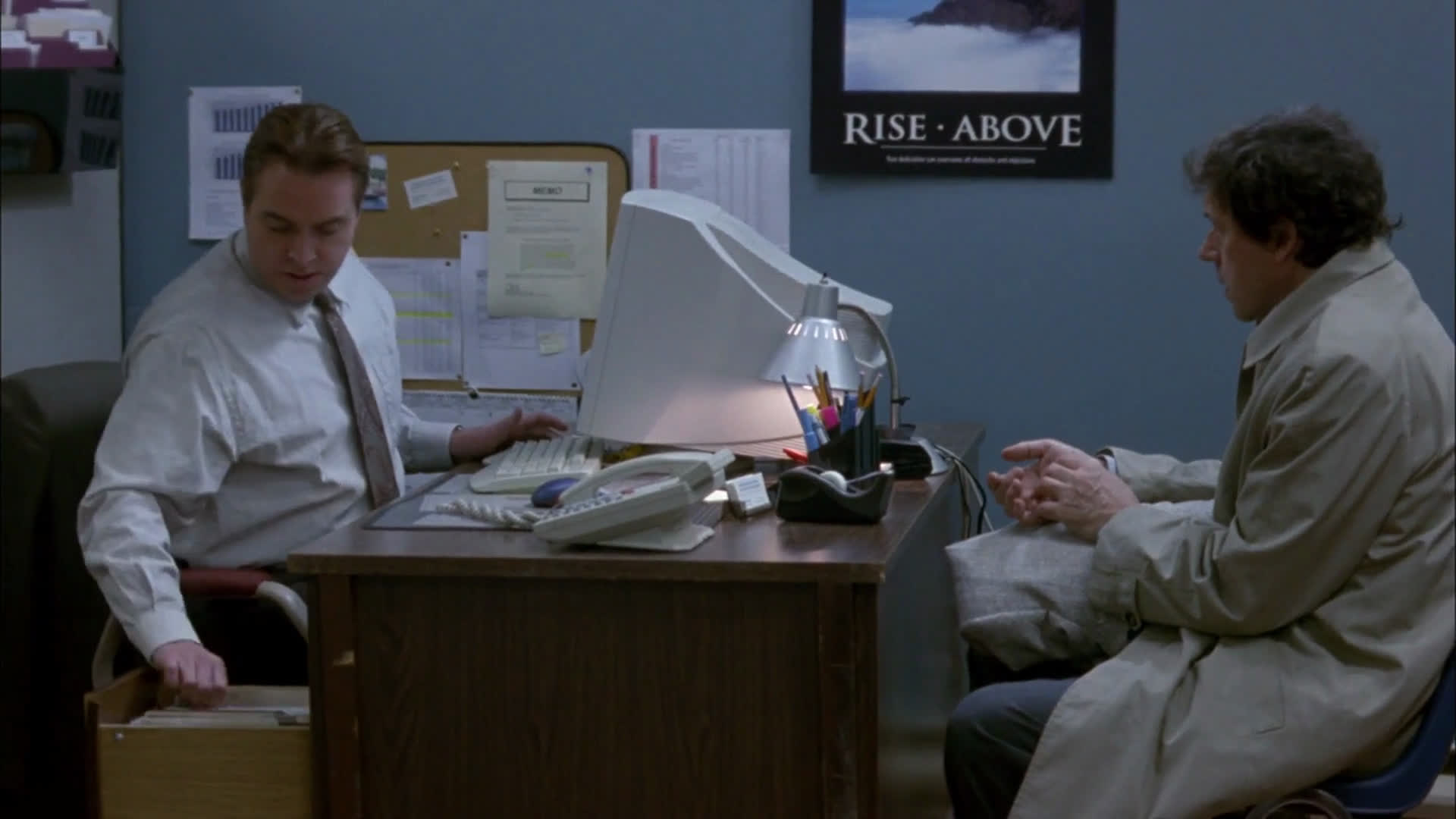} &
        \includegraphics[width=\linewidth]{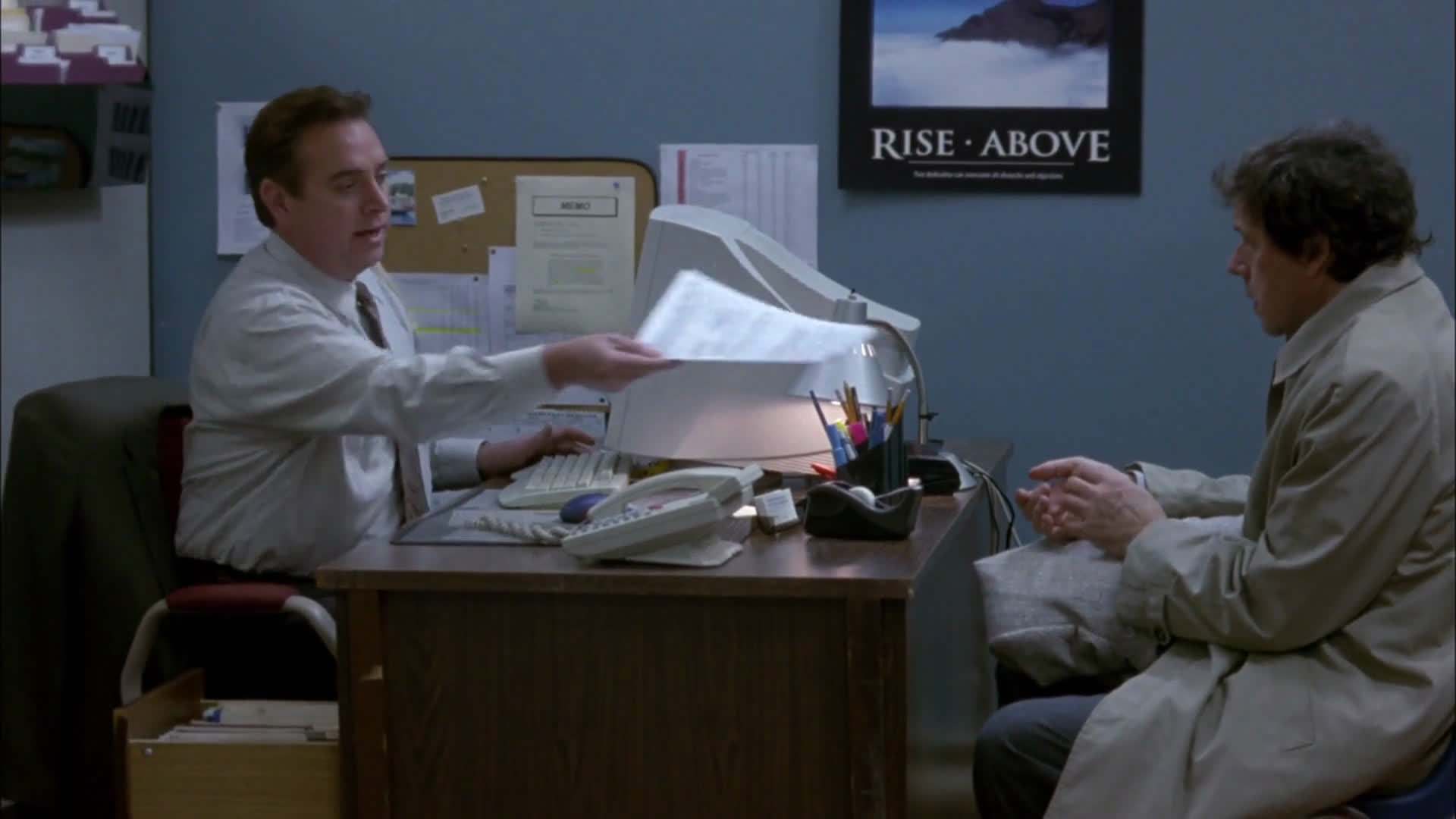} &
        \includegraphics[width=\linewidth]{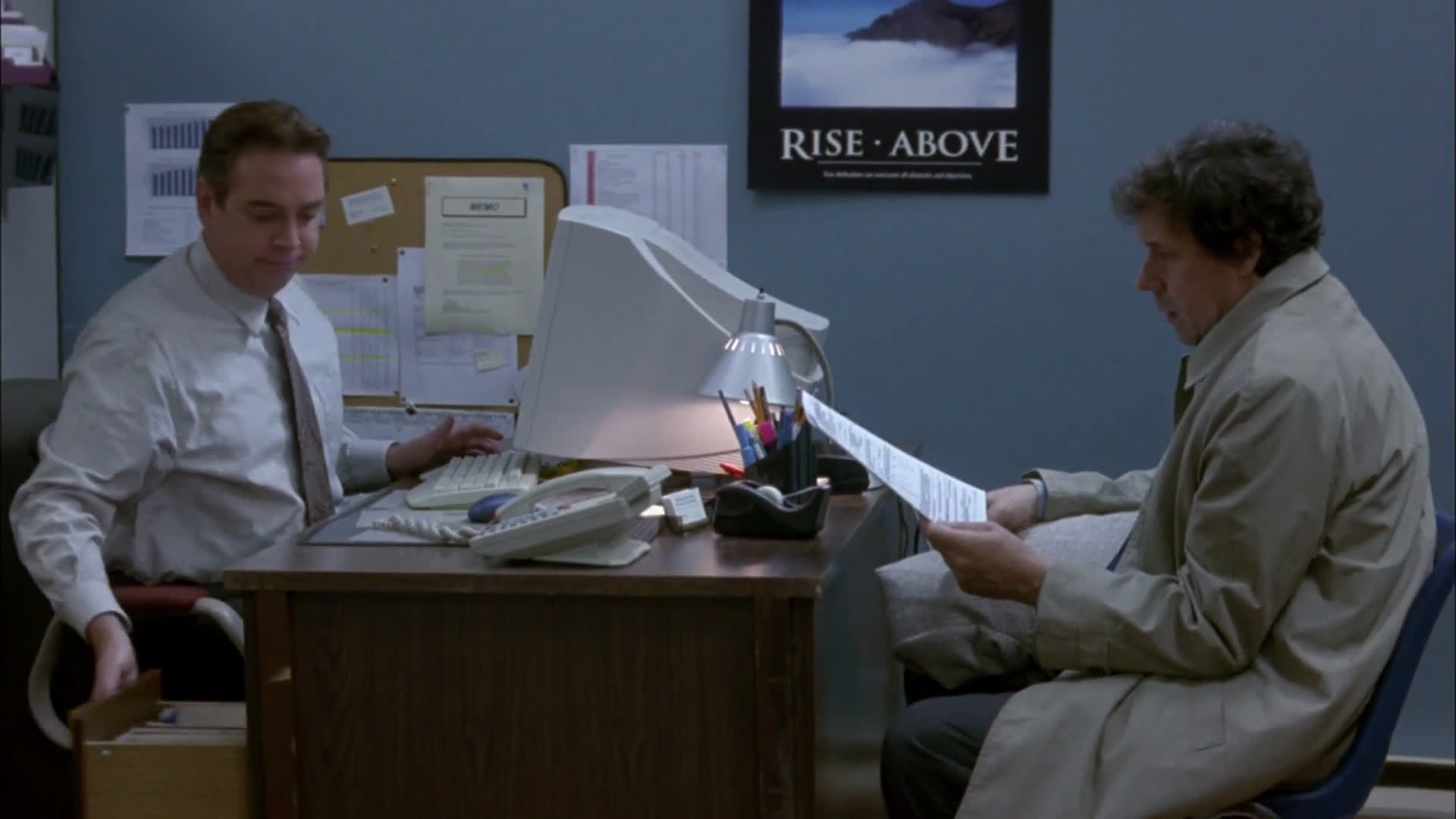} \\[2pt]

        \centering\rotatebox{90}{\textbf{Masked Video}} &
        \includegraphics[width=\linewidth]{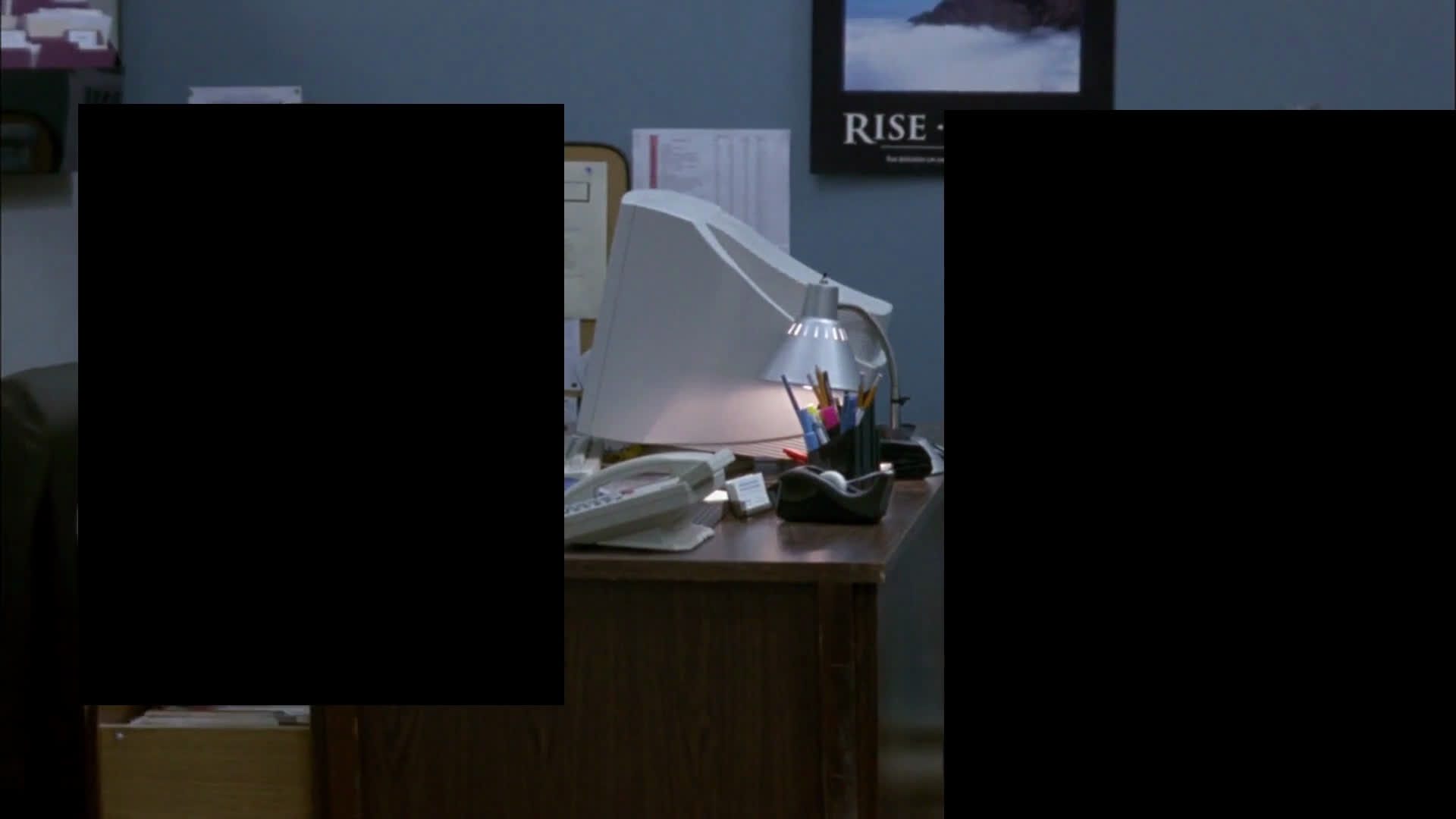} &
        \includegraphics[width=\linewidth]{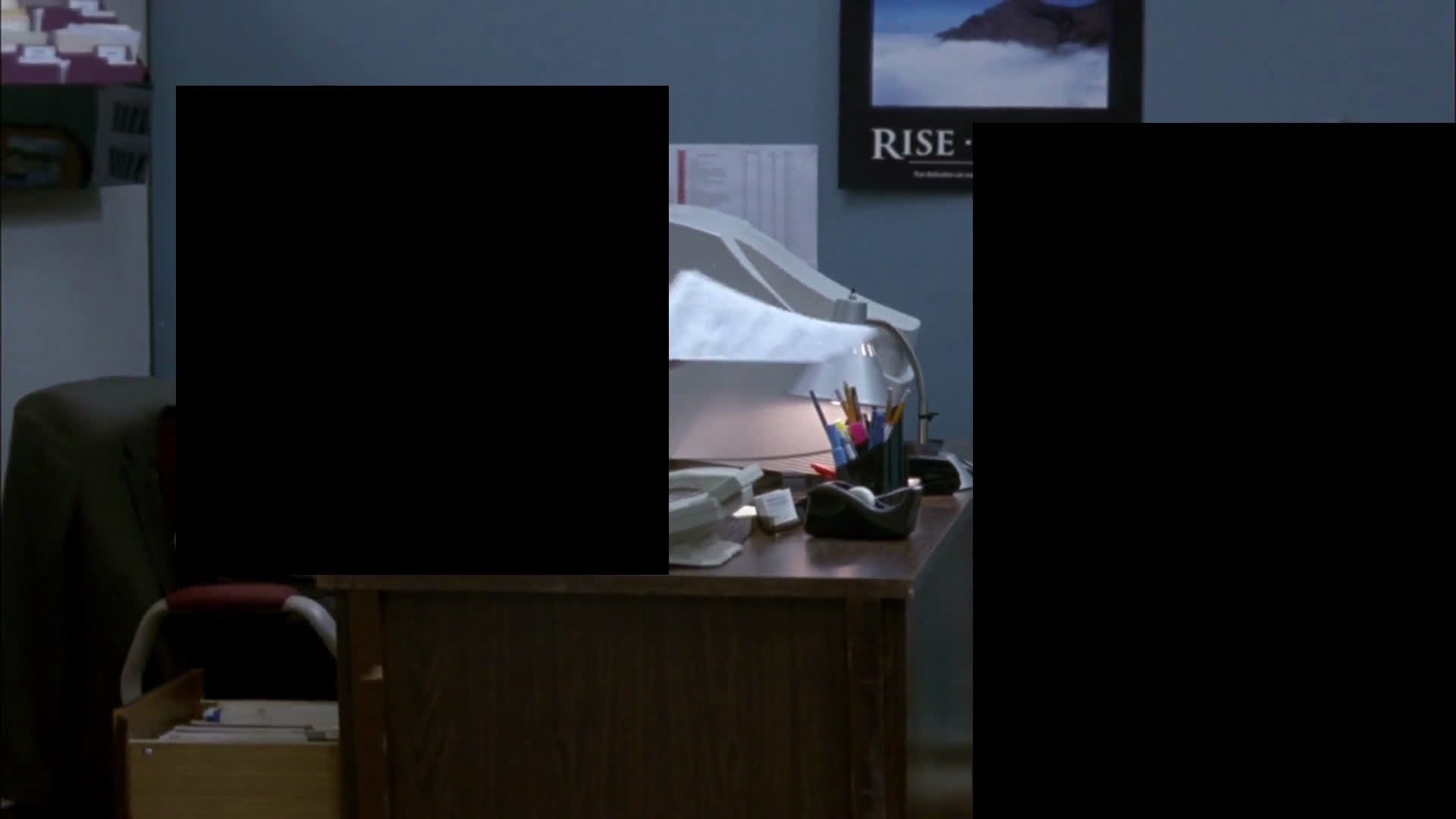} &
        \includegraphics[width=\linewidth]{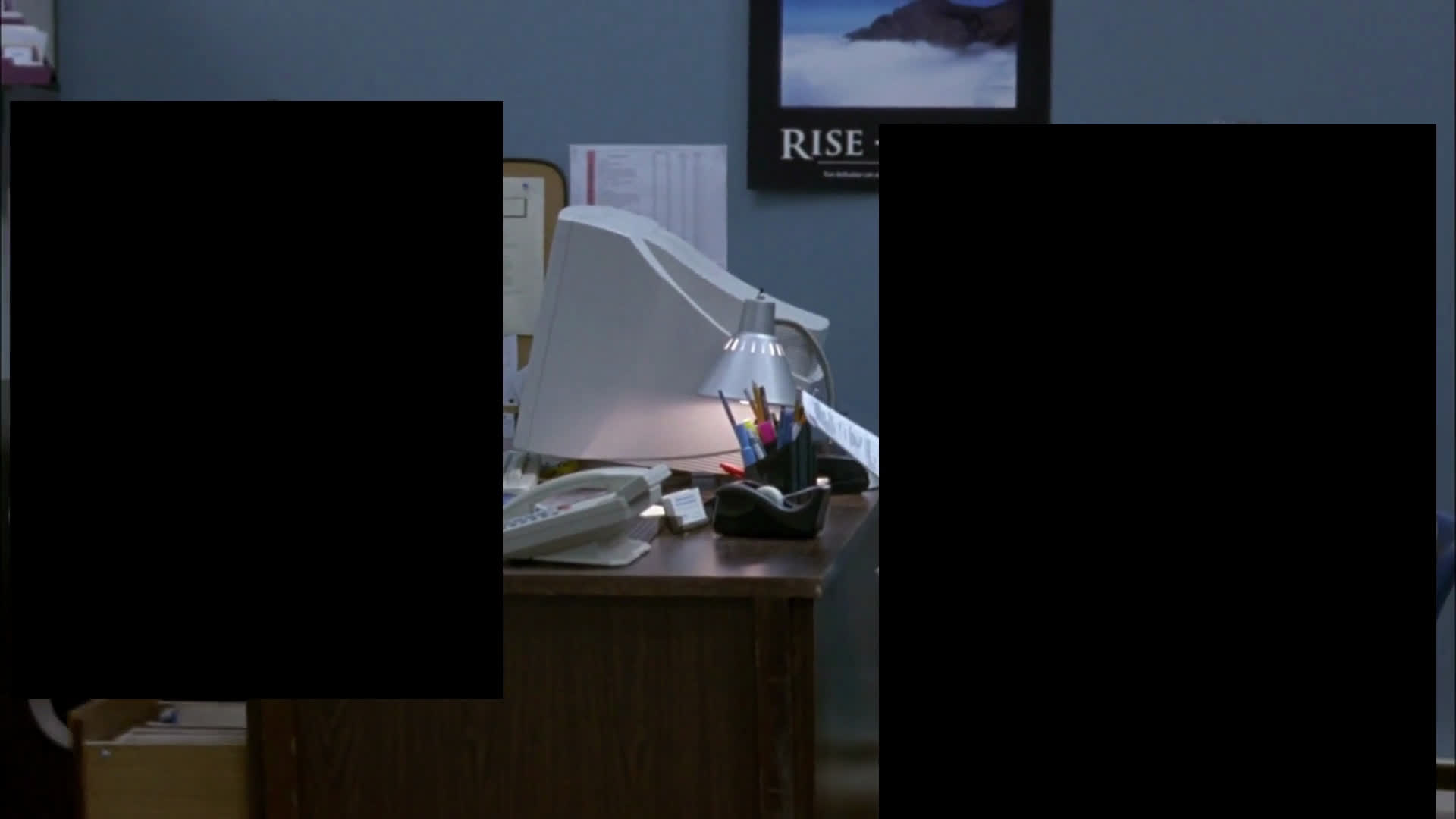} \\[2pt]

        \centering\rotatebox{90}{\textbf{Output Video}} &
        \includegraphics[width=\linewidth]{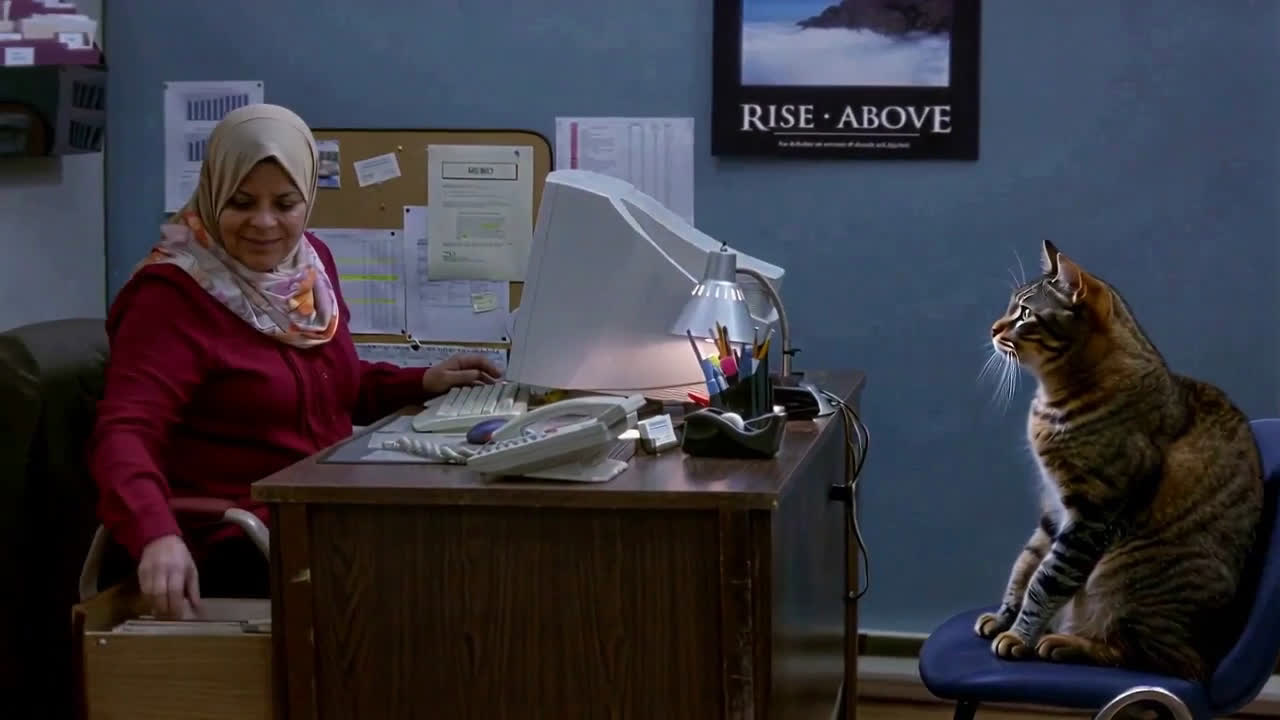} &
        \includegraphics[width=\linewidth]{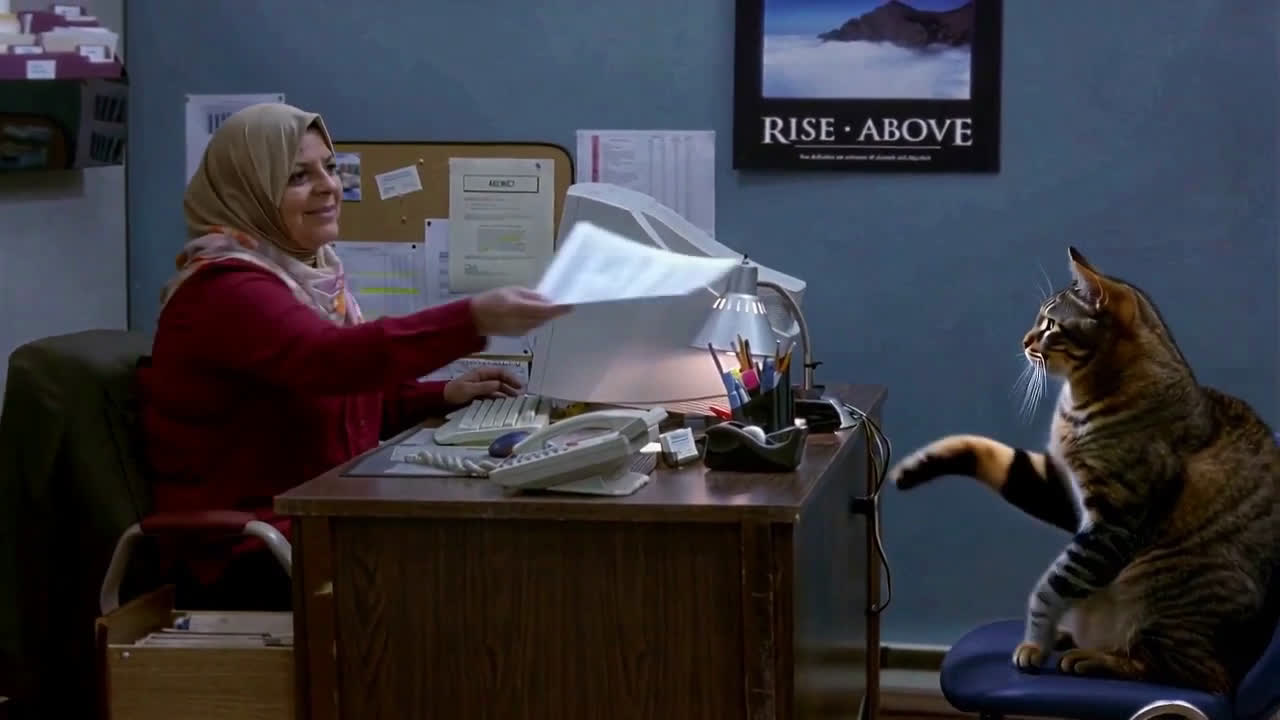} &
        \includegraphics[width=\linewidth]{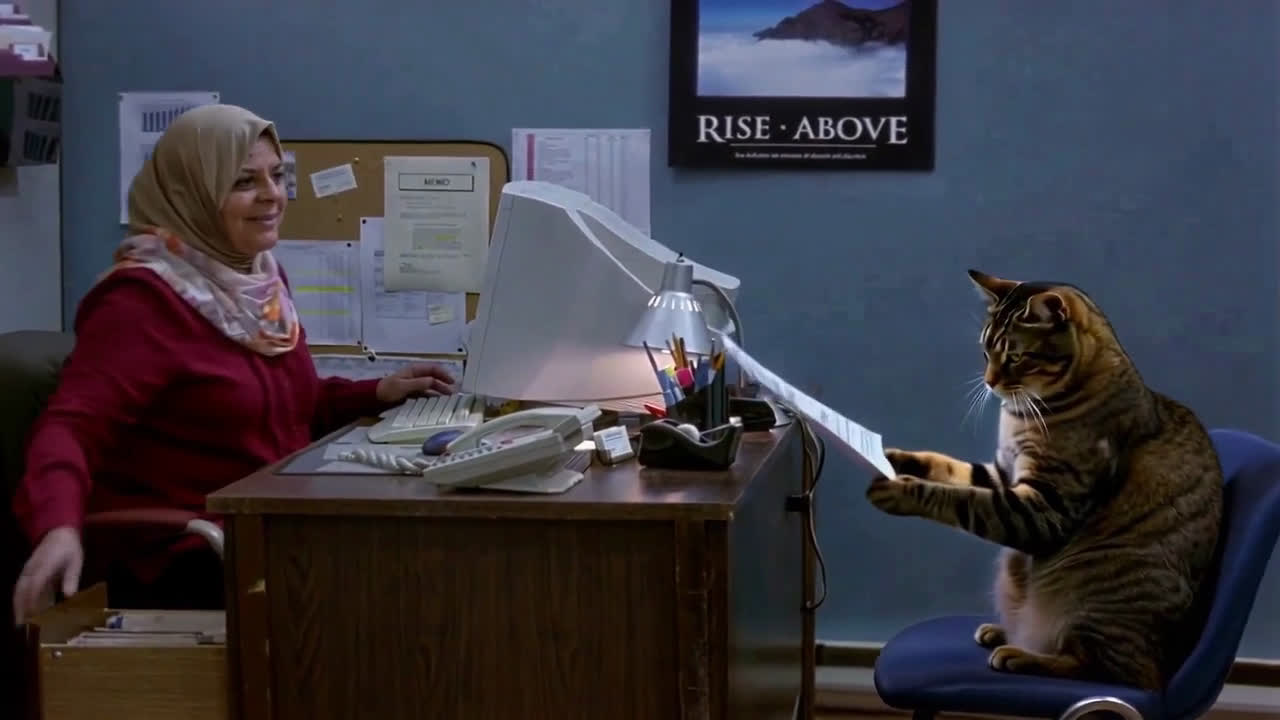} \\
    \end{tabular}
\end{flushleft}


\begin{figure}[h!]
    \centering
    \caption{Examples of image reference inpainting.}
    \label{fig:inpaint_img_ref_2}
\end{figure}

\newpage
\subsection{Video Editing}
\label{appendix:editing}

\subsubsection{Local Editing}
\label{appendix:local-editing}

The model enables fine-grained local video editing: subject, attribute, and element edits.

\paragraph{Watermark/Subtitle/Logo Removal} 
\label{appendix:removal}

The model can intelligently identify and remove watermarks, subtitles, logos, and other elements from videos while maintaining content coherence and naturalness.

\begin{flushleft}
    \textbf{Instruction:} \textit{Remove watermarks in @video\_1.}
    \begin{tabular}{m{1.2em} m{0.22\linewidth} m{0.22\linewidth} m{0.22\linewidth}}

        \centering\rotatebox{90}{\textbf{Input Video}} &
        \includegraphics[width=\linewidth]{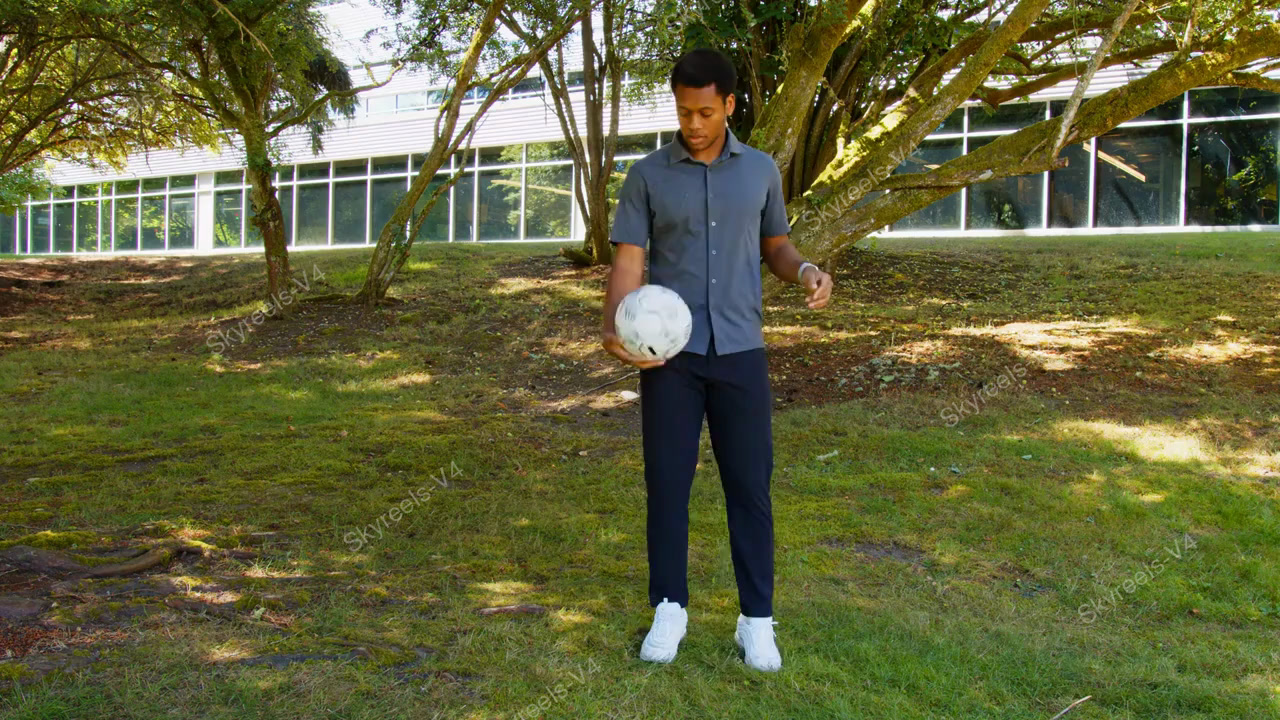} &
        \includegraphics[width=\linewidth]{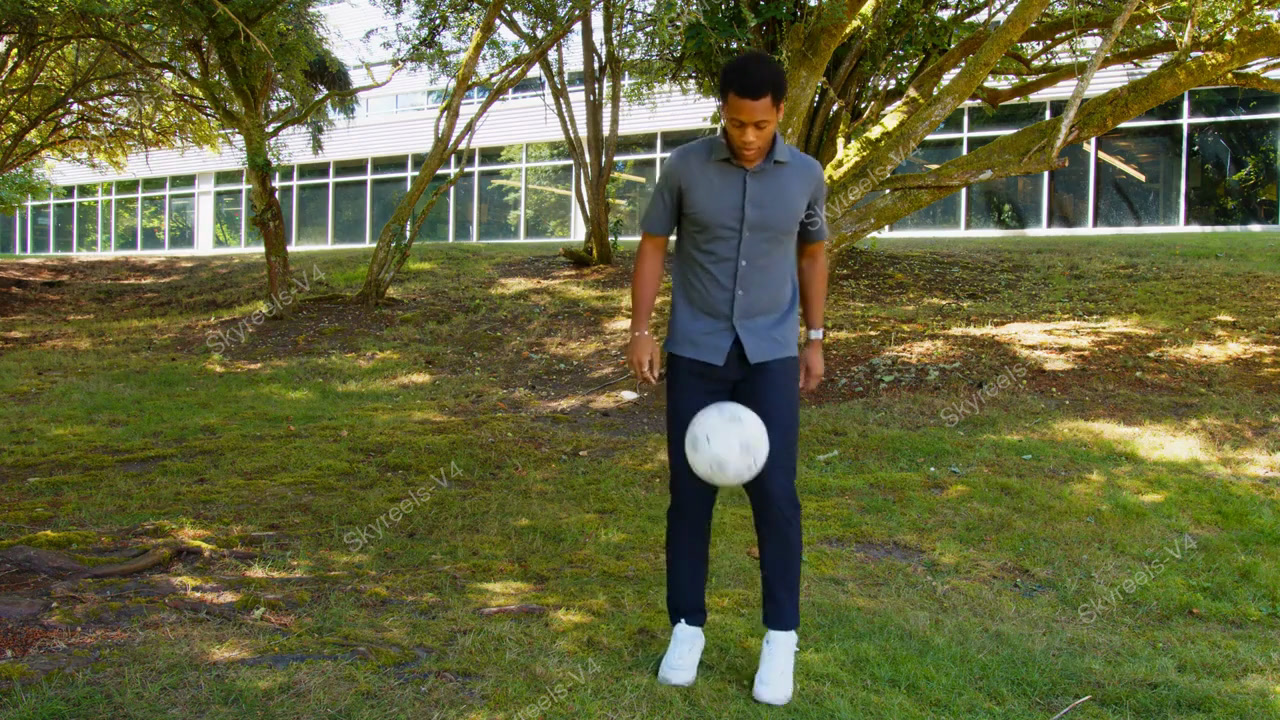} &
        \includegraphics[width=\linewidth]{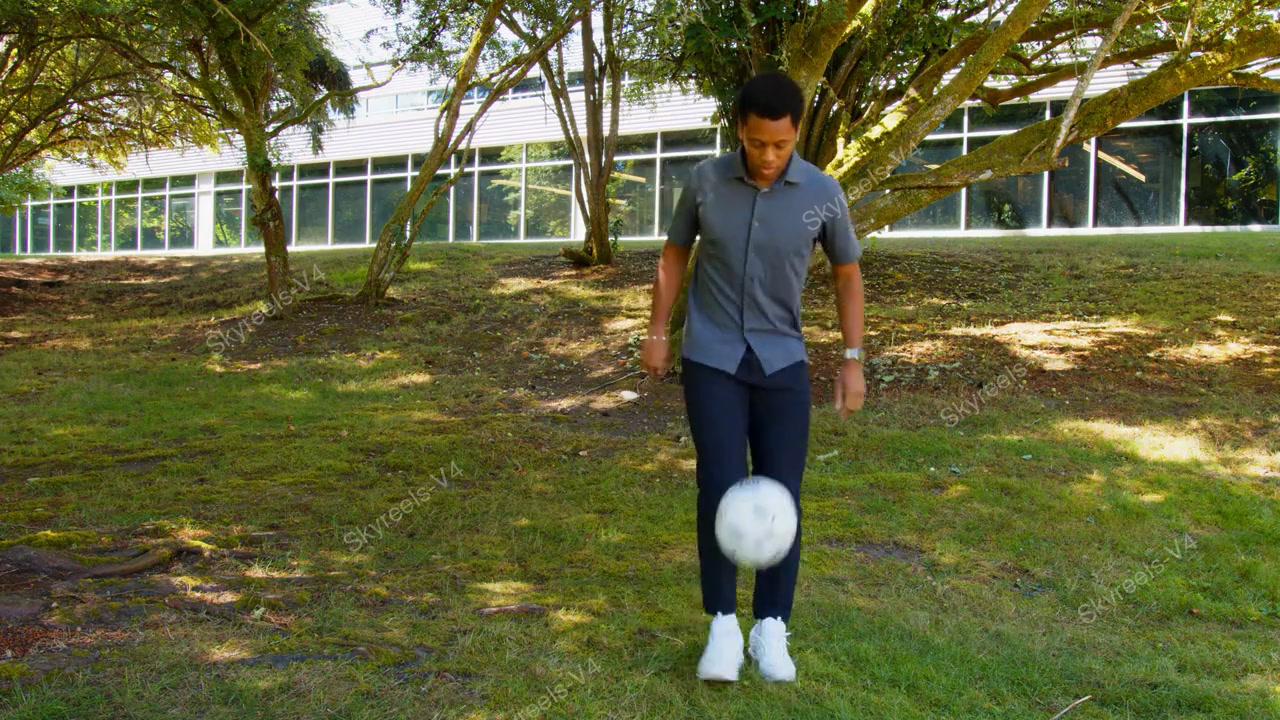} \\[2pt]

        \centering\rotatebox{90}{\textbf{Output Video}} &
        \includegraphics[width=\linewidth]{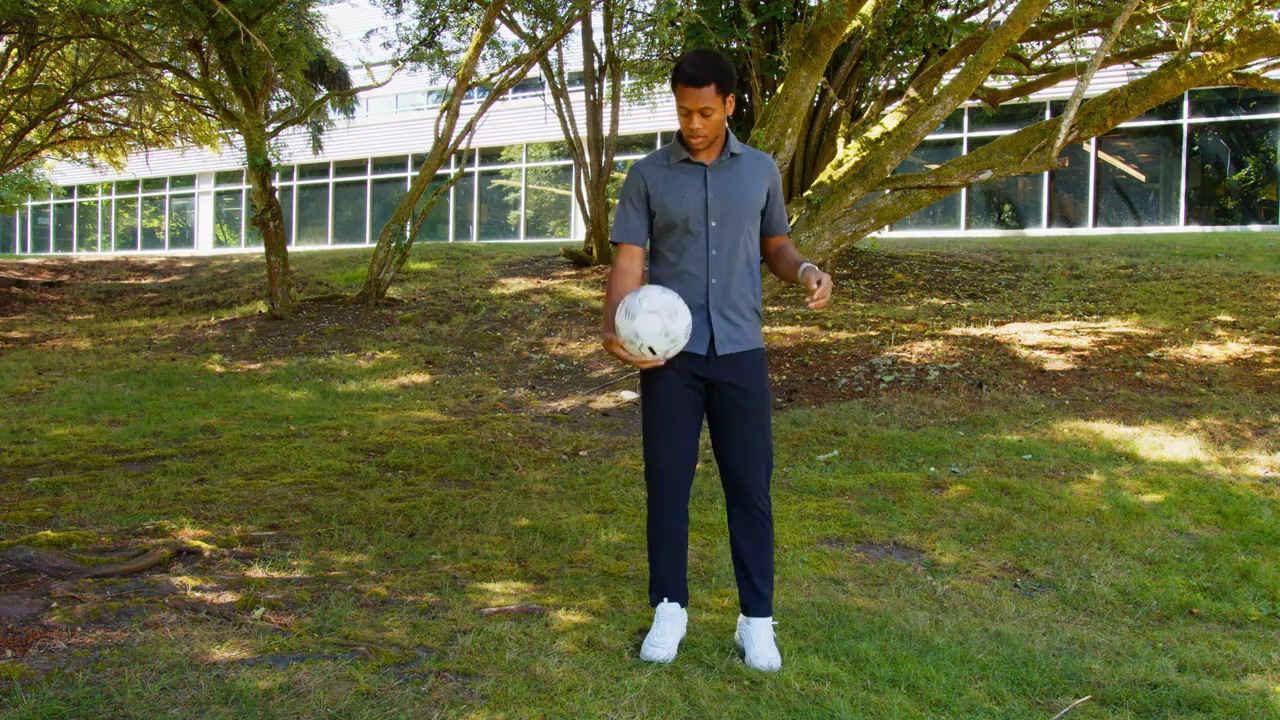} &
        \includegraphics[width=\linewidth]{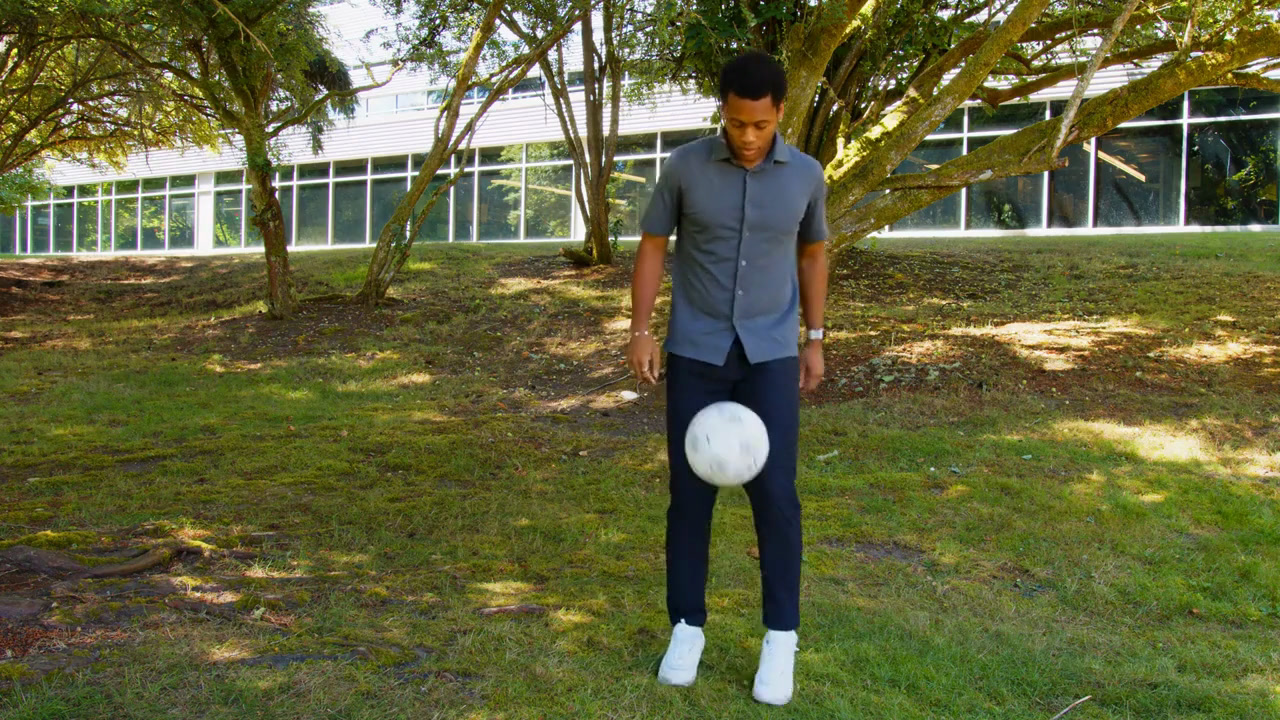} &
        \includegraphics[width=\linewidth]{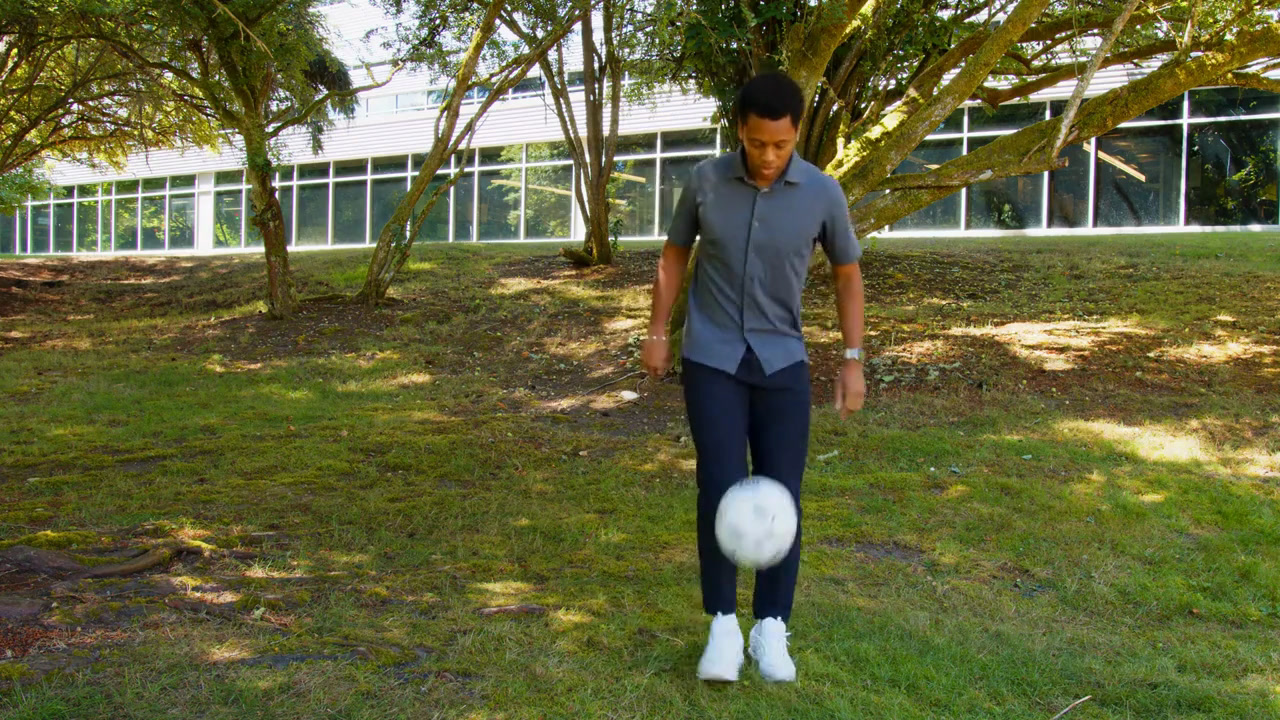} \\
    \end{tabular}
\end{flushleft}



\begin{flushleft}
    \textbf{Instruction:} \textit{Remove the text overlay at the bottom of @video\_1.}
    \begin{tabular}{m{1.2em} m{0.22\linewidth} m{0.22\linewidth} m{0.22\linewidth}}

        \centering\rotatebox{90}{\textbf{Input Video}} &
        \includegraphics[width=\linewidth]{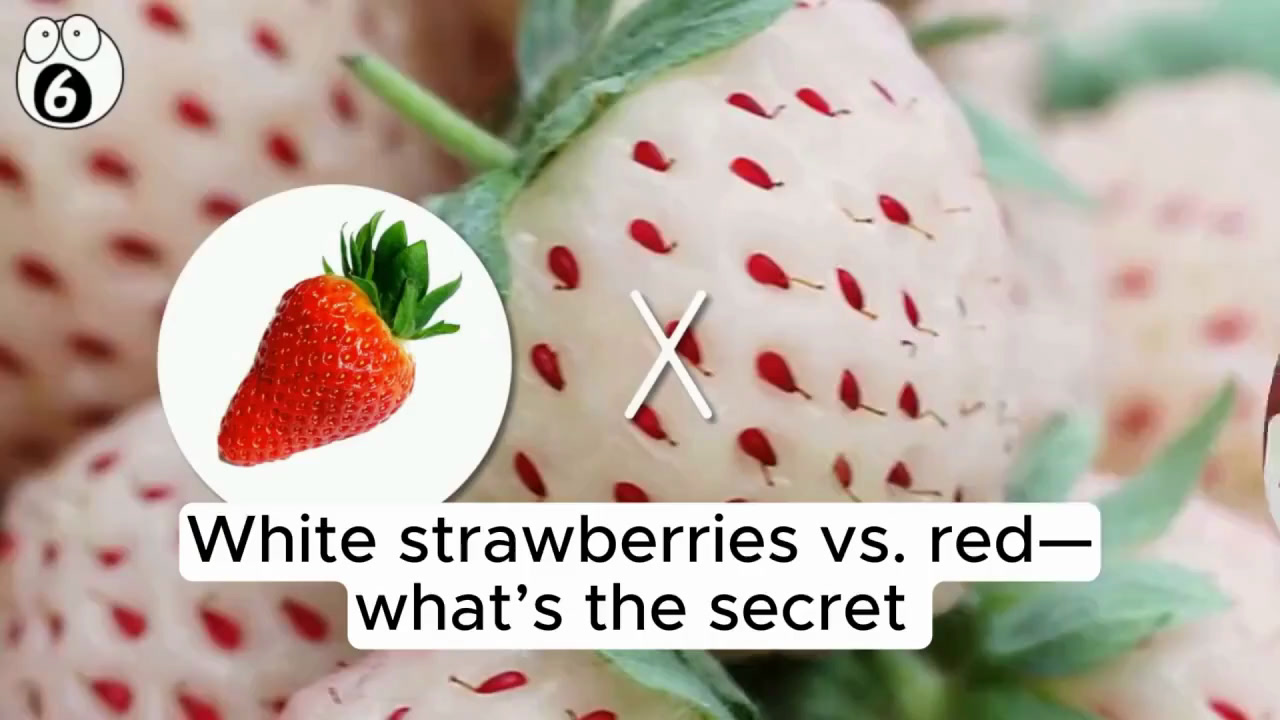} &
        \includegraphics[width=\linewidth]{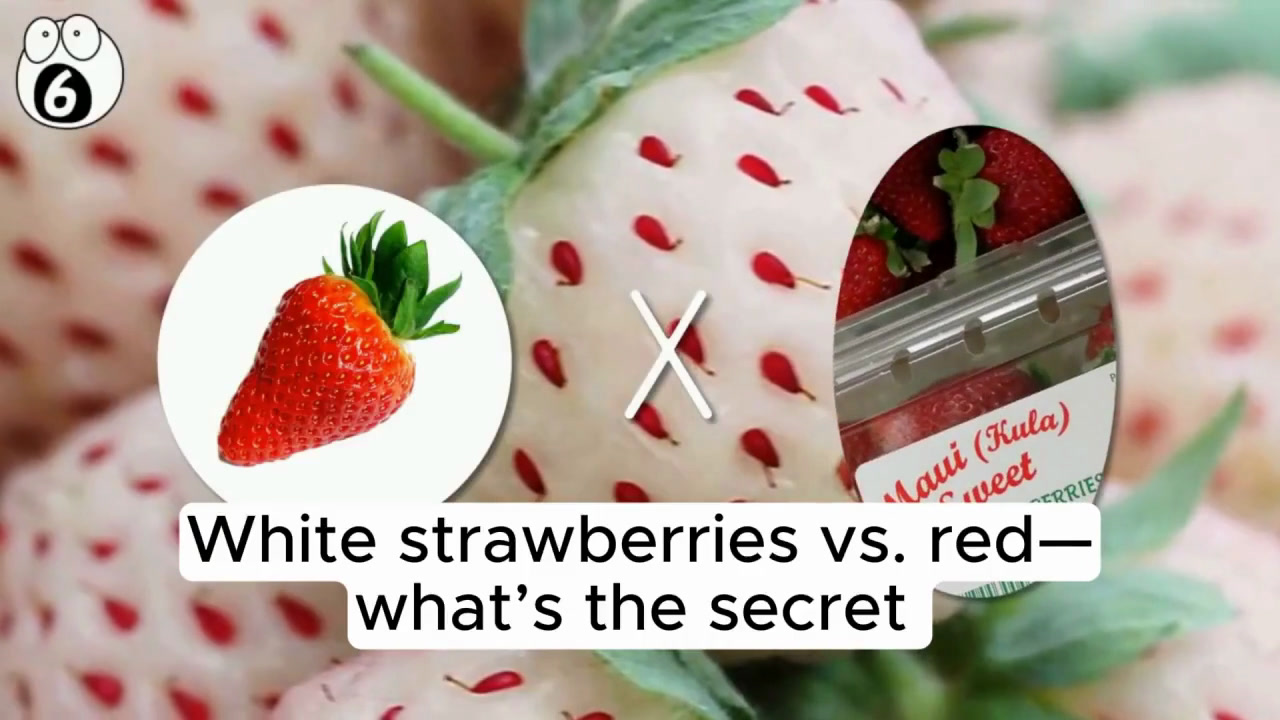} &
        \includegraphics[width=\linewidth]{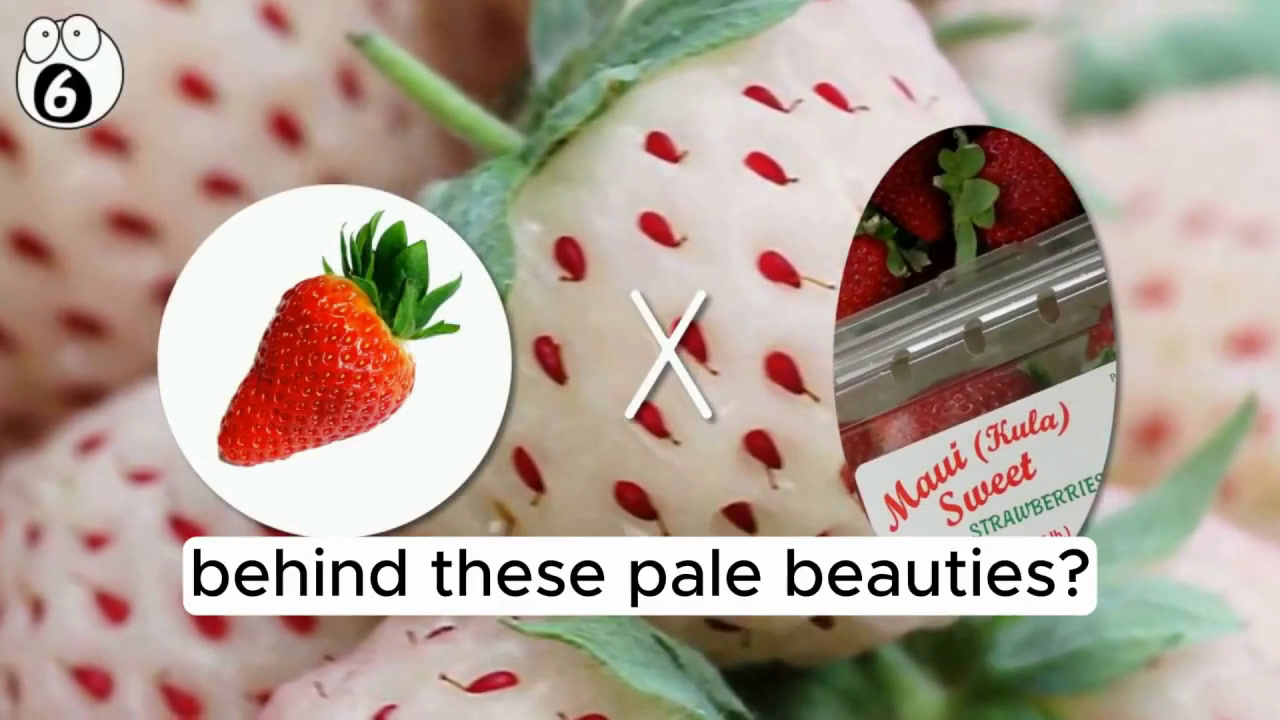} \\[2pt]

        \centering\rotatebox{90}{\textbf{Output Video}} &
        \includegraphics[width=\linewidth]{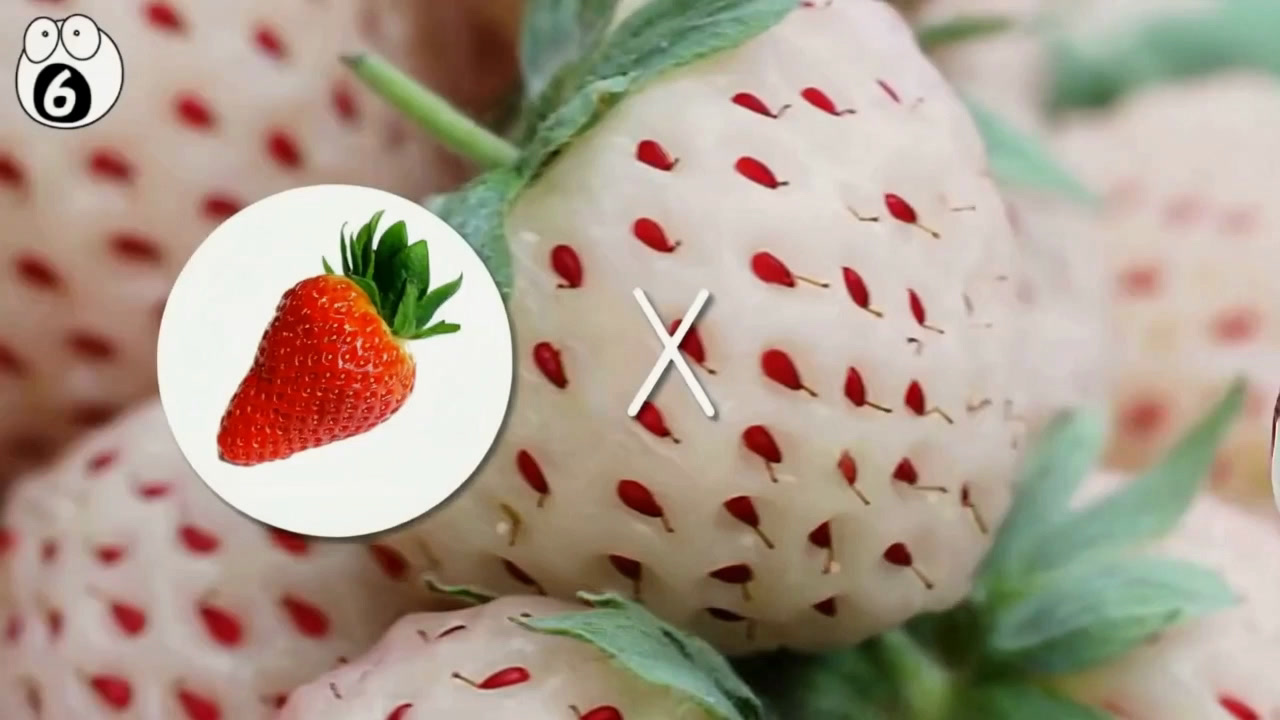} &
        \includegraphics[width=\linewidth]{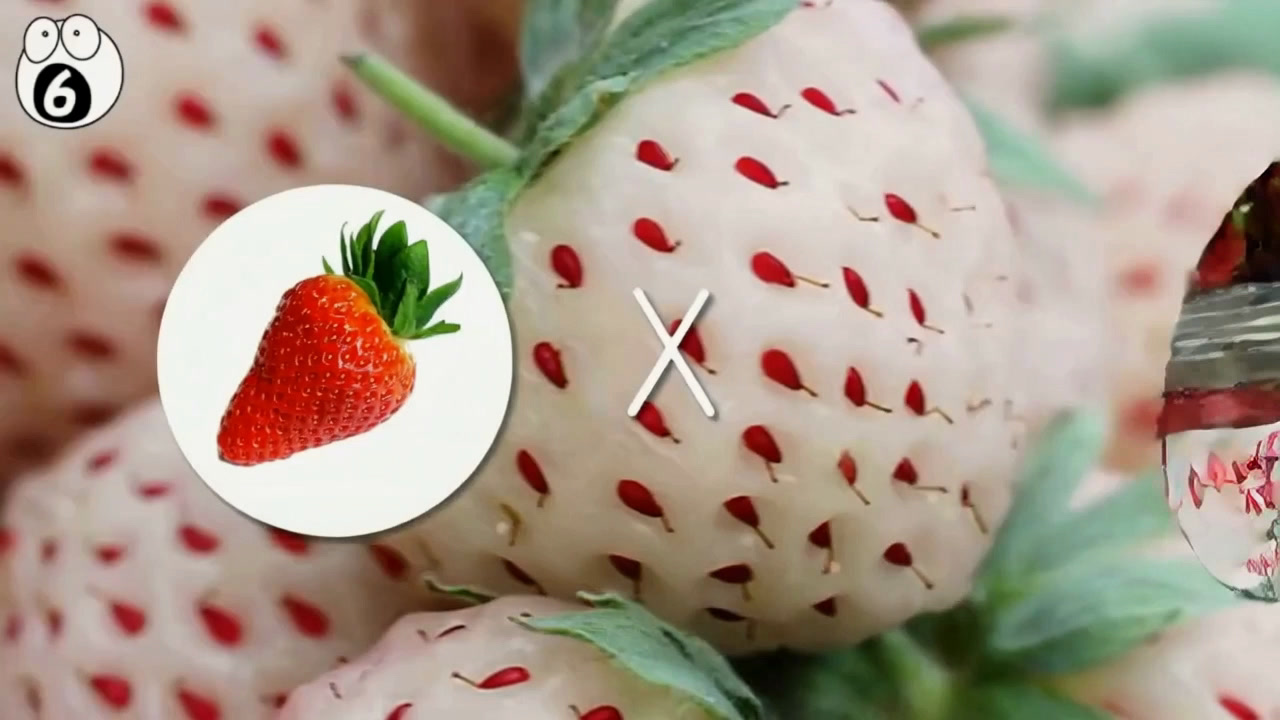} &
        \includegraphics[width=\linewidth]{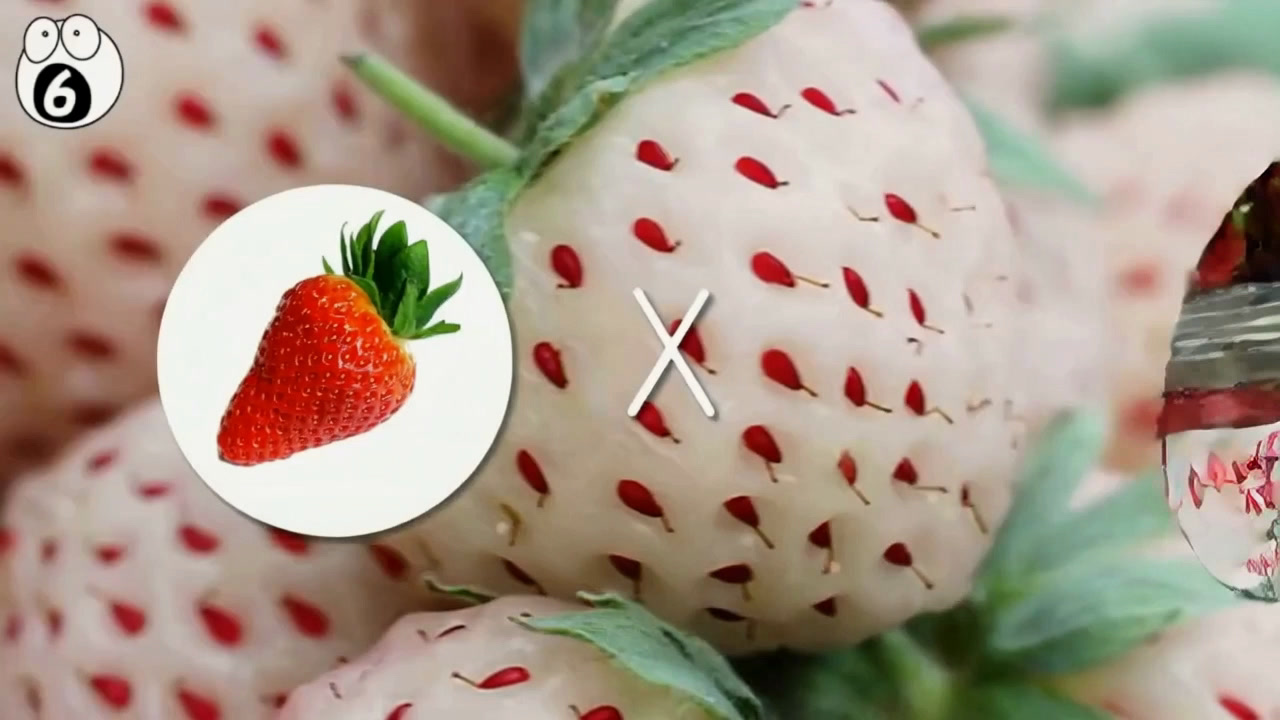} \\
    \end{tabular}
\end{flushleft}

\begin{flushleft}
    \textbf{Instruction:} \textit{Remove the logo in the upper right corner in @video\_1.}
    \begin{tabular}{m{1.2em} m{0.22\linewidth} m{0.22\linewidth} m{0.22\linewidth}}

        \centering\rotatebox{90}{\textbf{Input Video}} &
        \includegraphics[width=\linewidth]{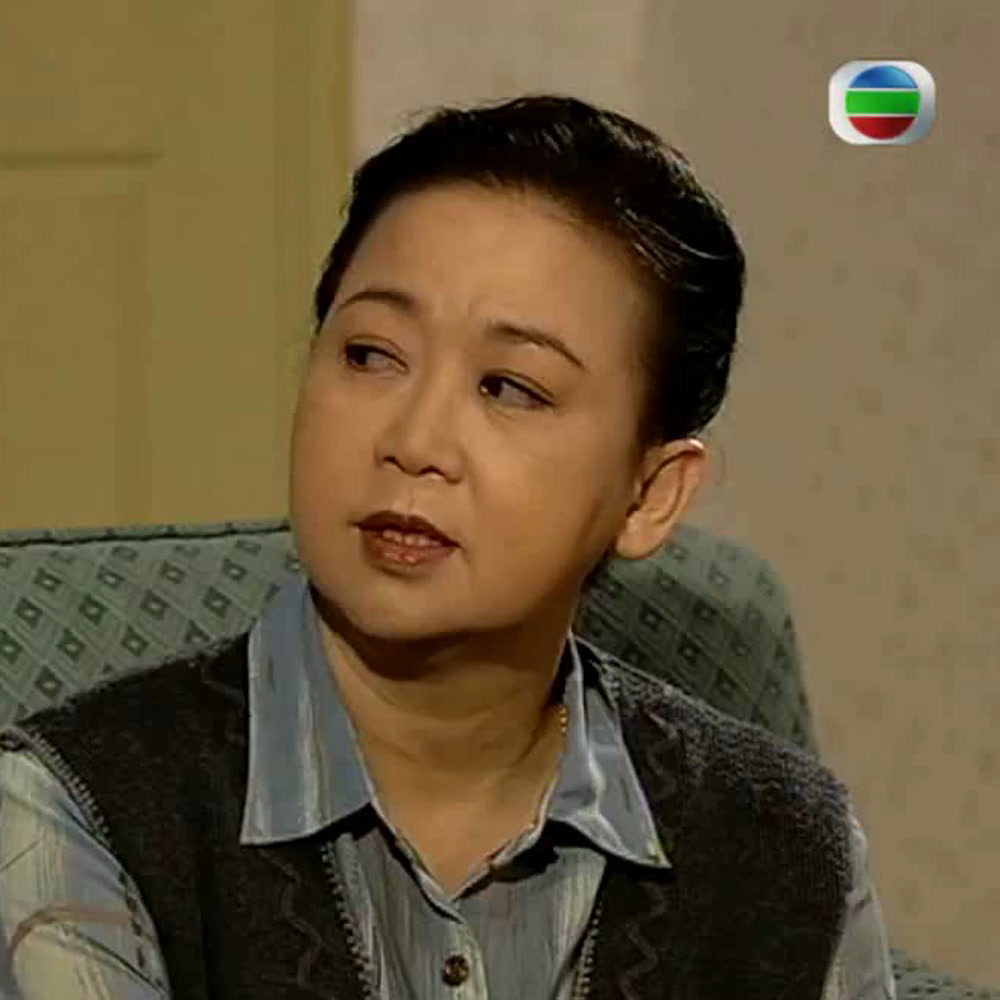} &
        \includegraphics[width=\linewidth]{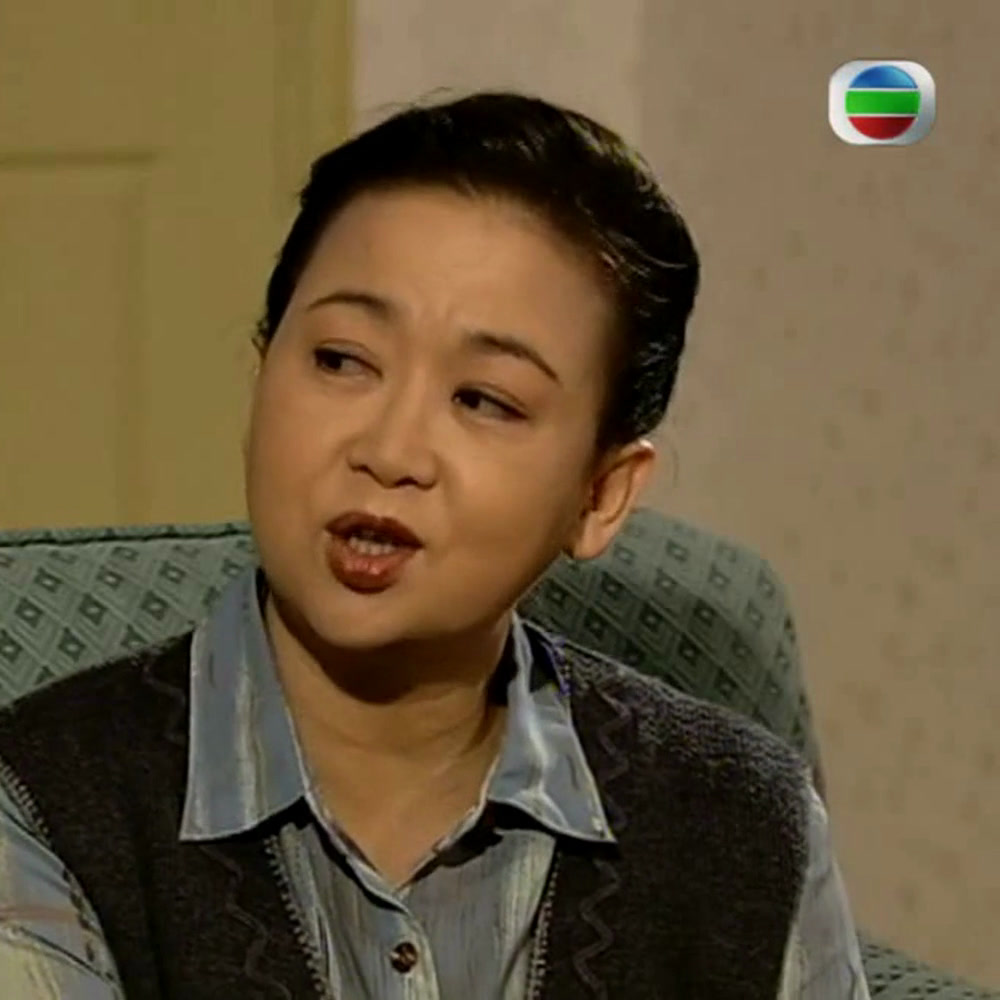} &
        \includegraphics[width=\linewidth]{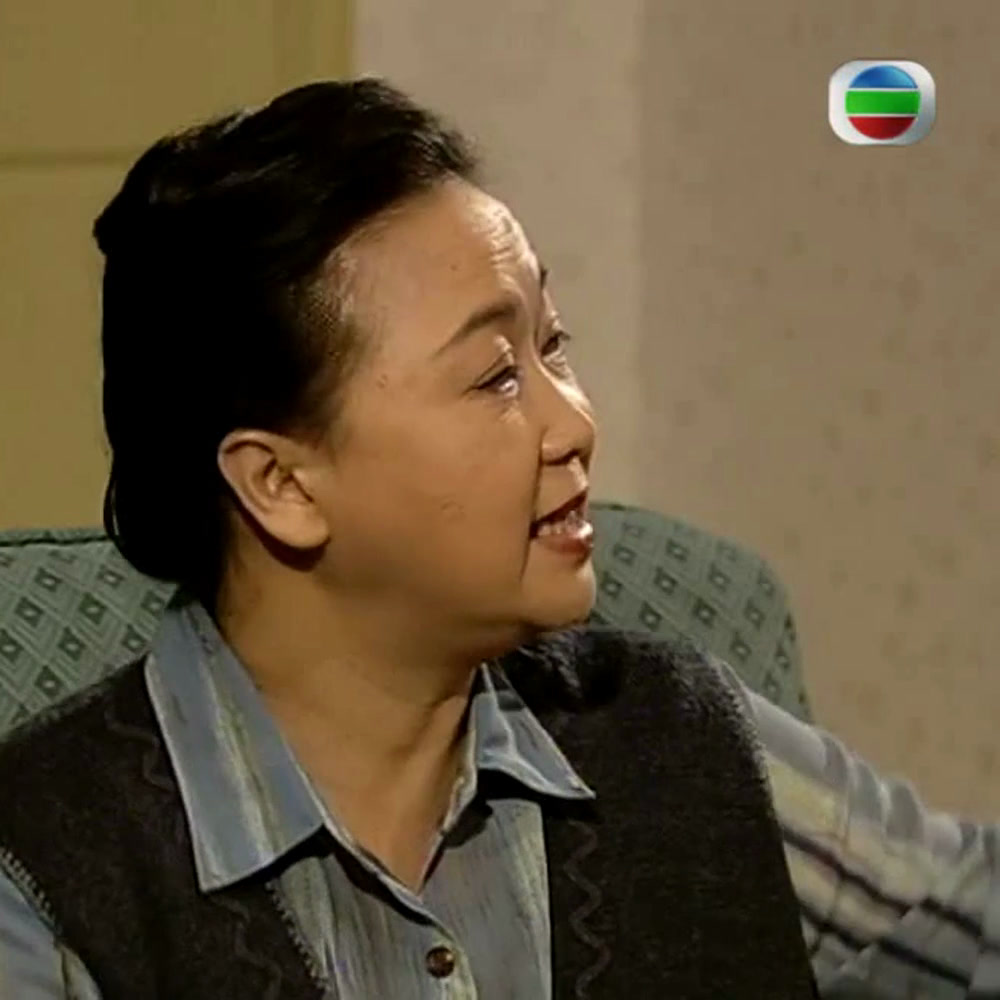} \\[2pt]

        \centering\rotatebox{90}{\textbf{Output Video}} &
        \includegraphics[width=\linewidth]{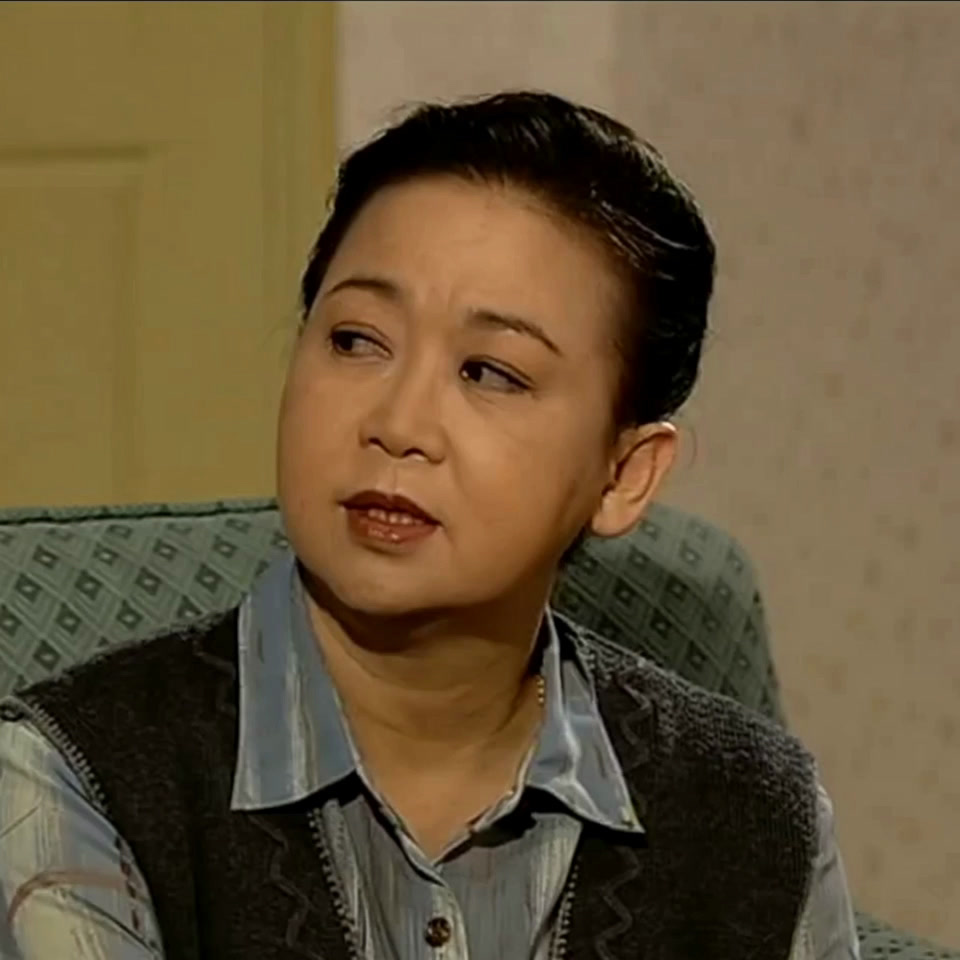} &
        \includegraphics[width=\linewidth]{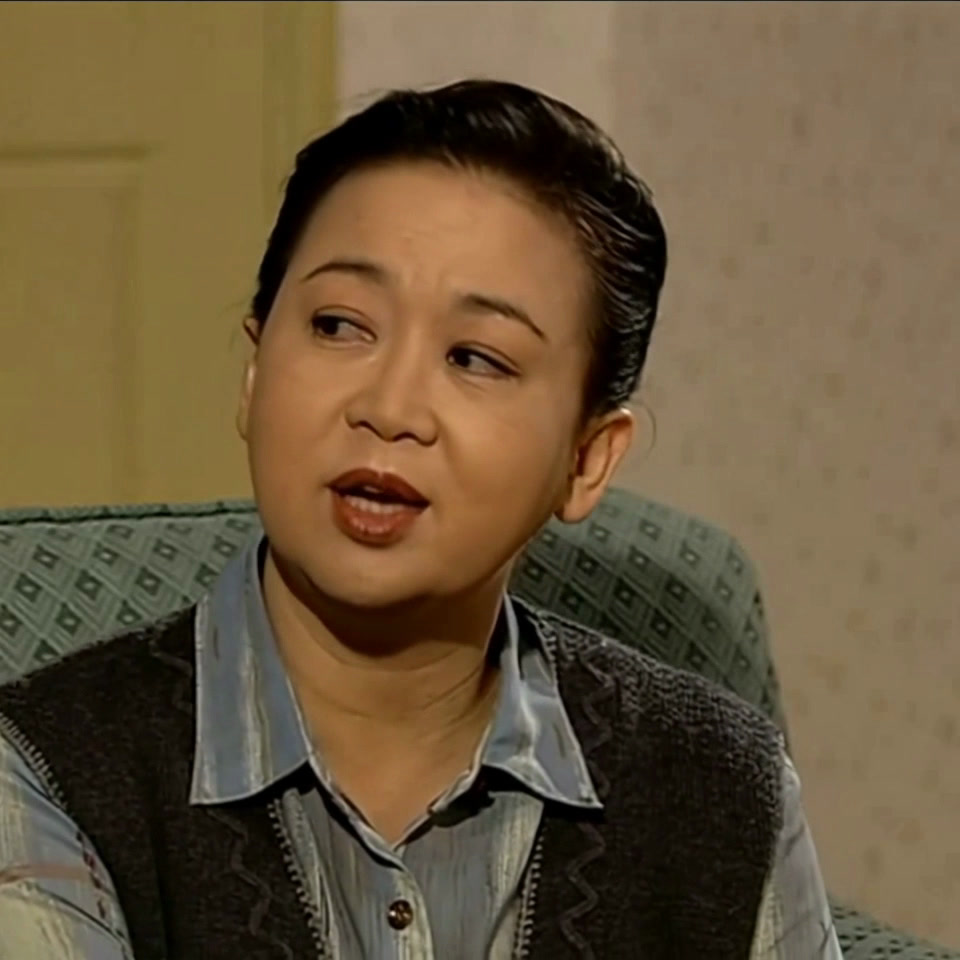} &
        \includegraphics[width=\linewidth]{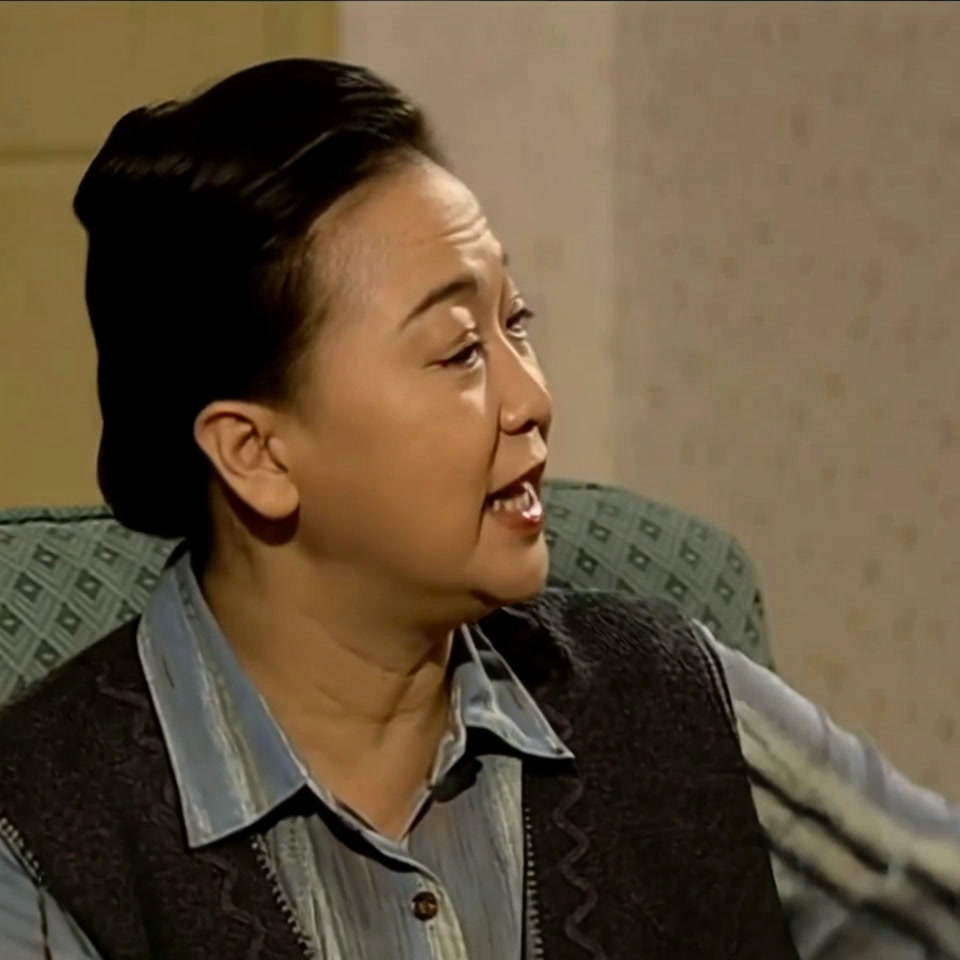} \\
    \end{tabular}
\end{flushleft}


\begin{figure}[h!]
    \centering
    \caption{Examples of watermark/subtitle/logo removal.}
    \label{fig:removal_logo_1}
\end{figure}

\newpage
\paragraph{Subject Manipulation}
\label{appendix:subject-manipulation}

The model supports adding, deleting, and modifying subjects in videos while maintaining temporal consistency.

\begin{flushleft}
    \textbf{Instruction:} \textit{Place a wooden bench with black metal armrests and legs on the grass next to the large rock on the right side of the path in @video\_1.}%
    \vspace{0.5em}
    \begin{tabular}{m{1.2em} m{0.22\linewidth} m{0.22\linewidth} m{0.22\linewidth}}

        \centering\rotatebox{90}{\textbf{Input Video}} &
        \includegraphics[width=\linewidth]{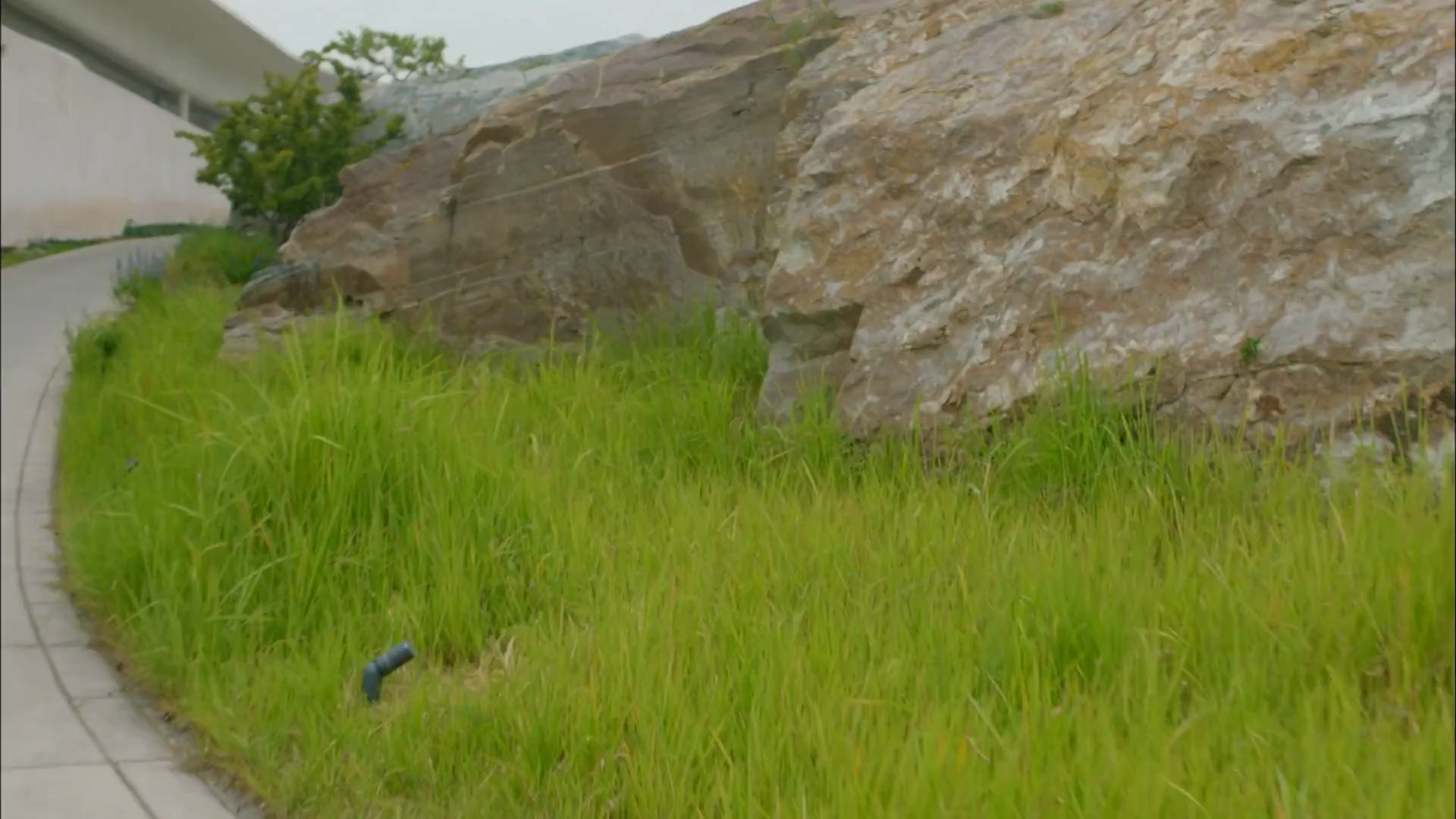} &
        \includegraphics[width=\linewidth]{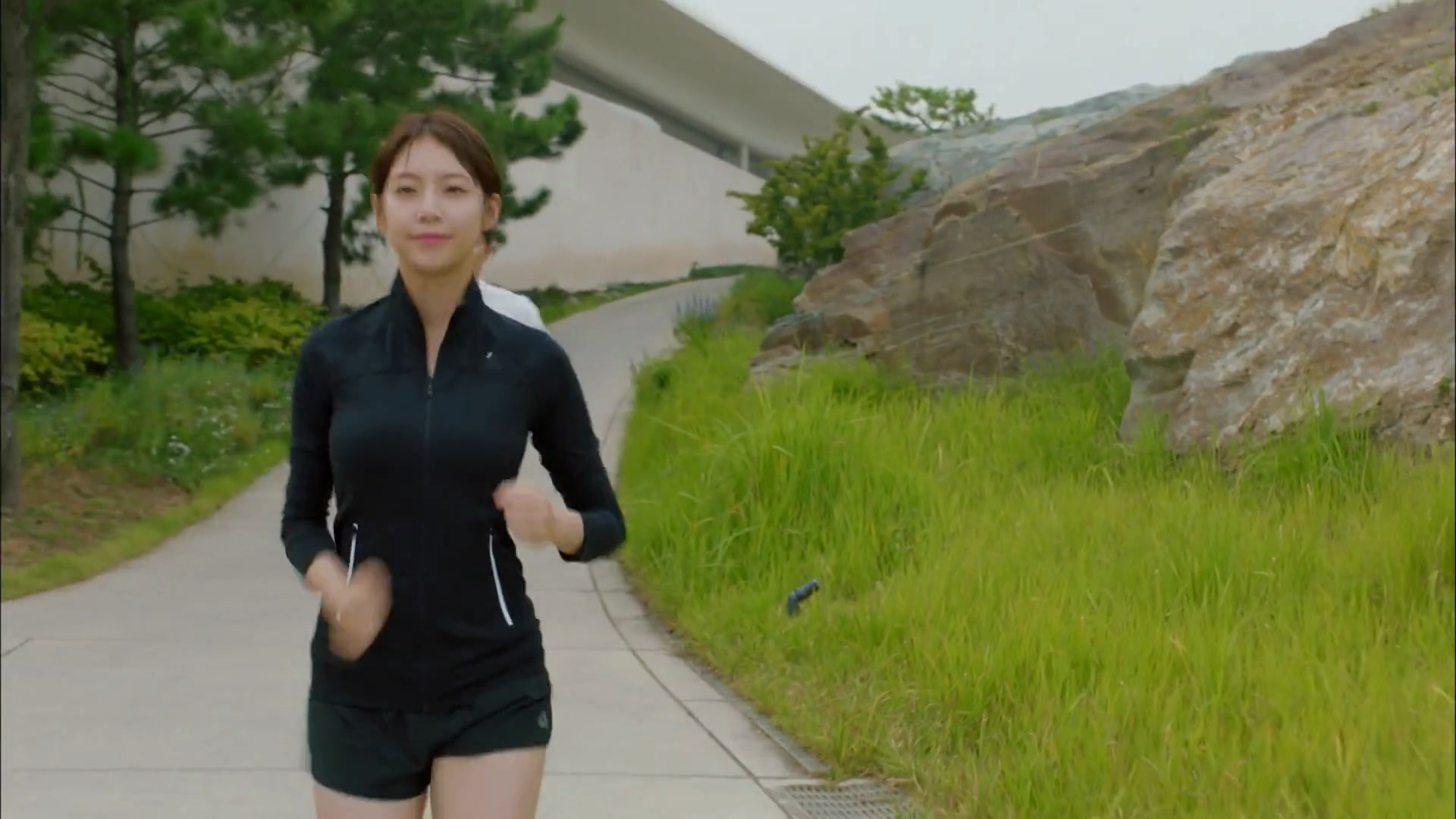} &
        \includegraphics[width=\linewidth]{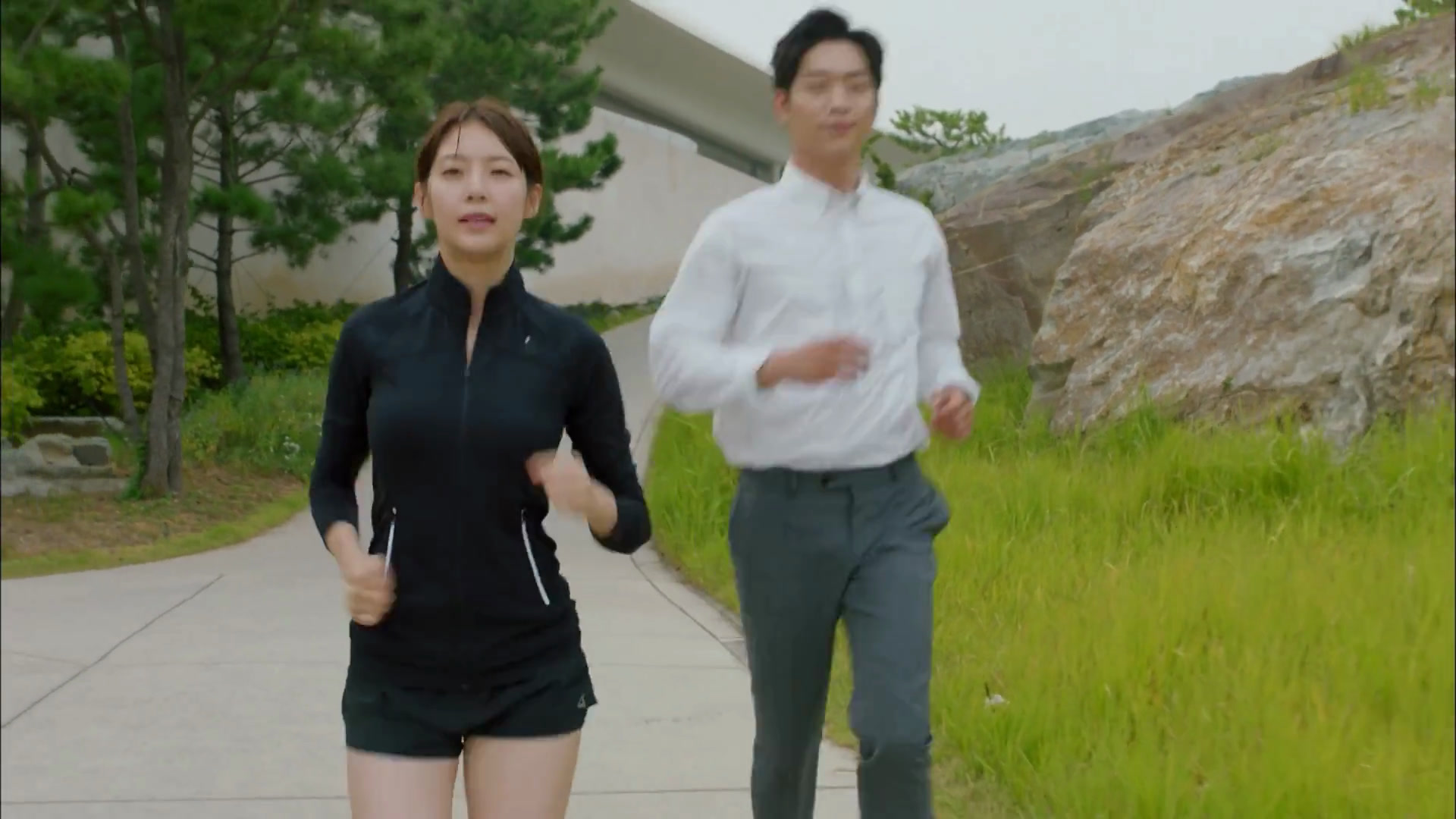} \\[2pt]

        \centering\rotatebox{90}{\textbf{Output Video}} &
        \includegraphics[width=\linewidth]{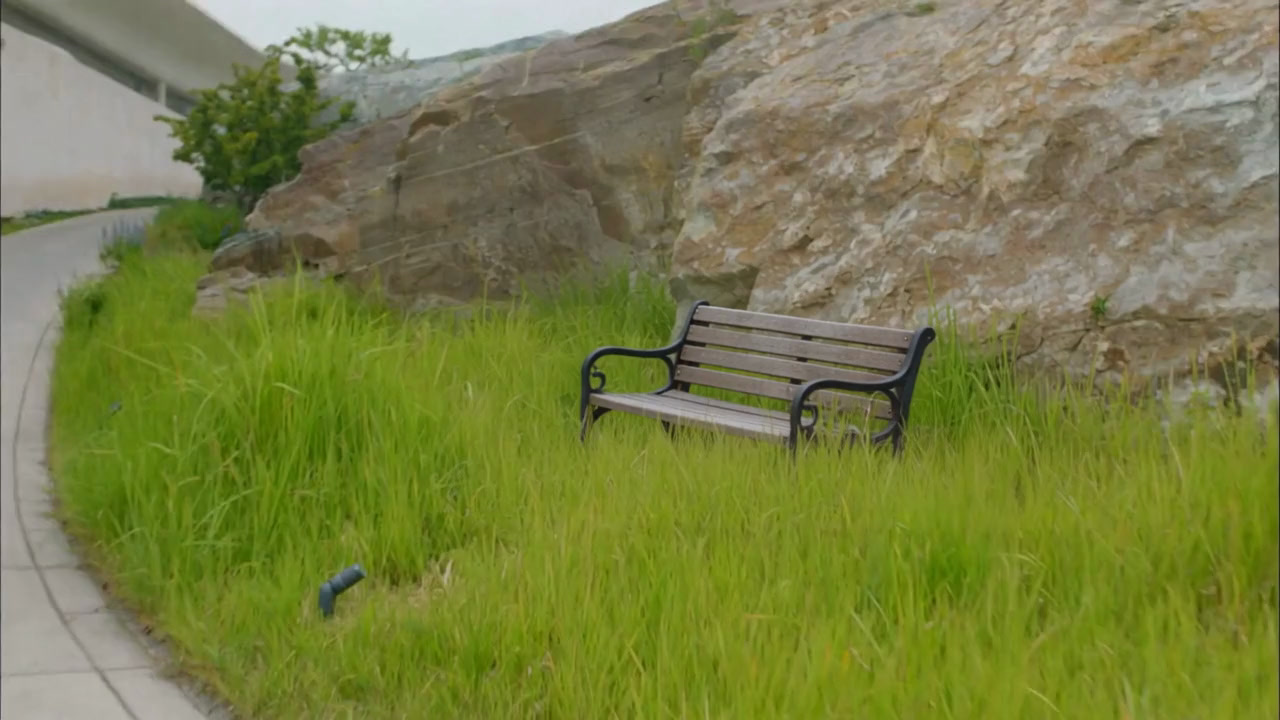} &
        \includegraphics[width=\linewidth]{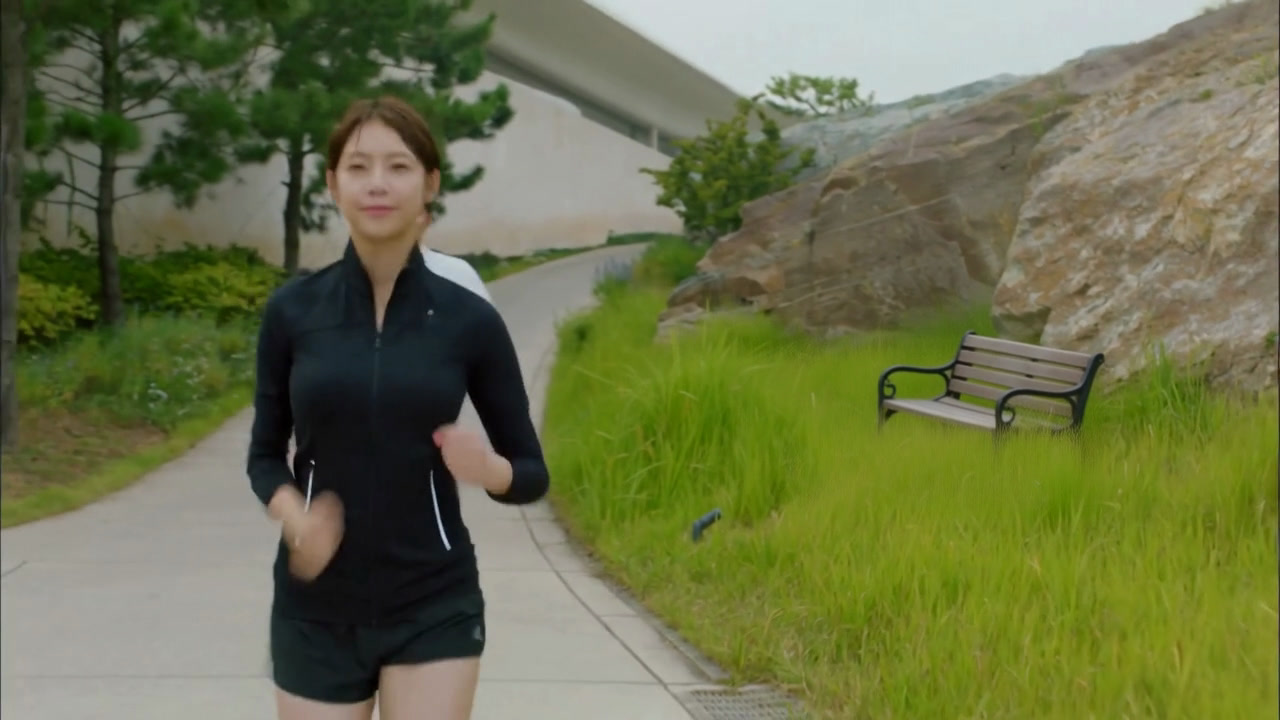} &
        \includegraphics[width=\linewidth]{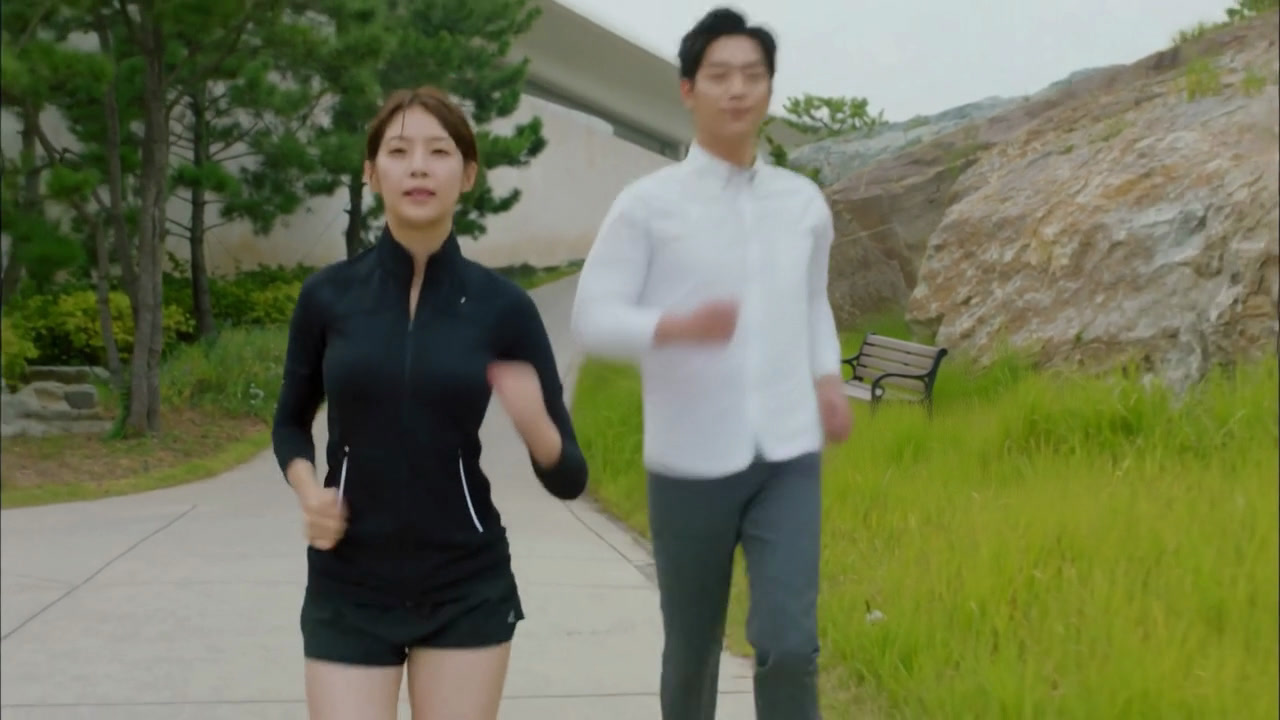} \\
    \end{tabular}
\end{flushleft}



\begin{flushleft}
    \textbf{Instruction:} \textit{Remove a bee from the center of @video\_1.}%
    \vspace{0.5em}
    \begin{tabular}{m{1.2em} m{0.22\linewidth} m{0.22\linewidth} m{0.22\linewidth}}
        
        \centering\rotatebox{90}{\textbf{Input Video}} &
        \includegraphics[width=\linewidth]{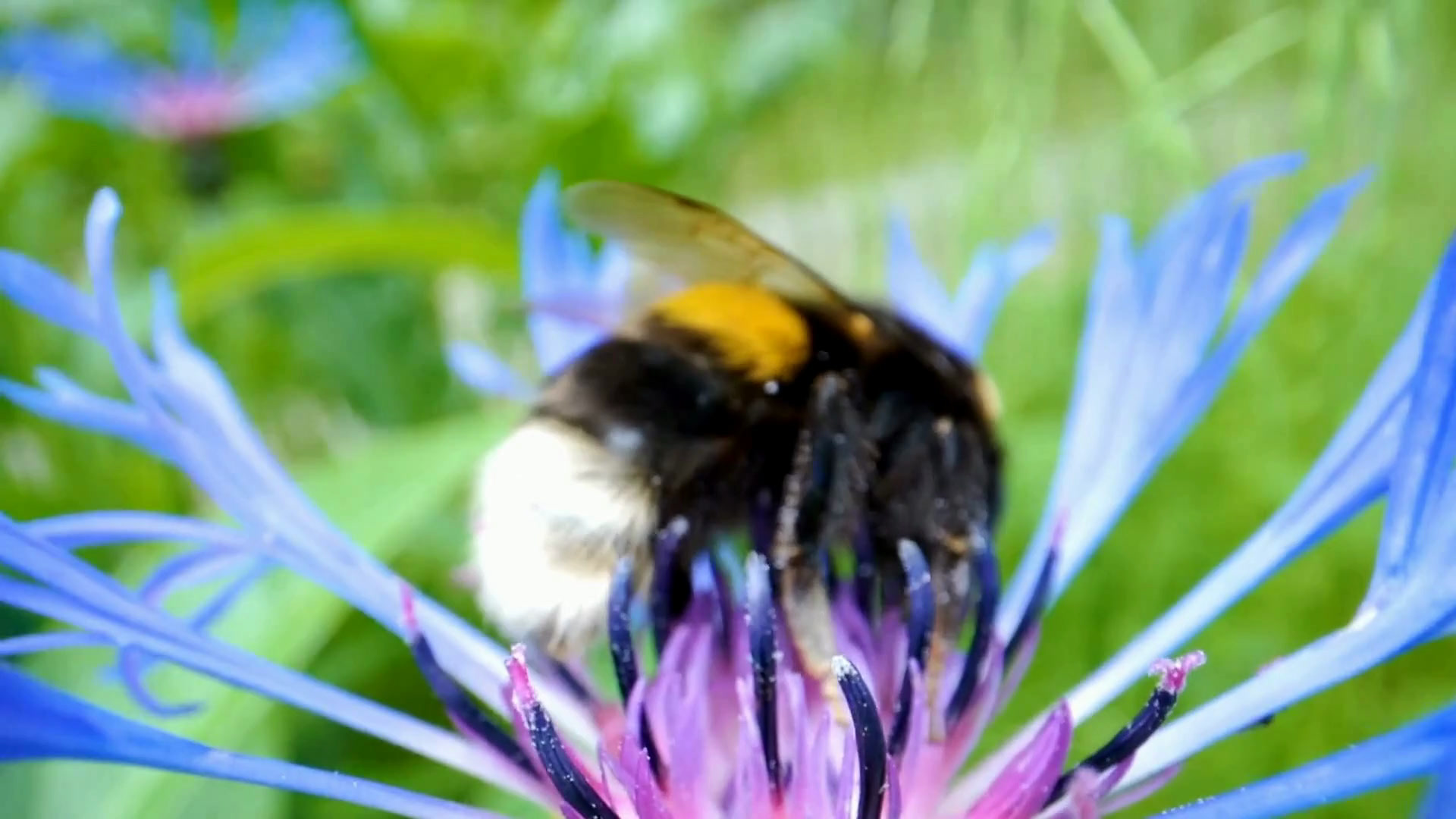} &
        \includegraphics[width=\linewidth]{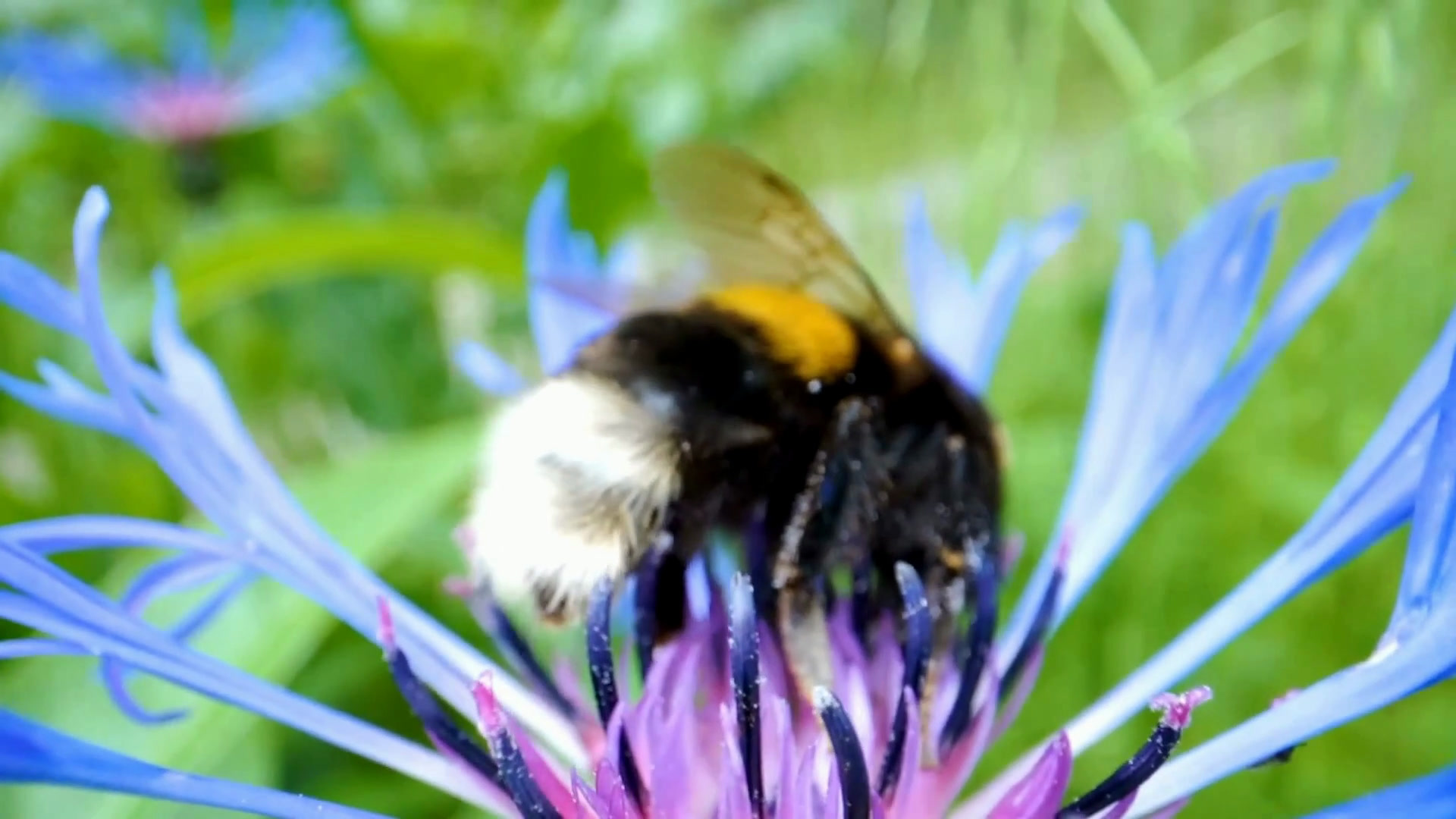} &
        \includegraphics[width=\linewidth]{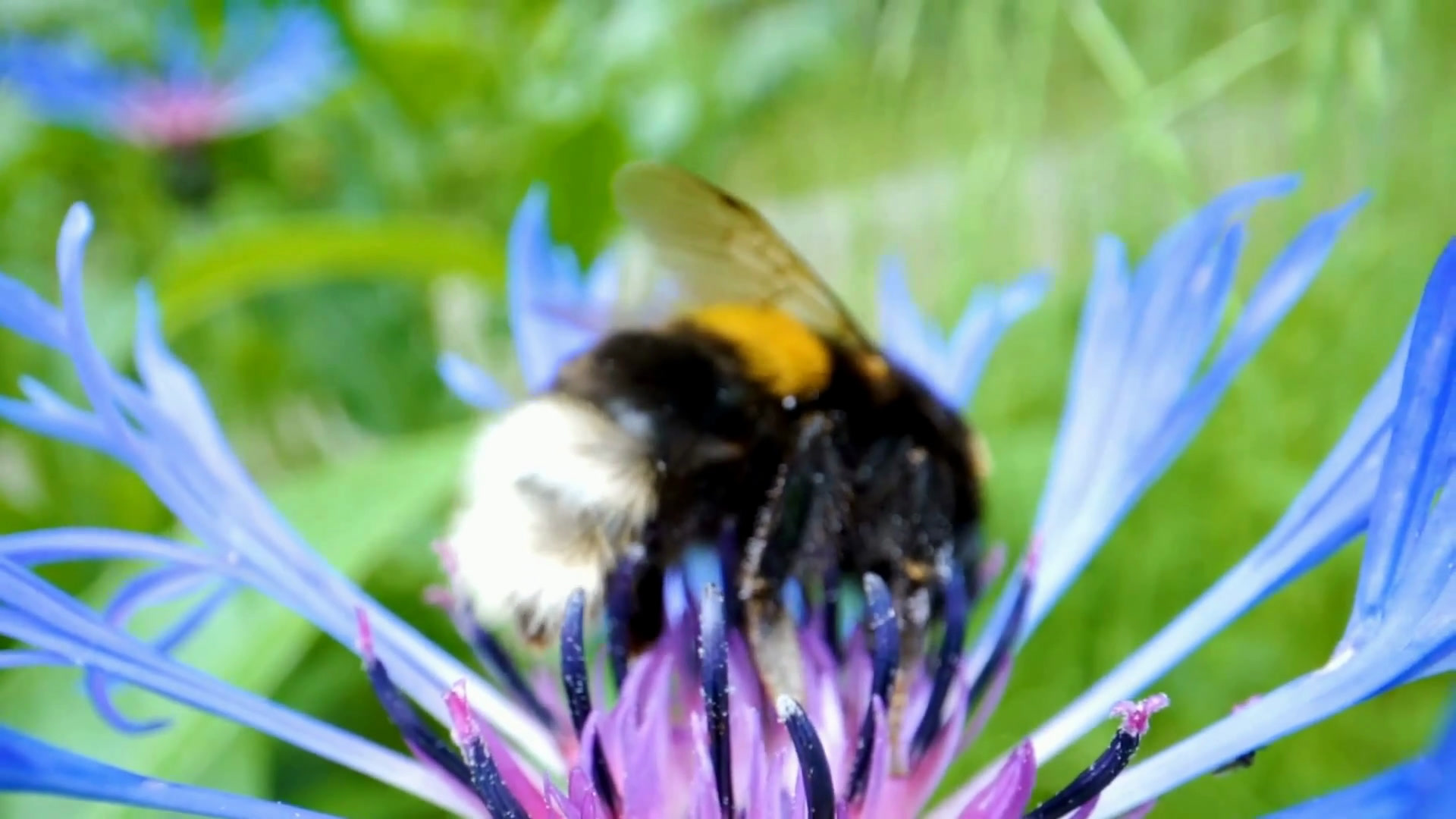} \\[2pt]

        \centering\rotatebox{90}{\textbf{Output Video}} &
        \includegraphics[width=\linewidth]{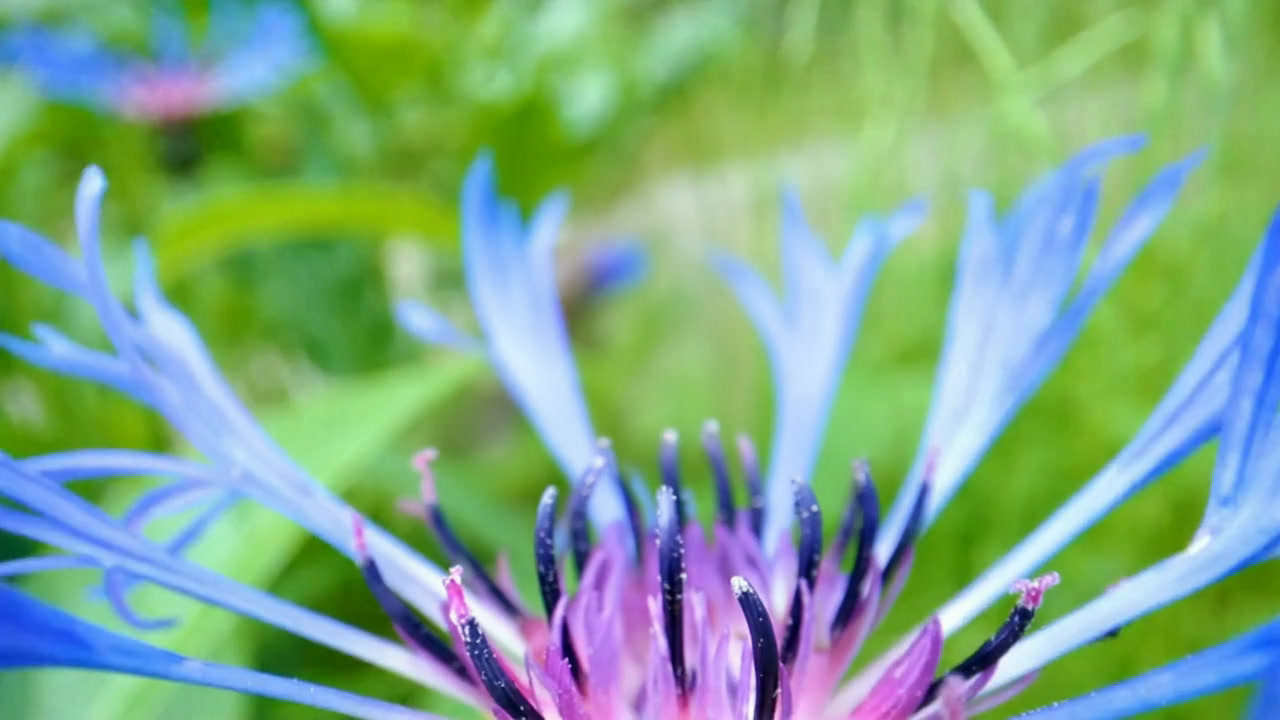} &
        \includegraphics[width=\linewidth]{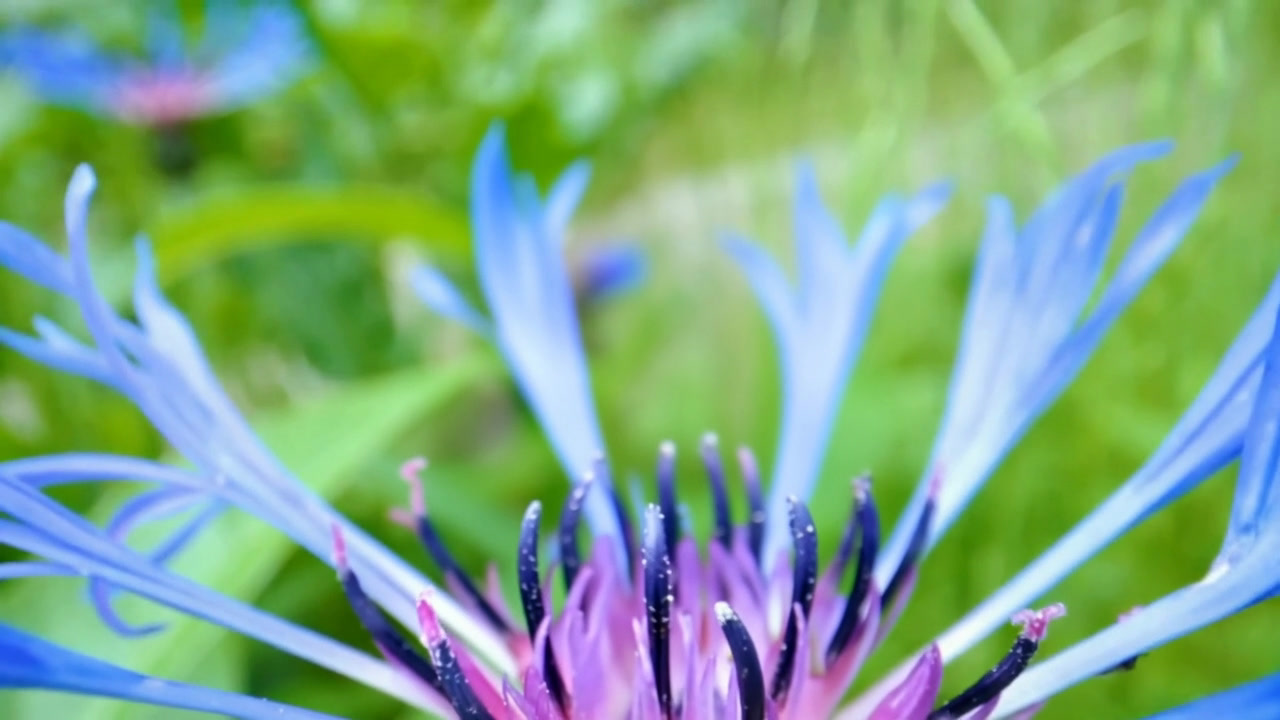} &
        \includegraphics[width=\linewidth]{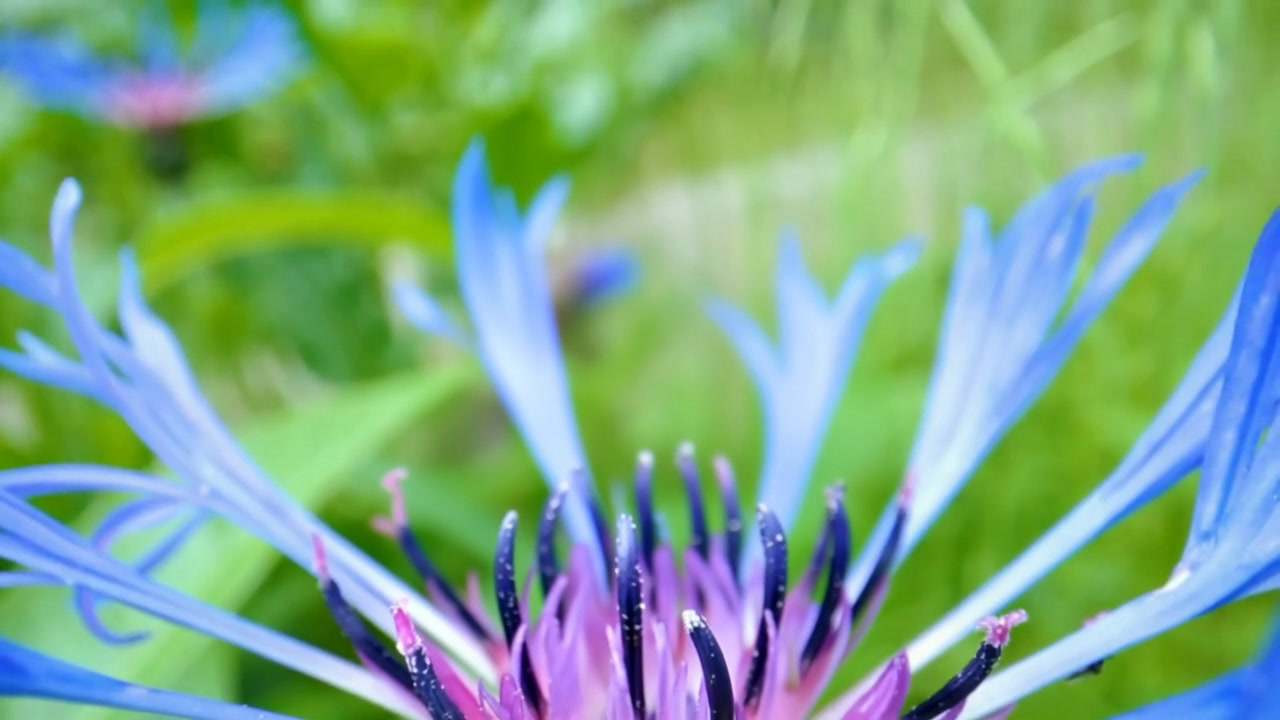} \\
    \end{tabular}
\end{flushleft}


\begin{figure}[h!]
    \centering
    \caption{Examples of subject manipulation.}
    \label{fig:edit_subject_2}
\end{figure}

\newpage
\paragraph{Local Attribute Editing}

The model can perform fine-grained editing of attributes for specific objects or regions in videos, such as color, texture, shape, etc.

\begin{flushleft}
    \textbf{Instruction:} \textit{Change the chair's color to black and replace its edges with wooden material in @video\_1.}%
    \vspace{0.5em}
    \begin{tabular}{m{1.2em} m{0.22\linewidth} m{0.22\linewidth} m{0.22\linewidth}}

        \centering\rotatebox{90}{\textbf{Input Video}} &
        \includegraphics[width=\linewidth]{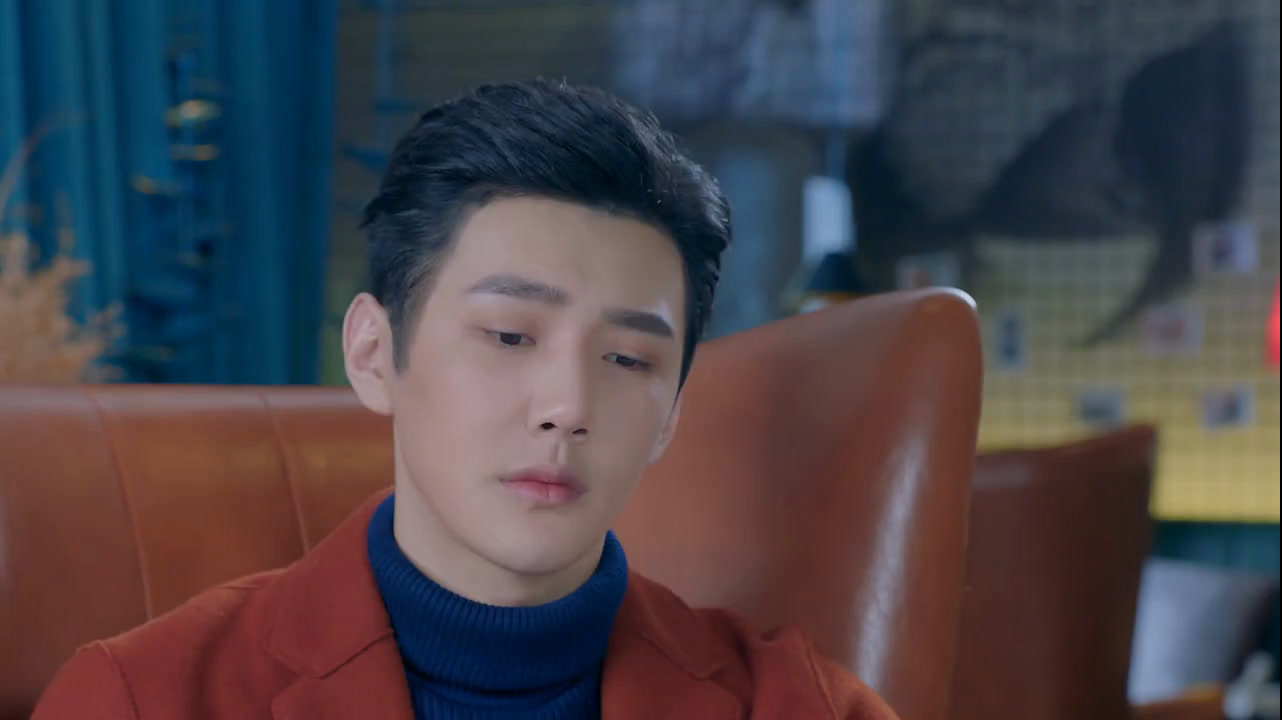} &
        \includegraphics[width=\linewidth]{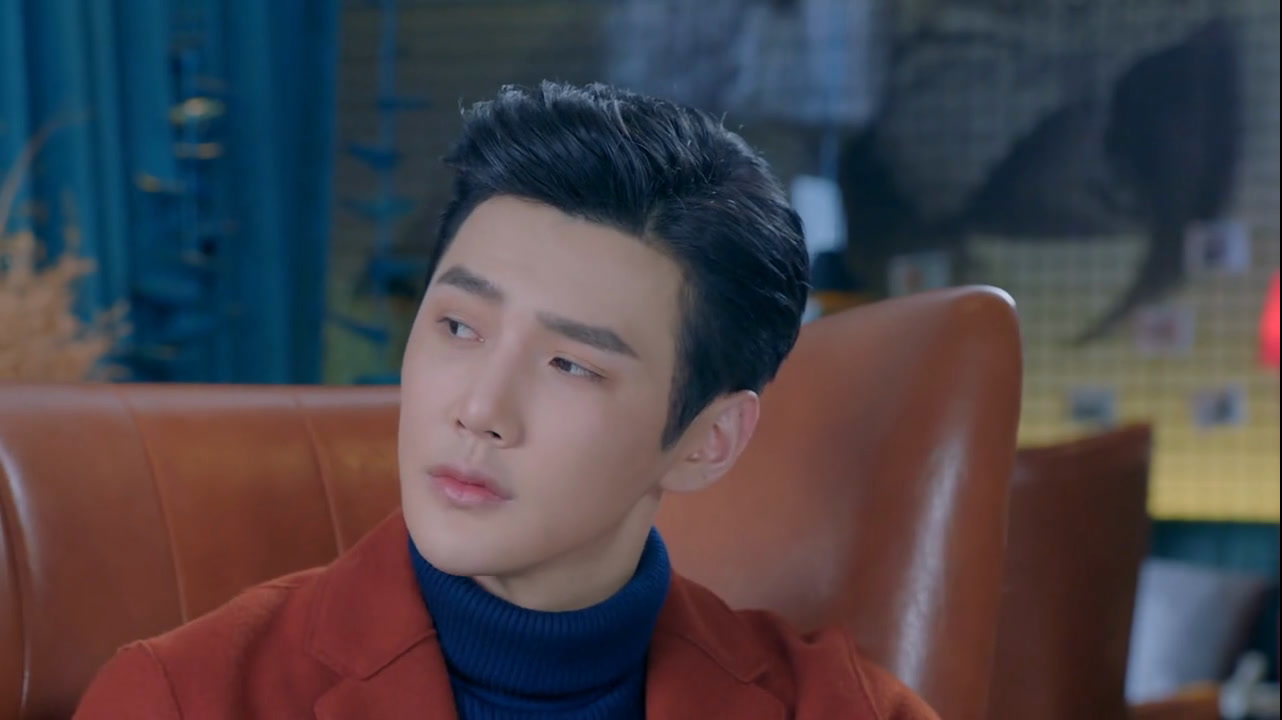} &
        \includegraphics[width=\linewidth]{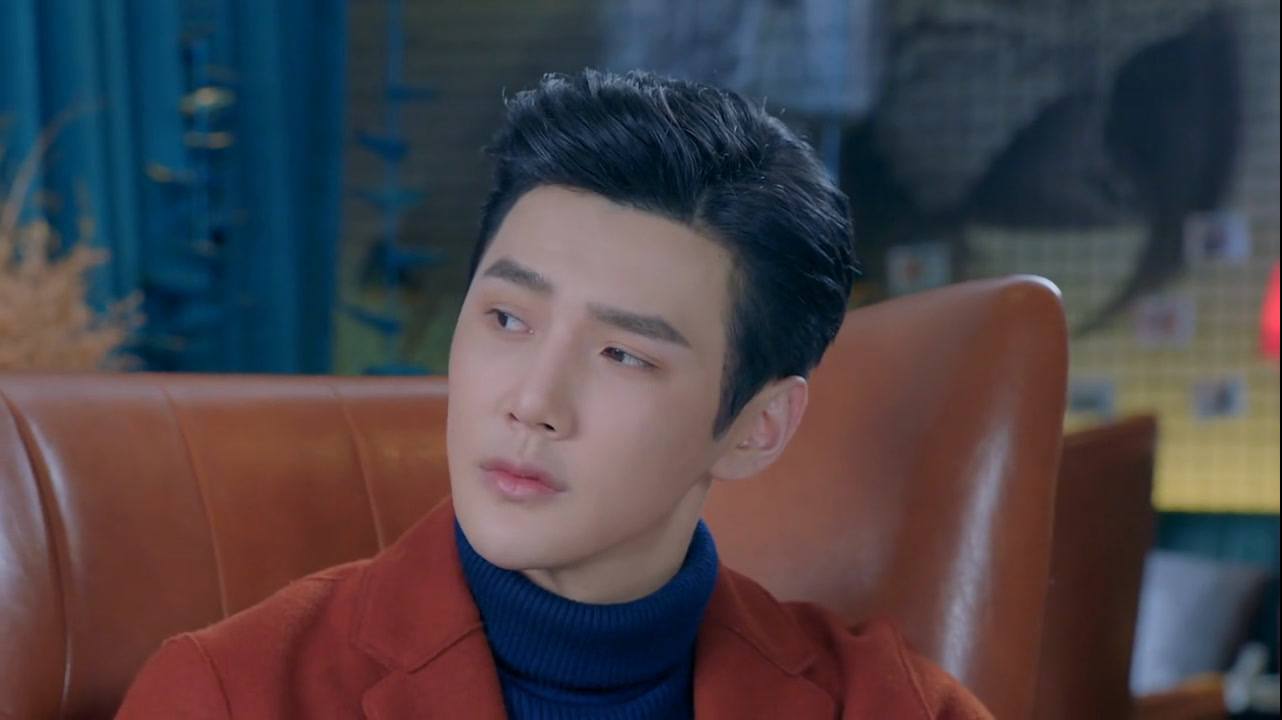} \\[2pt]

        \centering\rotatebox{90}{\textbf{Output Video}} &
        \includegraphics[width=\linewidth]{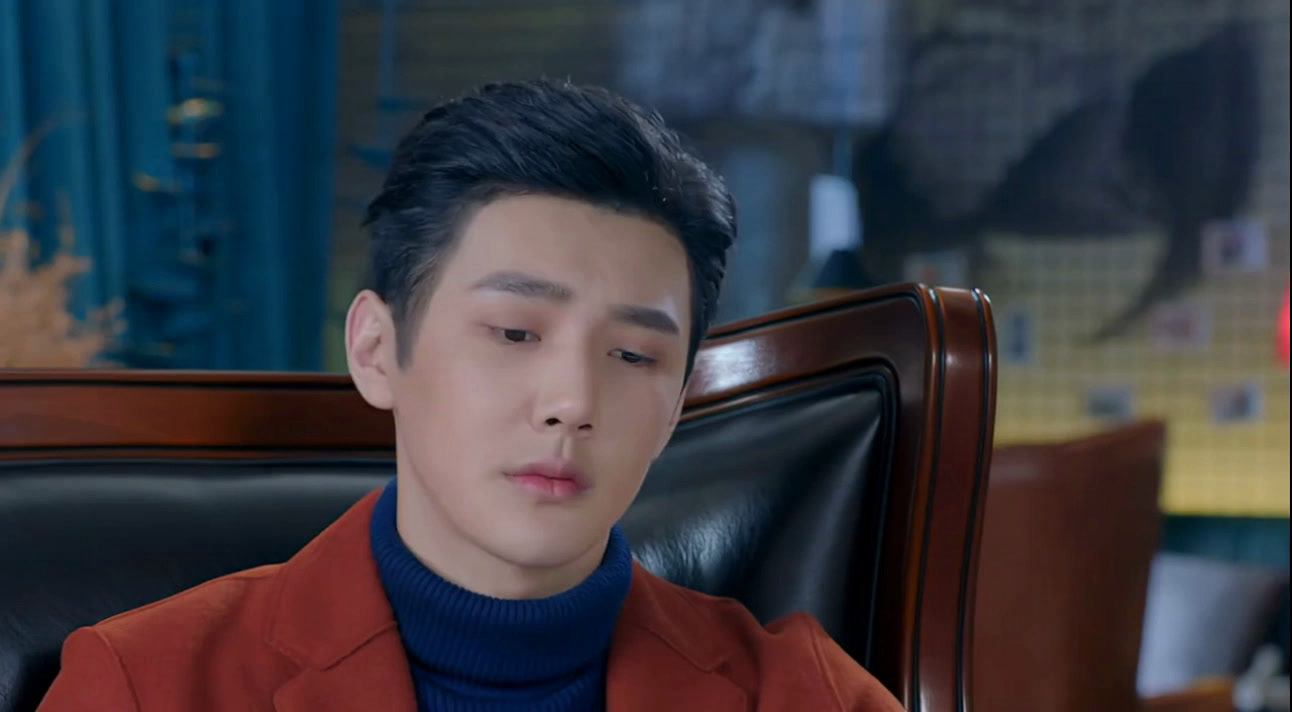} &
        \includegraphics[width=\linewidth]{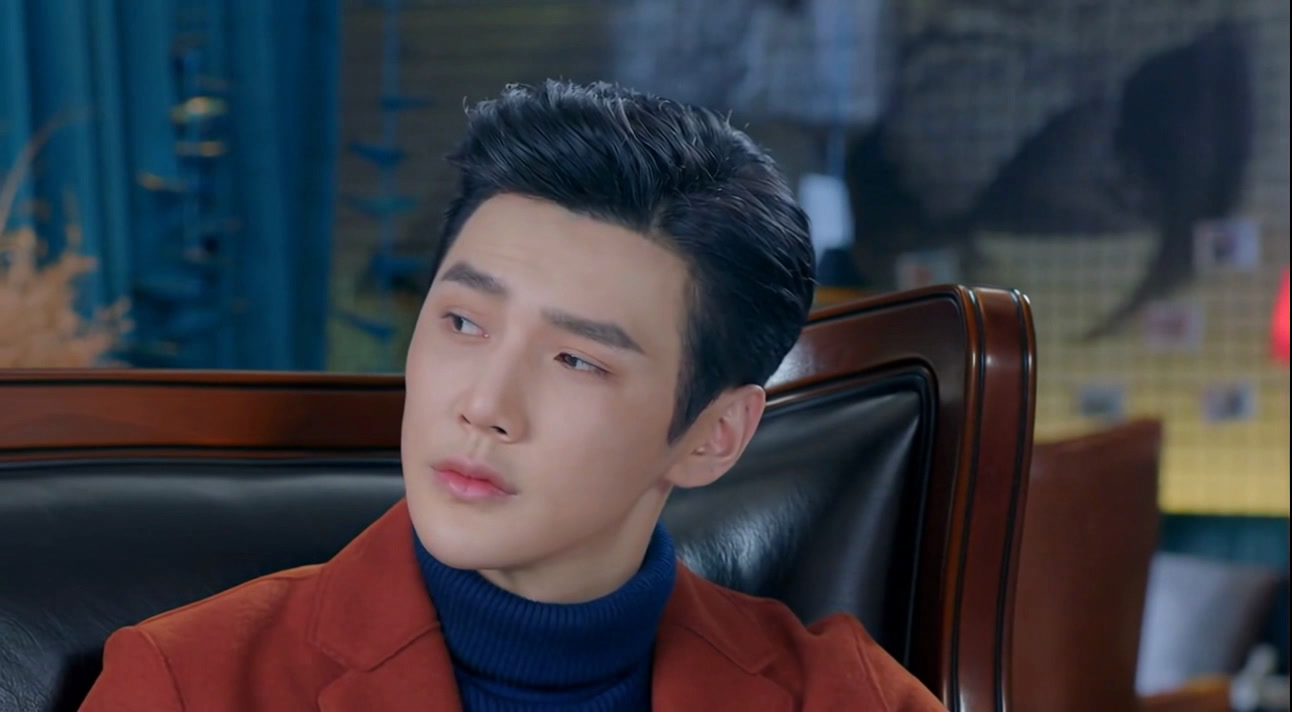} &
        \includegraphics[width=\linewidth]{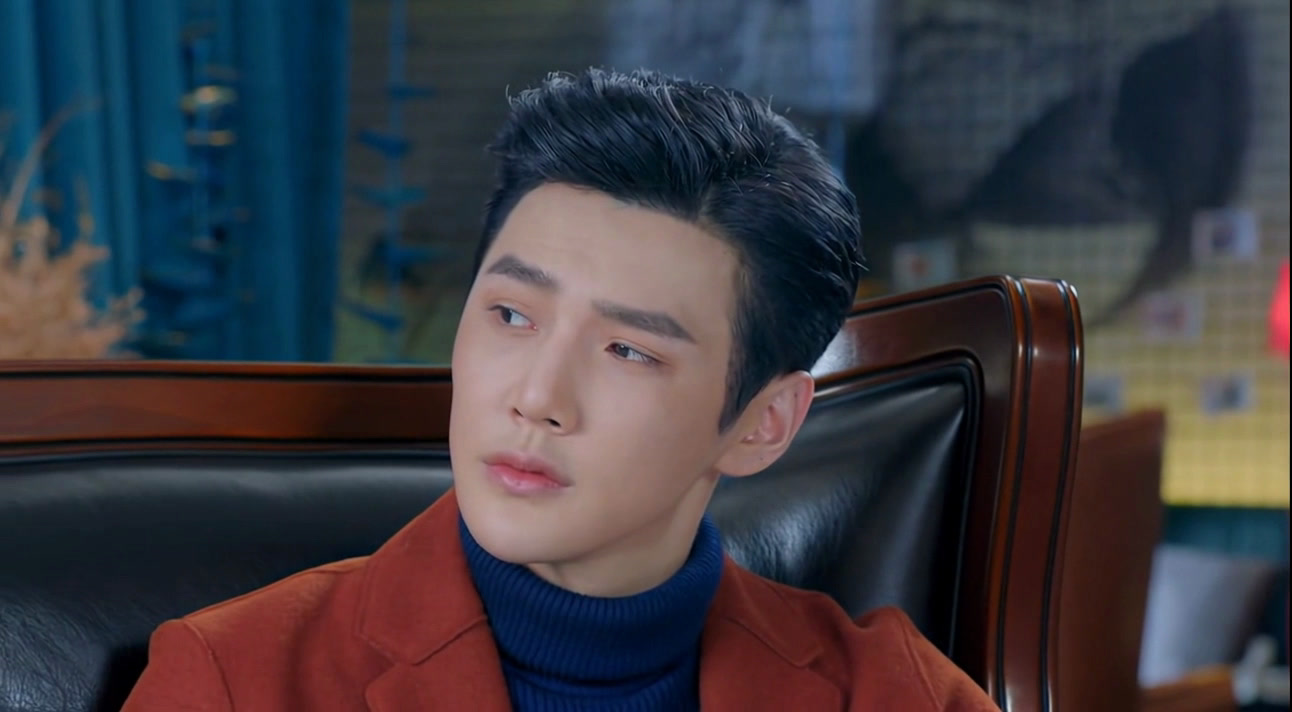} \\
    \end{tabular}
\end{flushleft}



\begin{flushleft}
    \textbf{Instruction:} \textit{Change the man's sleeveless shirt in @video\_1 to a blue Polo shirt style.}%
    \vspace{0.5em}
    \begin{tabular}{m{1.2em} m{0.22\linewidth} m{0.22\linewidth} m{0.22\linewidth}}

        \centering\rotatebox{90}{\textbf{Input Video}} &
        \includegraphics[width=\linewidth]{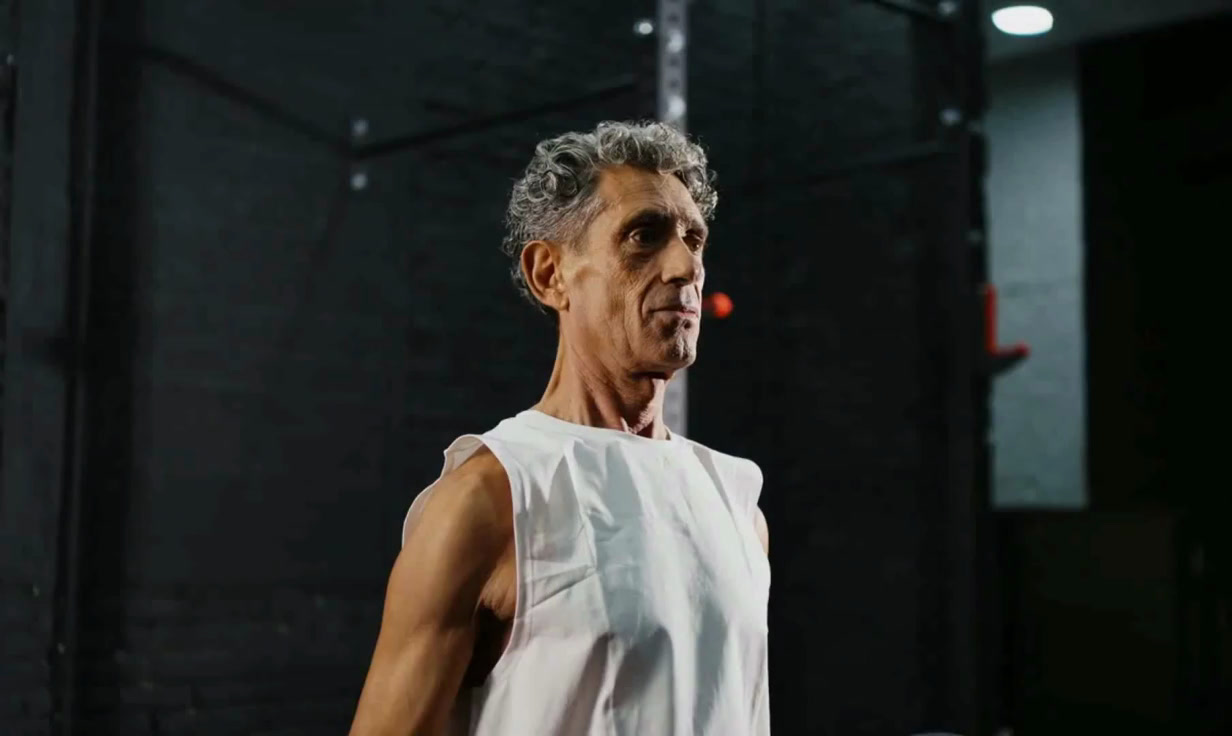} &
        \includegraphics[width=\linewidth]{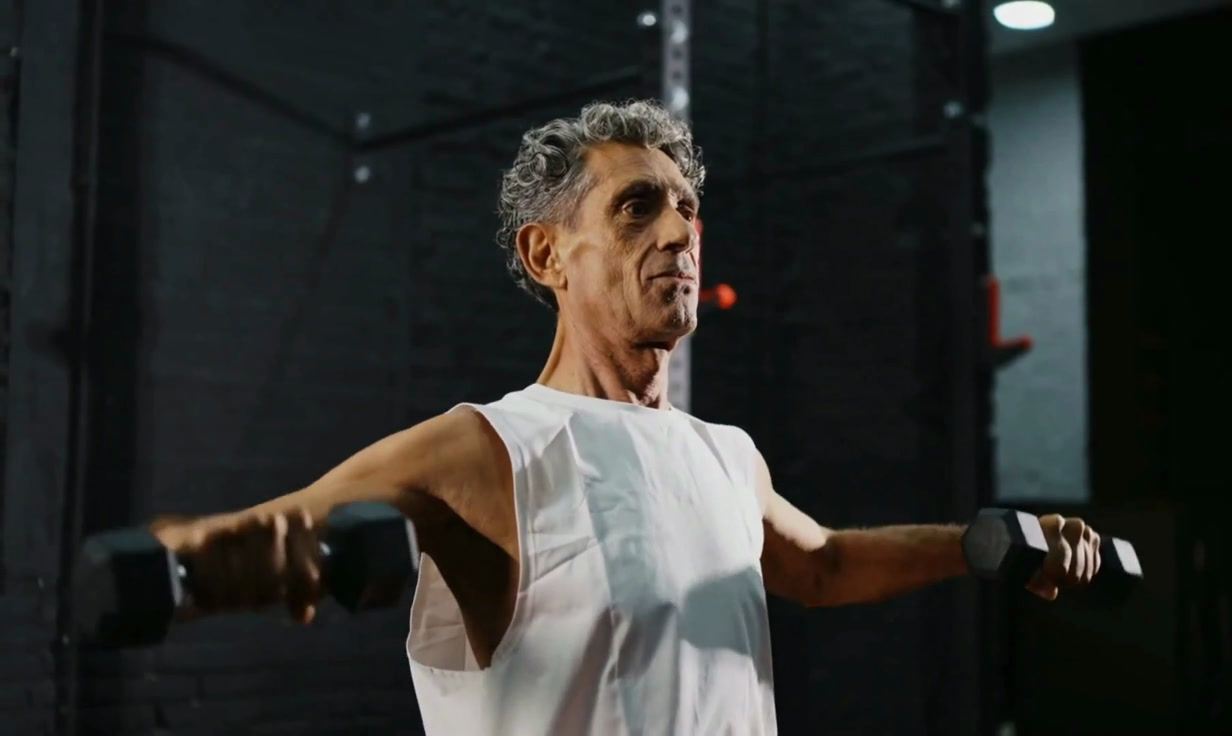} &
        \includegraphics[width=\linewidth]{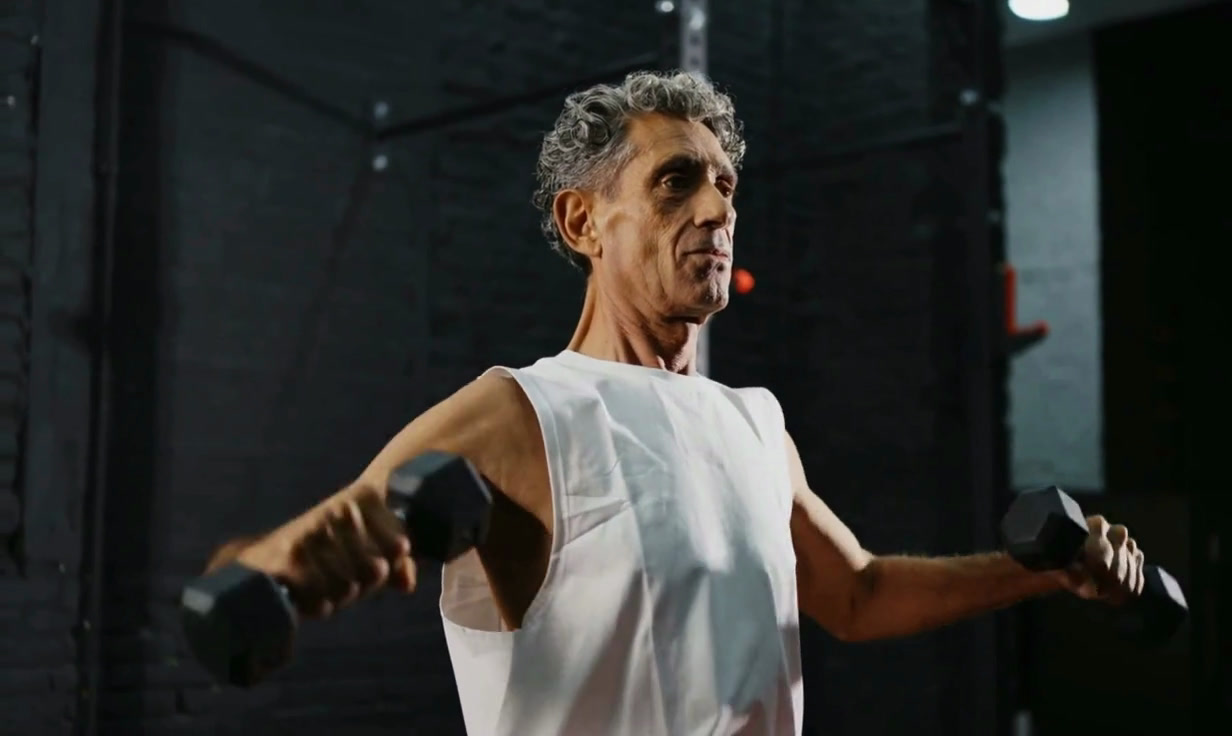} \\[2pt]

        \centering\rotatebox{90}{\textbf{Output Video}} &
        \includegraphics[width=\linewidth]{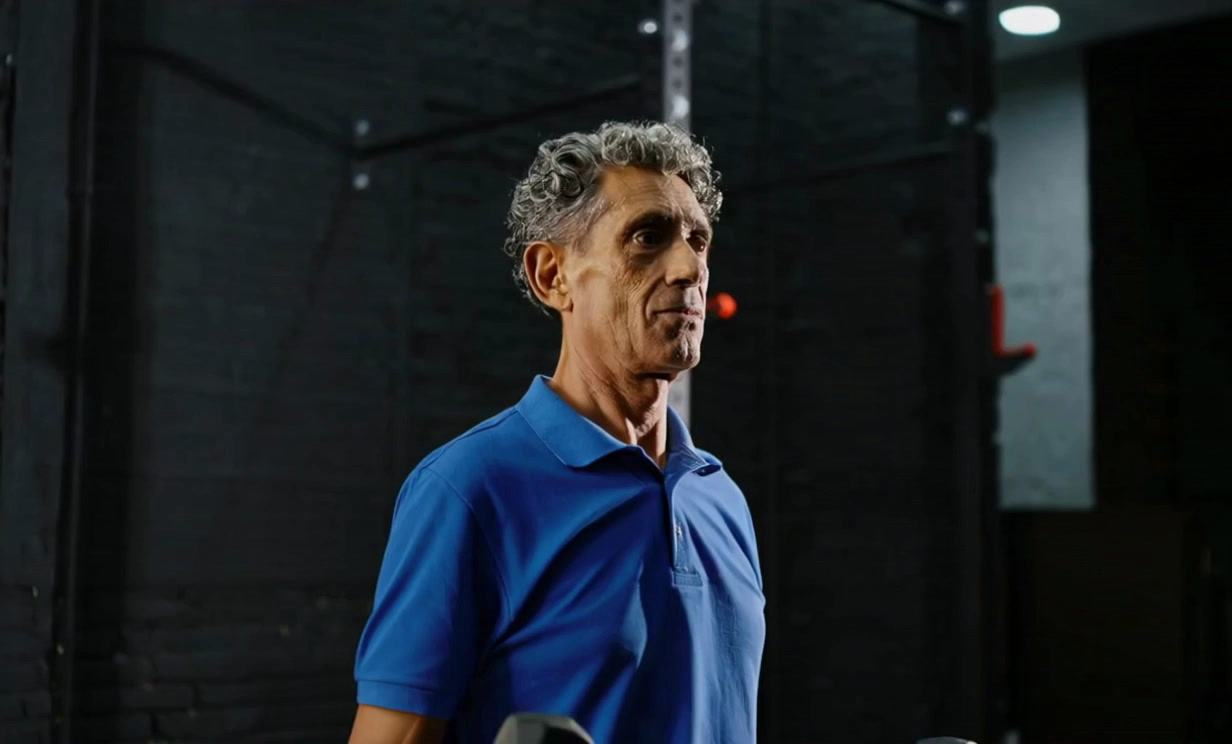} &
        \includegraphics[width=\linewidth]{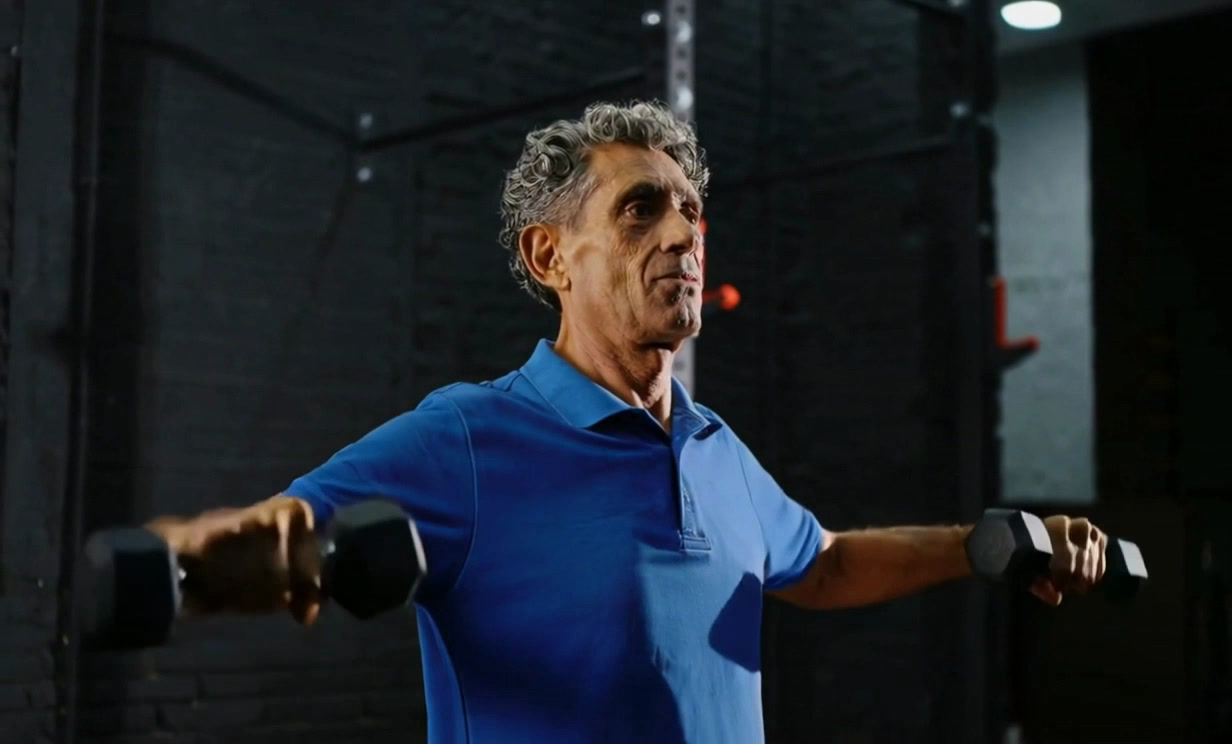} &
        \includegraphics[width=\linewidth]{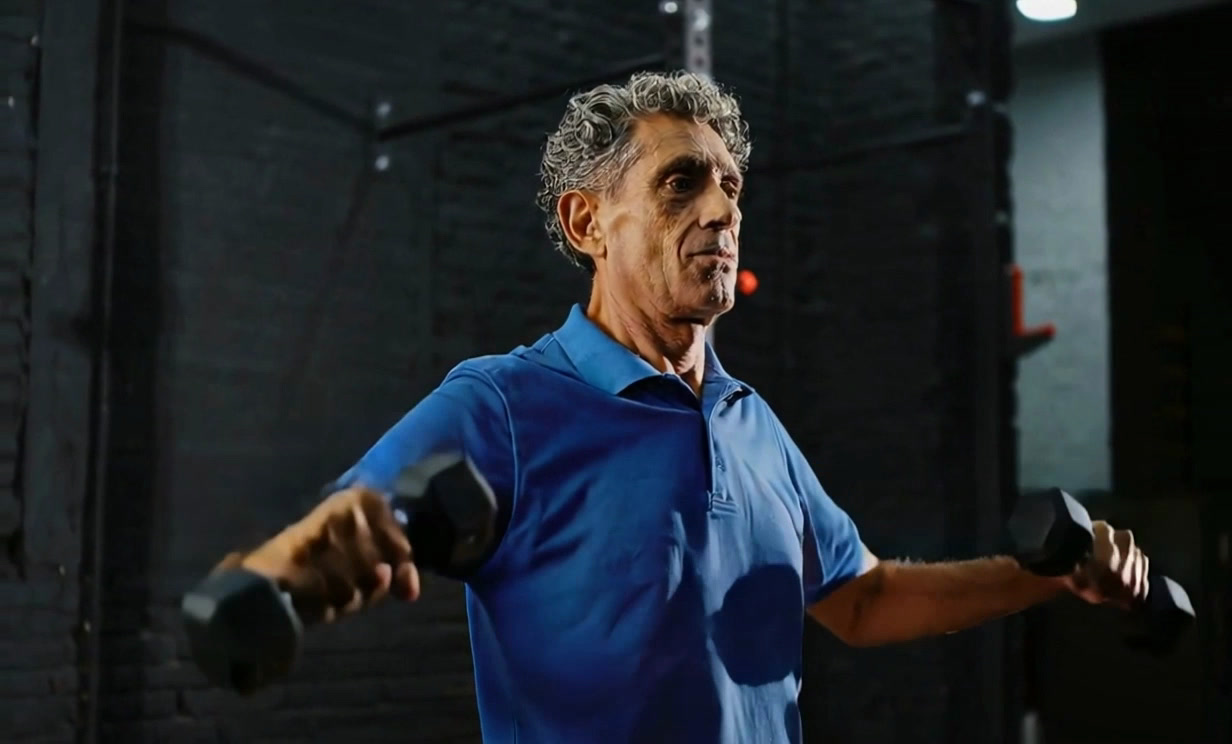} \\
    \end{tabular}
\end{flushleft}



\paragraph{Background Editing}

The model supports modifying background elements while preserving the foreground subjects.

\begin{flushleft}
    \textbf{Instruction:} \textit{Replace the background of @video\_1 with a post-rain European cobblestone street scene at dusk.}%
    \vspace{0.5em}
    \begin{tabular}{m{1.2em} m{0.22\linewidth} m{0.22\linewidth} m{0.22\linewidth}}

        \centering\rotatebox{90}{\textbf{Input Video}} &
        \includegraphics[width=\linewidth]{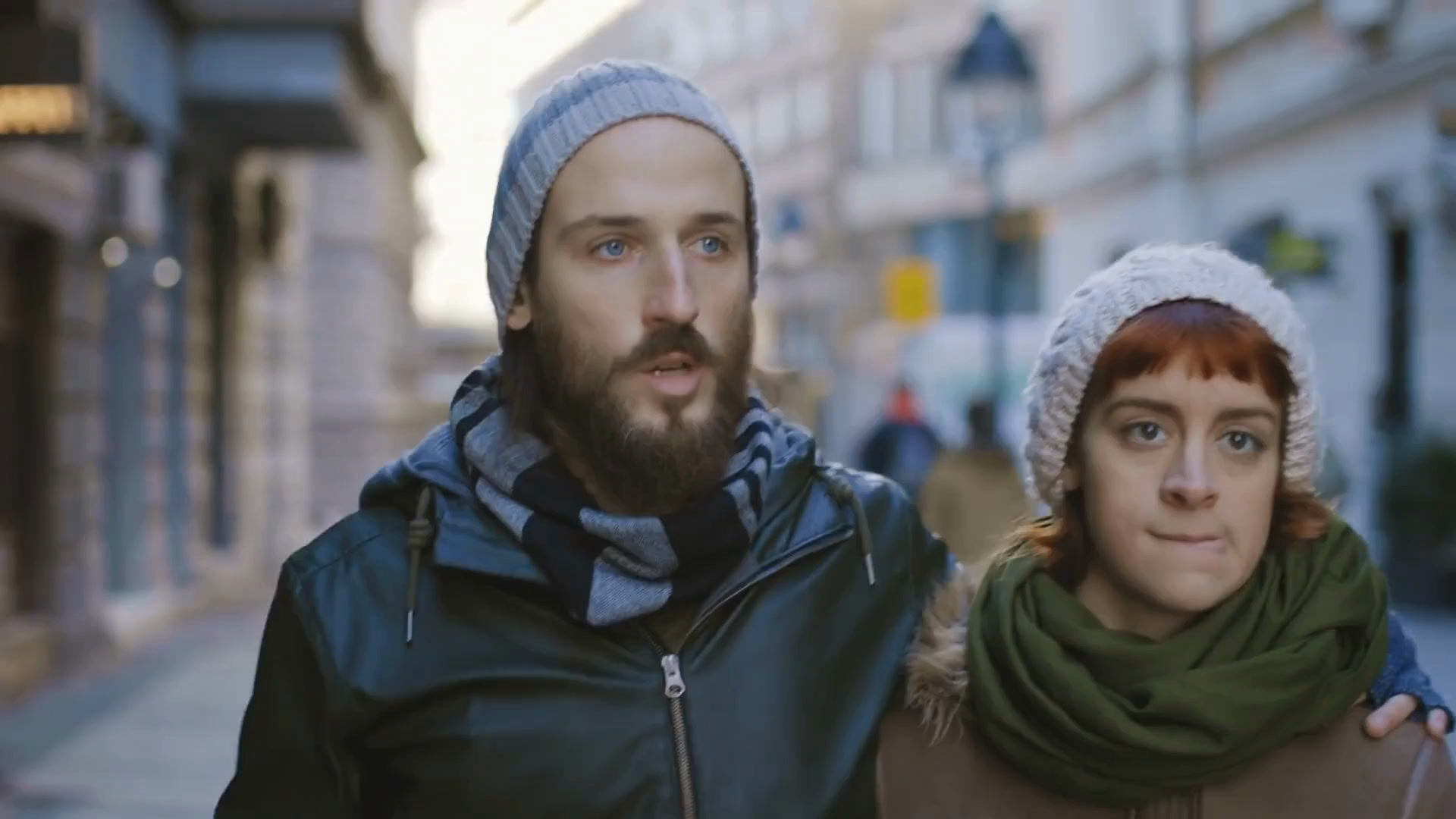} &
        \includegraphics[width=\linewidth]{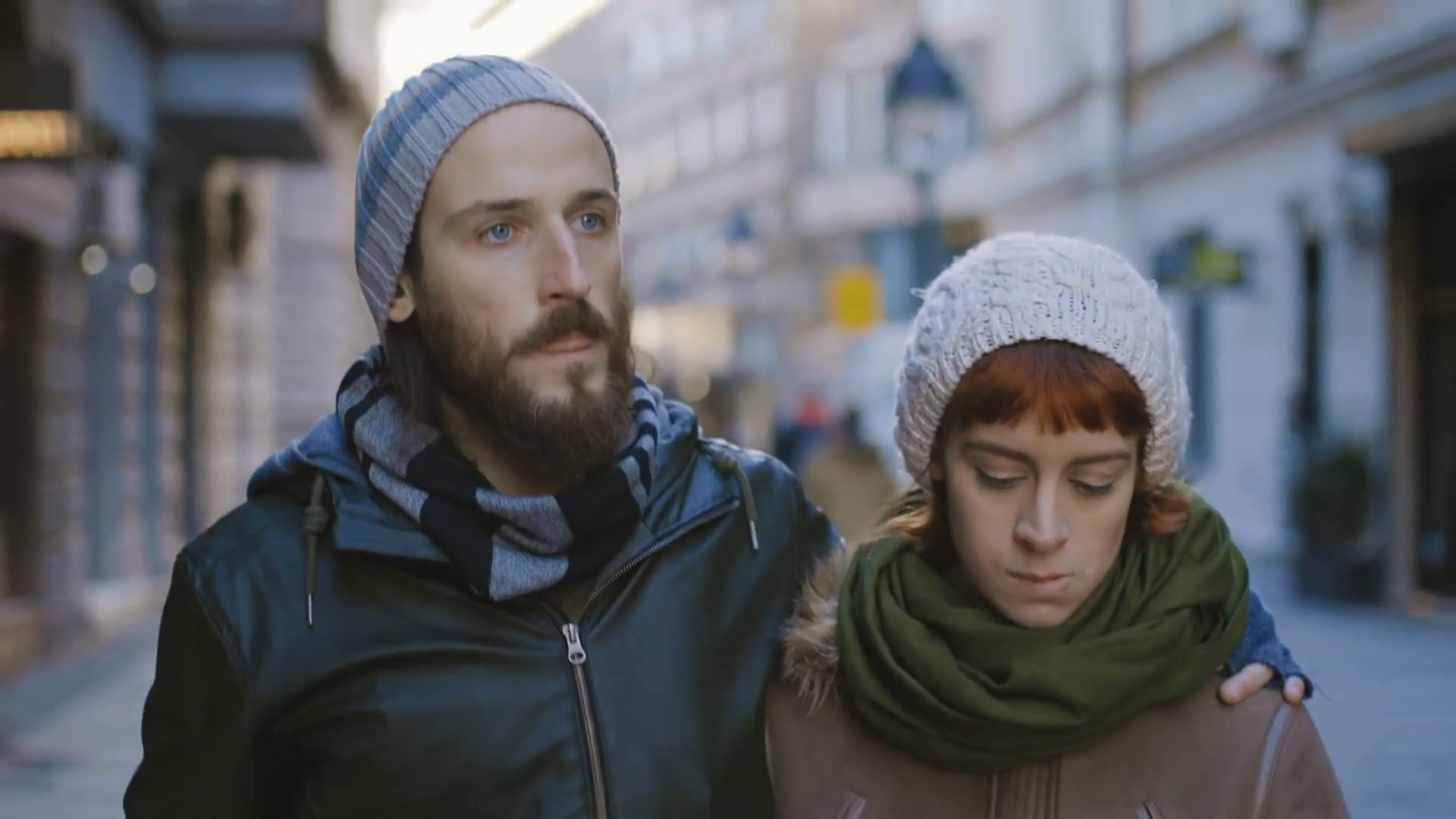} &
        \includegraphics[width=\linewidth]{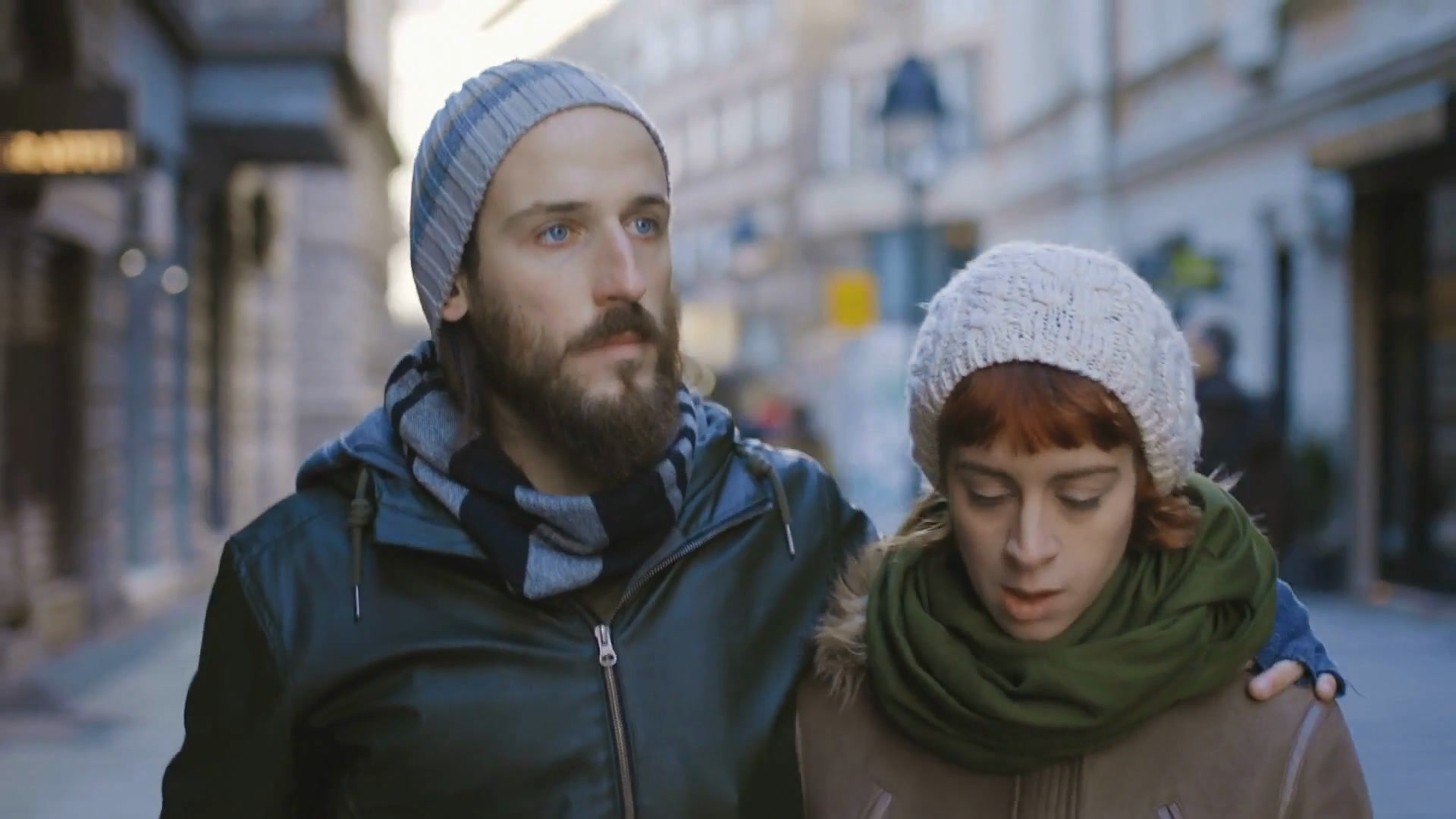} \\[2pt]

        \centering\rotatebox{90}{\textbf{Output Video}} &
        \includegraphics[width=\linewidth]{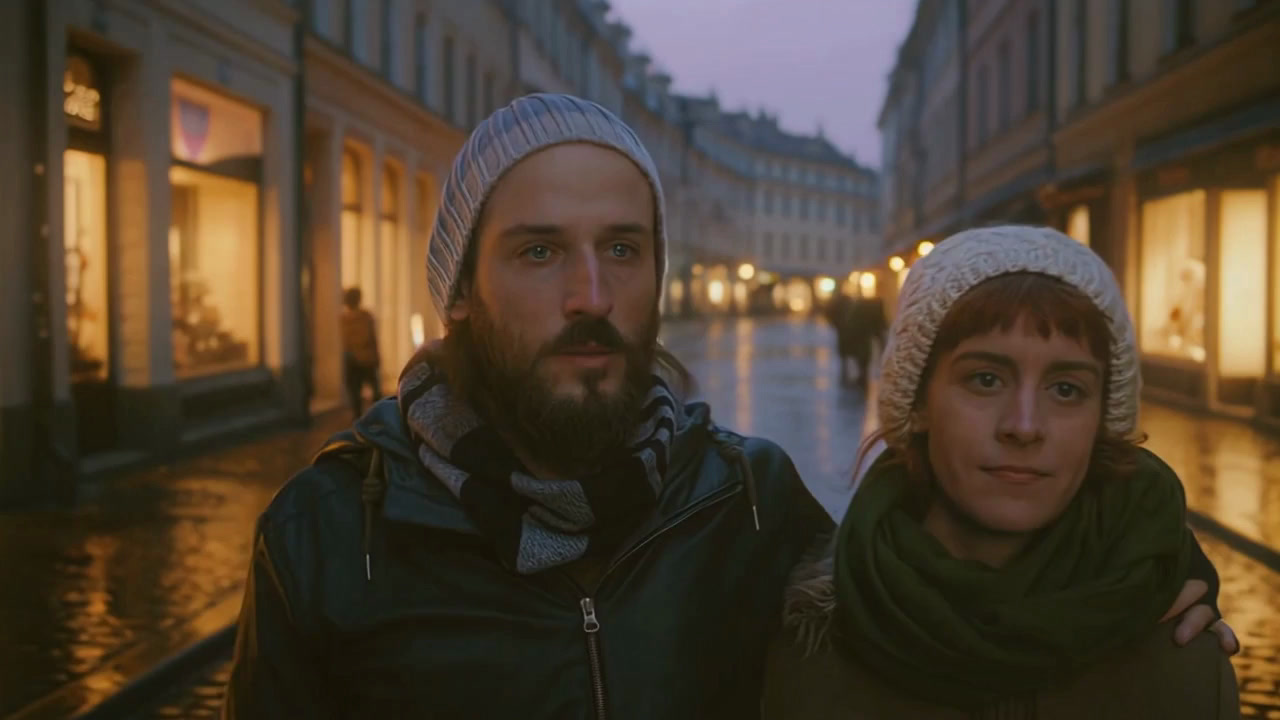} &
        \includegraphics[width=\linewidth]{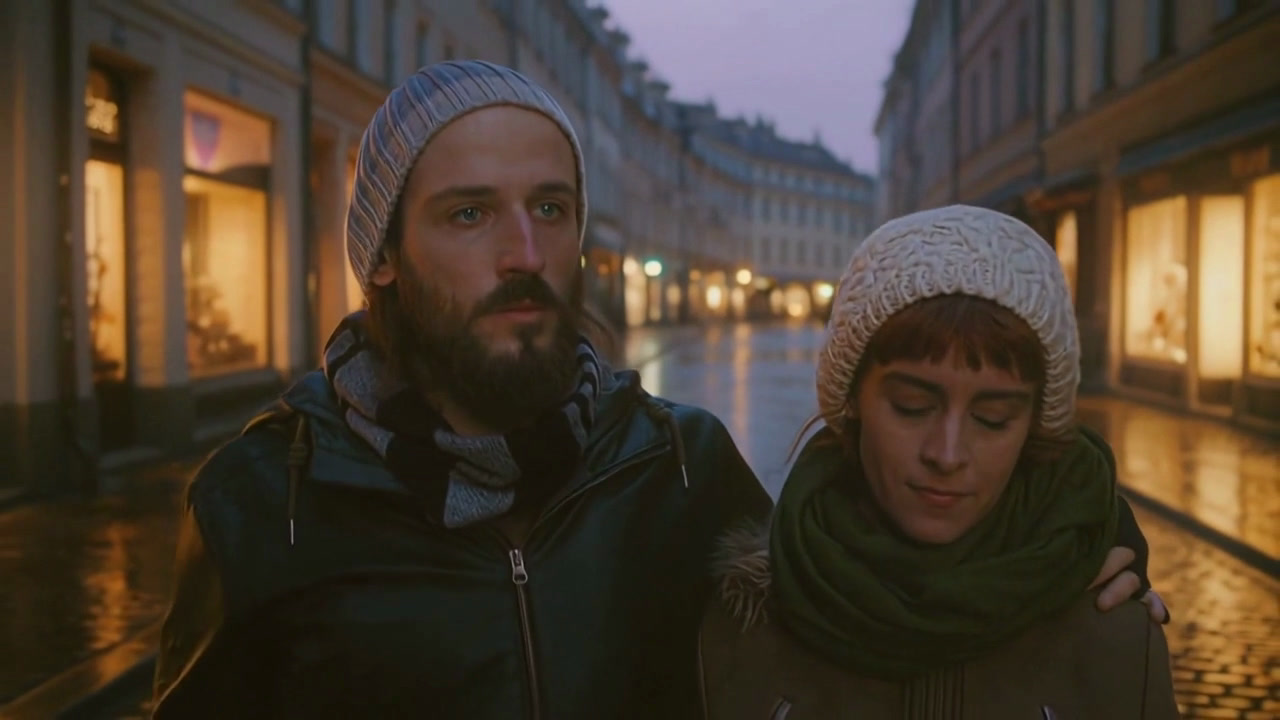} &
        \includegraphics[width=\linewidth]{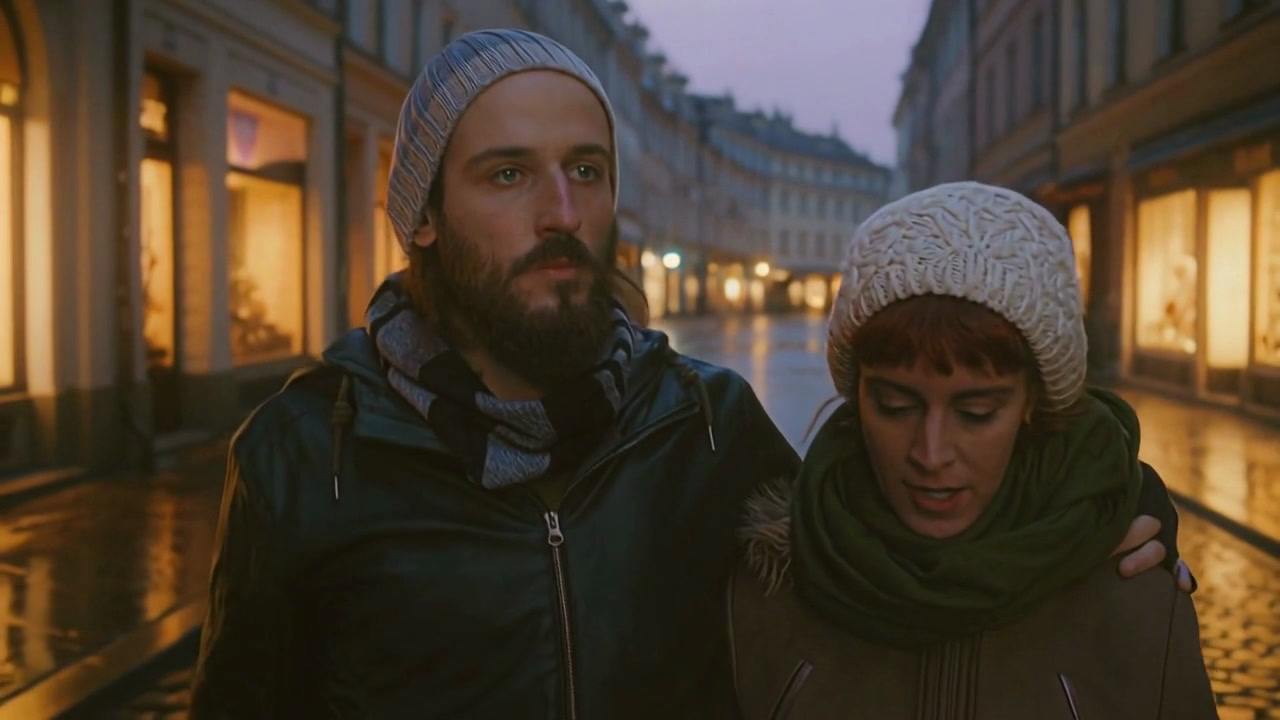} \\
    \end{tabular}
\end{flushleft}

\begin{figure}[h!]
    \centering
    \caption{Examples of local attribute editing.}
    \label{fig:local_attr_edit}
\end{figure}

\newpage
\subsubsection{Global Editing}
\label{appendix:global-editing}

The model supports global modifications that affect the entire video, including style, camera properties, and scene attributes.

\paragraph{Style Transfer}
\label{appendix:style-transfer}

The model can transform videos into different artistic or visual styles while maintaining semantic consistency of the video content.

\begin{flushleft}
    \textbf{Instruction:} \textit{Transform @video\_1 into Paper-Cutting style.}%
    \vspace{0.5em}
    \begin{tabular}{m{1.2em} m{0.22\linewidth} m{0.22\linewidth} m{0.22\linewidth}}

        \centering\rotatebox{90}{\textbf{Input Video}} &
        \includegraphics[width=\linewidth]{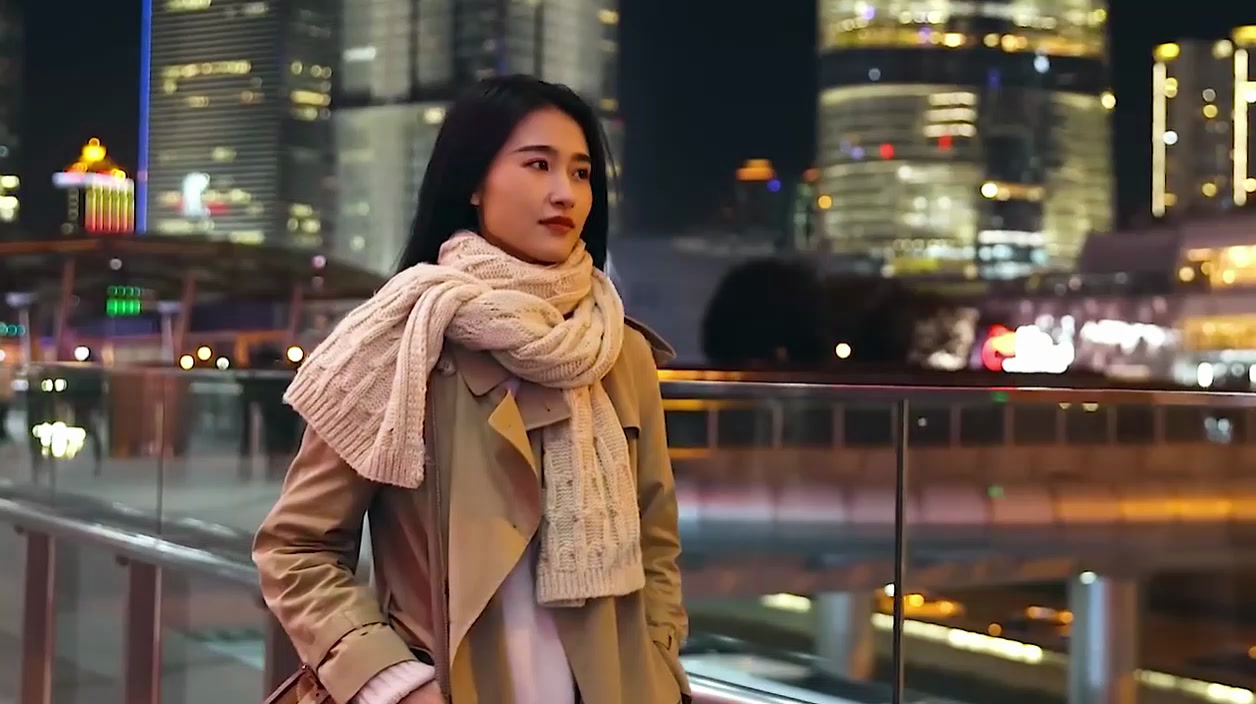} &
        \includegraphics[width=\linewidth]{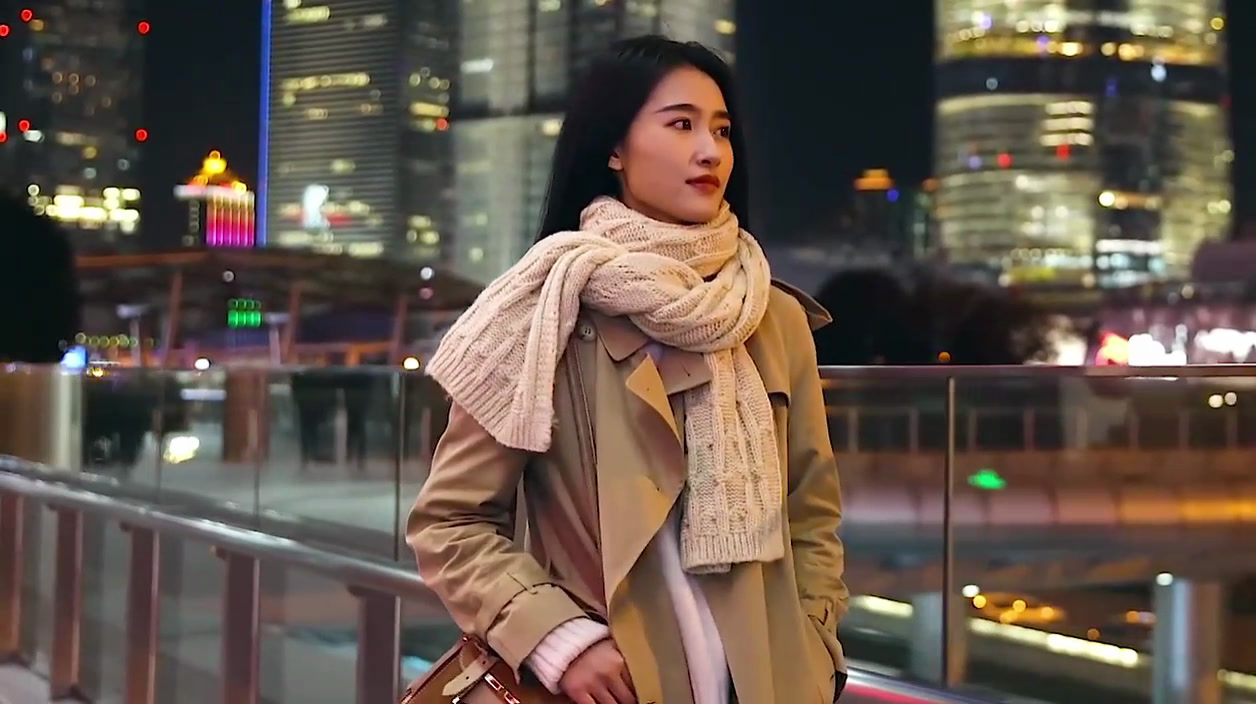} &
        \includegraphics[width=\linewidth]{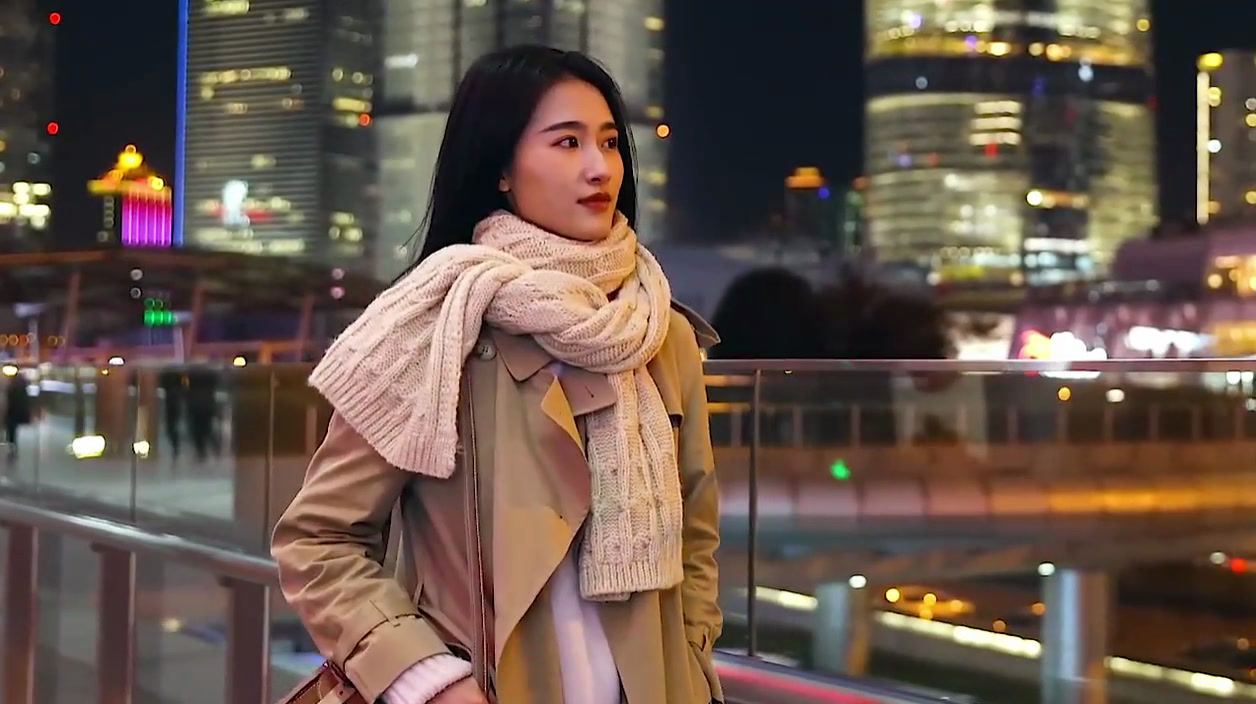} \\[2pt]

        \centering\rotatebox{90}{\textbf{Output Video}} &
        \includegraphics[width=\linewidth]{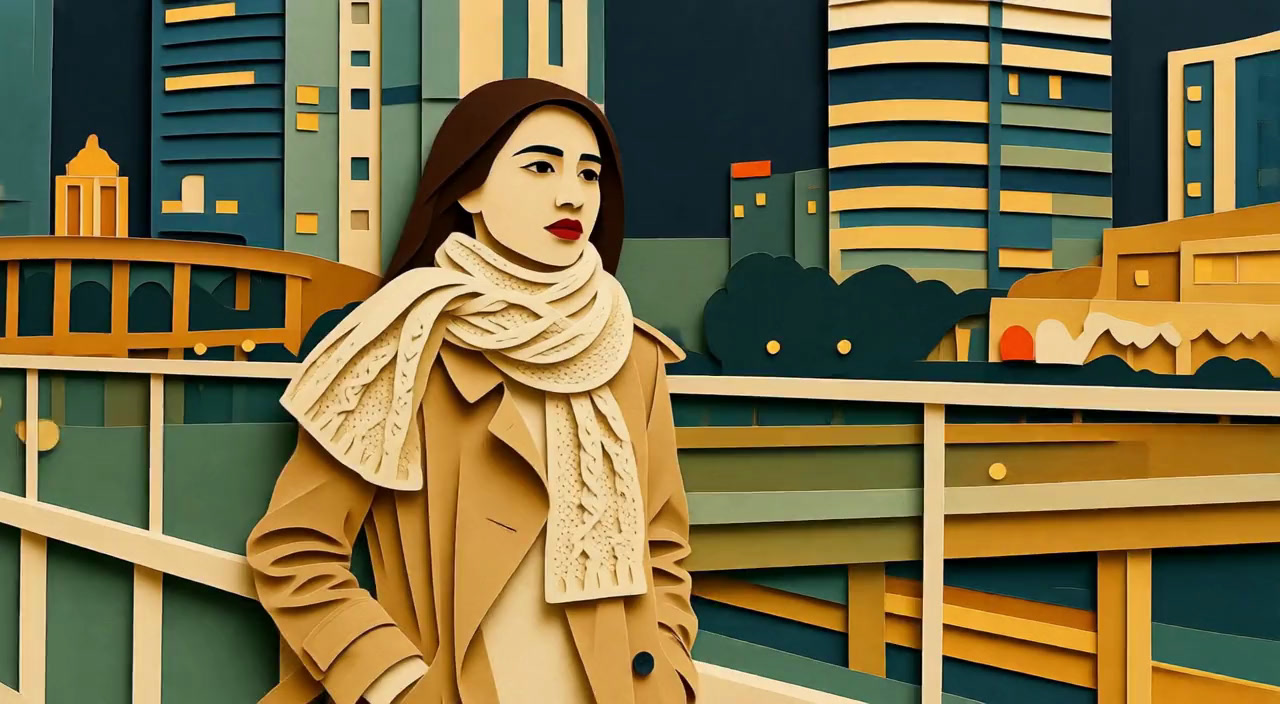} &
        \includegraphics[width=\linewidth]{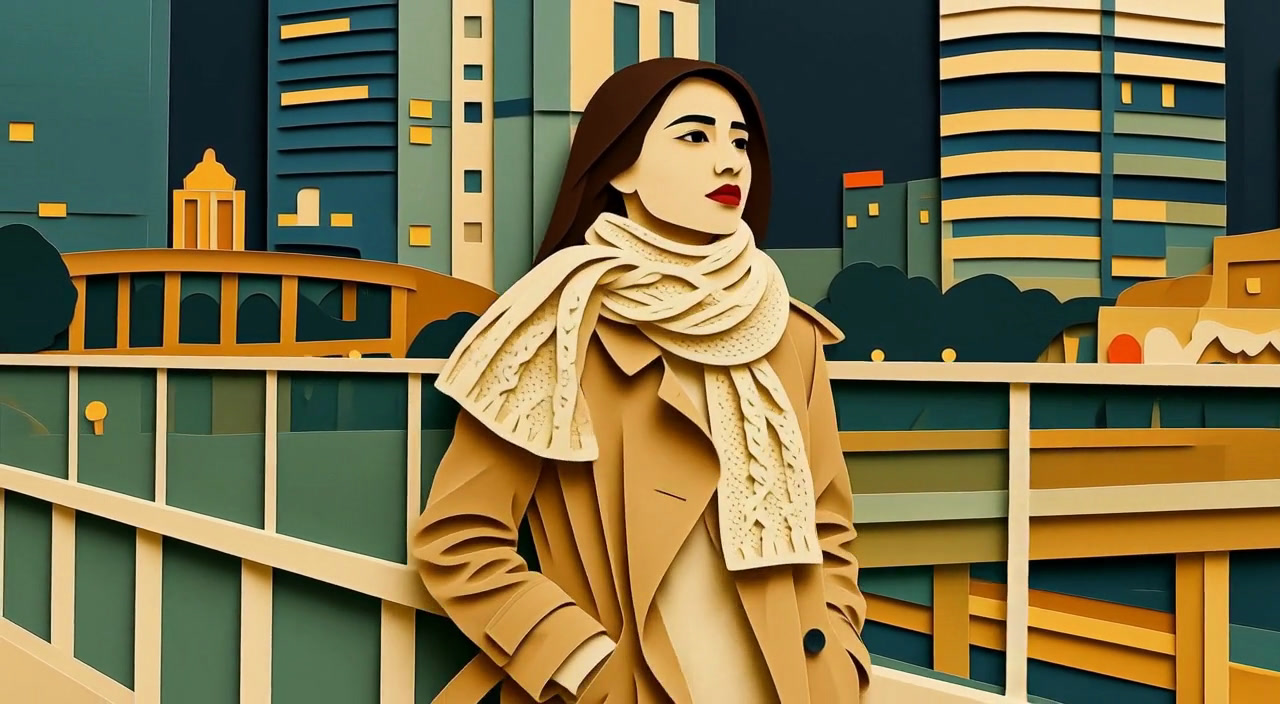} &
        \includegraphics[width=\linewidth]{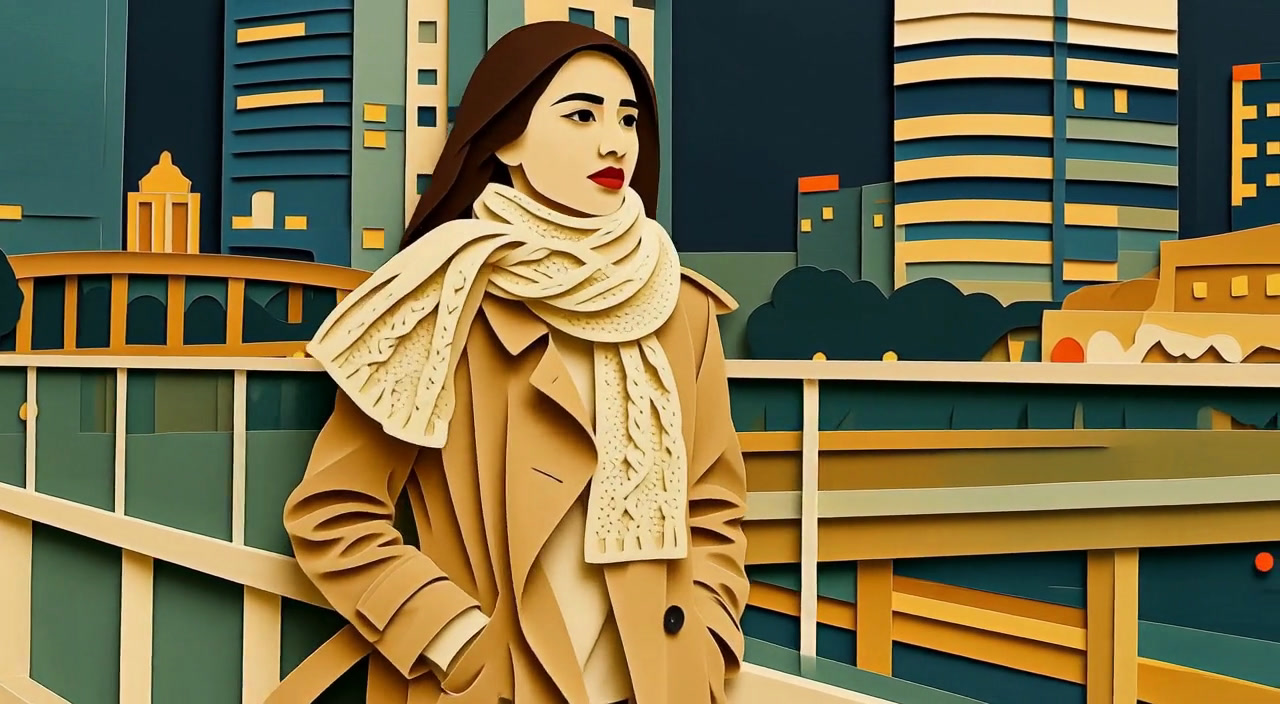} \\
    \end{tabular}
\end{flushleft}



\begin{flushleft}
    \textbf{Instruction:} \textit{Transform @video\_1 into LEGO style.}%
    \vspace{0.5em}
    \begin{tabular}{m{1.2em} m{0.22\linewidth} m{0.22\linewidth} m{0.22\linewidth}}

        \centering\rotatebox{90}{\textbf{Input Video}} &
        \includegraphics[width=\linewidth]{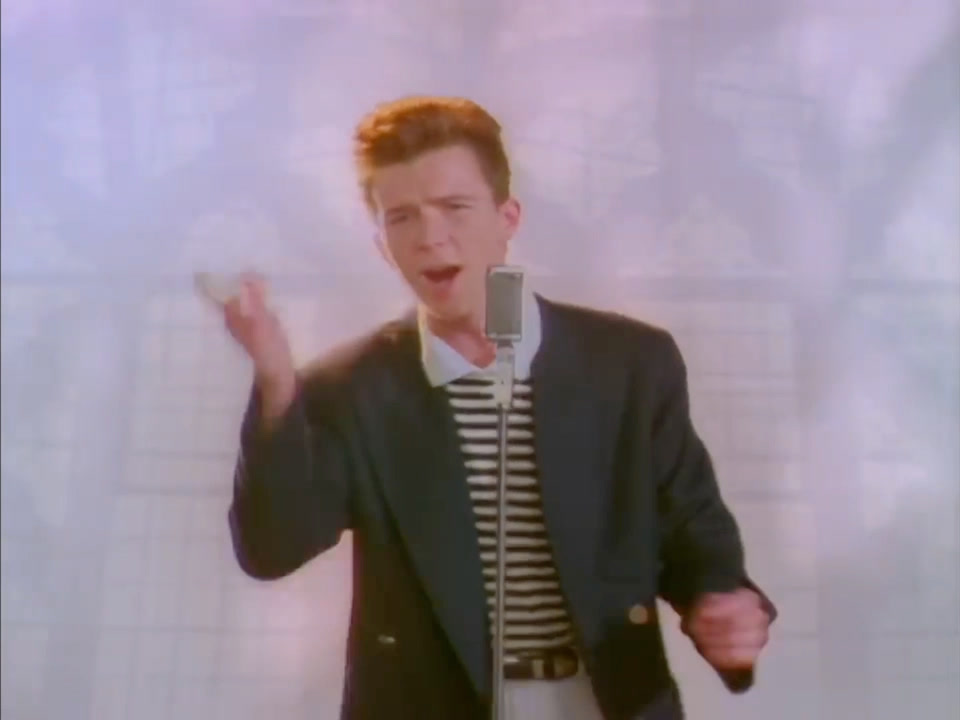} &
        \includegraphics[width=\linewidth]{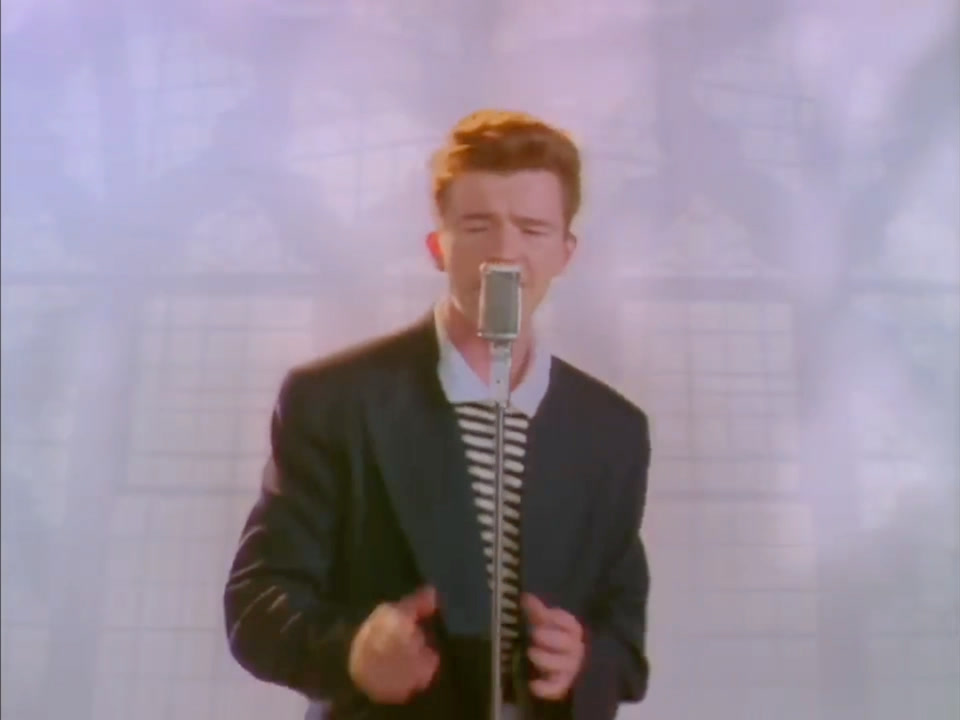} &
        \includegraphics[width=\linewidth]{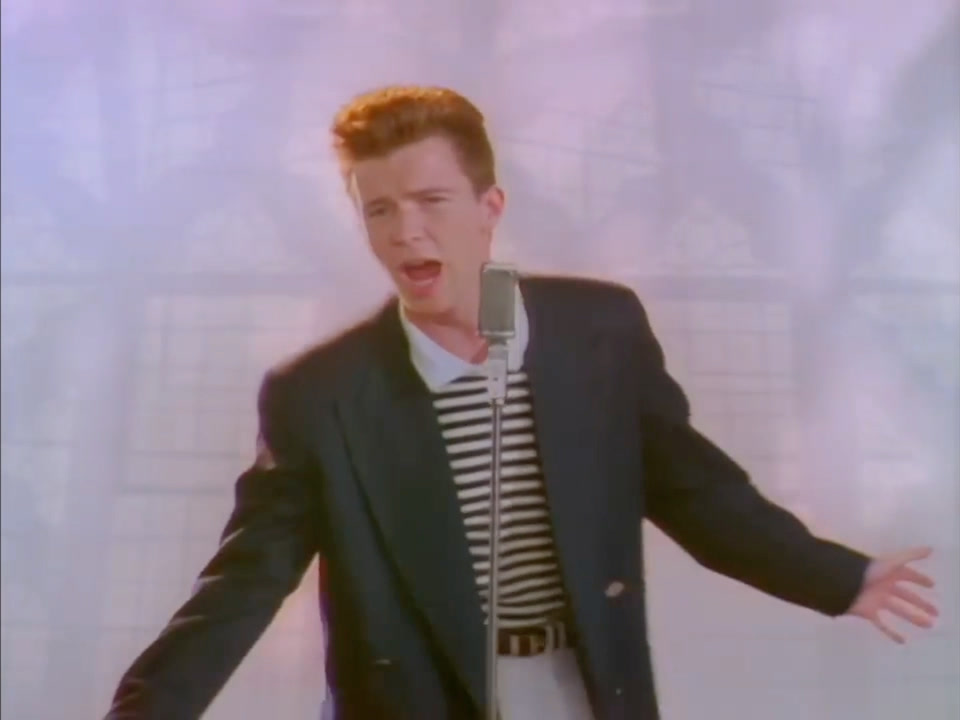} \\[2pt]

        \centering\rotatebox{90}{\textbf{Output Video}} &
        \includegraphics[width=\linewidth]{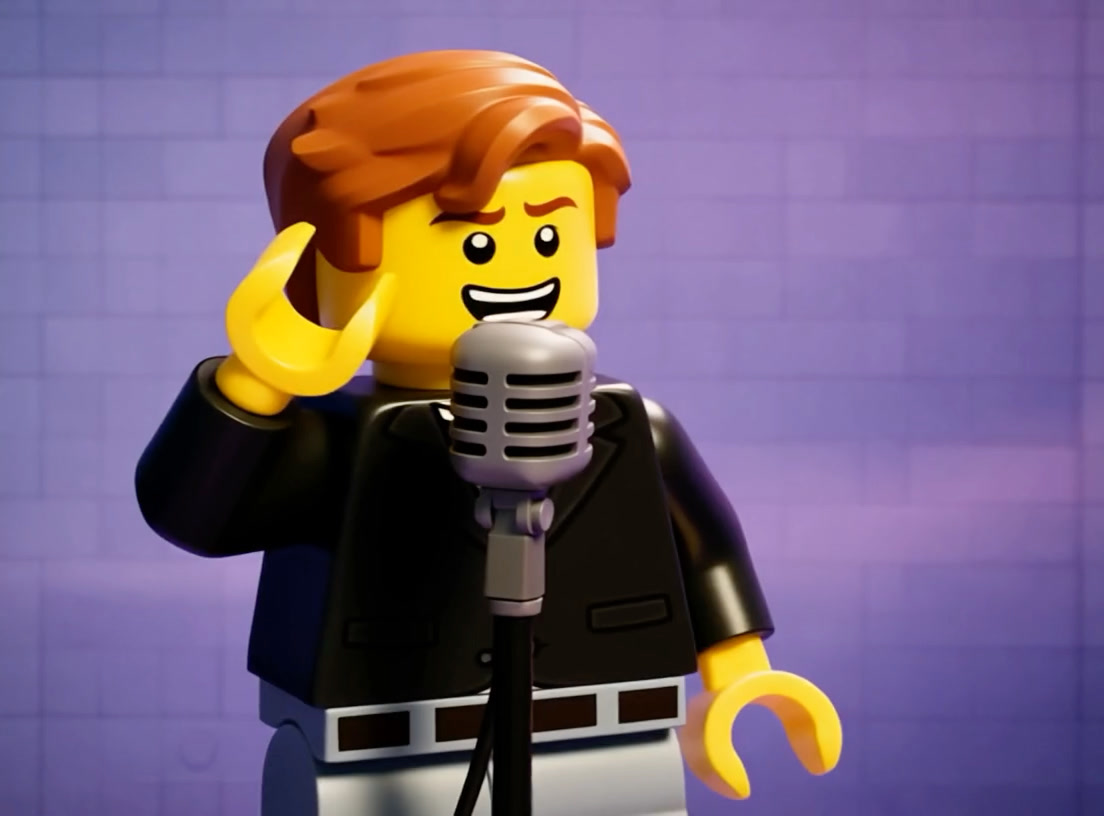} &
        \includegraphics[width=\linewidth]{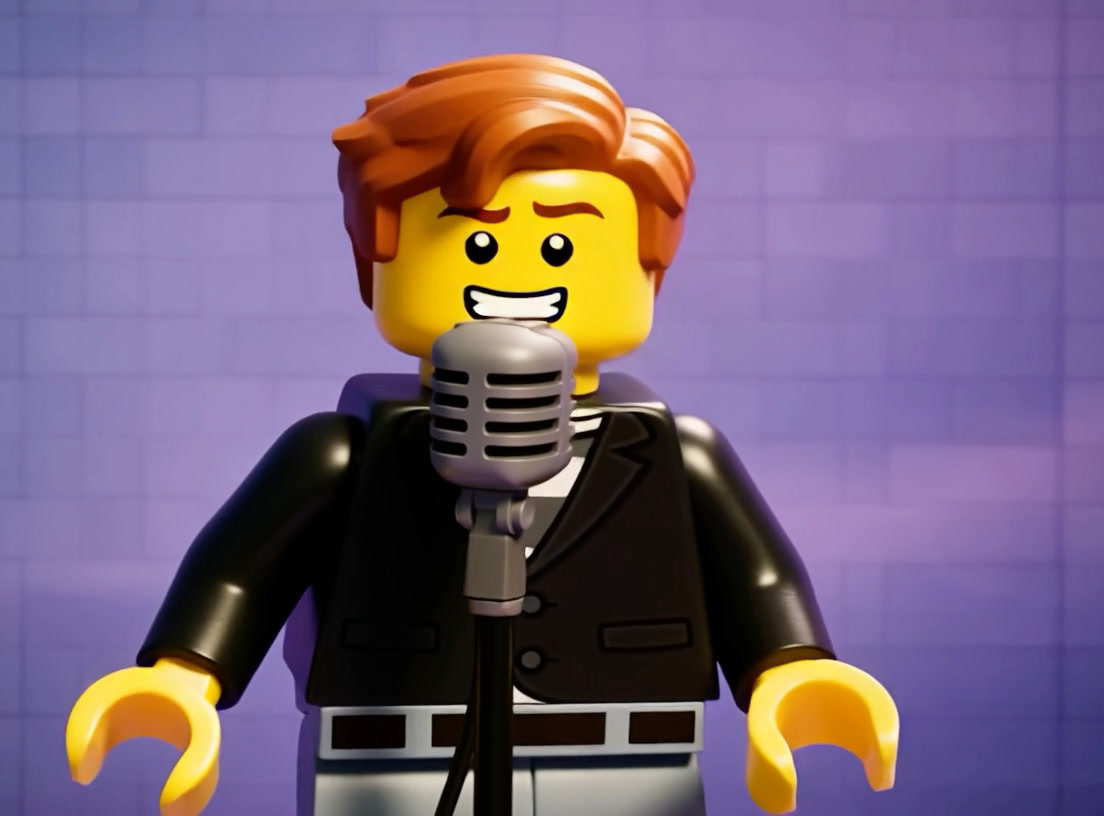} &
        \includegraphics[width=\linewidth]{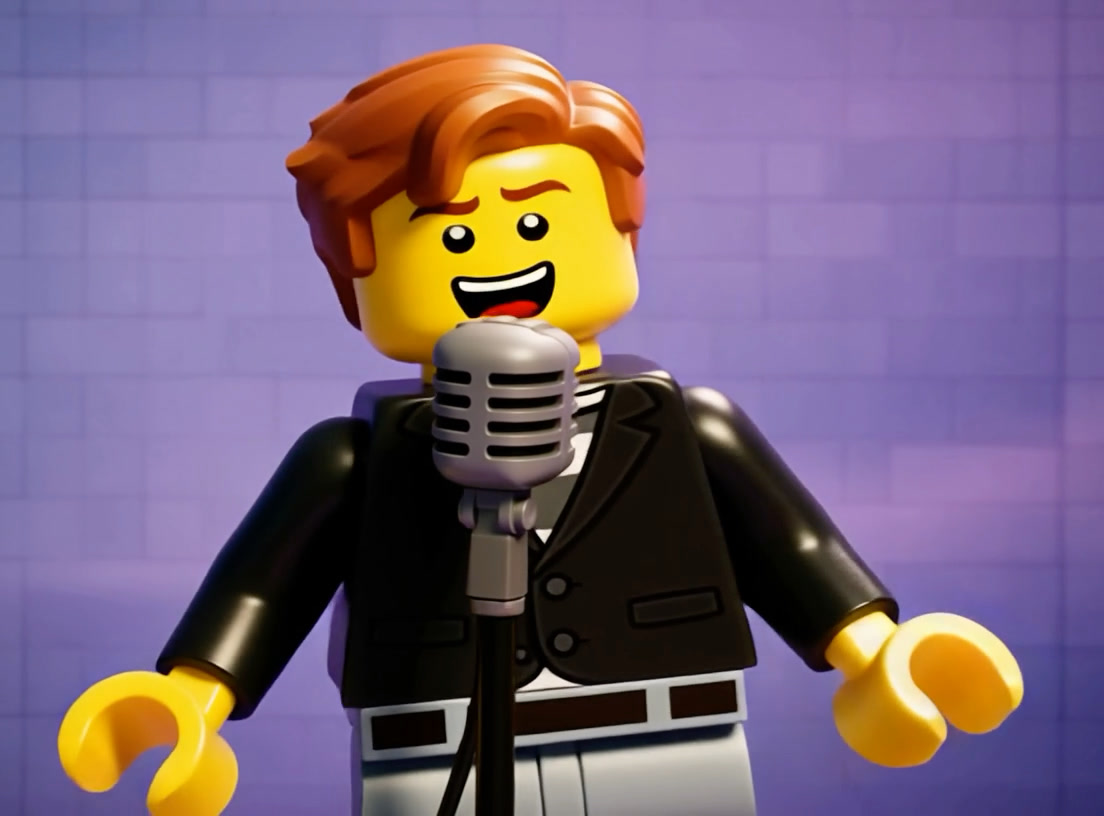} \\
    \end{tabular}
\end{flushleft}


\begin{figure}[h!]
    \centering
    \caption{Examples of style transfer.}
    \label{fig:edit_style}
\end{figure}

\newpage
\paragraph{Camera Control}

The model supports modifying camera properties including shot angle, shot type, and camera position.

\begin{flushleft}
    \textbf{Instruction:} \textit{Re-render @video\_1 with a Pan Right camera movement.}%
    \vspace{0.5em}
    \begin{tabular}{m{1.2em} m{0.22\linewidth} m{0.22\linewidth} m{0.22\linewidth}}

        \centering\rotatebox{90}{\textbf{Input Video}} &
        \includegraphics[width=\linewidth]{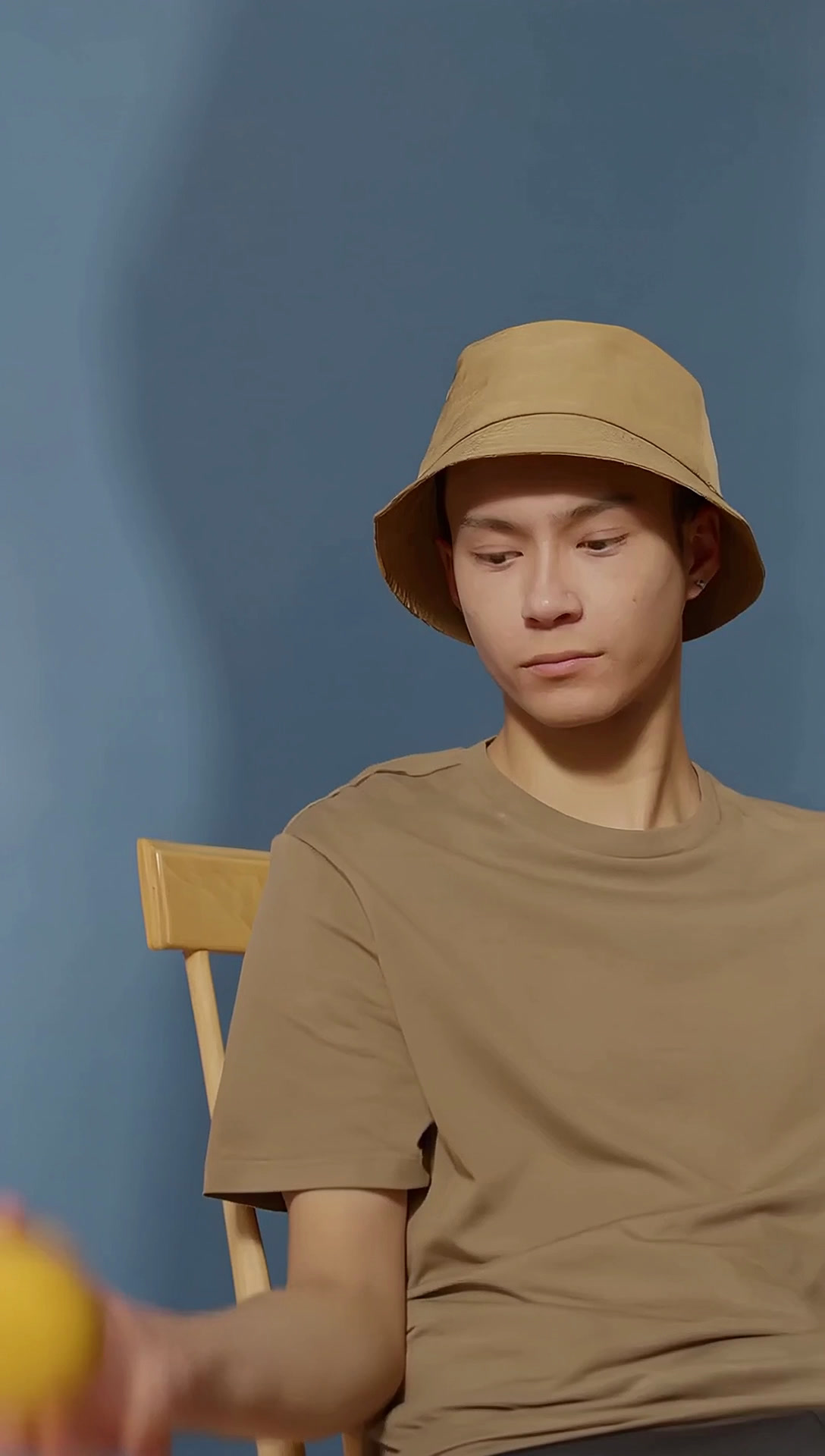} &
        \includegraphics[width=\linewidth]{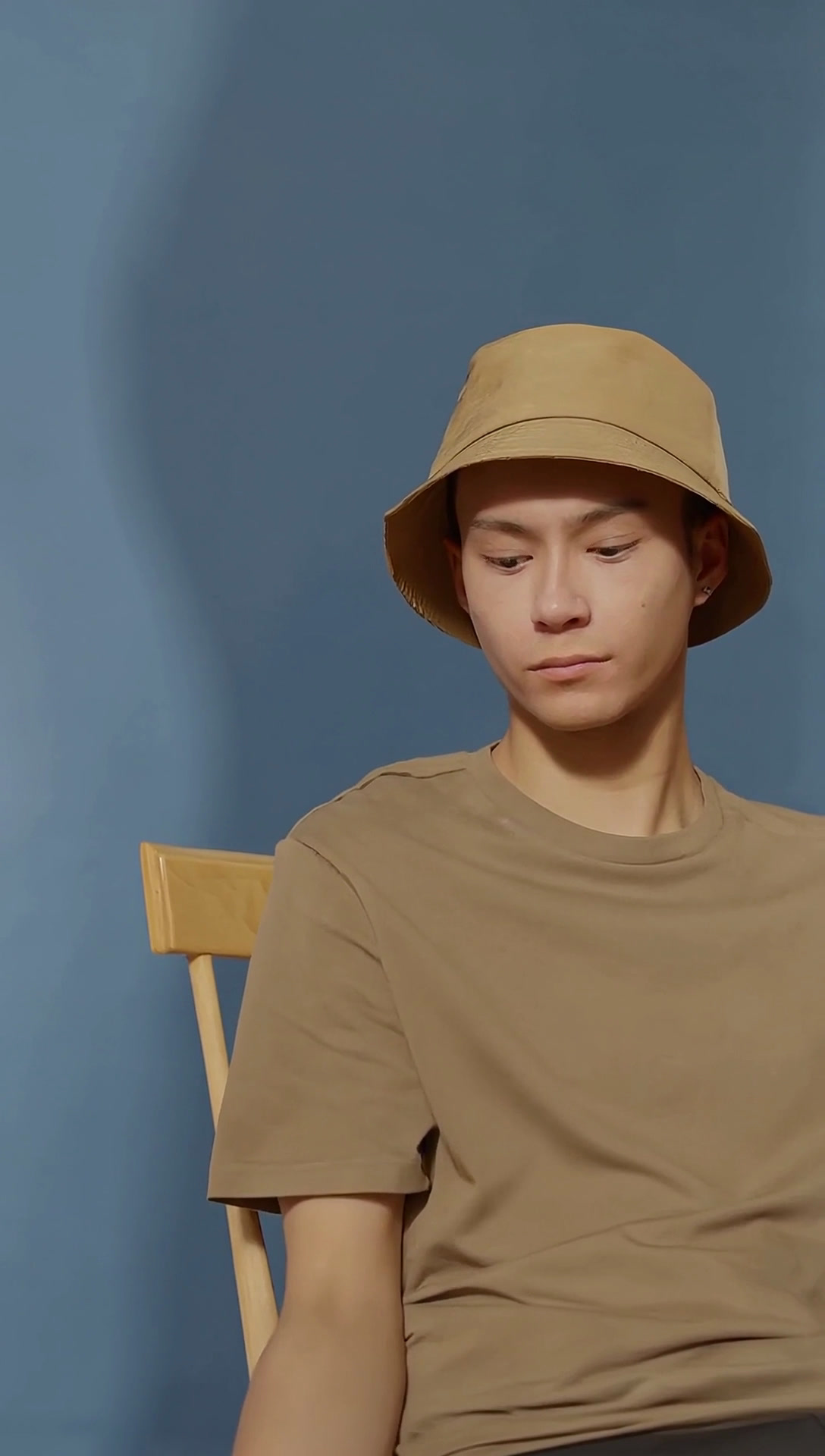} &
        \includegraphics[width=\linewidth]{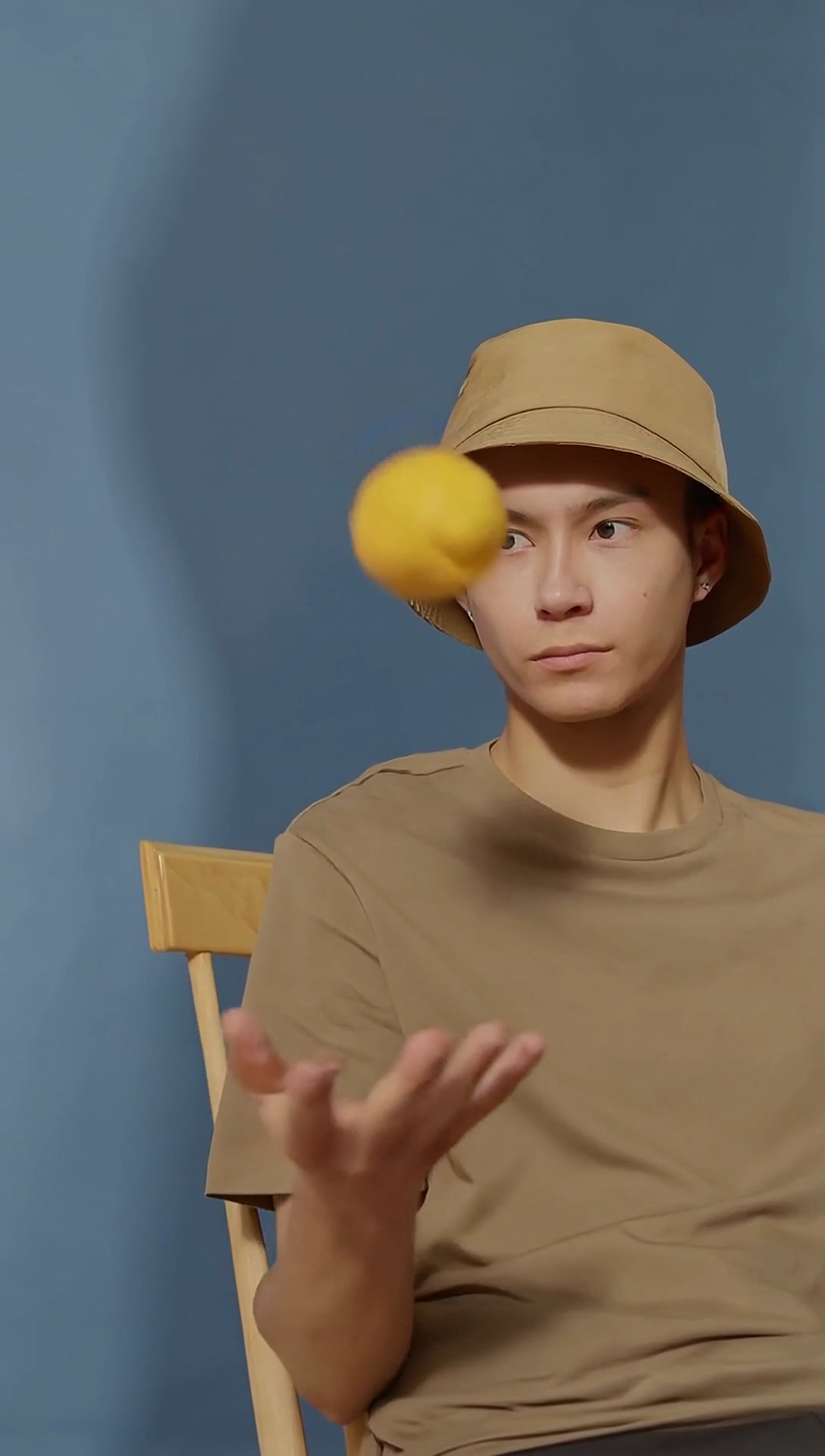} \\[2pt]

        \centering\rotatebox{90}{\textbf{Output Video}} &
        \includegraphics[width=\linewidth]{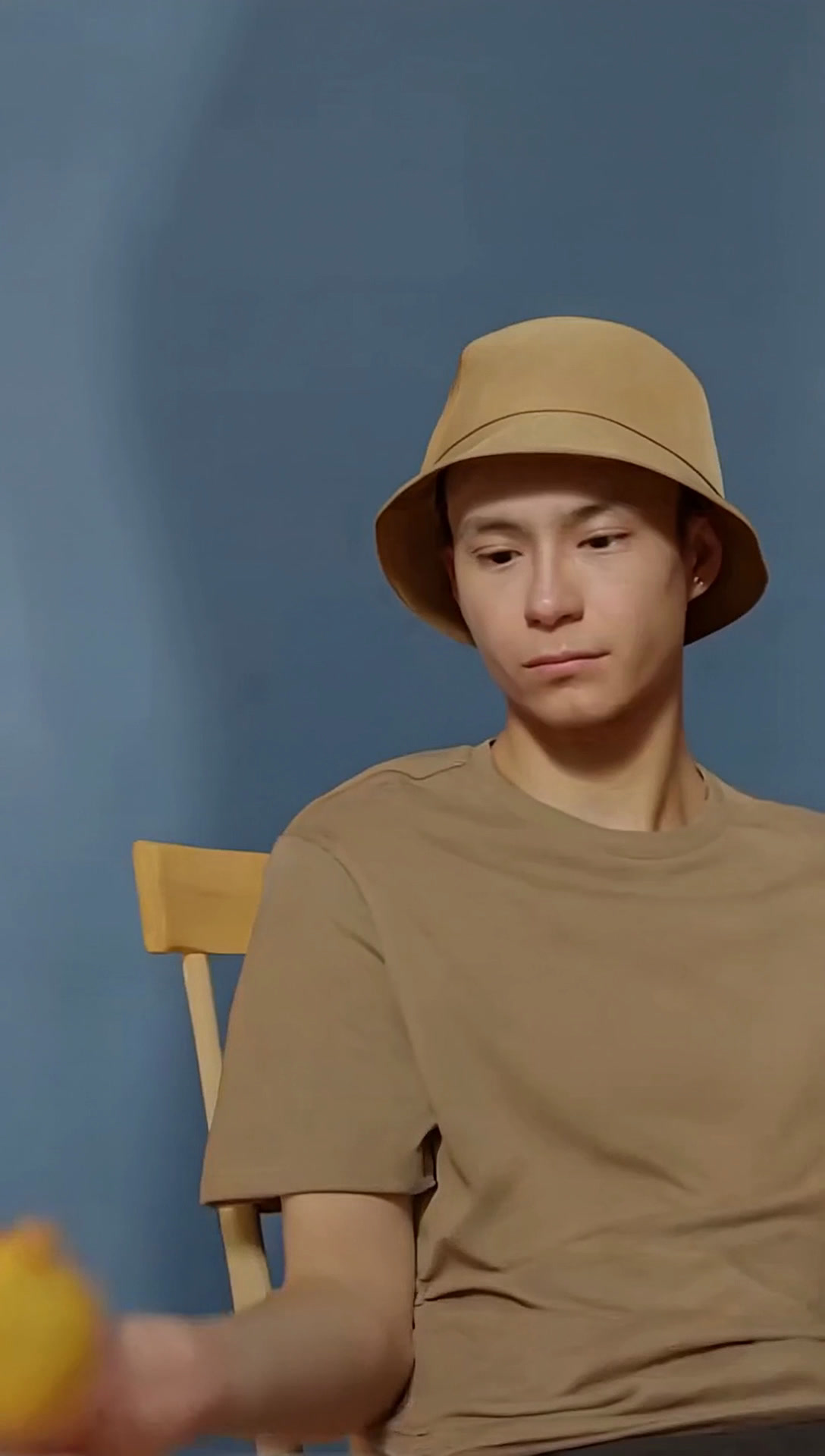} &
        \includegraphics[width=\linewidth]{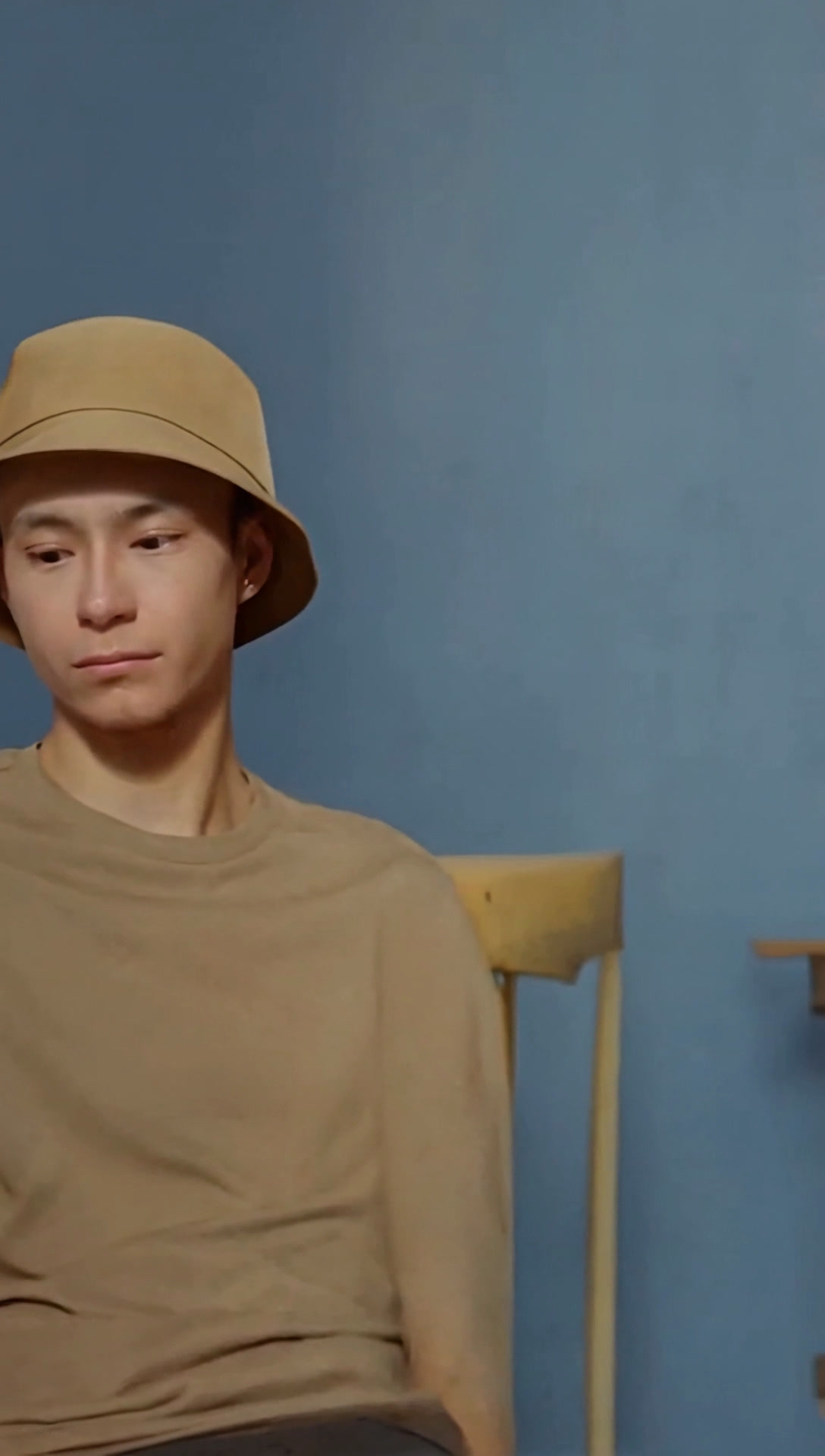} &
        \includegraphics[width=\linewidth]{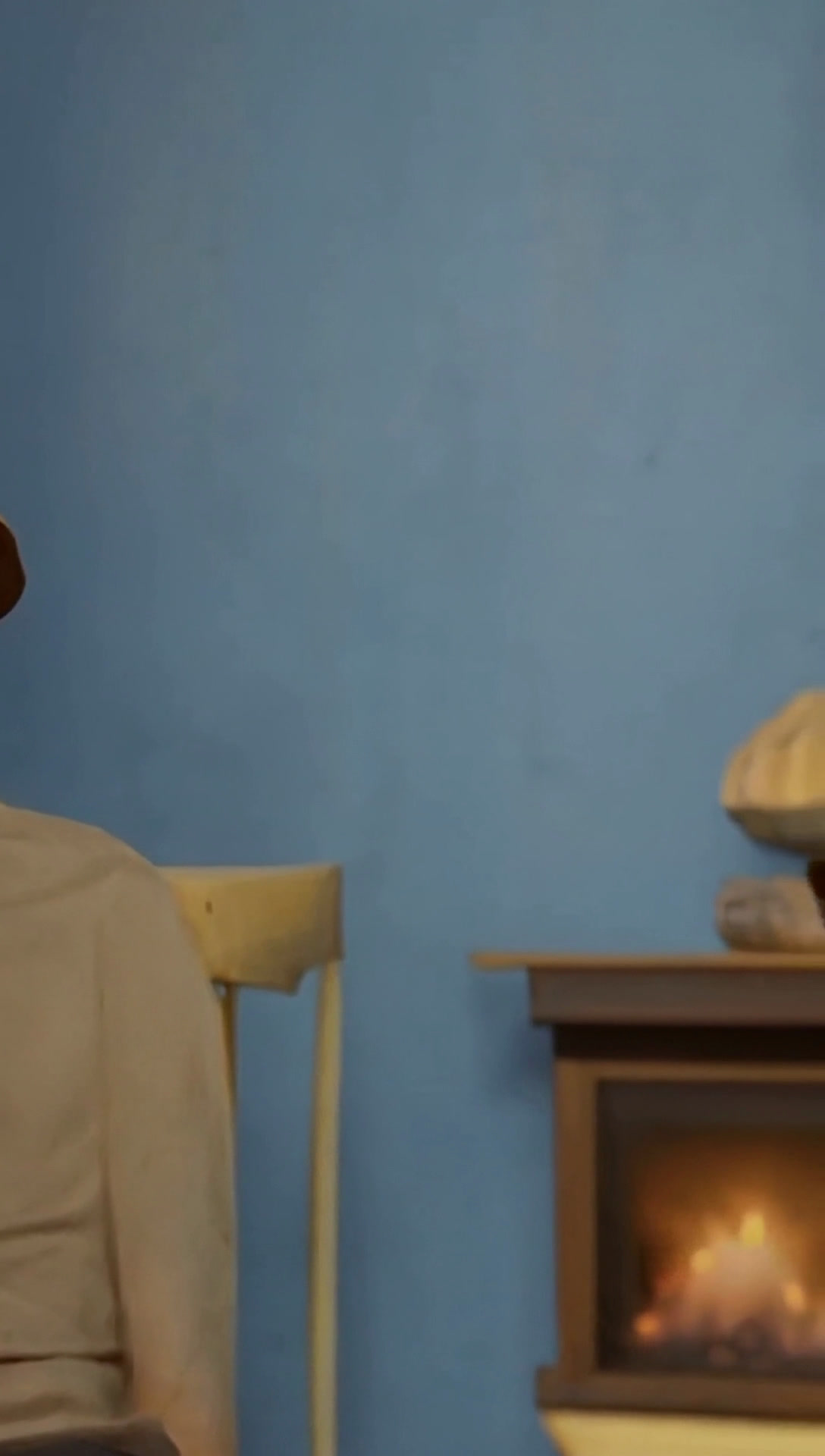} \\
    \end{tabular}
\end{flushleft}


\begin{figure}[h!]
    \centering
    \caption{Examples of camera control.}
    \label{fig:edit_camera}
\end{figure}

\newpage
\paragraph{Global Scene Attributes}
\label{appendix:scene-attributes}

The model supports transforming global scene attributes such as weather, lighting, color tone, and time of day.

\begin{flushleft}
    \textbf{Instruction:} \textit{Make @video\_1 nighttime.}%
    \vspace{0.5em}
    \begin{tabular}{m{1.2em} m{0.22\linewidth} m{0.22\linewidth} m{0.22\linewidth}}

        \centering\rotatebox{90}{\textbf{Input Video}} &
        \includegraphics[width=\linewidth]{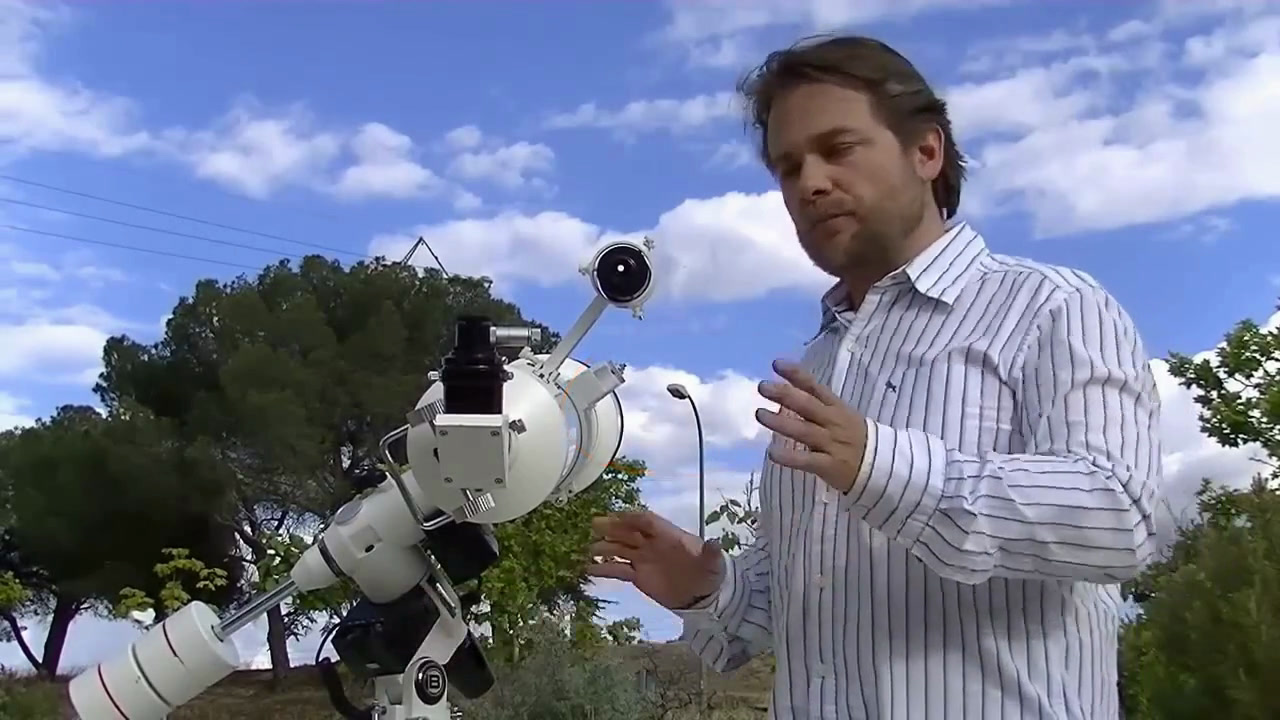} &
        \includegraphics[width=\linewidth]{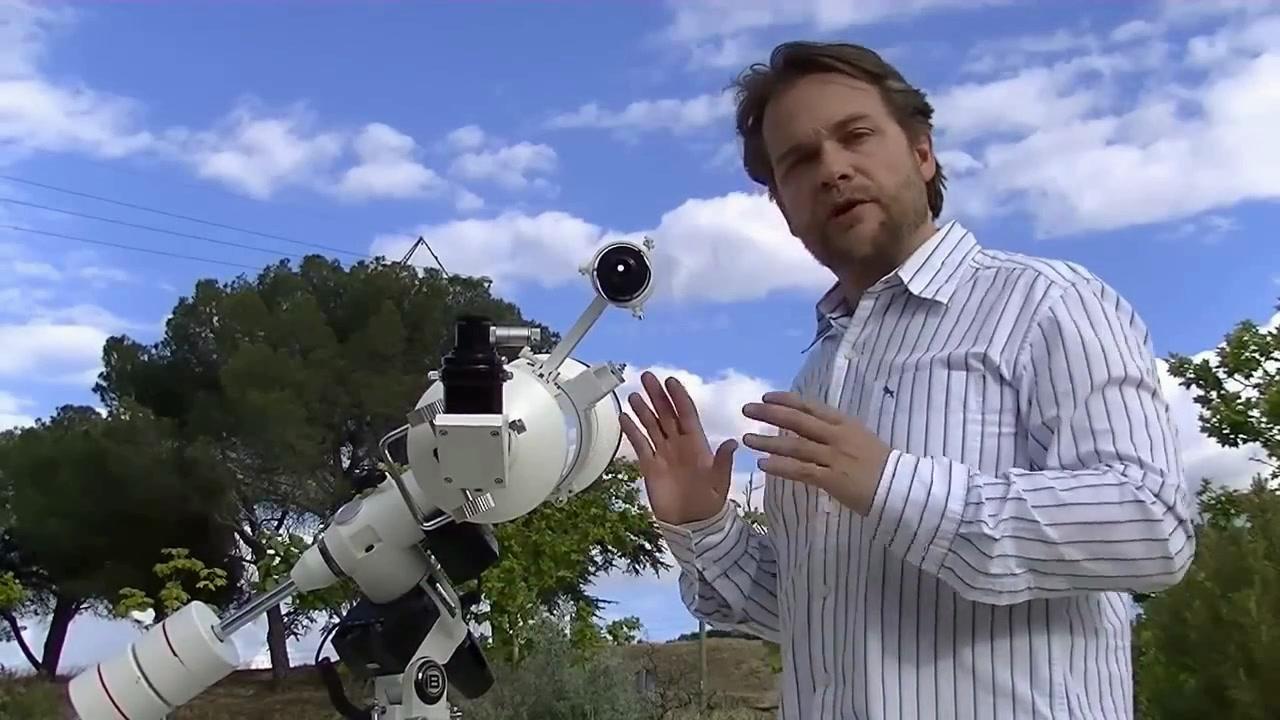} &
        \includegraphics[width=\linewidth]{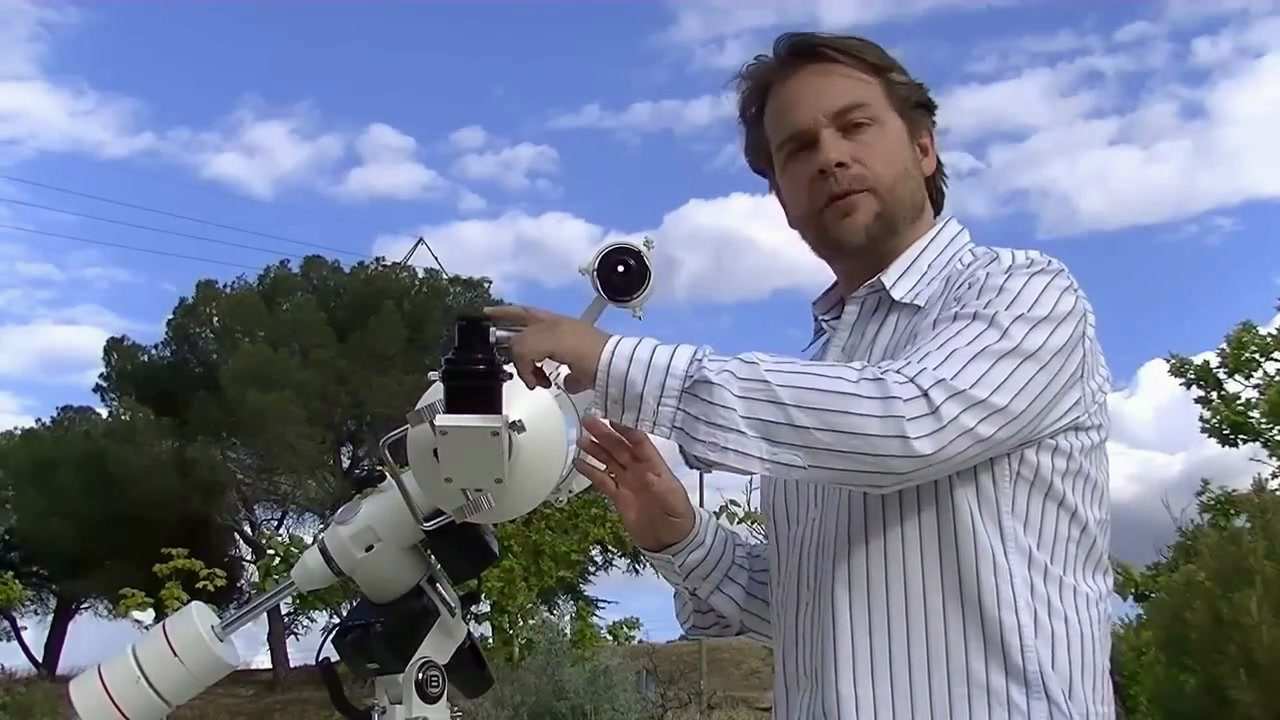} \\[2pt]

        \centering\rotatebox{90}{\textbf{Output Video}} &
        \includegraphics[width=\linewidth]{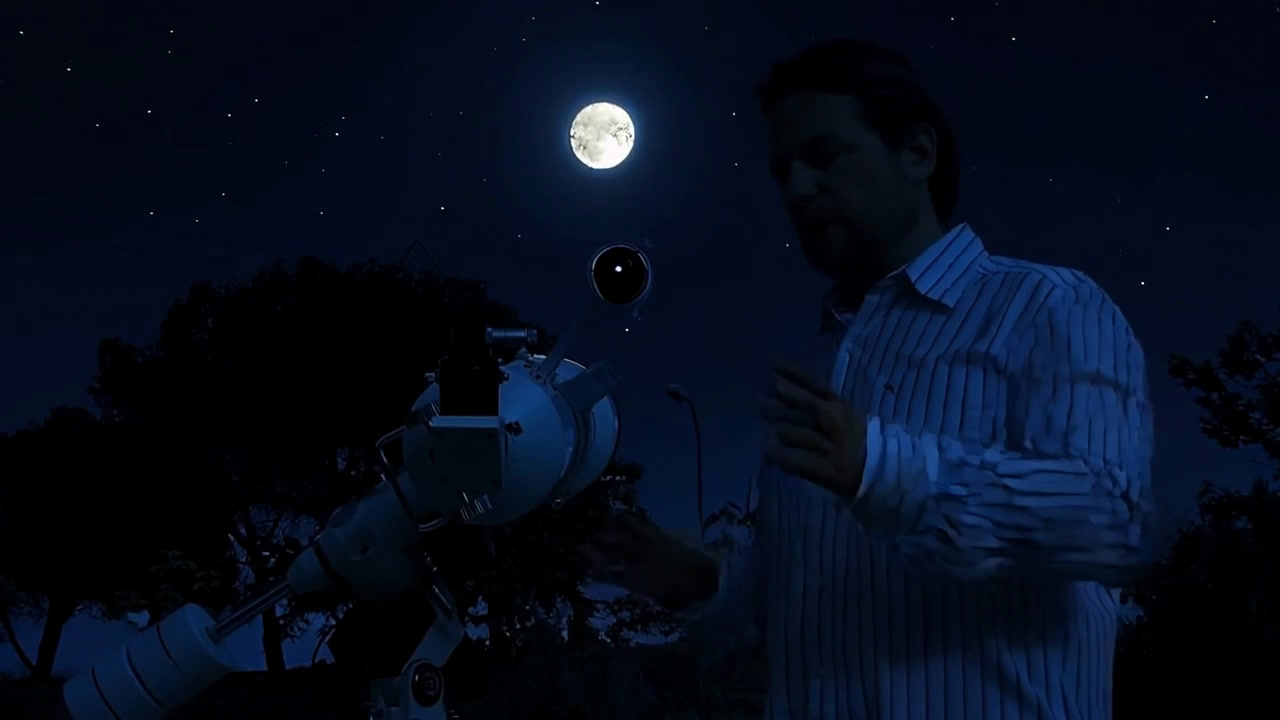} &
        \includegraphics[width=\linewidth]{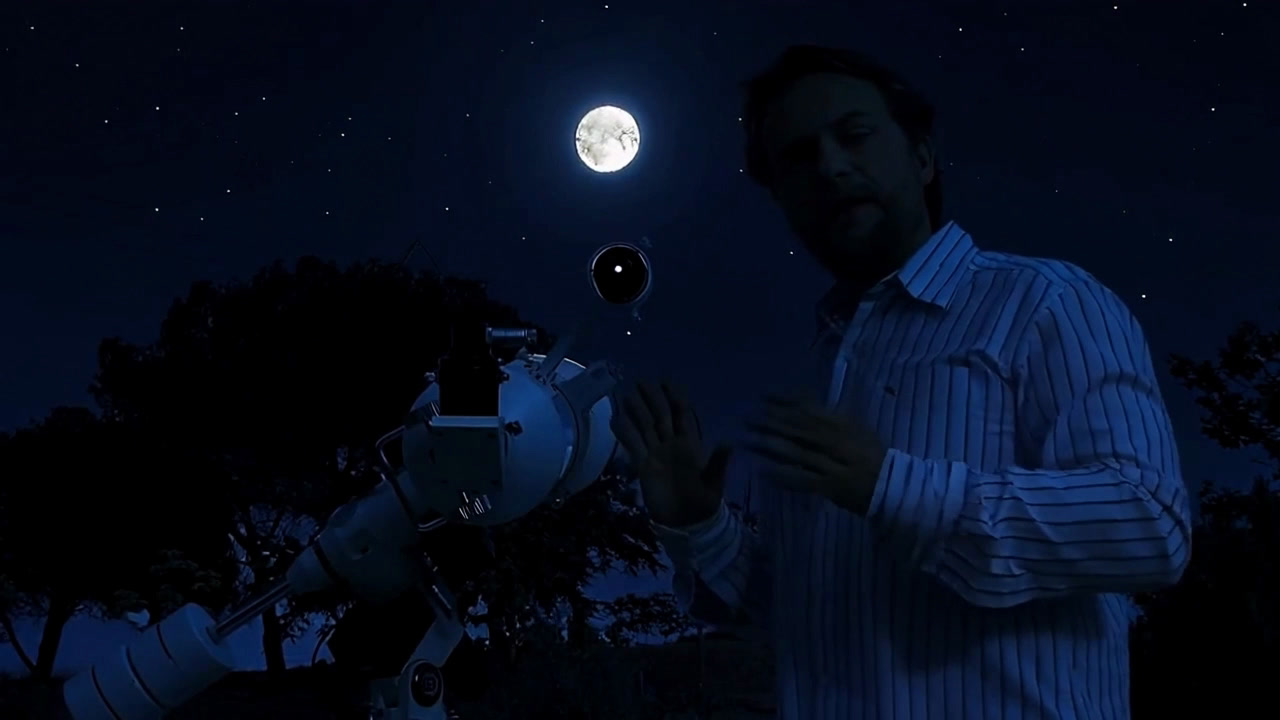} &
        \includegraphics[width=\linewidth]{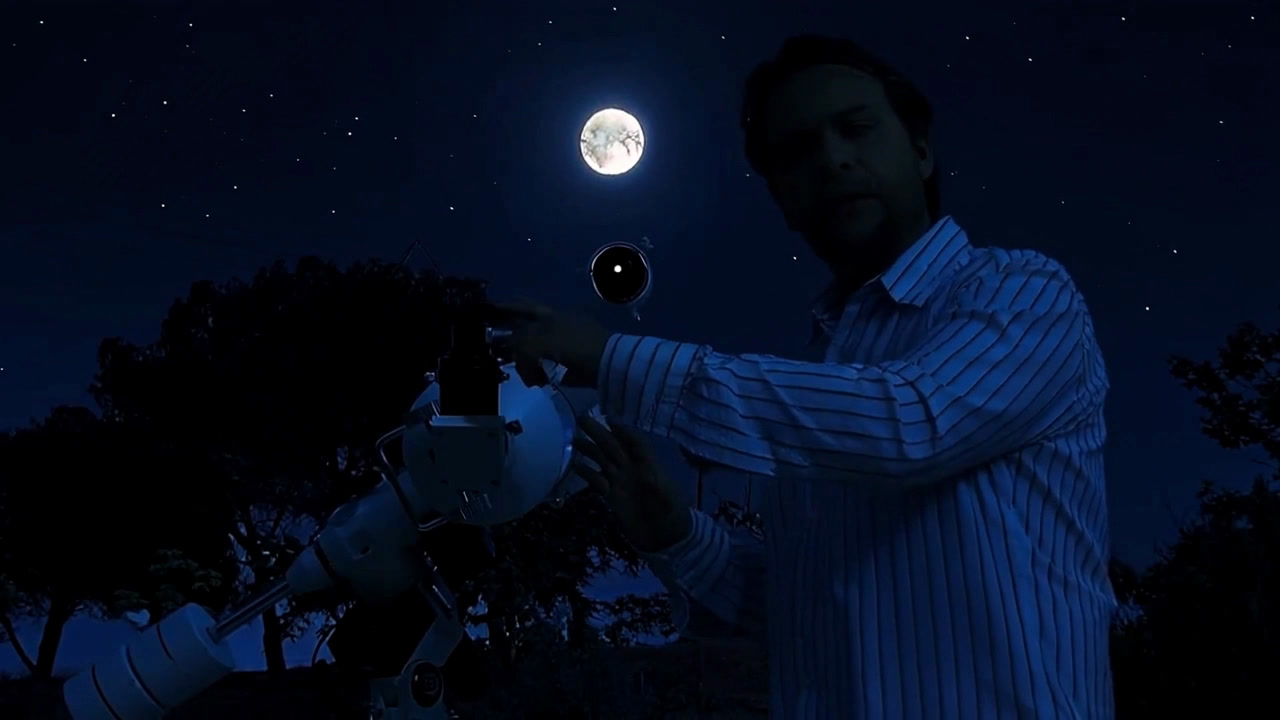} \\
    \end{tabular}
\end{flushleft}



\begin{flushleft}
    \textbf{Instruction:} \textit{Make @video\_1 snowy.}%
    \vspace{0.5em}
    \begin{tabular}{m{1.2em} m{0.22\linewidth} m{0.22\linewidth} m{0.22\linewidth}}

        \centering\rotatebox{90}{\textbf{Input Video}} &
        \includegraphics[width=\linewidth]{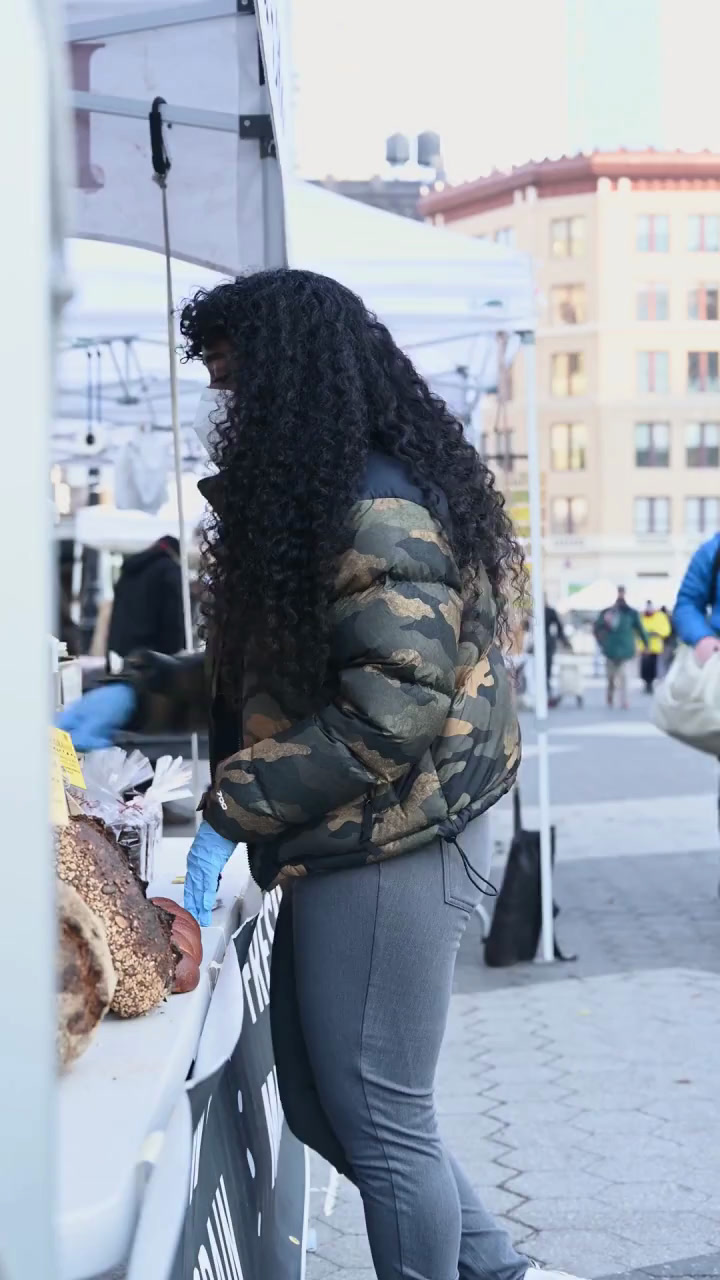} &
        \includegraphics[width=\linewidth]{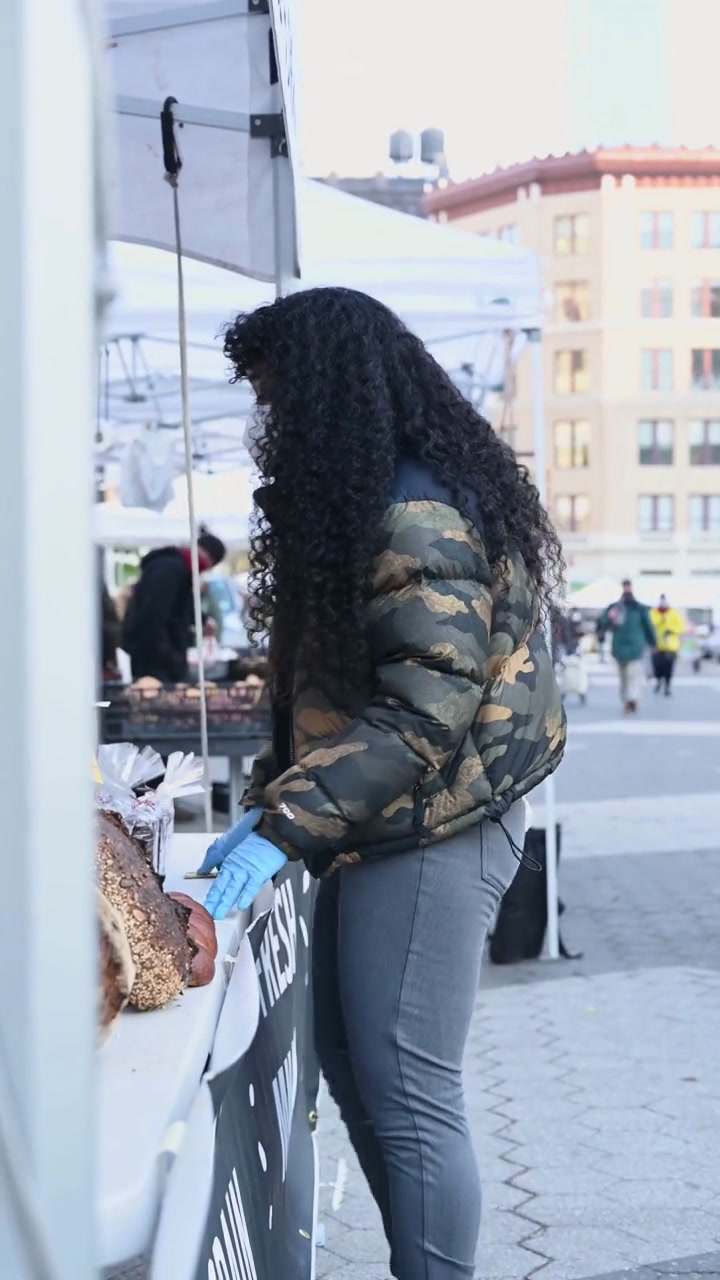} &
        \includegraphics[width=\linewidth]{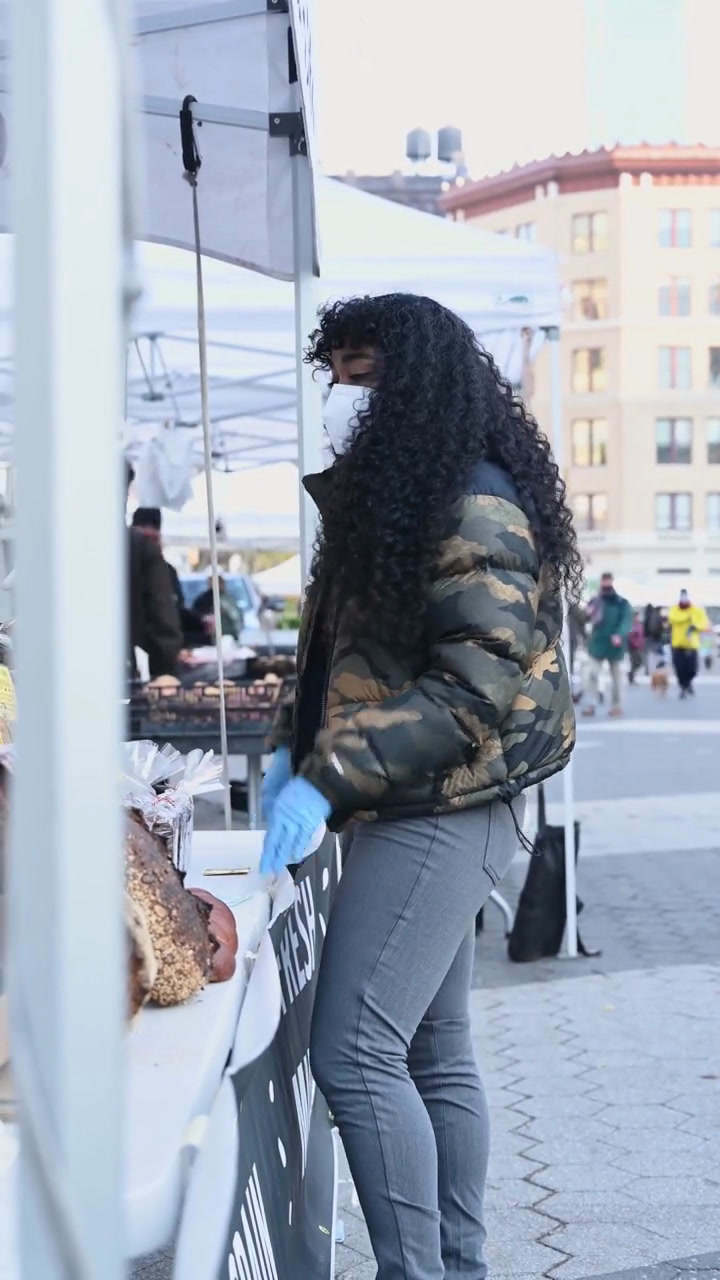} \\[2pt]

        \centering\rotatebox{90}{\textbf{Output Video}} &
        \includegraphics[width=\linewidth]{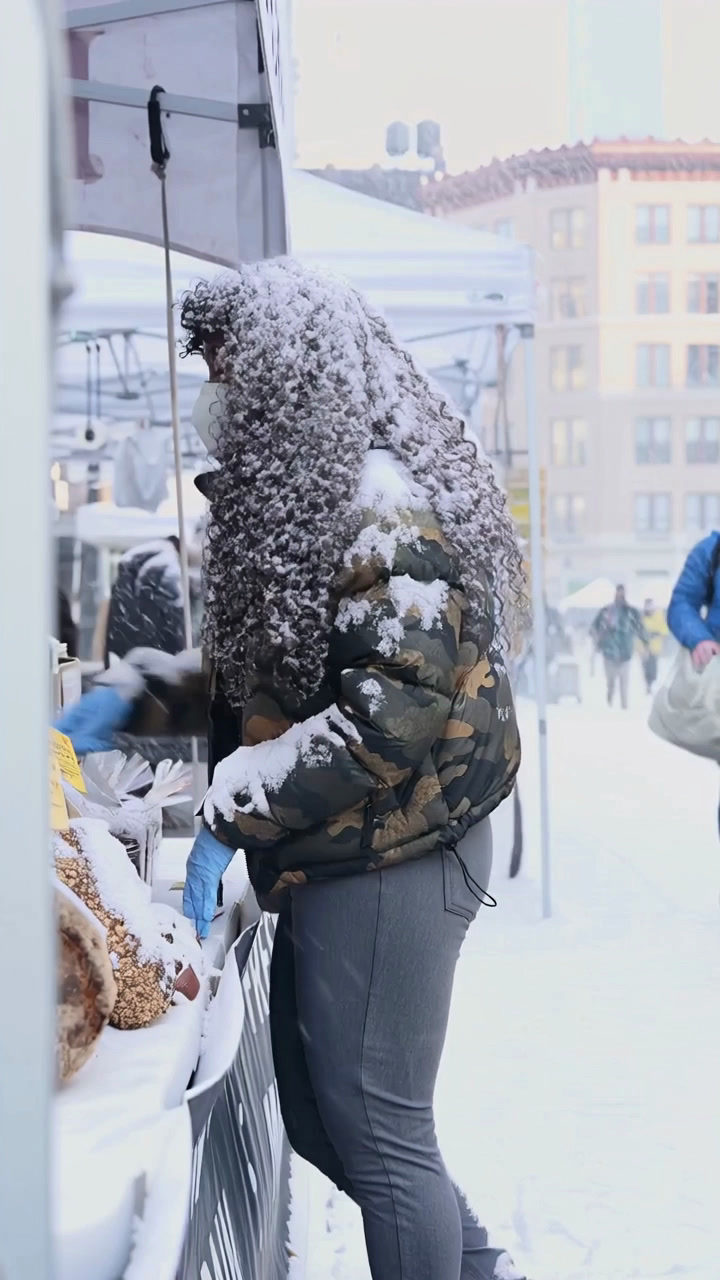} &
        \includegraphics[width=\linewidth]{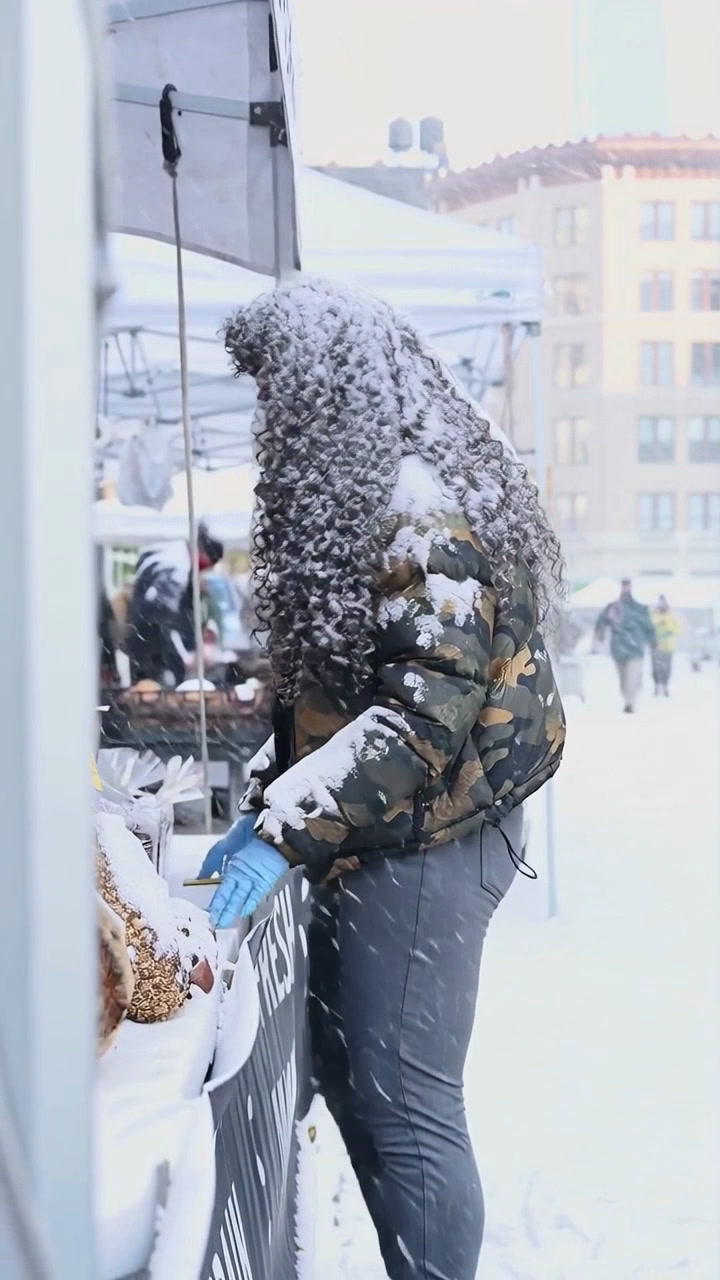} &
        \includegraphics[width=\linewidth]{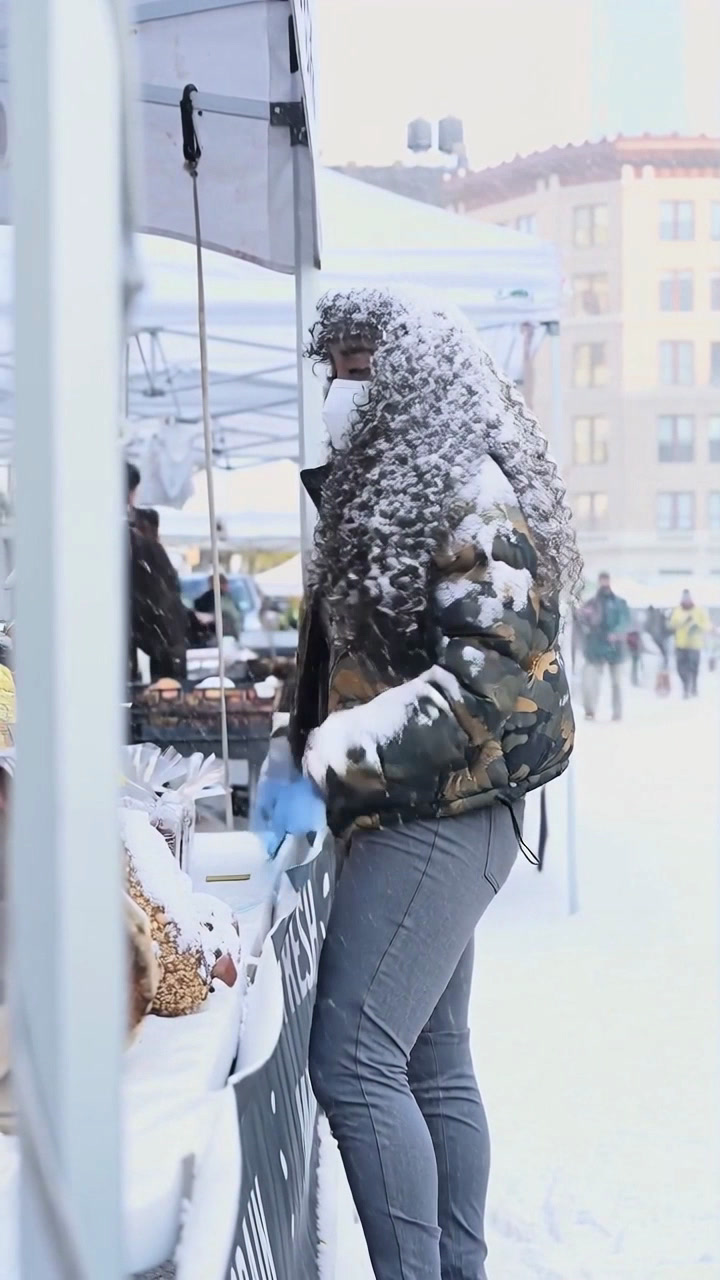} \\
    \end{tabular}
\end{flushleft}


\begin{figure}[h!]
    \centering
    \caption{Examples of global scene attributes.}
    \label{fig:edit_scene}
\end{figure}

\newpage
\subsubsection{Reference-Based Editing}
\label{appendix:reference-edit}

The model supports video editing based on image references, including subject reference, background reference, and first-frame reference, combined with reference videos to generate the final output. Reference videos can provide motion, expression, or visual effects guidance.

\paragraph{Subject Reference with Motion Reference}

The model can generate videos by combining a subject from a reference image with motion patterns from a reference video, matching action rhythm and trajectory.

\begin{flushleft}
    \textbf{Instruction:} \textit{Woman from @image\_1 mimics gestures from @video\_1 in its golden field background.}%
    \vspace{0.5em}
    \begin{tabular}{m{1.2em} m{0.22\linewidth} m{0.22\linewidth} m{0.22\linewidth}}

        \centering\rotatebox{90}{\textbf{Ref. Image}} &
        \includegraphics[width=\linewidth]{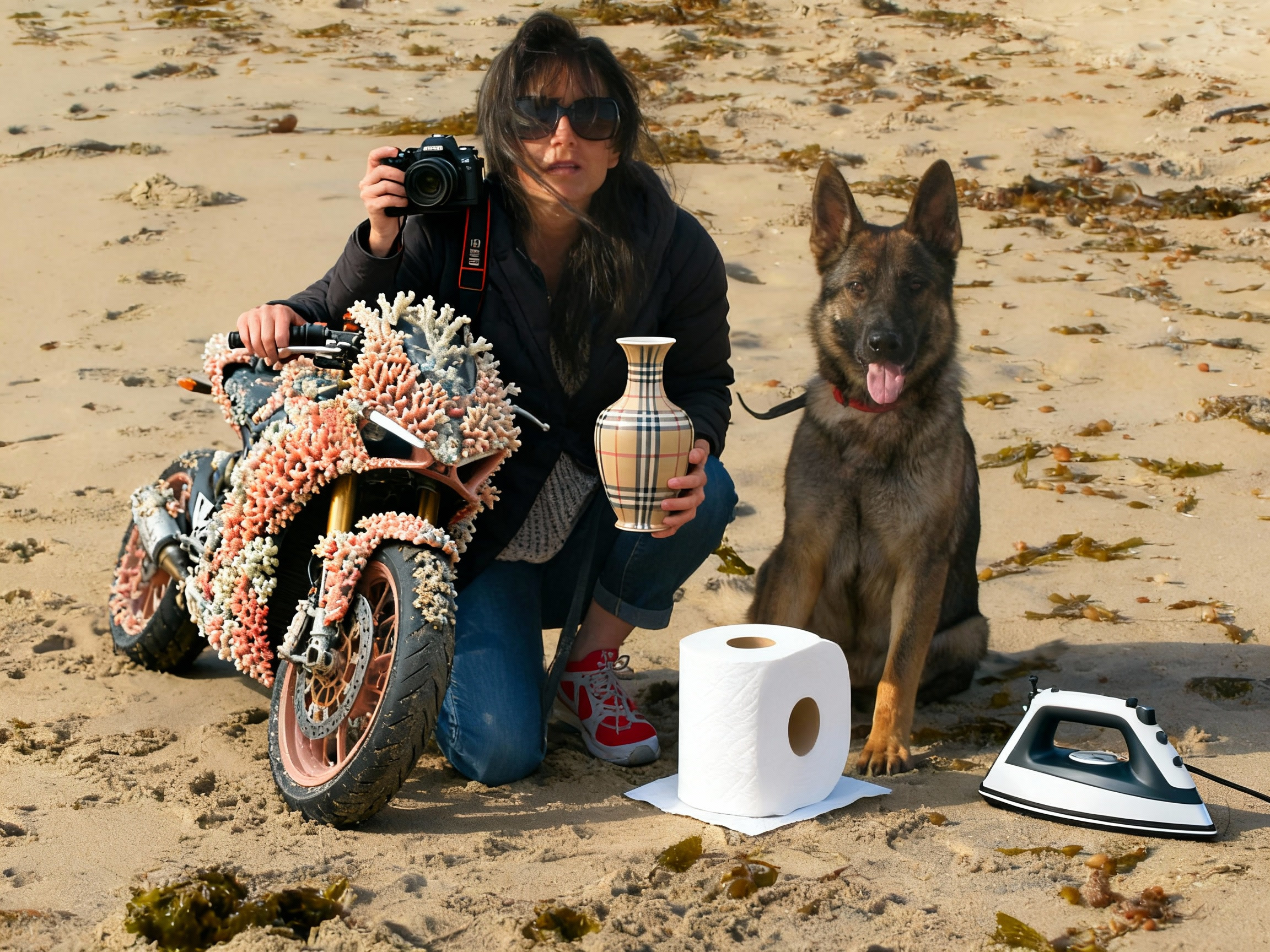} & & \\[2pt]

        \centering\rotatebox{90}{\textbf{Input Video}} &
        \includegraphics[width=\linewidth]{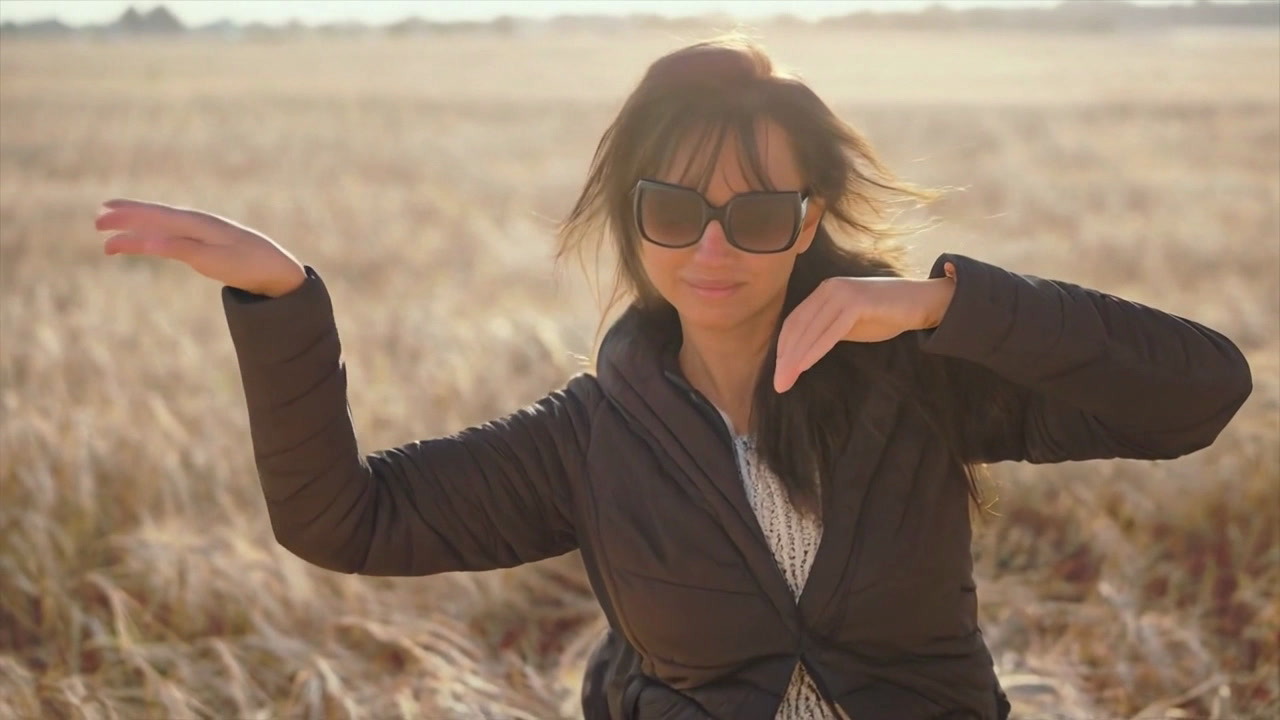} &
        \includegraphics[width=\linewidth]{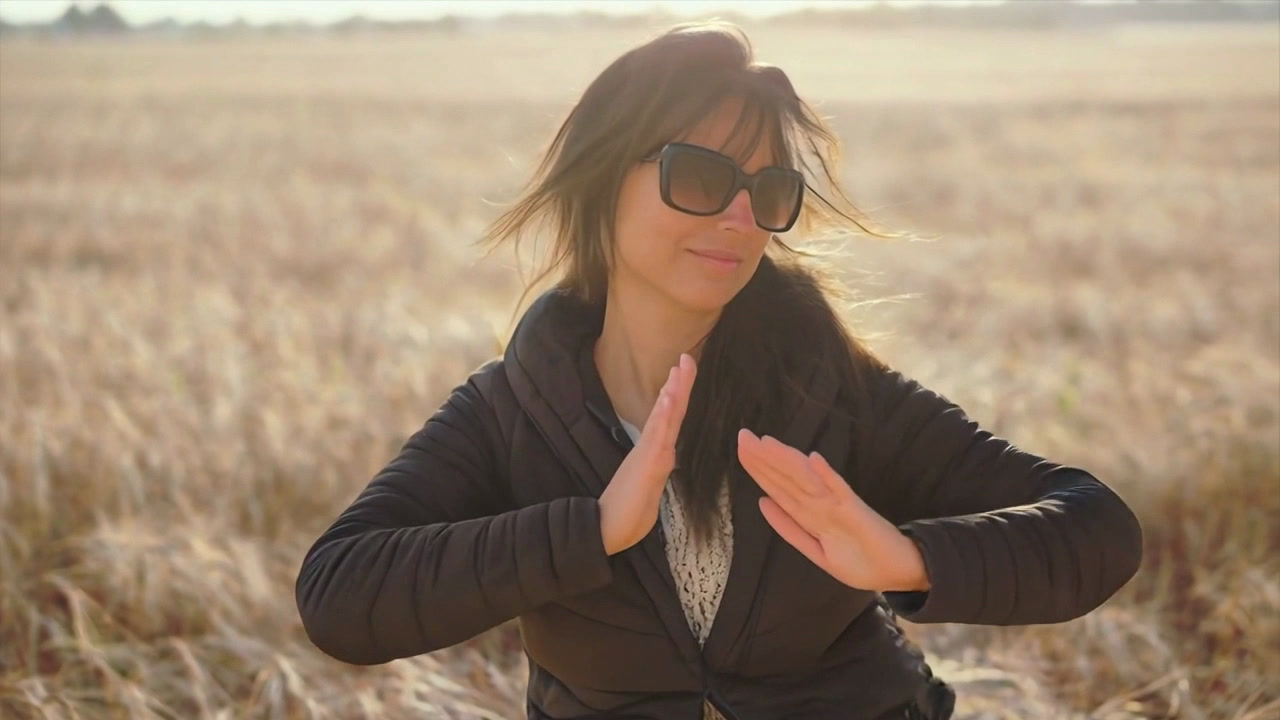} &
        \includegraphics[width=\linewidth]{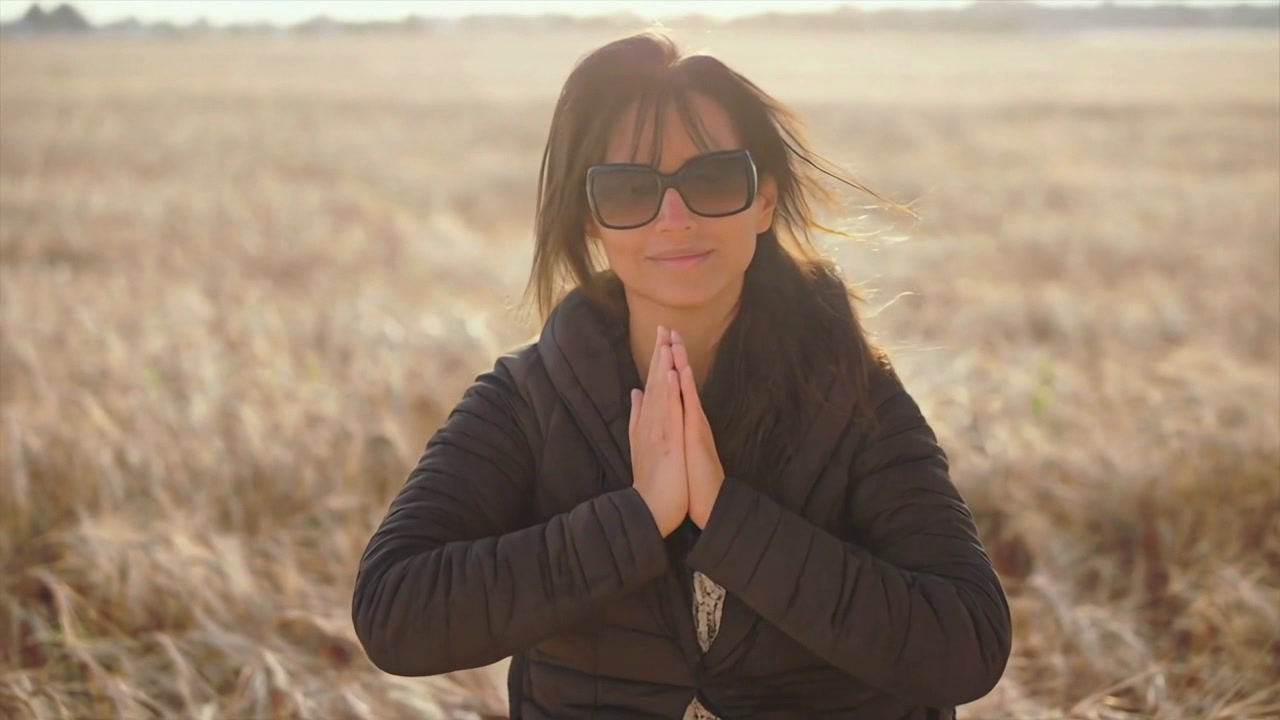} \\[2pt]

        \centering\rotatebox{90}{\textbf{Output Video}} &
        \includegraphics[width=\linewidth]{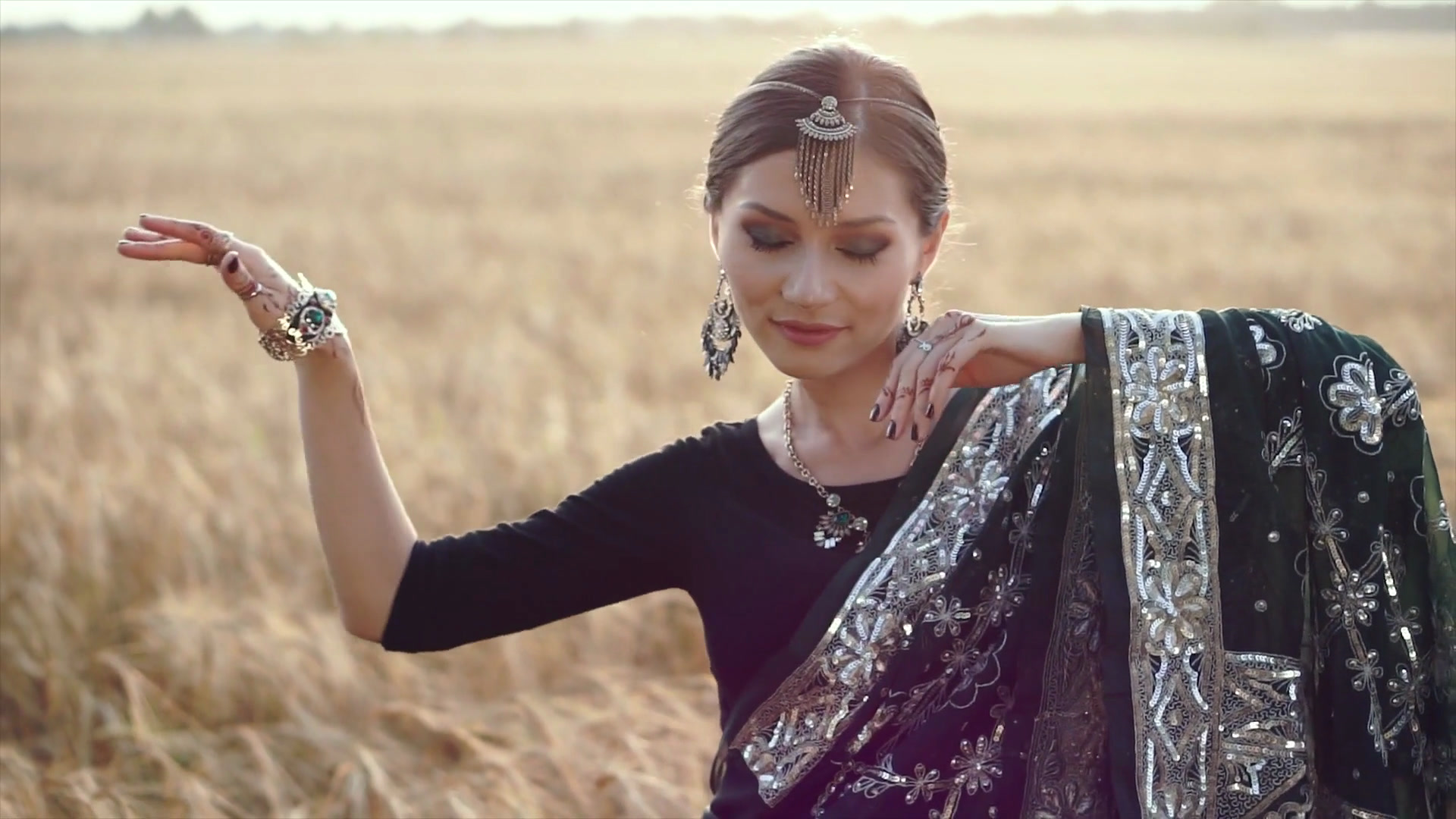} &
        \includegraphics[width=\linewidth]{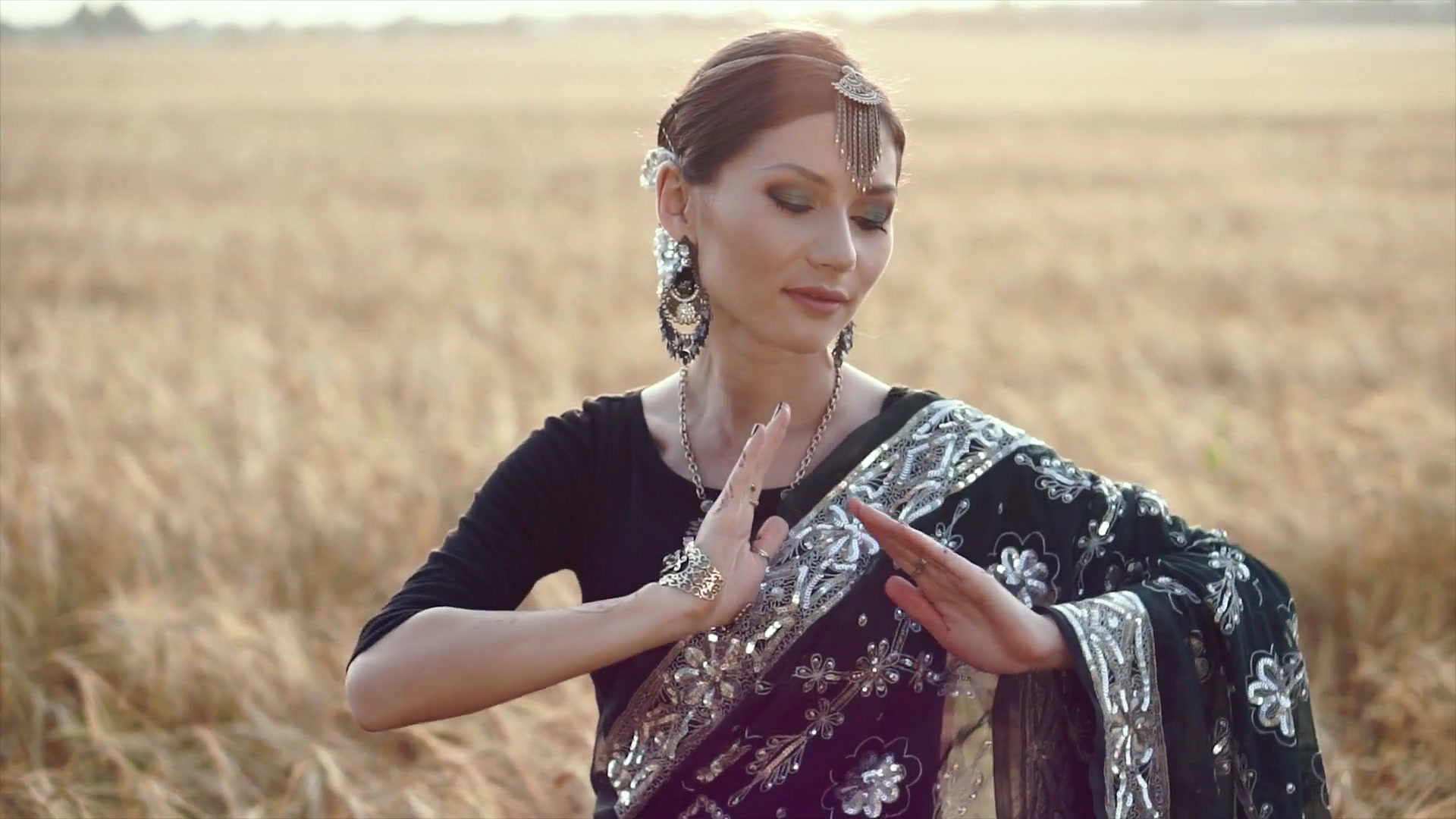} &
        \includegraphics[width=\linewidth]{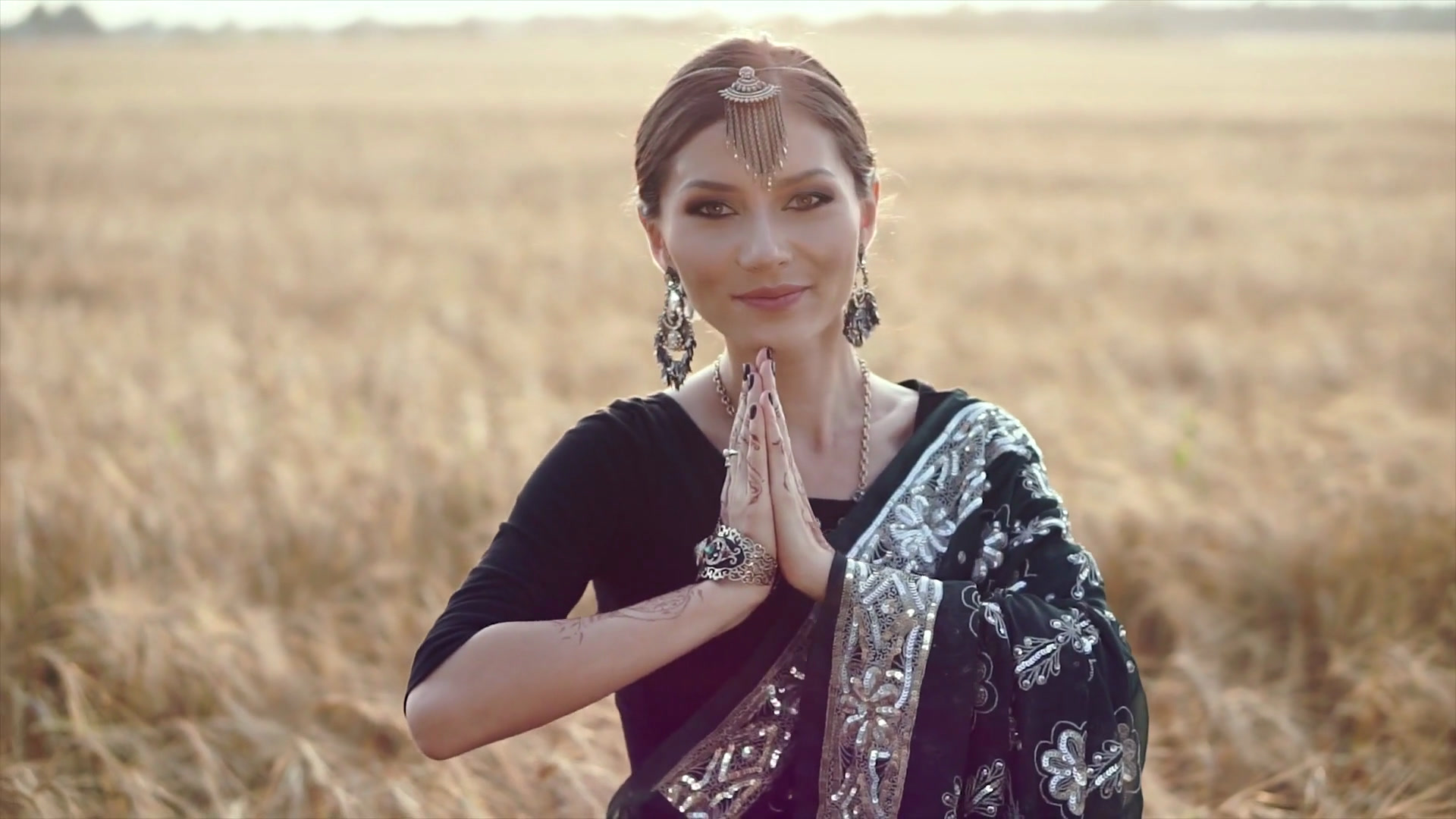} \\
    \end{tabular}
\end{flushleft}


\begin{figure}[h!]
    \centering
    \caption{Example of subject reference with motion reference.}
    \label{fig:edit_ref_subject_motion_1}
\end{figure}

\paragraph{Subject Reference with Expression Reference}

The model can transfer natural facial expressions from a reference video to a subject from a reference image.

\begin{flushleft}
    \textbf{Instruction:} \textit{Transfer the facial expressions from @video\_1 to the man in @image\_1.}%
    \vspace{0.5em}
    \begin{tabular}{m{1.2em} m{0.22\linewidth} m{0.22\linewidth} m{0.22\linewidth}}

        \centering\rotatebox{90}{\textbf{Ref. Image}} &
        \includegraphics[width=\linewidth]{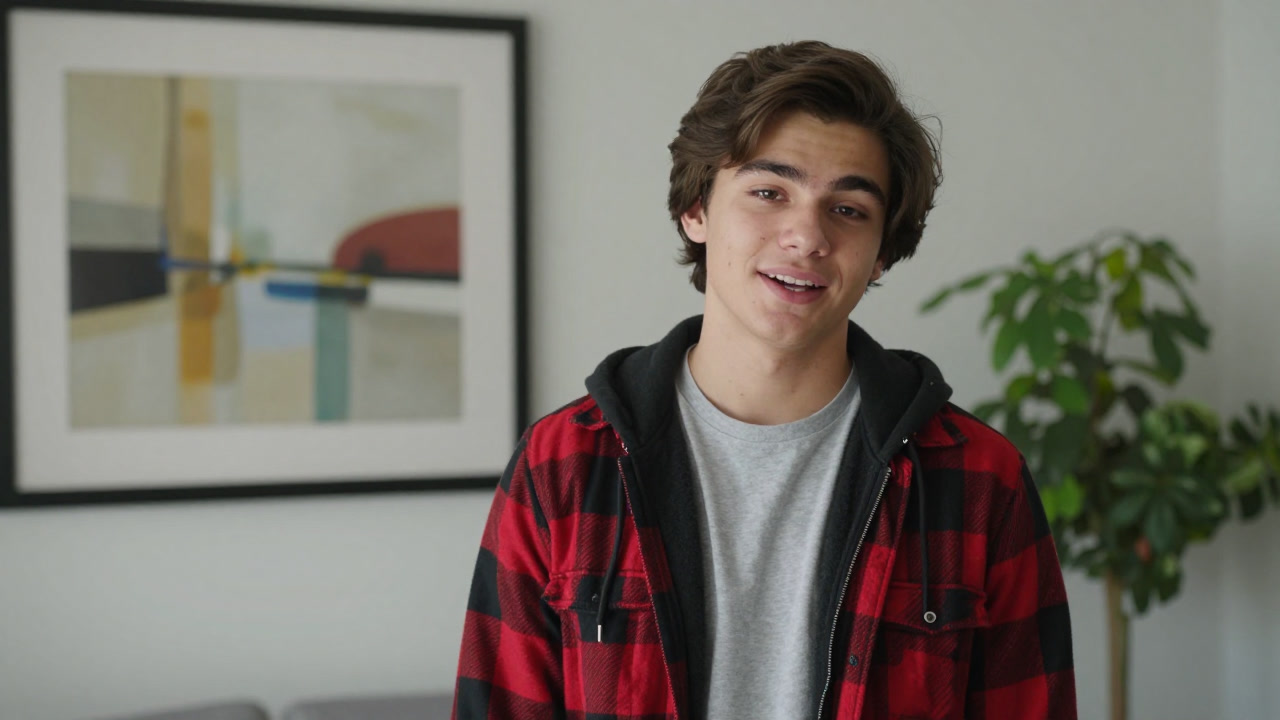} & & \\[2pt]

        \centering\rotatebox{90}{\textbf{Input Video}} &
        \includegraphics[width=\linewidth]{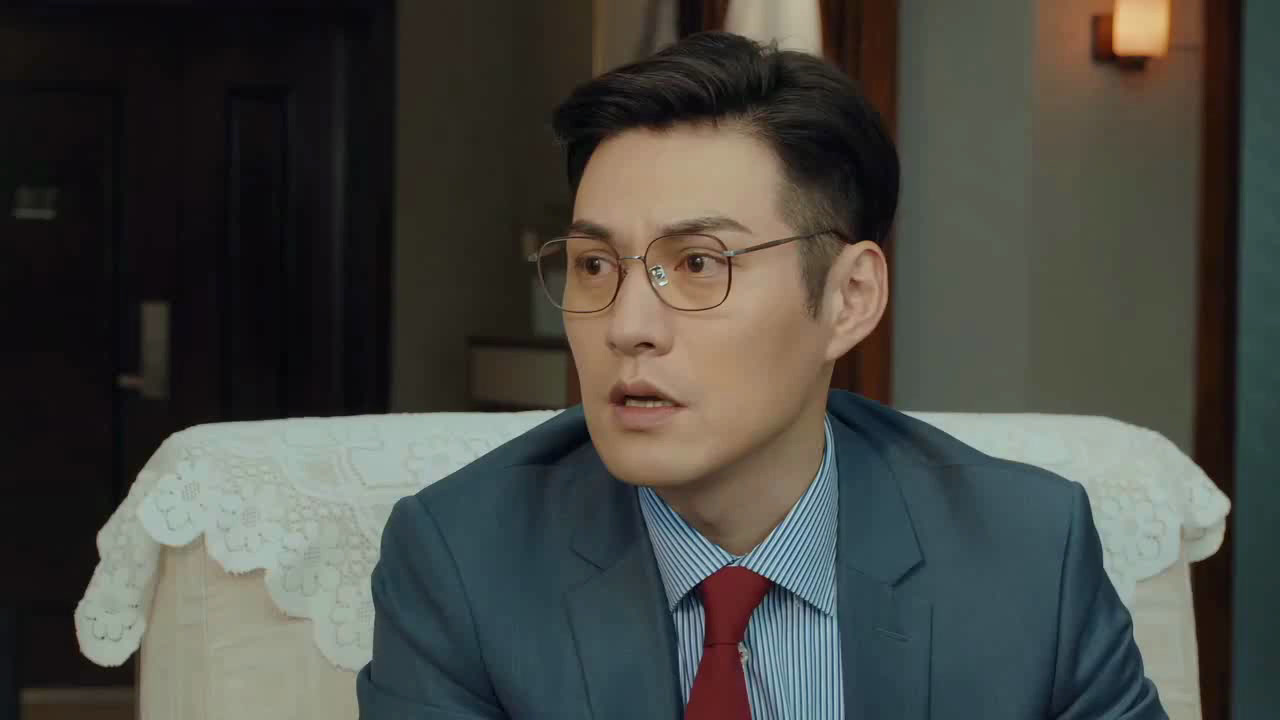} &
        \includegraphics[width=\linewidth]{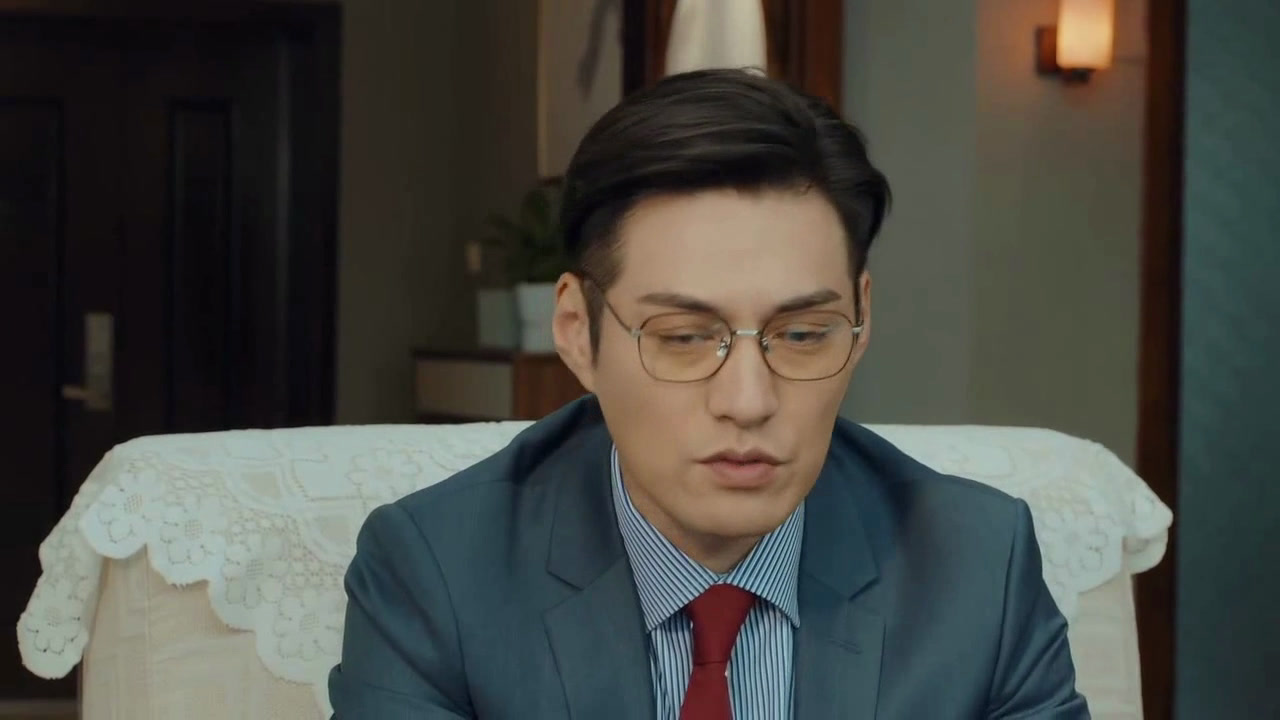} &
        \includegraphics[width=\linewidth]{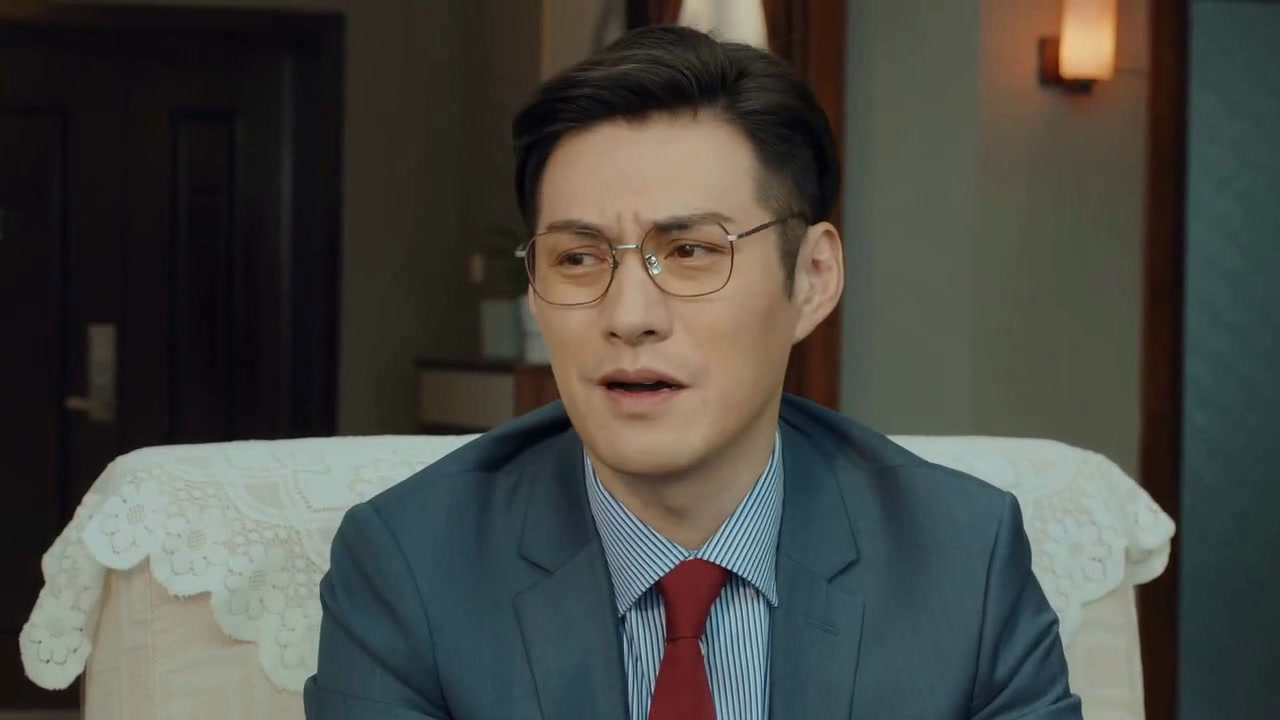} \\[2pt]

        \centering\rotatebox{90}{\textbf{Output Video}} &
        \includegraphics[width=\linewidth]{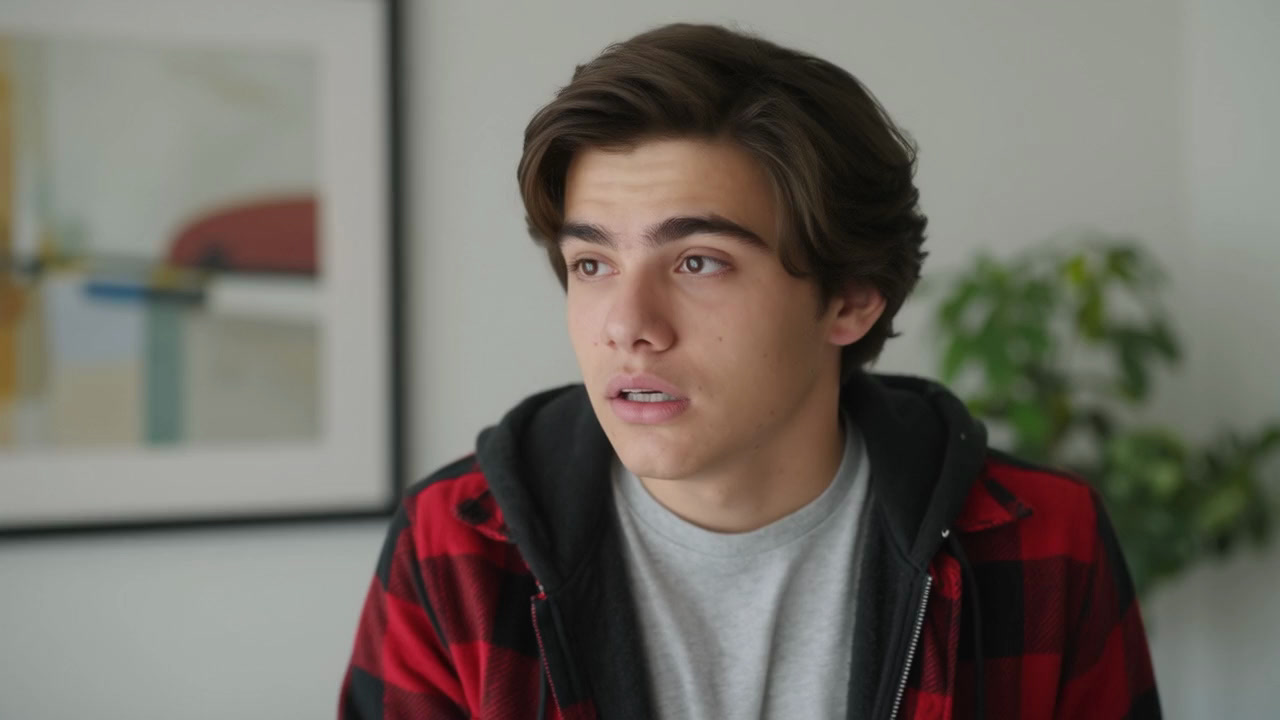} &
        \includegraphics[width=\linewidth]{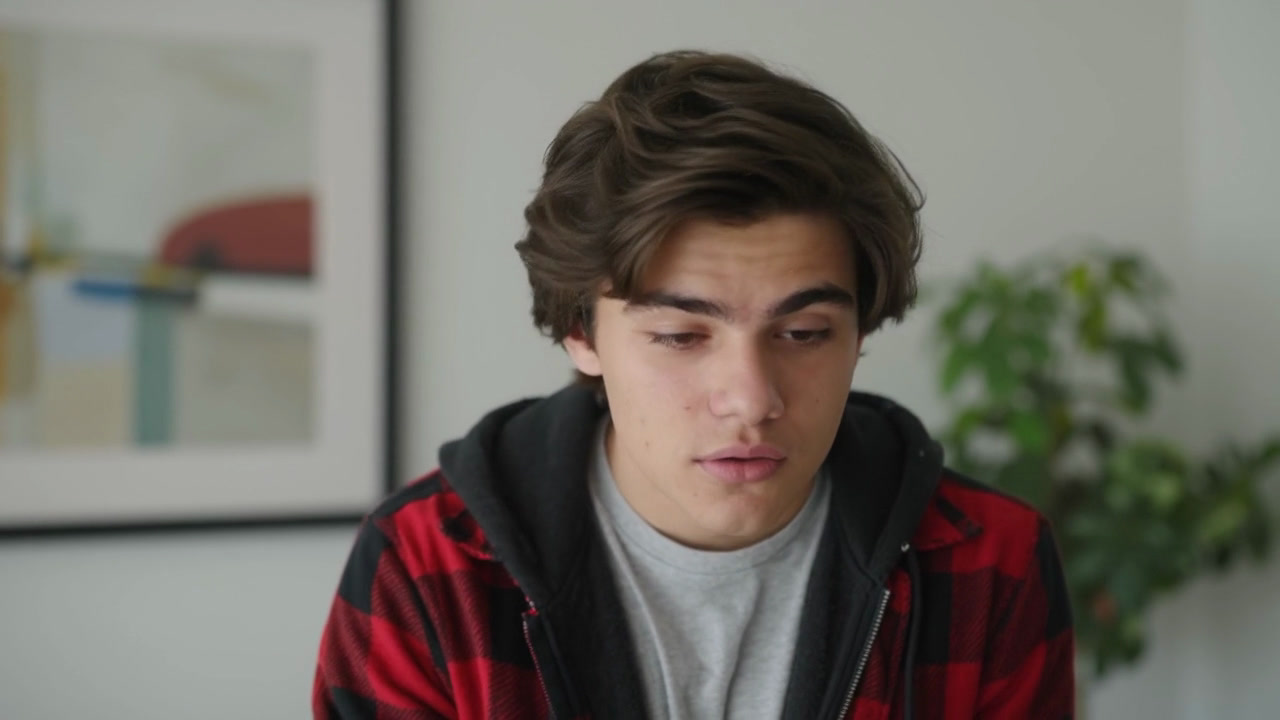} &
        \includegraphics[width=\linewidth]{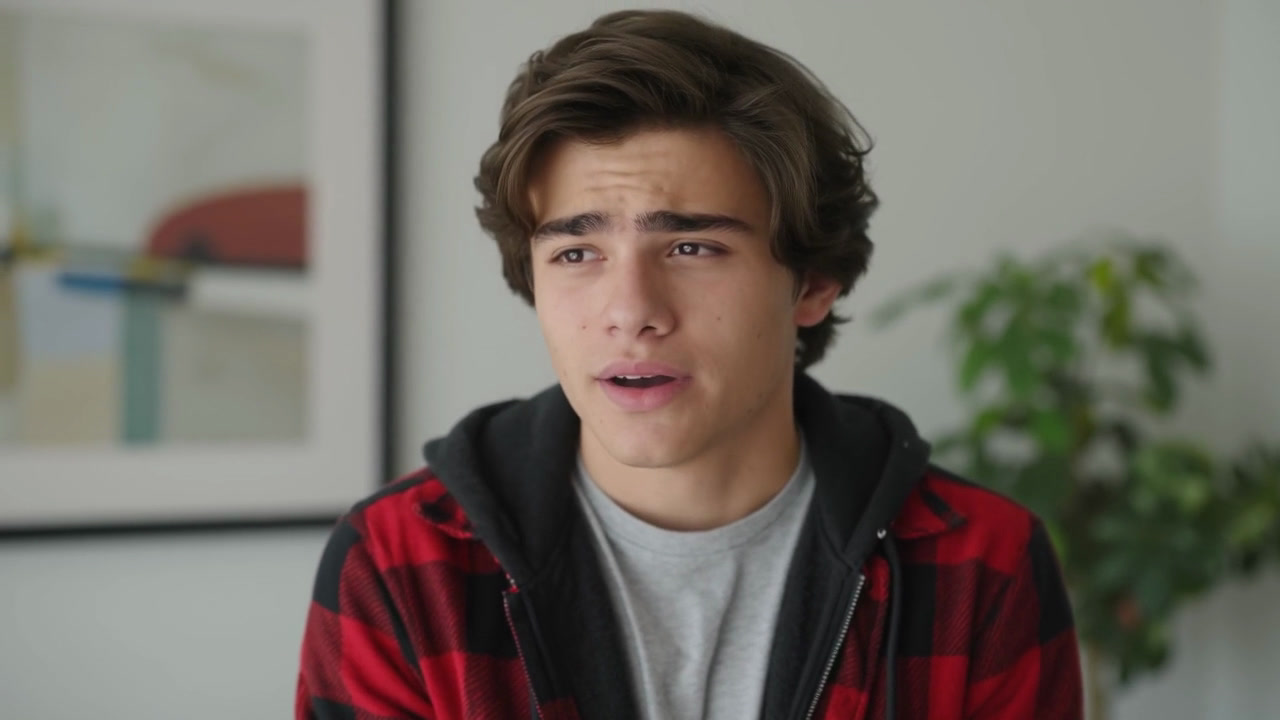} \\
    \end{tabular}
\end{flushleft}


\begin{figure}[h!]
    \centering
    \caption{Example of subject reference with expression reference}
    \label{fig:edit_ref_subject_expression_1}
\end{figure}

\newpage
\paragraph{Background Reference with Video Reference}

The model can combine a background from a reference image with content or motion from any reference video.

\begin{flushleft}
    \textbf{Instruction:} \textit{Replace the background of @video\_1 with @image\_1.}%
    \vspace{0.5em}
    \begin{tabular}{m{1.2em} m{0.22\linewidth} m{0.22\linewidth} m{0.22\linewidth}}

        \centering\rotatebox{90}{\textbf{Ref. Image}} &
        \includegraphics[width=\linewidth]{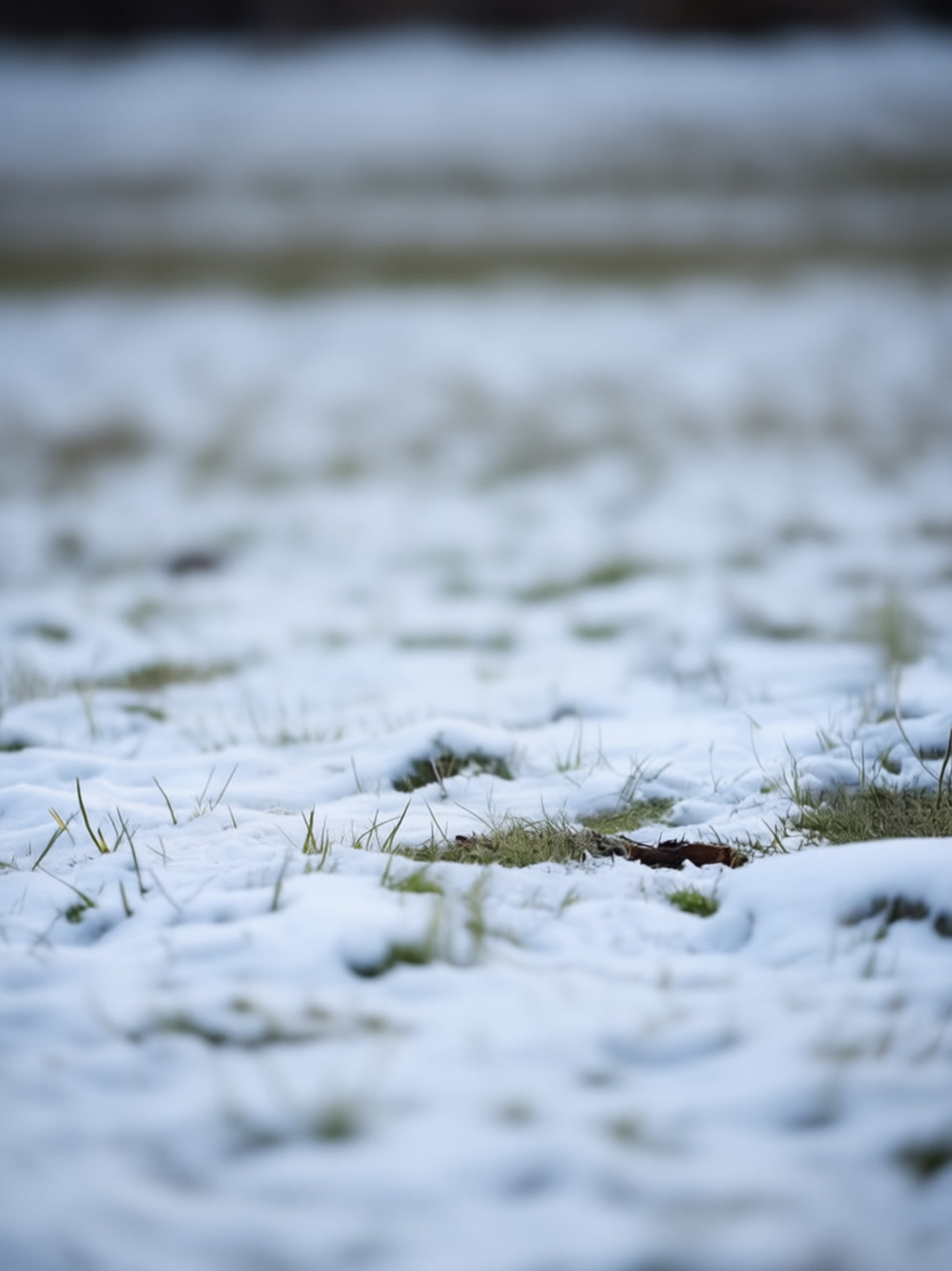} & & \\[2pt]

        \centering\rotatebox{90}{\textbf{Input Video}} &
        \includegraphics[width=\linewidth]{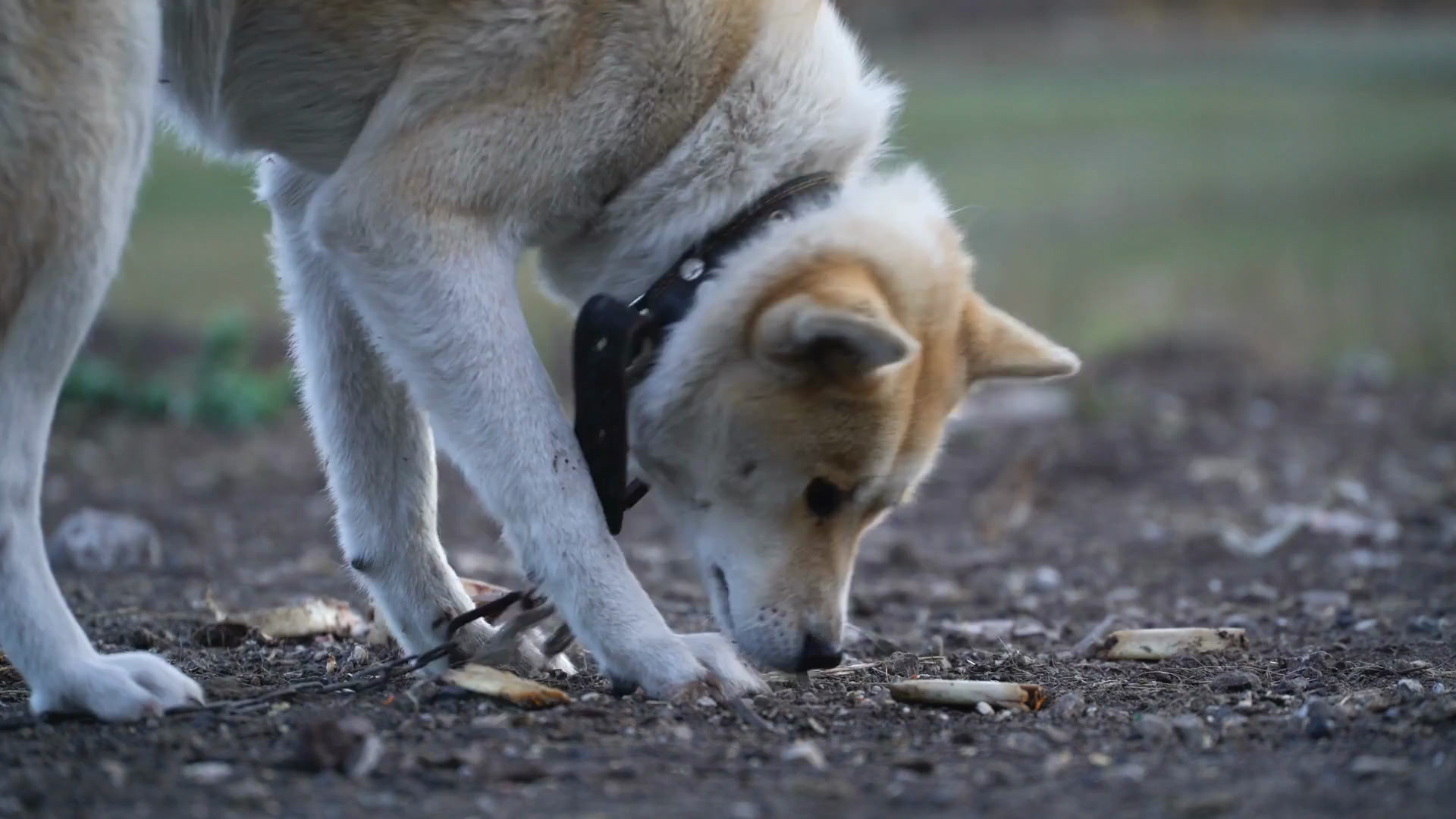} &
        \includegraphics[width=\linewidth]{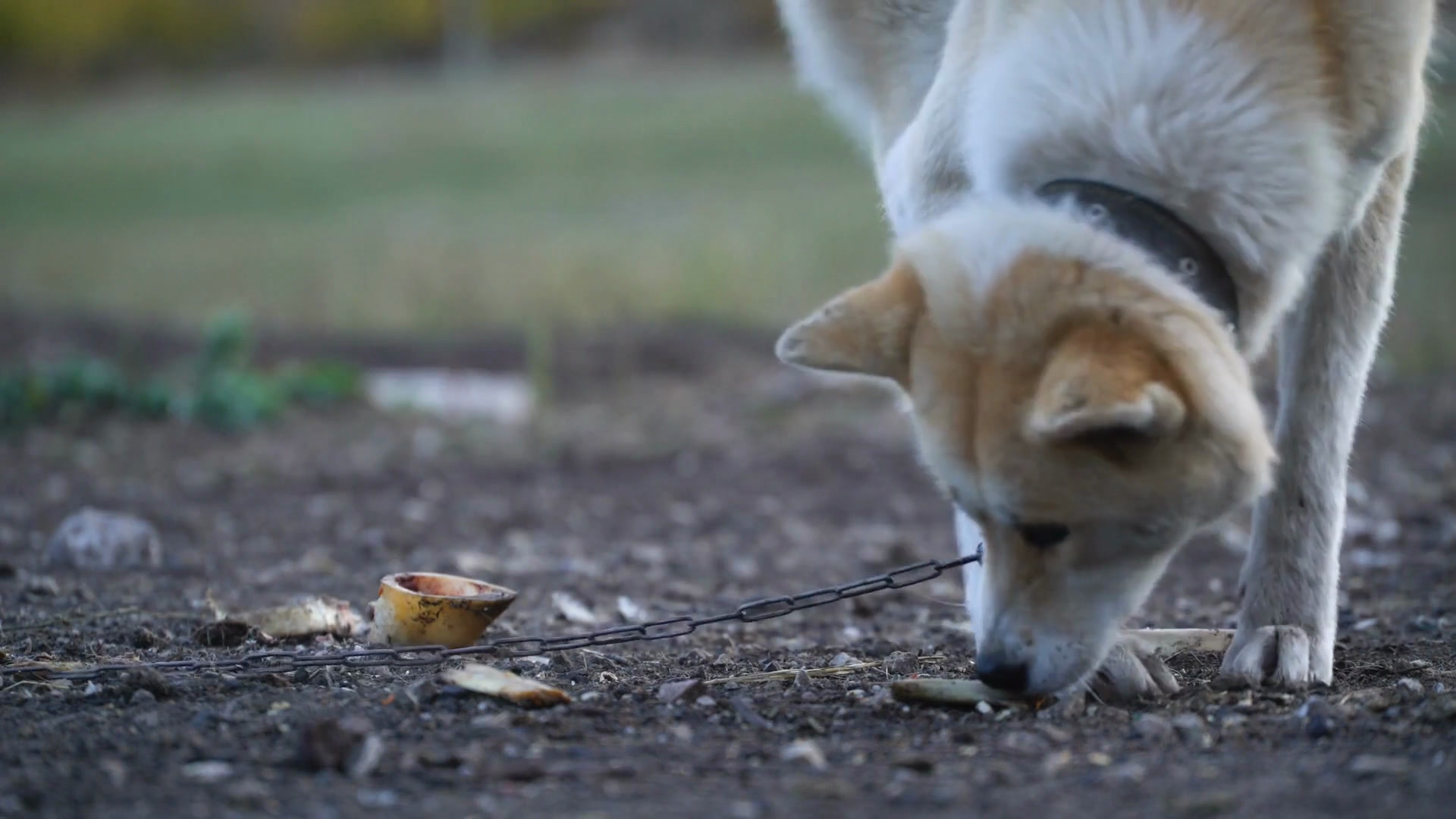} &
        \includegraphics[width=\linewidth]{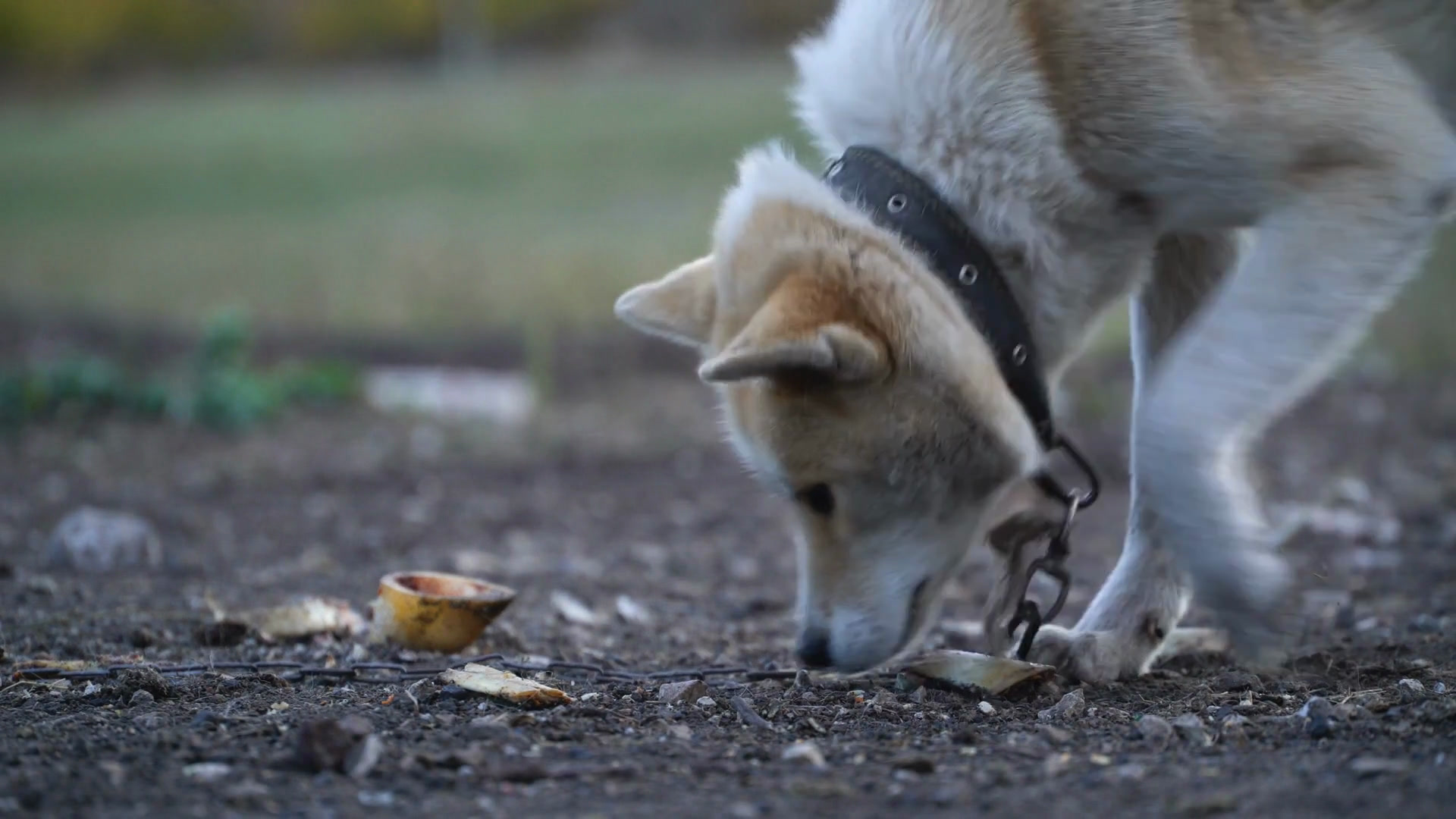} \\[2pt]

        \centering\rotatebox{90}{\textbf{Output Video}} &
        \includegraphics[width=\linewidth]{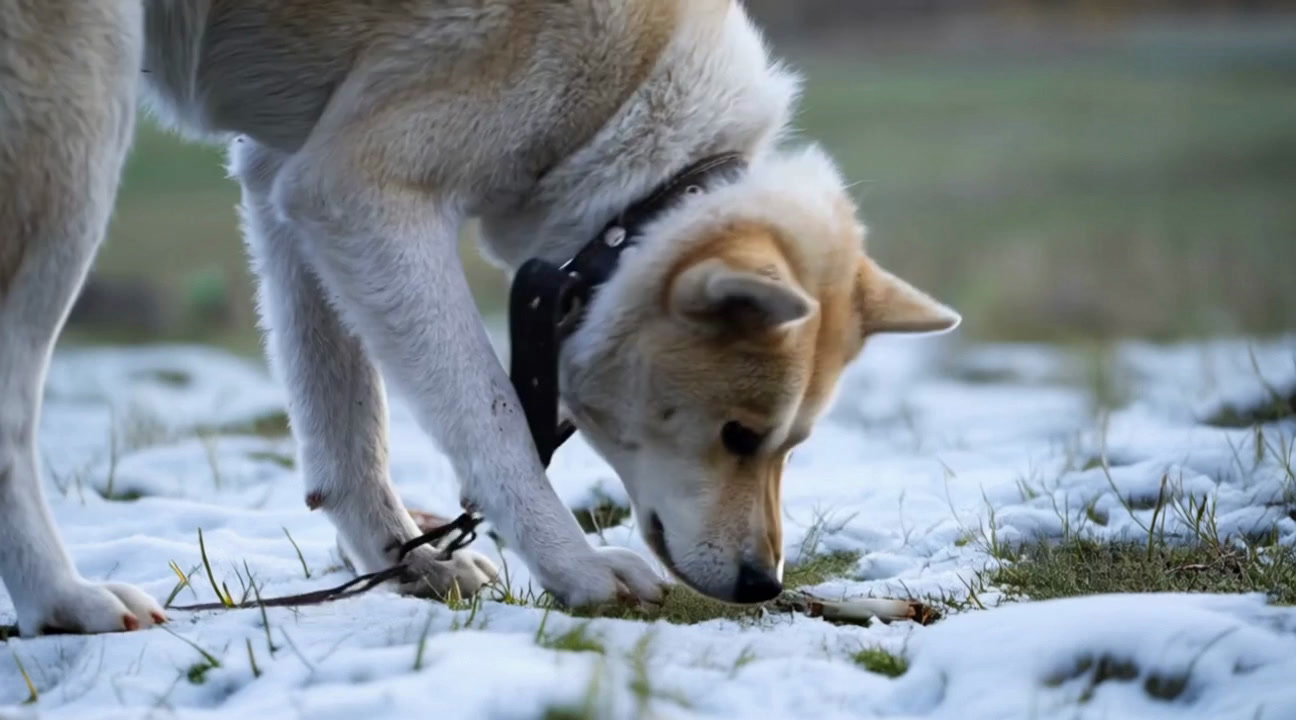} &
        \includegraphics[width=\linewidth]{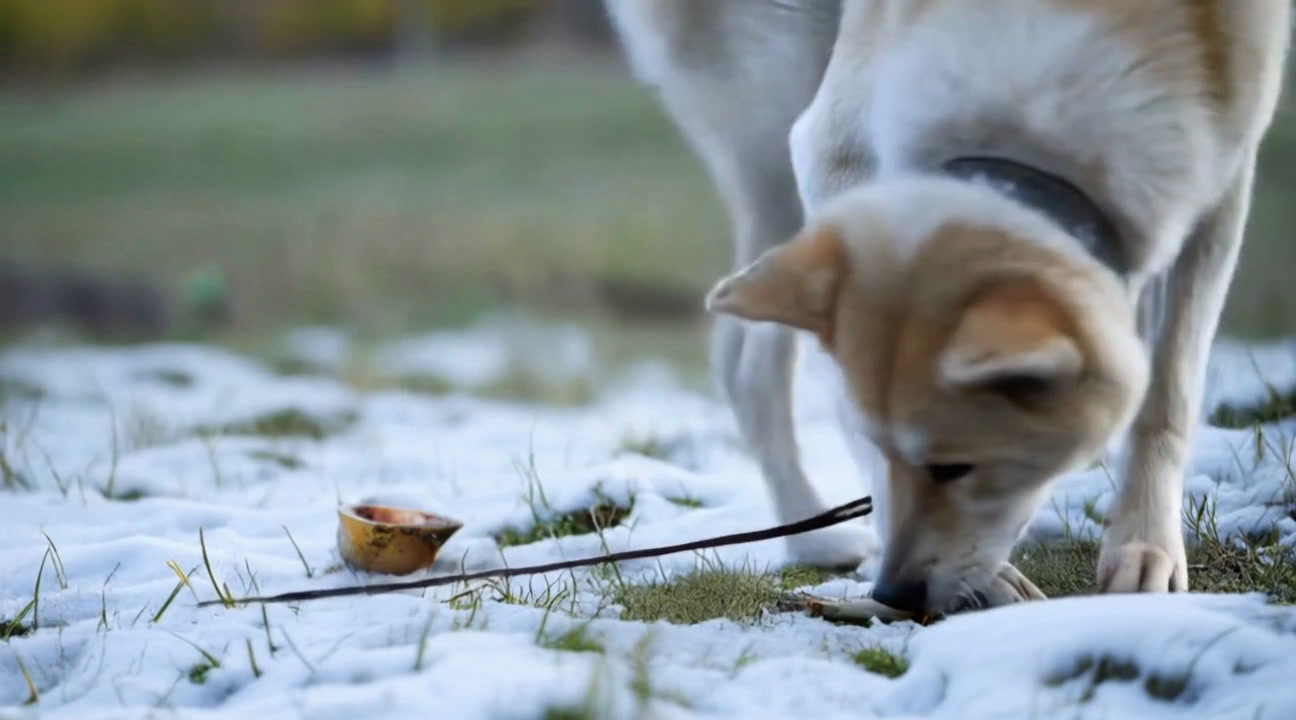} &
        \includegraphics[width=\linewidth]{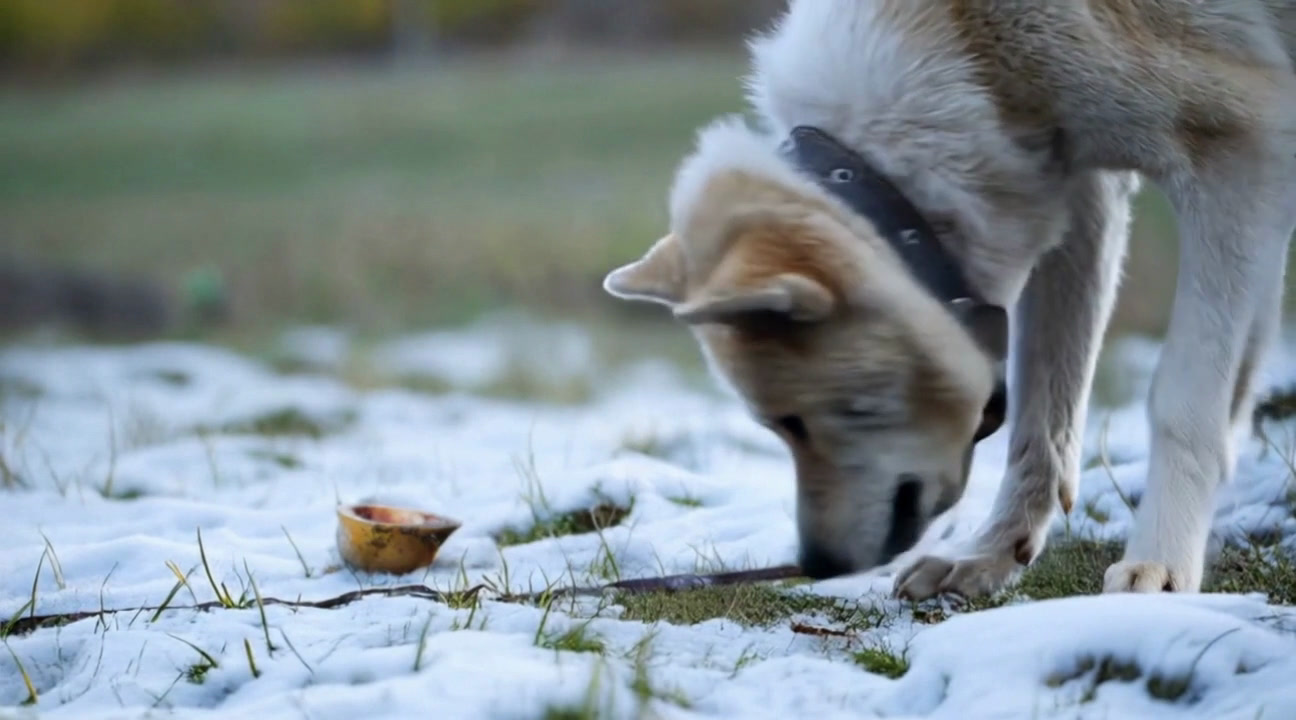} \\
    \end{tabular}
\end{flushleft}


\begin{figure}[h!]
    \centering
    \caption{Example of background reference with video reference.}
    \label{fig:edit_ref_background_1}
\end{figure}

\newpage
\paragraph{First-Frame Reference with Effect Reference}

The model can apply visual effects from a reference video to generate a video starting from a reference image, enabling effect transfer from the reference video.

\begin{flushleft}
    \textbf{Instruction:} \textit{Transfer the diamond morphing effect from @video\_1 onto the subject in @image\_1.}%
    \vspace{0.5em}
    \begin{tabular}{m{1.2em} m{0.22\linewidth} m{0.22\linewidth} m{0.22\linewidth}}

        \centering\rotatebox{90}{\textbf{Ref. Image}} &
        \includegraphics[width=\linewidth]{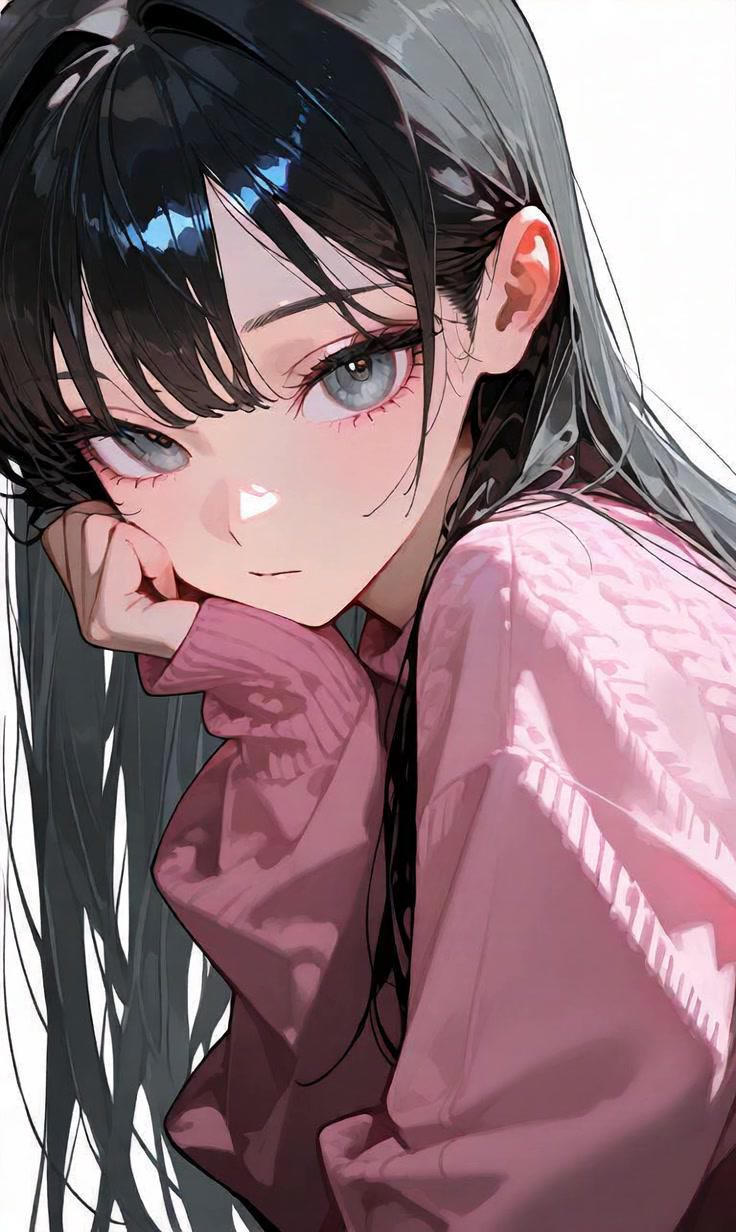} & & \\[2pt]

        \centering\rotatebox{90}{\textbf{Ref. Video}} &
        \includegraphics[width=\linewidth]{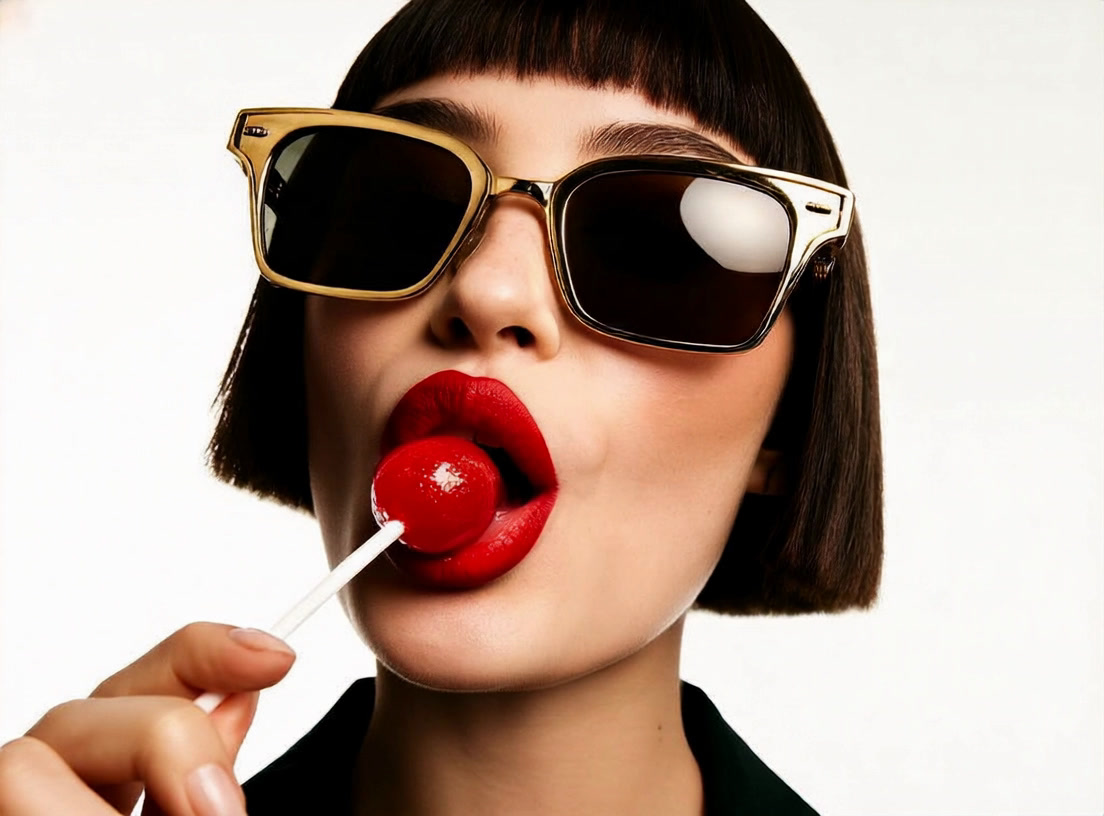} &
        \includegraphics[width=\linewidth]{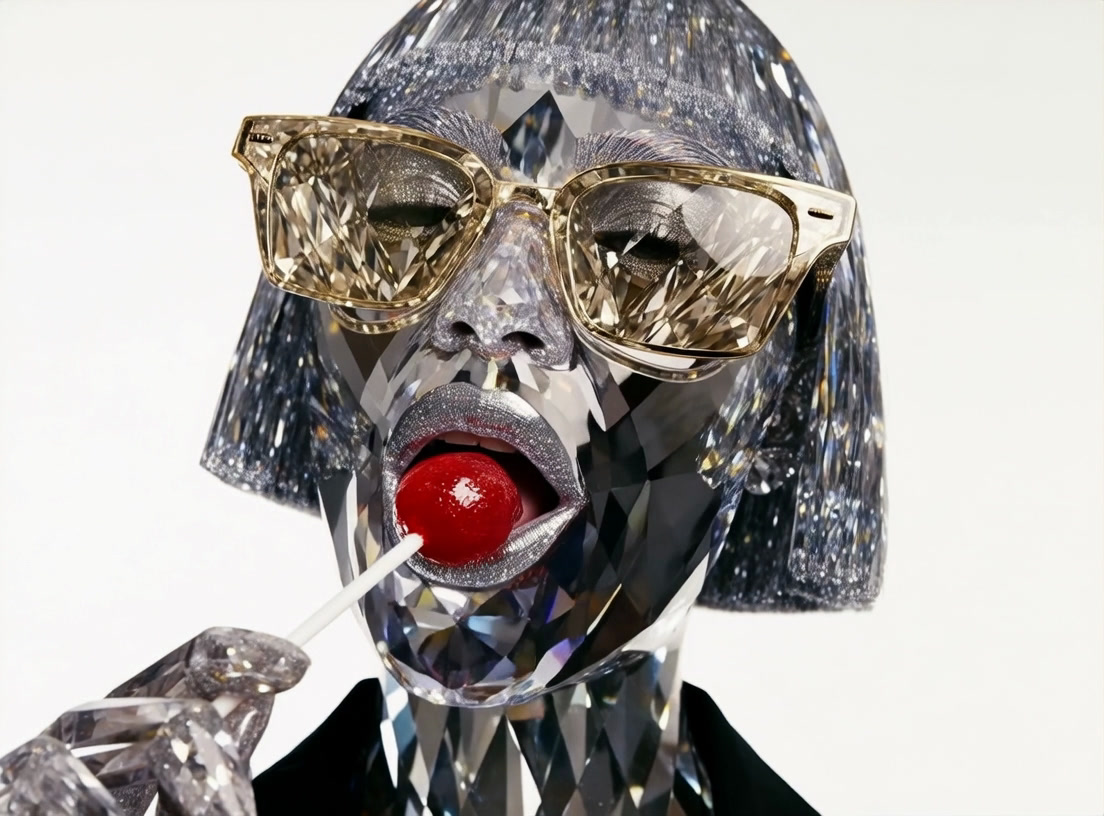} &
        \includegraphics[width=\linewidth]{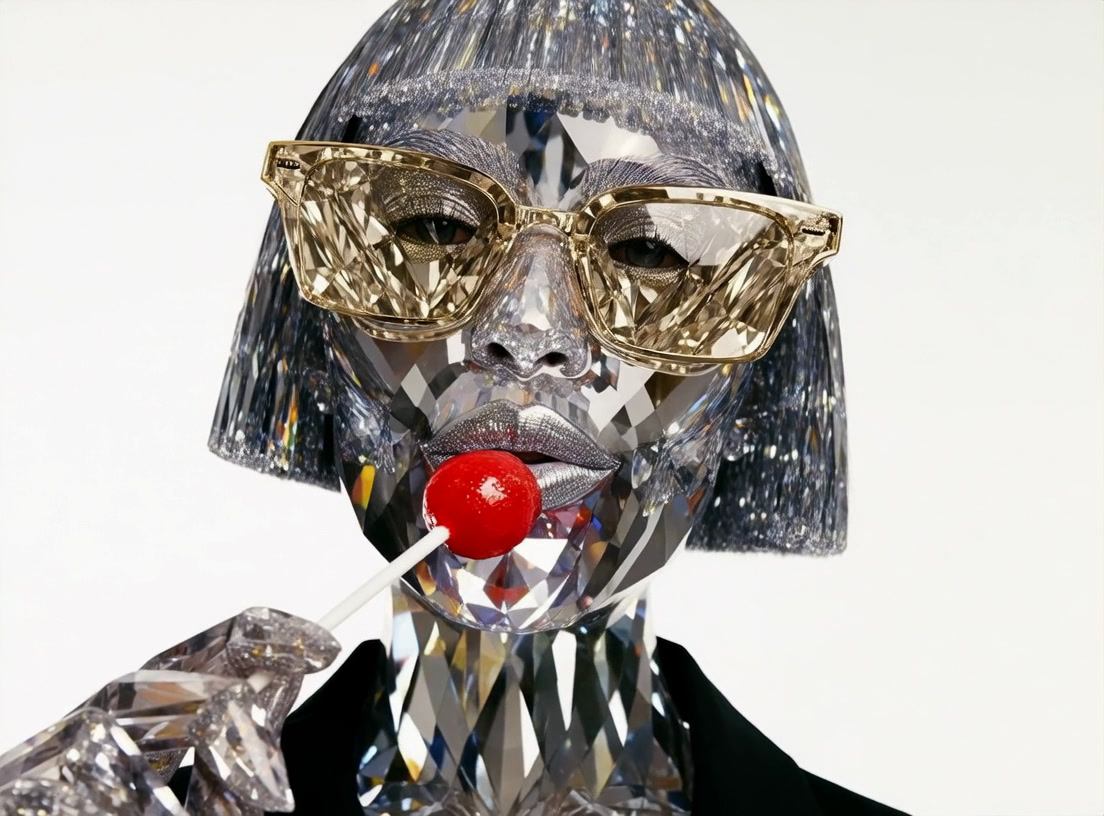} \\[2pt]

        \centering\rotatebox{90}{\textbf{Output Video}} &
        \includegraphics[width=\linewidth]{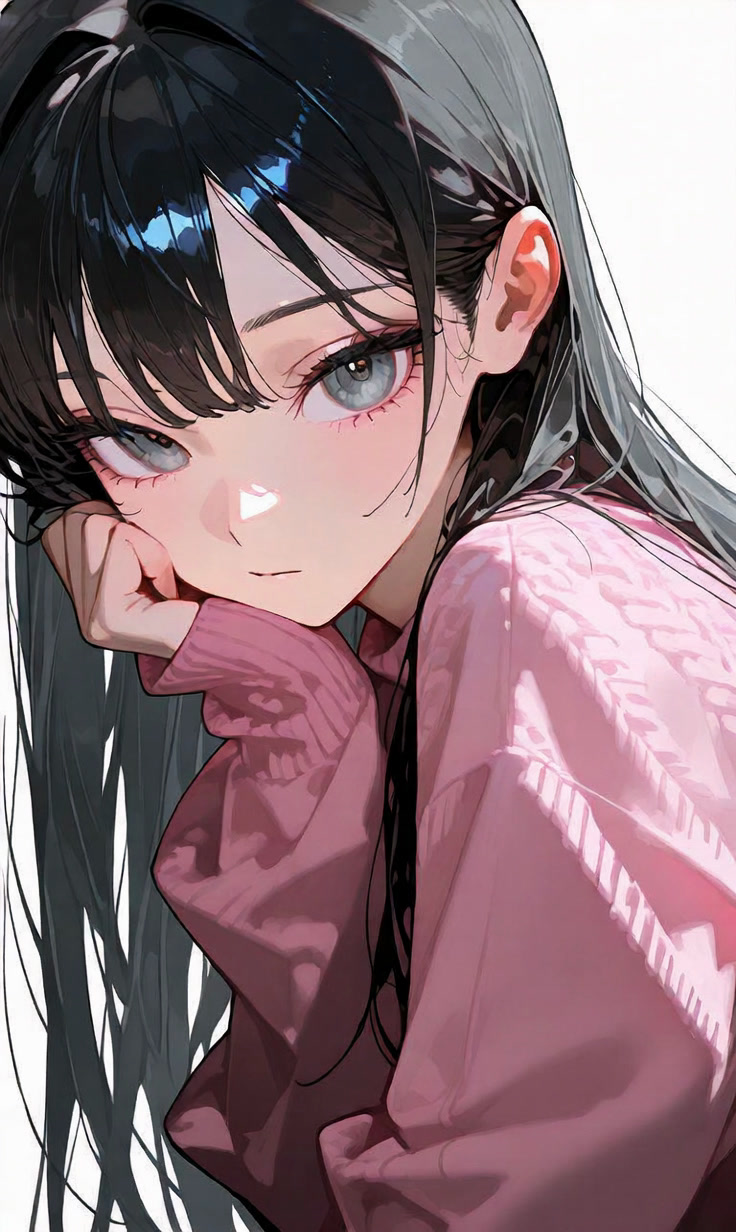} &
        \includegraphics[width=\linewidth]{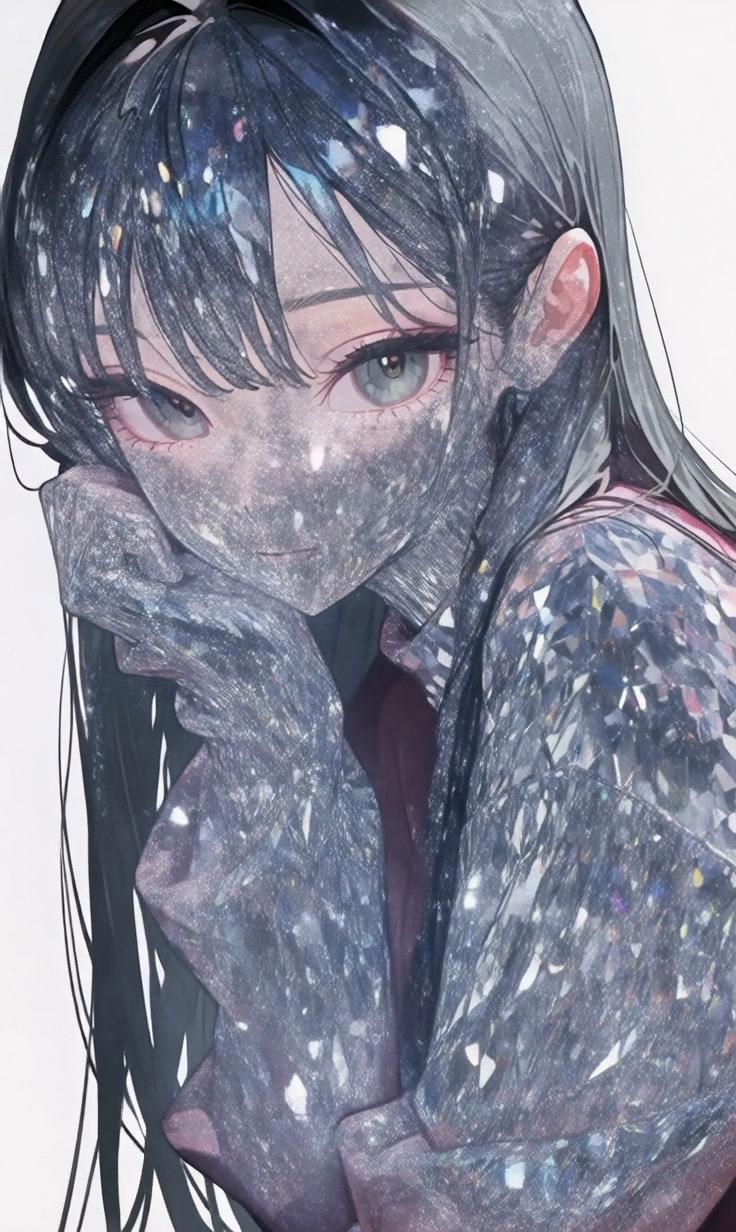} &
        \includegraphics[width=\linewidth]{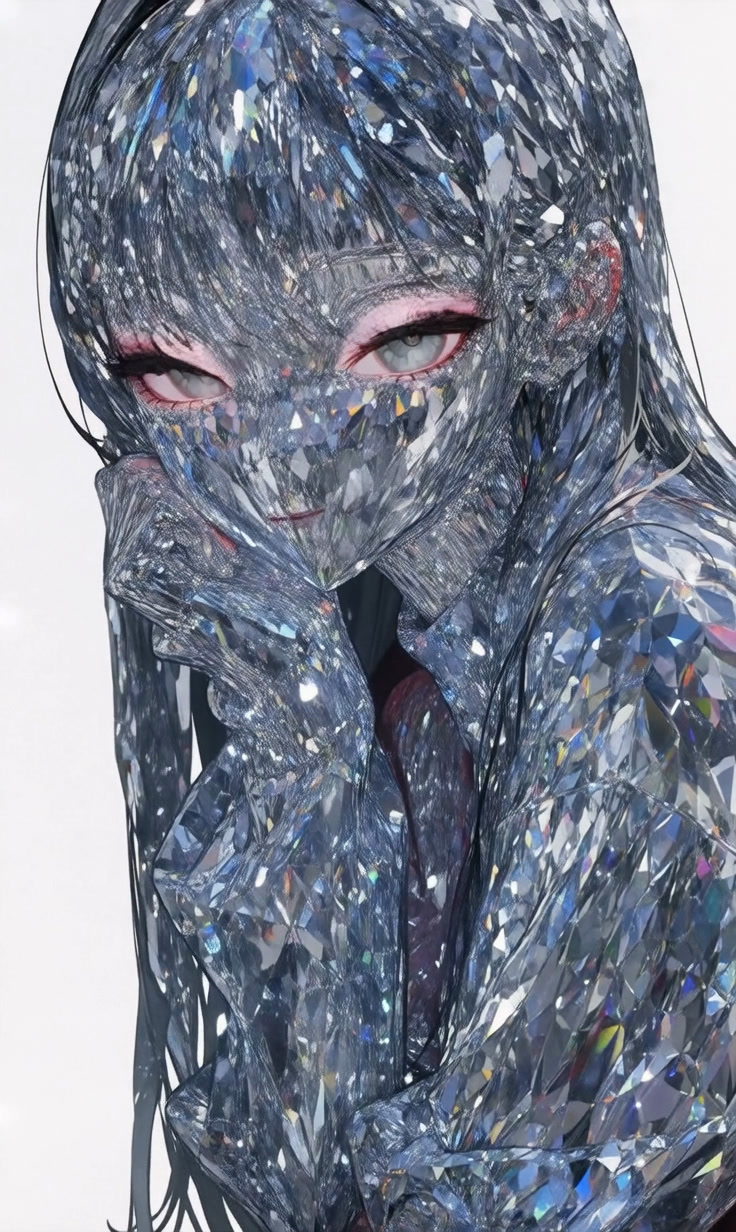} \\
    \end{tabular}
\end{flushleft}

\begin{figure}[h!]
    \centering
    \caption{Example of first-frame reference with effect reference.}
    \label{fig:edit_ref_firstframe_effect_1}
\end{figure}

\setcounter{table}{0} 
\renewcommand{\thetable}{A\arabic{table}} 
\setcounter{algorithm}{0}
\renewcommand{\thealgorithm}{A\arabic{algorithm}}

\label{app:skyreels-bench-detailed}

\end{document}